\pdfoutput=1
\documentclass{article}
\usepackage{iclr2027_conference,times}
\usepackage[T1]{fontenc}
\usepackage{amsmath,amssymb,amsthm}
\usepackage{booktabs}
\usepackage{array}
\usepackage{microtype}
\usepackage{xcolor}
\usepackage[hidelinks]{hyperref}
\hypersetup{pdftitle={Knowing the Store: What a Memory Backend Must Write Down Before an Agent Can Read It},pdfauthor={Ansuman Mullick, Eray T\"uz\"un}}
\usepackage{url}
\usepackage{tikz}
\usepackage{pgfplots}
\usepackage[section,above]{placeins}   %
\usepackage{float}
\usetikzlibrary{arrows.meta,positioning,shapes.misc,fit,calc}
\usepgfplotslibrary{groupplots}
\pgfplotsset{compat=1.18}
\definecolor{figmini}{HTML}{E69F00}
\definecolor{fighaiku}{HTML}{56B4E9}
\definecolor{figluna}{HTML}{009E73}
\definecolor{figsol}{HTML}{D55E00}
\definecolor{figopus}{HTML}{0072B2}
\definecolor{figcount}{HTML}{CC79A7}

\newtheorem{definition}{Definition}

\newcommand{\claim}[1]{\textbf{#1}}
\newcommand{\present}{\textsc{present}}
\newcommand{\absent}{\textsc{absent}}
\newcommand{\pn}{\textsc{pn}}
\newcommand{\trueidx}{\textsc{true}}
\newcommand{\believed}{\textsc{believed}}
\newcommand{\censusalias}{\label{tab:believed}}
\newcommand{\hd}[2]{\begin{tabular}[b]{@{}r@{}}#1\\#2\end{tabular}}
\newcommand{\hdl}[2]{\begin{tabular}[b]{@{}l@{}}#1\\#2\end{tabular}}
\newfloat{listing}{htbp}{lol}
\floatname{listing}{Listing}
\usetikzlibrary{patterns}

\title{Knowing the Store: What a Memory Backend Must Write Down Before an Agent Can Read It}
\author{Ansuman Mullick \& Eray T\"uz\"un \\
Bilkent University, Ankara, T\"urkiye \\
\texttt{ansuman.mullick@ug.bilkent.edu.tr}, \texttt{eraytuzun@cs.bilkent.edu.tr}}

\iclrfinalcopy
\begin{document}
\maketitle
\lhead{Preprint. Under review.}

\begin{abstract}
An agent with long-term memory can answer from a record it should no longer use, such as a plan the user later cancelled.
We ask what a memory store must expose for an agent to know this before retrieving anything, and we score that judgment on its own, as metamemory monitoring.
Readers see only a value-free summary: record counts by lifecycle state and a list of attribute names.
Each of $148$ questions is asked against three versions of one store that differ in one attribute, so wording cannot give the answer away.
What a model adds depends on the length of that list.
On the short lists the benchmark's ground truth produces ($2.6$ names on average), none of five language models reliably beats a cosine-similarity lookup over the names, and the three stronger ones are equivalent to it within $0.05$. Under a control that fixes counts and list length, none is reliably above it and none is shown equivalent.
On lists longer than the stores we tested write, padded to $60$ names, the lookup loses $0.18$; \mbox{GPT-5.6} Luna and Sol lose $0.06$ to $0.07$ and lead it by $0.13$ to $0.14$, while the other three fall with it. The lead holds for Sol against a sibling padded to the same length and counts.
Reworded to share no word with the names, the lookup loses $0.02$ and one stronger reader edges $0.04$ ahead.
On the short lists the three stronger readers lead only where the summary counts a state without naming it, and one count feature closes that lead within a question, though not when items are pooled.
The backends we tested omit or misstate this information: under FR-Bank's own metadata, \mbox{GPT-4.1} mini, Haiku 4.5 and the lookup fall from $0.73$, $0.82$ and $0.75$ to $0.60$, $0.71$ and $0.59$.
In these tests, what the store wrote down limited every reader we gave it to.
All stores are synthetic; control, whether a better judgment yields a better answer, is left to a later study.
\end{abstract}
\section{Introduction}\label{sec:intro}

An agent with a long memory must decide both whether its store covers a question and whether it holds something the agent must no longer use. A user plans a Saturday hike and then cancels it. If the agent retrieves the plan but misses the cancellation, it can answer from a plan that no longer applies. The relevant failure starts before the answer: the agent never asks for the record that would have stopped it.

We call the cancelled plan a \emph{forbidding record}: a non-active record whose use would make the answer wrong. The cancellation is the evidence that marks it. The \emph{obligation judgment} asks, before retrieval, whether the store holds such a record. The \emph{coverage judgment} asks whether the store covers the question at all. The second is the feeling of knowing (FOK) of the metamemory literature \citep{hart1965,nelson1990,koriat1993}; the first we call negative-obligation knowledge (NOK), a term this paper introduces. Metamemory separates monitoring a state from controlling its use \citep{nelson1990}; we study monitoring only, how accurately a reader judges the store from what it exposes. Whether the judgment then changes what is retrieved or answered is control, which we leave to a later study.

Most memory evaluations score what an agent retrieves or answers \citep{longmemeval,locomo,memoryagentbench,stale}. On MemoryAgentBench's multi-hop conflict-resolution questions, no memory mechanism exceeds $7\%$ accuracy and the best long-context model reaches $28\%$ \citep{memoryagentbench}. We instead ask what an agent can judge before retrieving any content, and what the store must expose for that judgment to be possible. A \emph{lifecycle summary} lists record counts by state and, per category, an attribute-name list with the values removed, so a store can expose it without disclosing stored values. In the hike example, the name can identify the plan without disclosing the plan itself. Figure~\ref{fig:framework} shows one such summary, abridged, beside a returned record with its tag.

Our diagnostic study separates the information in the summary from the reader's use of it. The central test holds the question fixed while changing the store. We compare language models with a name lookup that makes no generative-model call. We remove names or counts, change which states are named, and test questions whose intent licenses old content. We then examine backend writing and retrieval.
In these tests, every reader we tried was held back by what the store had written down.
Models differ further when the rubric states the intent rule.

Our contributions are:
\begin{itemize}\setlength{\itemsep}{1pt}
\item A counterfactual test of judgments made before retrieval. One question is asked of three versions of a store that differ in one attribute ($444$ items), so question wording cannot predict the label, and readers that make no generative-model call serve as baselines.
\item Evidence about what readers read. On the named state no model reliably exceeds a name lookup when the list is short, including under a control that holds counts and list length fixed, and one does on a $60$-name list with its length and counts fixed. The stronger readers' lead comes from the counts, which the two smaller models do not use.
\item A census of what four backends write and what it costs a reader. Under FR-Bank's own metadata mini, Haiku and the lookup fall to $0.60$, $0.71$ and $0.59$, and at retrieval a fetch keyed on the names reaches as far without a model's judgment as with it.
\end{itemize}

\begin{figure}[t]
\centering
\scriptsize
\begin{minipage}[t]{0.625\linewidth}
\textbf{Summary route}: read before retrieval\par\vspace{1pt}
\fbox{\begin{minipage}{0.96\linewidth}\ttfamily
LIFECYCLE TOTALS\\
\hspace*{1em}\makebox[7em][l]{active}\makebox[2em][r]{377}\\
\hspace*{1em}\makebox[7em][l]{superseded}\makebox[2em][r]{197}\\
\hspace*{1em}\makebox[7em][l]{retracted}\makebox[2em][r]{1}\\
\hspace*{1em}\makebox[7em][l]{expired}\makebox[2em][r]{49}\\
ATTRIBUTE NAMES BY CATEGORY (names only; abridged)\\
\hspace*{2ex}HOBBIES\_RECREATION: current\_reading\_list, hiking\_experience, hiking\_preference, ...\\
ATTRIBUTE NAMES CARRYING SUPERSEDED CONTENT (abridged)\\
\hspace*{2ex}..., exam\_date, ..., hiking\_preference, ...
\end{minipage}}
\end{minipage}\hfill
\begin{minipage}[t]{0.36\linewidth}
\textbf{Tag route}: read on returned records\par\vspace{1pt}
\fbox{\begin{minipage}{0.94\linewidth}\ttfamily
hiking\_preference [superseded]\\
``User prefers moderate difficulty hikes in the six to ten mile range.''
\end{minipage}}
\end{minipage}
\caption{A lifecycle summary as a reader sees it: counts by state and attribute names with the values removed. The summary route reads this before retrieval; the tag route reads lifecycle tags on returned records. Section~\ref{sec:framework} defines both.}
\label{fig:framework}
\end{figure}

\section{Related work}\label{sec:related}
Metamemory's monitoring and control distinction (Section~\ref{sec:intro}) carries over to an external store \citep{koriat1996}: coverage concerns whether relevant content exists, while obligation concerns content whose use would make the answer wrong.

Adaptive retrieval decides whether to query a static corpus using parametric knowledge or signals from generated content \citep{lewis2020,flare,mallen2023,selfrag,su2024}. Abstention work asks whether a question admits an answer \citep{rajpurkar2018,kamath2020,yang2024,abstentionbench}; calibration work evaluates confidence in an answer \citep{kadavath2022,lin2022,tian2023,xiong2024}. These decisions do not require a maintained account of an external store's lifecycle state. When they retrieve, they inspect a bounded view of content.

Memory systems train or implement writing, updating and querying \citep{automem,supersede,packer2023,zhong2024,zhang2025survey,mem0,amem}. Benchmarks evaluate recalled answers, obsolete-memory use or conflict resolution after retrieval \citep{longmemeval,locomo,memora,memoryagentbench}, and StateMemBench scores whether an answer reflects the current or a superseded state \citep{statemembench}. STALE is the closest comparison: it asks whether an agent recognizes that an already retrieved memory is no longer valid \citep{stale}. Its CUPMem prototype revises and consolidates state at write time. We ask what that structure must expose to support a judgment before retrieval.

Lifecycle information also appears in temporal databases, removal tombstones in replicated sets, and truth maintenance \citep{snodgrass1986,shapiro2011,doyle1979,zep,nissenbaum2004,cimemories,maini2024,gatemem,pmbench,hushbench,xu2024conflicts,longpre2021,dhingra2022}. Our target is the pre-retrieval judgment about the store, derived from the stored content's lifecycle state.

\section{Framework}\label{sec:framework}
\paragraph{Records and intent.}
Each stored record has content, a category, an attribute name, a lifecycle state, an optional event-time anchor and an ingest time \citep{fr}. The states are active, superseded, retracted and expired. We call the latter states \emph{non-active}. Expiry is derived from the event-time anchor, so an expired record can still carry an active tag. If several marks apply, labels take the first of retracted, superseded and expired. The question's intent is its time frame: current, upcoming, past or change. Intent determines whether old content is licensed for the answer. (Appendix~\ref{app:glossary})

\paragraph{Labels and judgments.}
\pn{} means present-negative: the store contains relevant content that must be suppressed for the answer to be right. We call a question with this label an \emph{obligation}. \present{} means licensed answering content exists and no relevant suppression obligation applies. \absent{} means neither exists. The obligation judgment estimates the probability of \pn{}; the coverage judgment estimates the probability of either \present{} or \pn{}.

\begin{definition}[Intent-conditional coverage]\label{def:coverage}
$\mathrm{Cov}(q,S)\in\{\present,\absent,\pn\}$.
The label is \pn{} when $S$ contains content that must be suppressed for the answer to be right under the question's intent; \present{} denotes licensed answering content with no obligation that matters for the answer, and \absent{} denotes neither. The grid below gives the label by the state of the relevant record and the intent.
\end{definition}

\begin{center}
\scriptsize
\setlength{\tabcolsep}{5pt}
\begin{tabular}{@{}lcccc@{}}
\toprule
Relevant record & \textsc{current} & \textsc{upcoming} & \textsc{past} & \textsc{change} \\
\midrule
active, licensed & \present{} & \present{} & \present{} & \present{} \\
superseded value & \pn{} & n.a. & n.a. & \present{} \\
retracted plan & \pn{} & \pn{} & n.a. & n.a. \\
expired event & \pn{} & \pn{} & \present{} & n.a. \\
none & \absent{} & \absent{} & \absent{} & \absent{} \\
\bottomrule
\end{tabular}
\par\vspace{1pt}{\scriptsize The \textsc{change} intent is never \pn{}, because both the old and the new value are licensed; n.a.\ means a combination the question construction never produces.}
\end{center}

For the Saturday hike, a still-valid plan supports \present{}; the cancelled plan creates \pn{} for a question about an upcoming hike; and no such record gives \absent{}. Intent prevents treating every old record as forbidden. A question about a change licenses both old and new values; a question about a past event can license the expired event.

\paragraph{Names and counts.}
The lifecycle summary is maintained outside query time. Its count table gives records by category and state, plus updates that superseded an entry. Its name map lists attributes by category and removes their values. A state is \emph{named} when its affected attributes are listed, and only \emph{counted} when the summary gives their count. The default summary names superseded content but only counts retracted and expired content. Cancelling the hike can therefore move a retraction count from zero to one without placing the hike's name in a retracted-name list. Naming that state exposes which attribute the count concerns.

The summary route reads this description before retrieval. The tag route reads returned records with their lifecycle tags. Masking non-active records removes the records whose tags could reveal an obligation. A rendered summary appears in Appendix~\ref{app:construction}.

\section{Setup}\label{sec:setup}
\begin{figure}[t]
\centering
\footnotesize
\setlength{\tabcolsep}{2pt}
\renewcommand{\arraystretch}{0.95}
\begin{tabular}{@{}>{\raggedright\arraybackslash}p{0.34\linewidth}>{\raggedright\arraybackslash}p{0.20\linewidth}>{\raggedright\arraybackslash}p{0.20\linewidth}>{\raggedright\arraybackslash}p{0.23\linewidth}@{}}
\multicolumn{4}{@{}l}{\textbf{A retraction triple.} Question: \textit{What should I expect during my next Brazil Carnival trip?}}\\
\toprule
summary line & \present{} version & \pn{} version & \absent{} version\\
\midrule
retraction total & \texttt{0} & \texttt{1} & \texttt{0}\\
category: entries, active, retracted & \texttt{55, 55, 0} & \texttt{56, 55, 1} & \texttt{54, 54, 0}\\
category names & \texttt{..., trip\_plan} & \texttt{..., trip\_plan} & \texttt{...} (no \texttt{trip\_plan})\\
superseded names & \texttt{(none)} & \texttt{(none)} & \texttt{(none)}\\
name lookup score & $0.00$ & $0.00$ & $0.00$\\
what changes & the trip is on & the trip is cancelled & no such record\\
\bottomrule
\end{tabular}
\caption{The Carnival retraction triple: one question, three store versions differing in one attribute. Only the count distinguishes \present{} from \pn{}; \absent{} also loses the category name. The default name lookup reads the superseded list, so it ties the versions. Only changed summary lines are shown (the supersession triple is in Appendix~\ref{app:counterfactual}).}
\label{fig:example}
\end{figure}
\paragraph{Stores and readers.}
The main backends are FR-Bank, Mem0 and A-MEM \citep{fr,mem0,amem}; Graphiti supplies a further comparison on the same personas \citep{zep}. FR-Bank tags every entry but masks non-active records in its stock retrieval configuration. Its released evaluation reports a $5\%$ pass rate on retraction questions with that mask. Our tag-route experiments turn it off. (Appendix~\ref{app:census})

The models are \mbox{GPT-4.1} mini, Haiku 4.5, \mbox{GPT-5.6} Luna, \mbox{GPT-5.6} Sol and Opus 5, abbreviated mini, Haiku, Luna, Sol and Opus in that order. We call the last three the stronger readers and the first two the smaller, by vendor tier. Kimi K3 (open weights) joins the cross-vendor comparison. Every model row in Table~\ref{tab:cf-main} averages five draws. Not every reader ran every condition, and run settings differ across rows; Appendix~\ref{app:laddertable} lists both.

\paragraph{Labels and summaries.}
The personas, their human-written manifests and sessions, the FR-Bank, Mem0 and A-MEM stores and part of the main question set come from the FR-Bank release (its LifecycleBench personas) \citep{fr}. Language models drafted our other questions, and we built the Graphiti graphs from the released FR-Bank records, with lifecycle fields removed. Those manifests provide the labels, with every \pn{} label checked by hand, and a presence control checks that targets are stored. Intent is fixed by question construction and is not supplied as a label to the reader. (Appendix~\ref{app:construction})

A \emph{ground-truth summary} uses the manifest-based ground truth; a \emph{store-written summary} uses fields the backend exposes. The same renderer caps each list at fifteen names, a cap that binds only on store-written lists, which Section~\ref{sec:backends} reports capped and uncapped. For Mem0 and A-MEM we also render the manifest under each backend's constraints; this constrained rendering is ours and differs from a summary built from their own fields. (Appendix~\ref{app:inversion})

\paragraph{Tests and comparisons.}
The \emph{counterfactual set} asks each of its questions of three versions of one store. It contains $444$ items from $148$ questions, and the versions differ only in the target attribute. Questions are written before any version exists, so wording cannot predict which version carries the obligation; Figure~\ref{fig:example} shows the Carnival example. The \emph{full obligation set}, $661$ questions about the original stores and their variants, supports the backend and retrieval comparisons.

Readers return probabilities under a fixed rubric. The scored step asks whether stale content about the question's target is present, which coincides with Definition~\ref{def:coverage}'s obligation on the counterfactual set, where every question has current or upcoming intent; Section~\ref{sec:intent} tests intent separately. A \emph{question-only} prompt omits the summary; a prompt \emph{with the summary} adds only that text. The \emph{name lookup} scores the largest cosine similarity between the question and a name on the superseded list, computed by a small sentence encoder, and makes no generative-model call. Under the summary naming all states, the same lookup reads all non-active names. (Appendix~\ref{app:baselines})

We score separation by AUROC, the chance that a positive scores above a negative \citep{hanley1982}. Chance is $0.50$. The pooled AUROC over all items was fixed in advance. Because the counterfactual set fixes the question, we also report the matched AUROC, the share of triples in which the \pn{} version outscores the \present{} version, ties counting half. It is post hoc, and we say where the two metrics disagree. A difference is \emph{reliable} when both the item-level paired and persona-grouped bootstrap intervals exclude zero. We apply no multiplicity correction; reliable describes each of the $71$ comparisons the main text makes on its own. \emph{Within tolerance} means a point difference no greater than $0.05$; it does not establish equivalence, which we test separately at $0.05$ (Appendix~\ref{app:counterfactual}). A gain compares the same questions with and without the summary.

Question-only probes can recover labels from the original question wording, which motivates paired comparisons and the counterfactual design. The predictions were written into the repository's commit history (Appendix~\ref{app:registration}), each before the run that tests it: $34$ of $106$ came out otherwise. They include follow-ups rather than a single plan fixed before all data collection. (Appendix~\ref{app:registration})

\section{What readers read}\label{sec:counterfactual}\label{sec:readers}
\subsection{Names and counts}\label{sec:names_counts}
\claim{Within a triple, a name or a count can identify the obligation.}
The \pn{} version differs from its siblings in one attribute, so its category holds one more non-active record. A privileged count reader, handed the question's category that the models must infer, wins every retraction and expiry triple, and the name lookup wins $82$ of the $88$ supersession triples and loses none. For $26$ of $30$ retraction pairs and $15$ of $30$ expiry pairs that count starts at zero, so a reader can identify the store version without identifying the attribute. What differs across readers is which of the two they use.

\claim{The two smaller models lose nothing when the counts are removed.}
With every count deleted, mini and Haiku score as before (Table~\ref{tab:robust}); given the counts alone, they fall by $0.14$. Told that one more plan was cancelled without being told which, they cannot use it: on retraction and expiry both stay reliably below a count reader that merely guesses the category.

\begin{figure}[t]
\centering
\pgfplotstableread[col sep=comma]{
x,state,mini,miniP,miniM,haiku,haikuP,haikuM,luna,lunaP,lunaM,sol,solP,solM,opus,opusP,opusM,lookup,lookupP,lookupM,count,countP,countM
0,Supersession,0.8920,0.0567,0.0692,0.9148,0.0467,0.0517,0.9602,0.0335,0.0427,0.9602,0.0293,0.0371,0.9773,0.0227,0.0341,0.9659,0.0285,0.0373,0.6932,0.0634,0.0644
1,Retraction,0.5667,0.1747,0.1761,0.5667,0.0940,0.0842,0.6500,0.1714,0.1985,0.5833,0.1667,0.1726,0.7667,0.1451,0.1556,0.5000,0.0000,0.0000,0.7667,0.0852,0.0770
2,Expiry,0.4000,0.1143,0.1143,0.5833,0.1283,0.1146,0.8667,0.1010,0.1081,0.8333,0.1061,0.1333,0.6833,0.1917,0.1833,0.5000,0.0000,0.0000,0.7500,0.0761,0.0887
}\figthreeSup
\pgfplotstableread[col sep=comma]{
x,state,mini,miniP,miniM,haiku,haikuP,haikuM,opus,opusP,opusM,lookup,lookupP,lookupM
0,Supersession,0.9205,0.0465,0.0569,0.8920,0.0492,0.0553,0.9659,0.0341,0.0519,0.9432,0.0378,0.0410
1,Retraction,0.9333,0.0667,0.1119,0.9333,0.0667,0.0657,0.8333,0.1099,0.1333,0.8167,0.0773,0.0846
2,Expiry,0.7833,0.1274,0.1404,0.9167,0.0661,0.0780,0.9667,0.0333,0.0738,0.9000,0.0545,0.0667
}\figthreeAll
\begin{tikzpicture}
\begin{groupplot}[group style={group size=2 by 1, horizontal sep=0.9cm, ylabels at=edge left, yticklabels at=edge left},
    width=0.41\linewidth, height=4.5cm, xmin=-0.5, xmax=2.5, ymin=0.35, ymax=1.0,
    xtick={0,1,2}, xticklabels={Supersession,Retraction,Expiry}, ytick={0.4,0.5,0.6,0.7,0.8,0.9,1.0},
    ylabel={matched AUROC}, ylabel style={font=\scriptsize}, tick label style={font=\scriptsize}, title style={font=\scriptsize, yshift=-3pt},
    ymajorgrids, grid style={black!12}, axis line style={black!60}, clip=false,
    error bars/y dir=both, error bars/y explicit, error bars/error mark options={mark size=0.8pt, rotate=90}, error bars/error bar style={line width=0.35pt},
    legend style={at={(1.04,0.5)}, anchor=west, font=\scriptsize, draw=none, fill=none, row sep=-1pt, nodes={anchor=west}}]
  \nextgroupplot[title={only supersession named}]
      \draw[dashed, black!60] (axis cs:-0.5,0.5) -- (axis cs:2.5,0.5);
      \addplot[only marks, mark=*, mark size=2.0pt, figmini, xshift=-9pt] table[x=x, y=mini, y error plus=miniP, y error minus=miniM] {\figthreeSup};
      \addplot[only marks, mark=square*, mark size=1.9pt, fighaiku, xshift=-6pt] table[x=x, y=haiku, y error plus=haikuP, y error minus=haikuM] {\figthreeSup};
      \addplot[only marks, mark=triangle*, mark size=2.3pt, figluna, mark options={fill=white}, xshift=-3pt] table[x=x, y=luna, y error plus=lunaP, y error minus=lunaM] {\figthreeSup};
      \addplot[only marks, mark=diamond*, mark size=2.3pt, figsol, mark options={fill=white}, xshift=0pt] table[x=x, y=sol, y error plus=solP, y error minus=solM] {\figthreeSup};
      \addplot[only marks, mark=pentagon*, mark size=2.2pt, figopus, xshift=3pt] table[x=x, y=opus, y error plus=opusP, y error minus=opusM] {\figthreeSup};
      \addplot[only marks, mark=x, mark size=2.6pt, black, thick, xshift=6pt] table[x=x, y=lookup, y error plus=lookupP, y error minus=lookupM] {\figthreeSup};
      \addplot[only marks, mark=+, mark size=2.6pt, figcount, thick, xshift=9pt] table[x=x, y=count, y error plus=countP, y error minus=countM] {\figthreeSup};
  \nextgroupplot[title={all states named}]
      \draw[dashed, black!60] (axis cs:-0.5,0.5) -- (axis cs:2.5,0.5);
      \addplot[only marks, mark=*, mark size=2.0pt, figmini, xshift=-9pt] table[x=x, y=mini, y error plus=miniP, y error minus=miniM] {\figthreeAll};
      \addplot[only marks, mark=square*, mark size=1.9pt, fighaiku, xshift=-6pt] table[x=x, y=haiku, y error plus=haikuP, y error minus=haikuM] {\figthreeAll};
      \addlegendimage{only marks, mark=triangle*, mark size=2.3pt, figluna, mark options={fill=white}}
      \addlegendimage{only marks, mark=diamond*, mark size=2.3pt, figsol, mark options={fill=white}}
      \addplot[only marks, mark=pentagon*, mark size=2.2pt, figopus, xshift=3pt] table[x=x, y=opus, y error plus=opusP, y error minus=opusM] {\figthreeAll};
      \addplot[only marks, mark=x, mark size=2.6pt, black, thick, xshift=6pt] table[x=x, y=lookup, y error plus=lookupP, y error minus=lookupM] {\figthreeAll};
      \addlegendimage{only marks, mark=+, mark size=2.6pt, figcount, thick}
      \addlegendimage{dashed, black!60}
      \legend{mini (five draws), Haiku, Luna (left only), Sol (left only), Opus, name lookup, count reader (left only), chance}
\end{groupplot}
\end{tikzpicture}
\caption{Per-state matched AUROC on the counterfactual set (\pn{} against \present{} within the triple; post hoc); the dashed line is chance. Mini uses five draws, matching Table~\ref{tab:cf-main}. The count reader guesses the category. Luna and Sol ran only the default summary. Every value is in the tables of Appendix~\ref{app:counterfactual}; bars: persona-grouped 95\% intervals.}
\label{fig:perstate}
\end{figure}
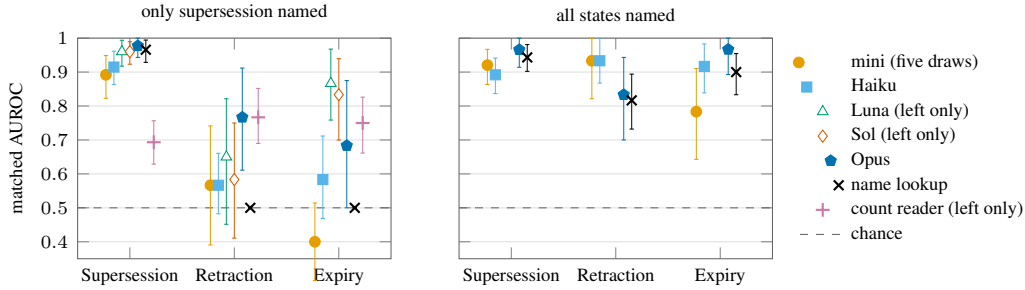

\claim{The three stronger readers also read the counts.}
On retraction and expiry, which the default summary only counts, none of Luna, Sol and Opus sits reliably below the guessing count reader (Figure~\ref{fig:perstate}). When the obligation's extra count moves to a sibling version with the names unchanged, all three follow it, while mini and Haiku show no reliable change (Table~\ref{tab:robust}). The falls are reliable on the matched metric and, pooled, on expiry for all three and on retraction for Opus alone. Cancelling the hike changes only a count in the default summary, and these three readers follow it. (Appendix~\ref{app:counterfactual})

\begin{table}[t]
\caption{Counterfactual obligation AUROC: $88$ supersession, $30$ retraction and $30$ expiry positives. Brackets: persona intervals. The matched column is the within-triple comparison of \pn{} against \present{}, $(W+T/2)/148$; it and W/T/L are post hoc; persona intervals keep triples intact. Question-only AUROC is $0.50$. Differences at four decimals. (Appendix~\ref{app:counterfactual})}
\label{tab:cf-main}
\centering
\scriptsize
\setlength{\tabcolsep}{3pt}
\begin{tabular}{@{}llrlrl@{}}
\toprule
Reader & Named states & AUROC & persona interval & matched & W/T/L \\
\midrule
mini$^{a}$ & superseded & $0.76$ & $[0.72,0.80]$ & $0.73$ & $100/15/33$ \\
mini$^{a}$ & all & $0.83$ & $[0.79,0.87]$ & $0.90$ & $130/5/13$ \\
Haiku & superseded & $0.76$ & $[0.71,0.80]$ & $0.78$ & $89/52/7$ \\
Haiku & all & $0.88$ & $[0.84,0.91]$ & $0.91$ & $125/18/5$ \\
Luna & superseded & $0.84$ & $[0.80,0.88]$ & $0.88$ & $129/2/17$ \\
Luna & all & -- & -- & -- & -- \\
Sol & superseded & $0.85$ & $[0.81,0.88]$ & $0.86$ & $126/2/20$ \\
Sol & all & -- & -- & -- & -- \\
Opus & superseded & $0.84$ & $[0.80,0.88]$ & $0.88$ & $128/3/17$ \\
Opus & all & $0.91$ & $[0.88,0.94]$ & $0.94$ & $139/0/9$ \\
\midrule
\multicolumn{6}{@{}l}{\textit{Readers with no generative-model call}}\\
Name lookup & superseded & $0.77$ & $[0.72,0.81]$ & $0.78$ & $82/66/0$ \\
Name lookup & all & $0.88$ & $[0.84,0.91]$ & $0.91$ & $121/27/0$ \\
Count reader, guessed category (share) & superseded & $0.66$ & $[0.63,0.70]$ & $0.72$ & $65/83/0$ \\
Count reader, given category (count) & superseded & $0.60$ & $[0.58,0.64]$ & $0.70$ & $60/88/0$ \\
Lookup with one count feature & superseded & $0.79$ & $[0.75,0.82]$ & $0.89$ & $114/34/0$ \\
\bottomrule
\end{tabular}
\par\vspace{2pt}
\begin{tabular}{@{}llrll@{}}
\toprule
All states named: Opus minus name lookup & metric & difference & item interval & persona interval \\
\midrule
 & pooled & $+0.0323$ & $[+0.0060,+0.0617]$ & $[+0.0077,+0.0608]$ \\
 & matched & $+0.0304$ & $[-0.0068,+0.0643]$ & $[-0.0038,+0.0648]$ \\
\bottomrule
\end{tabular}
\par\smallskip
\begin{minipage}{0.97\linewidth}\scriptsize
$^{a}$Mini uses the provider's default temperature here. Its earlier one-draw rows at temperature $0$ score $0.72$ and $0.80$; both draw count and temperature change (Table~\ref{tab:mini-k5}). The difference row's pooled comparison is from Table~\ref{tab:counterfactual}, its matched comparison from Table~\ref{tab:cf-matched}. A dash marks a condition the model did not run.
\end{minipage}
\end{table}

\subsection{The lookup against the models}\label{sec:lookup}
\claim{On the named state, no model reliably beats the lookup on the short list; on a long one, Sol does with list length and counts fixed.}
Readers are compared within each triple on the matched metric, the pooled metric named where the two disagree. The default summary names only supersession; on its triples the lookup scores $0.97$, Opus is $0.01$ above it, Luna and Sol $0.01$ below, and only mini reliably below. Pooled, the lookup scores $0.92$ and the three $0.93$ to $0.94$, none reliably above it. Within $0.05$, Luna, Sol and Opus are equivalent to the lookup on the short list; mini sits reliably below it on the matched metric only; for Haiku neither holds (Appendix~\ref{app:counterfactual}). At that ceiling equivalence is the expected outcome; the sibling-name control is where a reader could show more, and none does. A sibling-name control removes list length and counts with a fourth store version listing a different attribute of the same category, with the same counts and as many names. The lookup and every model pick the \pn{} version over it at $0.88$ or higher, none reliably above the lookup (Opus, the closest, $+0.02$ $[-0.02,+0.07]$) and none shown equivalent, while meaningless tokens and the count reader sit near chance. %
Table~\ref{tab:robust} lists the checks we ran to break this result and what each removes. With the superseded-name list padded to $60$ names (Table~\ref{tab:robust}; the default lengths in Appendix~\ref{app:length}), the lookup loses $0.18$ and mini, Haiku and Opus fall with it, ending at or below it with no reliable difference. Luna and Sol lose $0.06$ to $0.07$ and lead the lookup by a reliable $0.13$ to $0.14$ pooled and $0.09$ to $0.10$ within triples. The pad is identical across a triple. The \pn{} list is one to three names longer than its siblings' in every supersession triple, on the short list and the padded one alike; a reader that outputs only the list length separates them in every such triple. The sibling-name control removed that cue on the short list. A sibling padded to the same length and counts, run on Luna, Sol and mini, removes that cue: the lookup scores $0.82$, and Sol leads it by a reliable $0.09$. Luna's lead holds on the pooled metric only ($0.11$, reliable) and not within triples ($+0.04$; Appendix~\ref{app:counterfactual}).

\begin{table}[t]
\caption{The checks we ran to break the lookup result, what each removes, and what happened. Short ground-truth lists do not establish robustness to longer store-written ones. In row 9 retraction and expiry do not move, since the list names neither. (Appendices~\ref{app:counterfactual} and~\ref{app:census})}
\label{tab:robust}
\centering
\scriptsize
\setlength{\tabcolsep}{3pt}
\renewcommand{\arraystretch}{0.95}
\begin{tabular}{@{}p{0.255\linewidth}p{0.19\linewidth}p{0.525\linewidth}@{}}
\toprule
What we tried to break & What it removes & What happened \\
\midrule
One question, three store versions & Question-wording leakage & A question-only reader scores $0.50$ by construction \\
Random tokens for the names$^{p}$ & Name meaning, length kept & $0.65$; the real lookup is $0.13$ higher \\
Sibling-name control & List length and counts & Lookup $0.95$, models $0.88$ to $0.97$, none reliably above the lookup \\
Lifecycle totals permuted & The summary's two totals & No model moves by more than $0.01$, none reliably \\
Count moved to a sibling version$^{p}$ & Where the count sits & Luna, Sol, Opus fall $0.28$ to $0.33$; mini, Haiku no reliable change \\
Every count deleted & The counts & Mini and Haiku within $0.01$; Luna loses $0.06$, Opus $0.04$ \\
Questions reworded, no shared word & Word overlap with names & Word lookup at $0.50$; embedding lookup loses $0.02$; mini, Haiku lose more \\
Two other sentence encoders & The choice of encoder & Both within $0.01$ of the lookup \\
List padded to 60 names & Short, clean lists & Lookup $0.92$ to $0.74$; mini, Haiku, Opus fall with it; Luna, Sol lose $0.06$ to $0.07$, lead it by $0.13$ to $0.14$ \\
The store's own metadata (FR-Bank) & The ground-truth summary & Mini, Haiku, lookup: $0.73$, $0.82$, $0.75$ to $0.60$, $0.71$, $0.59$ \\
\bottomrule
\end{tabular}
\end{table}

\claim{The result survives rewording.}
With the questions rewritten to share no content word with any name ($115$ pairs), the embedding lookup loses $0.02$ and the two smaller models lose more and sit below it (Table~\ref{tab:robust}). Luna, added later, edges $0.04$ above it on the pooled metric only. Under the all-states summary the lookup loses $0.06$ pooled and $0.02$ within triples.

\claim{The stronger readers' lead comes from counts, and a count feature closes it within a question.}
Across all $148$ triples Luna, Sol and Opus score a reliable $0.08$ to $0.10$ above the lookup. All of that comes from retraction and expiry, where the summary lists no name and the lookup ties every triple. A lookup given one more feature, the guessed category's share of non-active records, scores $0.89$ beside the three models' $0.86$ to $0.88$; equivalence within $0.05$ is neither shown nor excluded (Appendix~\ref{app:counterfactual}). Pooled across questions the gap remains: the same reader scores $0.79$ and the models stay a reliable $0.05$ to $0.06$ above it (Table~\ref{tab:cf-main}).

\claim{Naming all states lifts every reader that ran it.}
With the retracted and expired attributes listed too, every reader that ran the summary rises (Table~\ref{tab:cf-main}), Haiku by a reliable $0.35$ on the two states that were only counted. Opus's $0.03$ lead over the all-states lookup is reliable on the pooled metric and not on the matched one, so with every state named no model is reliably ahead on both.

\subsection{When intent licenses old content}\label{sec:intent}
\claim{The stated intent rule separates the stronger readers.}
The supersession crossover holds the store fixed and changes the question from the current value to how it changed, using the default summary; the change question licenses the old content. When the rubric states that rule, Luna, Sol and Opus score $0.88$ to $0.99$, while mini and Haiku, told the same rule, score $0.48$ and $0.55$. A fixed cue list is not reliably below the best model ($+0.01$ $[-0.02,+0.03]$), so the test shows rule-following on cued questions and no general model advantage over deterministic intent handling.

\claim{Only Opus lowers its judgment without being told.}
Under the main-study rubric, mini, Haiku, Luna and Sol score $0.35$ to $0.44$ on the same contrast, against Opus's $0.90$, scoring the change question higher. The main-study rubric's scored step asks whether stale content is present, which is not the intent-conditional target of Definition~\ref{def:coverage}; the explicit-intent rubric asks the defined question.

\claim{The explicit rule also separates readers on expired events.}
The expiry crossover uses the summary naming all states; its past-event question licenses the record forbidden by the upcoming-event question. Under the explicit rule Luna, Sol and Opus score $0.92$ to $1.00$ and mini and Haiku do not separate the intents; Sol also separates them under the main-study rubric, at $0.64$. (Appendix~\ref{app:counterfactual})

\section{What backends write}\label{sec:backends}\label{sec:inversion}\label{sec:stores}
\claim{The backends we tested omit or misstate the lifecycle information readers need.}
FR-Bank marks far more entries stale than the reference does, yet retains obsolete assertions as active and conflates retraction with supersession. Mem0 overwrites at the retrieval surface we evaluated, and A-MEM's links resolve to no note in our integration (Table~\ref{tab:census}).

\begin{table}[h]
\caption{Backend census: counts of disagreement with the manifest-and-alignment reference, without hand adjudication or sampling intervals. Lower panel: coverage losses and rebuilt-summary scores. (Appendices~\ref{app:inversion} and~\ref{app:census})}
\label{tab:census}\censusalias%
\centering
\scriptsize
\setlength{\tabcolsep}{2pt}
\begin{tabular}{@{}p{0.10\linewidth}p{0.08\linewidth}p{0.80\linewidth}@{}}
\toprule
Backend & Mode & Census measurement \\
\midrule
FR-Bank & Over-mark & Of $18{,}936$ entries and $109$ events, marks $5{,}968$ entries stale against ${\sim}102$ in the reference (about $58\times$) and retains an active obsolete assertion after $72\%$ of supersession events, at the stored-versus-obsolete matching threshold of the census. \\
Mem0 & Overwrite & At the retrieval surface we evaluated (\texttt{search} and \texttt{get\_all}, \texttt{mem0ai} 1.0.5), $12.7\%$ of $10{,}074$ entries expose \texttt{updated\_at} and nothing else, because \texttt{UPDATE} overwrites the record it exposes and ingestion is silently lossy; its per-memory history call was not consulted. \\
A-MEM & Unresolved links & In our integration (\texttt{agentic-memory} 0.0.1), all $23{,}186$ link references across $11{,}319$ notes resolve to $0$ note identifiers, on the $40$ scored personas (not checked against the reference implementation). \\
\bottomrule
\end{tabular}
\par\vspace{4pt}
\begin{tabular}{@{}p{0.70\linewidth}rr@{}}
\toprule
Coverage judgment, \mbox{GPT-4.1} mini & Mem0 & A-MEM \\
\midrule
loss under the constrained rendering & $0.10$ & $0.11$ \\
score with the names rebuilt from ground truth & $0.88$ & $0.90$ \\
\midrule
\multicolumn{3}{@{}p{0.98\linewidth}@{}}{On the $66$ questions pooled across both stores, \mbox{GPT-5.6} Luna (raw API) loses $0.29$ and \mbox{GPT-5.6} Sol (agent tool) $0.10$.}\\
\bottomrule
\end{tabular}
\end{table}

\claim{FR-Bank's metadata leave mini and the lookup near chance.}
Given only what the store wrote, its capped summary on the full obligation set leaves mini at $0.60$ and the lookup at $0.59$. Haiku, at $0.71$, keeps more of its gain. Lifting the cap raises all three without restoring their ground-truth scores. The loss is no artifact of the obligations the alignment misses: on the ones it marks, the capped summary still costs each reader a reliable $0.13$ to $0.20$. The cancelled hike is only as findable as the store's own marks make it. (Appendix~\ref{app:census})

\claim{Graphiti writes part of what the judgment needs, and part of its loss is extraction.}
Of the $34$ full-store obligations it could express, its invalidation marks $13$, and $11$ were never extracted into the graph. Retraction is inexpressible under this backend's mapping. On the $23$ obligations whose content the graph retains, mini's capped store-written score of $0.69$ against $0.84$ under ground truth is still a reliable loss. (Appendix~\ref{app:posthoc-graphiti})

\claim{Mem0 and A-MEM lose coverage when their summaries lose names.}
Under the constrained rendering mini loses coverage on both stores and Luna and Sol lose on the questions pooled across them (Table~\ref{tab:census}). Omitting or rewording the unavailable-field marker does not remove the Mem0 loss, and Luna loses $0.27$ to $0.34$ under every rendering tested through the raw API. Store-written summaries remain within tolerance of question only, and rebuilding name-bearing summaries from ground truth restores mini's coverage on both stores (Table~\ref{tab:census}). (Appendix~\ref{app:inversion})

\section{What retrieval puts in view}\label{sec:downstream}\label{sec:routes}\label{sec:retrieval}
\claim{The tagged view complements the summary for mini.}
FR-Bank's cosine search returns records without a lifecycle mask on the full obligation set. Mini scores $0.73$ with the default summary, $0.69$ with the retrieved records and their ground-truth tags, and $0.83$ with both, a reliable gain over the summary alone. The two routes mirror each other by state: the view helps mini on expiry, where the summary only counts, and hurts it on supersession, where the summary names, both effects reliable (Table~\ref{tab:routes}). Haiku's score with both stays within tolerance of its summary alone; answers are not scored here (Appendix~\ref{app:resp}).

\claim{FR-Bank's own tags help mini despite over-marking.}
With the store's native tags in place of the ground-truth ones, mini's view-only score rises reliably to $0.81$. Their precision over retrieved lines is only $0.14$, and the rise comes from their marking more of the forbidding records already in view. On Graphiti, native tags on the same retrieved edges bring no reliable improvement. (Appendix~\ref{app:posthoc-graphiti})

\claim{The lifecycle fetch supplies the added reach, gated or not.}
The fetch adds the store's non-active records for an attribute matched to the default name list. It raises reach from $0.52$ to $0.82$, the always-on fetch reaching as far as the model-gated one at the matched budget, and the name-lookup rule reaches the same level (Table~\ref{tab:reach}). Unmatched, the always-on fetch spends $1.20$ times the lines per question; truncated to the gated budget it loses licensed records, a reliable $0.04$.

\begin{table}[h]
\caption{Reach on the full obligation set: forbidding records among the $104$ obligations, mean lines over the $661$ questions, and licensed records among the $174$ items. Matched rows interpolate between depths (Table~\ref{tab:gatereach}, Appendix~\ref{app:resp}). No answer is scored here.}
\label{tab:reach}
\centering\scriptsize\setlength{\tabcolsep}{4pt}\renewcommand{\arraystretch}{0.9}
\begin{tabular}{@{}lrrrr@{}}
\toprule
Policy & \hd{Forbidding records}{in view (of $104$)} & rate & \hd{Mean lines}{per question} & \hd{Licensed record}{in view (of $174$)} \\
\midrule
Top-five retrieval & $54$ & $0.52$ & $5.00$ & $0.37$ \\
Similarity, matched budget & -- & $0.54$ & $5.43$ & $0.39$ \\
Newest per attribute, matched & -- & $0.23$ & $5.43$ & $0.29$ \\
Always-on fetch, matched & -- & $0.82$ & $5.43$ & $0.34$ \\
Lookup rule & $85$ & $0.82$ & $5.46$ & $0.37$ \\
Model rule & $85$ & $0.82$ & $5.43$ & $0.37$ \\
Oracle & $99$ & $0.95$ & $5.33$ & $0.37$ \\
\bottomrule
\end{tabular}
\end{table}

\claim{Keeping the newest record misses every tested retraction obligation.}
At the matched budget, newest-per-attribute retrieval reaches none of the $26$ retraction obligations (Table~\ref{tab:reach}). The cancelled hike shows why: its cancellation is not the newest record of the attribute it cancels.

\claim{Active copies complicate the use of tagged records.}
Among $89$ scorable obligation views, $35$ contain an active line asserting the stale value. The answers the exploratory detector flags trace to these untagged copies even when the forbidding record is present and tagged (Appendix~\ref{app:resp}). An intervention that tags the copies uses the obligations' reference information, so it diagnoses where the copies sit rather than supplying a deployable repair. Propagating a lifecycle tag to matching lines under the same attribute changes $1$ of $452$ assembled views under the reference tags. Under FR-Bank's own tags the same rule changes $90$ views and fully tags $15$ of the $35$ duplicate views, but it also propagates falsely ($69$ of $2{,}333$ licensed active lines). But $187$ of $272$ audited stale lines sit under a different attribute key from their tagged twin, which leaves the write path. A reference-free repair for copies under other keys remains open. (Appendix~\ref{app:resp})

\section{Replication and the coverage judgment}\label{sec:populations}\label{sec:replication}\label{sec:calibration}\label{sec:coverage}\label{sec:discrimination}\label{sec:calibration-secondary}\label{sec:ladder}
The summary's obligation benefit replicates across vendors. On the shared questions every model whose question-only score varies gains $0.15$ to $0.45$ with it, Kimi included (Appendix~\ref{app:laddertable}).

On the full obligation set, no completed model reliably beats the lookup. Haiku, Luna and Opus score $0.82$ to $0.83$ against the lookup's $0.75$, but their margins fail the persona-grouped interval. On the easier coverage questions, mini's advantage over the superseded-name lookup disappears against a lookup over the whole name map. (Appendix~\ref{app:baselines})

\section{Limitations}\label{sec:limitations}\label{sec:discussion}
The controlled labels come from synthetic stores because complete lifecycle ground truth is not observable in real user data. All tests use one bank of $40$ synthetic personas; transfer to a second benchmark is untested. We change no backend's write path: the counterfactual test bypasses writing, every intervention is read-side, and no backend tested here is shown maintaining the summary, so the reach results describe what a store as shipped affords a reader; a write-path intervention is a separate study. Retraction and expiry are small groups ($30$ counterfactual pairs each; $26$ and $15$ obligations on the full set), so state-specific results give only a direction. Counts cannot say which record changed: two stores that cancel different plans in one category can share a summary, so only a named state identifies the obligation.

The matched metric and every comparison made with it are post hoc; pooled AUROC was fixed in advance. Sol did not run the names-and-counts ablation, and among the stronger readers only Opus ran the all-states summary. The paraphrase check covers $115$ of the $148$ questions, its rewrites are vaguer than the originals, and Opus ran it with one draw. On the $60$-name list Opus 5 ran one draw against its five-draw short-list row, and the two readers that keep their lead there are from one model family. The padded sibling control ran on Luna, Sol and mini; Haiku and Opus were not run on it. Luna's lead is not separated from the cue. The intent questions contain explicit cues; questions without them are untested.
The counterfactual set itself was rerun under the explicit-intent rubric for mini and Luna only; the crossover ran every reader under both rubrics. (Appendix~\ref{app:counterfactual}) The $22$ obligations the manifest-to-record alignment does not mark are invisible to every summary (mini scores $0.52$ on them under ground truth).

Control is outside this paper's scope. The answer stage in Appendix~\ref{app:resp} is exploratory: its token detector can flag mentions rather than use, and its human validation is specified in Appendix~\ref{app:registration}, with its workbooks built and no label collected at submission; nothing here shows that a better judgment changes an answer. (Appendix~\ref{app:resp})

\section{Conclusion}\label{sec:conclusion}
In these monitoring tests, no reader we tried made up for what the store had failed to write down. On a named state no model reliably exceeds the name lookup on the short list; the three stronger readers are equivalent to it within $0.05$, though not under the control that fixes list length. When the questions are reworded only one stronger reader edges past it. On a $60$-name list Luna and Sol lead the lookup by $0.13$ to $0.14$ and, with list length and counts fixed, Sol by $0.09$; the other three fall with it. The stronger readers' lead comes from counts the smaller models do not use. Told the intent rule, only the stronger readers apply it, and a fixed cue list is not reliably below the best of them.

Missing or wrong names weaken the summaries agents actually receive. At retrieval, the fetch keyed on the names reaches as far ungated as gated, at a budget trade-off.

\begin{center}
\fbox{\begin{minipage}{0.97\linewidth}
\textbf{For system builders.} Keep non-active records reachable with their lifecycle tags where deletion or access revocation is not required. Maintain names for superseded, retracted and expired content. Compare the name lookup with a model on the actual store: in our tests a model added something only on counted states, on explicit intent questions, on a $60$-name list and, pooled only, in two further conditions (Sections~\ref{sec:lookup} and~\ref{sec:backends}); whether the judgment pays for itself is unresolved (Appendix~\ref{app:utility-completion}).
\end{minipage}}\label{main-text-end}
\end{center}

\subsection*{AI use statement}

We used generative AI tools to draft questions and for implementation.
The persona bank, its sessions and the FR-Bank, Mem0 and A-MEM stores are a released synthetic benchmark \citep{fr}.
Language models drafted the question sets we added, and Graphiti's language-model pipeline built our graphs from the released FR-Bank store's records, rendered one episode per session without its lifecycle fields; the benchmark's human-written manifests are the ground truth against which every label is scored.
An AI assistant pre-read the paraphrase sheet and flagged rows with reasons; an author read every row and made every verdict.
For the intent crossover's questions an AI model pre-flagged candidate failures before an author's check; every verdict was an author's, entered from a dictated list.
Agentic coding assistants implemented the pipelines and executed the experimental protocols under human-directed designs with frozen decision rules, including the cross-vendor and question-only probes; independent implementations reproduced the computations and checked every printed number in both directions.

Language models also assisted with experiment-design feedback, hypothesis refinement, data cleaning, discussion of results, drafting, and \LaTeX{} preparation.

We have reviewed all AI-assisted work.
Human researchers made every design and reporting decision and reviewed every reported claim.
Every present-negative label in the main study was verified by hand from its constituent judgments, $248$ judgments yielding $104$ confirmed labels.
A separate pass, the one that built the balanced $36$-question set, adjudicated $33$ further candidates by hand: $12$ were confirmed, $11$ were finalized as rejections, and $10$ were routed for further review; all ten are excluded from every reported population.
The $12$ confirmed rows are the present-negative items of that $36$-question set, and none of the $33$ is among the $104$.

We take responsibility for the final content of this work, including text, claims, and artifacts produced with the aid of generative AI.

\subsection*{Reproducibility statement}

Code and data are released at \url{https://github.com/ansh348/knowing-the-store} (archive sha256 \texttt{685e398ceed7bda2\allowbreak 8a2fb6aad4fcde74\allowbreak 60d633397e15b114\allowbreak 198855dad189500e}); the supplement is an ancillary file of this arXiv record.

\appendix
\noindent\textbf{A reader's guide to the appendices.}
The appendices keep the vocabulary of the registered texts, which are quoted verbatim; in the main text, blind is \emph{question only}, indexed is \emph{with the summary}, route A is \emph{the summary route}, route B is \emph{the tag route}, NOK is \emph{the obligation judgment}, FOK is \emph{the coverage judgment}, the monitor is \emph{the reader}, and the structureless stores are \emph{Mem0 and A-MEM} (the rows of Appendix~\ref{app:glossary}'s terms table).
The first finding, that the two smaller models read only the summary's names and the three stronger ones also read its counts, rests on Table~\ref{tab:cf-main} (every reader on the counterfactual set, both metrics), Table~\ref{tab:cf-names-counts} with Table~\ref{tab:cf-nc-strong} (the summary with its counts or its names removed), Table~\ref{tab:v2rcat} (the moved-count control) and Table~\ref{tab:cf-matched-ablations} (the same under the matched metric).
The second, that on a named state no model reliably exceeds a name lookup and three models pull ahead only where the summary counts, rests on Table~\ref{tab:sup-margins} (each model's supersession-only margin), Table~\ref{tab:counterfactual} and Table~\ref{tab:cf-matched} (every margin over the lookup under both metrics), Table~\ref{tab:cf-sibling} (the sibling-name control), Table~\ref{tab:cf-hybrid} (the lookup with one count feature) and Table~\ref{tab:cf-intent} (the intent crossover).
The third, that the backends write the summary poorly, rests on Table~\ref{tab:census} (the census), Table~\ref{tab:levels} with Table~\ref{tab:levels-strata} (the store-written summary's levels, and by what the store retains) and Table~\ref{tab:graphiti-completed} with Table~\ref{tab:graphiti-strata} (the fourth backend).
The reach of a lifecycle fetch at retrieval is Tables~\ref{tab:reach} and~\ref{tab:gatereach}.
Every headline number's population, label source, tag source and evaluator are in Table~\ref{tab:provenance} (Appendix~\ref{app:provenance}).
The registration record, with each freeze's hash and its day in the registration order, is Appendix~\ref{app:registration}.

\section{Exact capability table}\label{app:laddertable}

This appendix carries the cross-model set in full; Table~\ref{tab:setup} defines every population the body names, with the models that ran on each, and Table~\ref{tab:runs} records which model ran on which question set.

\begin{table}[H]
\caption{The populations, with their sizes and positives and the models that ran on each (calls per question in parentheses). The body refers to a population by these names only.}
\label{tab:setup}
\centering
\scriptsize
\setlength{\tabcolsep}{2pt}
\renewcommand{\arraystretch}{0.9}
\begin{tabular}{@{}p{0.165\linewidth}p{0.505\linewidth}p{0.31\linewidth}@{}}
\toprule
Population & $n$ (positives) & Models (calls per question) \\
\midrule
the counterfactual set & $444$ items: $148$ questions asked of three versions of one store that differ in one attribute ($148$ obligations: $88$ supersession, $30$ retraction, $30$ expiry) & GPT-4.1 mini (1, and 5), Haiku 4.5 (5), Opus 5 (5), GPT-5.6 Sol (5), GPT-5.6 Luna (5) \\
the expiry extension & $42$ items: the $14$ eligible expiry pairs the first draw left undrawn, built by the same construction ($44$ expiry pairs pooled with the set's $30$) & GPT-4.1 mini (1), Haiku 4.5 (5), Opus 5 (1) \\
the paraphrase set & $345$ items: the $115$ of the $148$ questions whose rewrite, sharing no content word with any printed name, passed a word screen and an author's read, asked of the same three store versions ($115$ obligations: $76$ supersession, $17$ retraction, $22$ expiry) & GPT-4.1 mini (1), Haiku 4.5 (5), Opus 5 (1), GPT-5.6 Luna (5), GPT-5.6 Sol (5); the last two under the default summary only \\
the counterfactual set, lists padded to $60$ names & $444$ items: the same $148$ questions asked of the same three store versions with the superseded-names list padded to $60$ names (148 obligations: $88$ supersession, $30$ retraction, $30$ expiry) & GPT-4.1 mini (5), Haiku 4.5 (5), Opus 5 (1), GPT-5.6 Luna (5), GPT-5.6 Sol (5); the default summary only \\
the supersession pairs, a sibling padded to $60$ names & $176$ items: the $88$ supersession pairs' padded \pn{} version against a fourth version with one name swapped for a same-category sibling (identical pad and counts) & GPT-4.1 mini (5), GPT-5.6 Luna (5), GPT-5.6 Sol (5); Haiku 4.5 and Opus 5 not run \\
the full obligation set & $661$ questions ($104$ obligations: $63$ supersession, $26$ retraction, $15$ expiry): the hard tier's $307$ and the easy tier's $354$ together & GPT-4.1 mini (1); Haiku 4.5, Opus 5 and GPT-5.6 Luna (5); GPT-5.6 Sol and Kimi K3 stopped at the budget \\
the coverage set & $307$ questions ($35$ non-\absent{}): the hard tier, on which every coverage positive is an obligation & the same \\
the easy tier & $354$ questions ($243$ non-\absent{}, $69$ obligations): absent items made by removing a whole category from the store & the same \\
the cross-model set & $84$ questions ($54$ FOK-positive, $40$ obligations): the same questions asked of every model & all six; GPT-4.1 mini (1), the others (5) \\
the authorship-matched set & $36$ questions ($12$ obligations): every question rewritten in one voice & GPT-4.1 mini (3)$^{a}$, Haiku 4.5 (5), Opus 5 (5) \\
the downstream items & $174$ licensed items and $89$ scorable obligations: the items the answer stage can answer and the obligations its detector can score & GPT-4.1 mini and Haiku 4.5 as answerers (1) \\
\bottomrule
\end{tabular}
\par\smallskip
\begin{minipage}{0.97\linewidth}\scriptsize
$^{a}$GPT-4.1 mini ran three draws per question under each prompt on this set and one call per question on every other set; the score resolution of the one-call and five-call rows is in Appendix~\ref{app:counterfactual}.
\end{minipage}
\end{table}

\begin{table}[H]
\caption{Run coverage, model by question set, in the words of the run records. Two runs stopped at the budget on the full populations, so their complete rows are on the cross-model set; five models ran on the counterfactual set, and five ran its paraphrase set (Table~\ref{tab:setup}): GPT-4.1 mini one call, Haiku 4.5 five, Opus 5 one, GPT-5.6 Luna and GPT-5.6 Sol five through the raw interface under the default summary only; the same five ran it again with the superseded-names list padded to $60$ names, GPT-4.1 mini, Haiku 4.5, GPT-5.6 Luna and GPT-5.6 Sol at five calls and Opus 5 at one. $^{h}$: elicited through the vendor's agent harness rather than the raw API (Appendix~\ref{app:laddertable}).}
\label{tab:runs}
\centering
\tiny
\setlength{\tabcolsep}{2pt}
\begin{tabular}{@{}llllllp{0.24\linewidth}@{}}
\toprule
Model & Vendor & Counterfactual ($444$) & Full obligation ($661$) & Coverage ($307$) & Cross-model ($84$) & Scoring \\
\midrule
GPT-4.1 mini & OpenAI & ran & ran (the main study) & ran & ran & one call, temperature $0$; on the counterfactual set also five calls at the provider default (Table~\ref{tab:mini-k5}) \\
Haiku 4.5 & Anthropic & ran & ran & ran & ran & five calls, temperature $0$ \\
Opus 5 & Anthropic & ran & ran & ran & ran & five calls, provider default; one call on the expiry extension \\
GPT-5.6 Luna & OpenAI & ran & ran & ran & ran$^{h}$ & five calls, provider default; the cross-model set through the agent harness at reasoning effort low \\
GPT-5.6 Sol & OpenAI & ran & stopped at the budget & stopped at the budget & ran$^{h}$ & five calls; the cross-model set through the agent harness at reasoning effort low, the counterfactual set through the raw API \\
Kimi K3 & Moonshot & not run & stopped at the budget & stopped at the budget & ran & five calls \\
\bottomrule
\end{tabular}
\end{table}
\paragraph{Coverage notes for the main text.}
Each note below records a run setting that differs from its neighbours in the same sentence of the main text, or a comparison set that is incomplete.
(i) Section~\ref{sec:names_counts}, the count-removal result for GPT-4.1 mini: the names-and-counts ablation ran on its one-call row at temperature $0$ (Table~\ref{tab:cf-names-counts}); its five-draw row of Table~\ref{tab:cf-main} was not rerun under the ablation.
(ii) Section~\ref{sec:names_counts}, the count-removal result for the stronger models: GPT-5.6 Luna at five draws through the raw API and Opus 5 at one draw; GPT-5.6 Sol was not run, dropped by the registration's budget rule (Table~\ref{tab:cf-nc-strong}).
(iii) Section~\ref{sec:lookup}, the summary naming all states: GPT-4.1 mini, Haiku 4.5 and Opus 5 ran it; GPT-5.6 Luna and GPT-5.6 Sol ran the default summary only (Table~\ref{tab:cf-matched}).
(iv) Section~\ref{sec:intent}, the intent crossover: GPT-4.1 mini, Opus 5 and GPT-5.6 Sol ran one draw and Haiku 4.5 and GPT-5.6 Luna five, the body reading their draw-1 column; Sol was added after an interim read of three models' rows (Table~\ref{tab:cf-intent}).
(v) Section~\ref{sec:populations}, the full obligation set: Haiku 4.5, GPT-5.6 Luna and Opus 5 completed it; the GPT-5.6 Sol and Kimi K3 runs stopped at the budget and are not scored as sets (Table~\ref{tab:x4-expansion}).

GPT-4.1 mini's blind obligation score is flat on the full population, two values with $604$ of the $661$ questions at one of them, so its indexed-minus-blind difference is not a gain.
Its gain is defined only on the counterfactual set, where it is $+0.22$ and was predicted before the run (Section~\ref{sec:counterfactual}).
It gives six models on the same $84$ questions, the five models with a varying blind obligation score on the full population, and the five capability items registered before the cross-model run.
Across the five models whose blind obligation score carries information, the blind span of $0.3307$ contracts to $0.0389$ when indexed.
The indexed scores sit between $0.8199$ and $0.8588$ across three vendors, so the summary sets the level and the model sets little of it.
The Anthropic pair spans two model generations, so its ordering compares price and capability rather than a within-generation change.
GPT-4.1 mini's obligation row on the full population reads blind $0.4946$ and indexed $0.7339$.
That difference of $+0.2393$ has item-level interval $[+0.1801,+0.2949]$ and $[+0.1552,+0.3237]$ with questions grouped by persona.
The blind score takes two distinct values on the $661$ questions, with or without the rubric's detection step.
The difference is therefore not read as a gain over a blind judgment, and the main text compares the indexed level with the name lookup instead.
On coverage, model scale improves the blind guess, because blind levels include what the question text gives away.
GPT-5.6 Sol's blind FOK reaches $0.8131$ on the $150$-question stratified set and $0.7099$ on the cross-model set.
On the two largest models the coverage gain includes zero (GPT-5.6 Sol $+0.0364$, Kimi K3 $+0.0571$), while the obligation gain excludes zero on both.
GPT-4.1 mini's blind obligation score is constant at $0.0$ on all $84$ items, so its $0.5000$ is an all-tie artifact rather than calibrated uncertainty, and its clustered interval is not reported for the same reason.
Forty-eight distinct values are spanned by GPT-5.6 Luna, which still fails to discriminate.
The GPT-4.1 mini $\Delta$ interval entries carry a dash because those rows are the recorded main-study elicitation, which predates the study's paired-interval machinery, and the persona-clustered interval is shown where defined.

The GPT-5.6 Luna and GPT-5.6 Sol rows on the cross-model set and on the pooled $66$ structureless questions, marked $^{h}$ here and in Tables~\ref{tab:ladder-predictions}, \ref{tab:inversion} and~\ref{tab:runs}, were elicited in August through the vendor's command-line coding agent (\texttt{codex exec}, reasoning effort low, ephemeral sessions, subscription-billed), which prepends its own system text and tool schemas and passes the frozen system and user texts to the model as one flattened user turn.
No token telemetry survives those runs.
When the full-population runs tried to complete GPT-5.6 Sol through the same agent, two twelve-call probe gates returned no parseable reply, $0$ of $12$ and $0$ of $12$ (the deviation record of those runs), so the agent transport is not one condition with the raw API; the same models' rows on the full populations are raw API calls under the provider default.

\begin{table}[H]
\caption{The exact capability table records the cross-model set on FR-Bank with $n=84$; luna, sol, and the three new models use $k=5$, while mini is the original primary elicitation; $^{h}$ marks the rows elicited through the vendor's agent harness (the paragraph above). It shows tight indexed FOK across all six models and tight indexed NOK across the five models whose blind NOK varies. The clustered column reports the persona-clustered sensitivity uniformly for the whole set; the item-level interval remains the preregistered primary endpoint.}
\label{tab:ladder-exact}
\centering
\scriptsize
\setlength{\tabcolsep}{1.6pt}%
\begin{tabular}{@{}llrrrrrl@{}}
\toprule
Model & Judgment & $n$ & Positive & Blind & Indexed & $\Delta$ ($95\%$ CI) & Clustered CI \\
\midrule
mini & FOK & $84$ & $54$ & $0.4488$ & $0.7090$ & $+0.2602$ --- & $[+0.0913,+0.3912]$ \\
mini & NOK & $84$ & $40$ & $0.5000^{\dagger}$ & $0.7139$ & ---$^{\dagger}$ & ---$^{\dagger}$ \\
GPT-5.6 Luna$^{h}$ & FOK & $84$ & $54$ & $0.5735$ & $0.7417$ & $+0.1682$ $[+0.0340,+0.3033]$ & $[+0.0367,+0.3001]$ \\
GPT-5.6 Luna$^{h}$ & NOK & $84$ & $40$ & $0.5585$ & $0.8526$ & $+0.2940$ $[+0.1696,+0.4172]$ & $[+0.1984,+0.4051]$ \\
GPT-5.6 Sol$^{h}$ & FOK & $84$ & $54$ & $0.7099$ & $0.7463$ & $+0.0364$ $[-0.0930,+0.1600]$ & $[-0.1095,+0.1609]$ \\
GPT-5.6 Sol$^{h}$ & NOK & $84$ & $40$ & $0.5290$ & $0.8588$ & $+0.3298$ $[+0.1994,+0.4470]$ & $[+0.2117,+0.4500]$ \\
\addlinespace
Opus 5 & FOK & $84$ & $54$ & $0.5855$ & $0.7068$ & $+0.1213$ $[+0.0047,+0.2449]$ & $[-0.0136,+0.2368]$ \\
Opus 5 & NOK & $84$ & $40$ & $0.6994$ & $0.8523$ & $+0.1528$ $[+0.0659,+0.2509]$ & $[+0.0923,+0.2392]$ \\
Haiku 4.5 & FOK & $84$ & $54$ & $0.3136$ & $0.7287$ & $+0.4151$ $[+0.2336,+0.5712]$ & $[+0.1339,+0.6129]$ \\
Haiku 4.5 & NOK & $84$ & $40$ & $0.3688$ & $0.8199$ & $+0.4511$ $[+0.3265,+0.5741]$ & $[+0.2925,+0.5981]$ \\
K3 & FOK & $84$ & $54$ & $0.6651$ & $0.7222$ & $+0.0571$ $[-0.0954,+0.2052]$ & $[-0.1235,+0.2282]$ \\
K3 & NOK & $84$ & $40$ & $0.5983$ & $0.8588$ & $+0.2605$ $[+0.1456,+0.3866]$ & $[+0.1653,+0.3933]$ \\
\bottomrule
\multicolumn{8}{@{}p{0.96\linewidth}@{}}{\footnotesize $^{\dagger}$GPT-4.1 mini rows: the blind obligation score is an all-tie artifact, so the indexed-minus-blind difference on that row ($+0.2139$) is not a gain and is not shown; the dash convention is explained in the paragraph above the table.}\\
\end{tabular}
\end{table}

The $84$-item cross-model set spans $33$ of the $40$ personas.
Under persona clustering, nine of the ten rows carrying both intervals keep the verdict they had at item level.
The exception is Opus 5 coverage, whose item-level interval excludes zero by $+0.0047$ and whose clustered interval includes it ($[-0.0136,+0.2368]$).
Leave-one-persona-out shows zero sign changes on that row, so the point estimate is stable and the interval is what widens.
Every obligation row survives clustering, including all three models from the other two vendors.

A preregistered expansion re-elicited the blind and indexed conditions on the full FR-Bank populations for every run that completed under the budget stop (Table~\ref{tab:x4-expansion}).
Those populations are obligation $n=661$ with $104$ \pn{} and coverage $n=307$.
The $84$-item cross-model set remains the primary, because its predictions were fixed first.
GPT-4.1 mini's rows are the recorded main-study elicitation, re-derived exactly by the full-population scorer.

\begin{table}[H]
\caption{The full-population runs: blind and indexed FOK and NOK for every model that completed, with the paired gain and both intervals. The paired separation replicates on every completed run, and the spread of indexed NOK scores across models ($0.0988$) is again smaller than the spread of blind scores ($0.2279$). All five registered predictions held. The GPT-5.6 Sol and Kimi K3 runs were stopped by a budget directive before completion ($1{,}422$ and $231$ raw rows retained, disclosed, and not scored as sets). GPT-4.1 mini's blind obligation condition spans two distinct values on this population, so its $0.4946$ rests on a condition that is otherwise flat and its difference ($+0.2393$, intervals $[+0.1801,+0.2949]$ and $[+0.1552,+0.3237]$, recorded in the text above) is not shown as a gain; on the $84$-item cross-model set of Table~\ref{tab:ladder-exact} the same condition is constant and scores $0.5000$ exactly.}
\label{tab:x4-expansion}
\centering
\scriptsize
\setlength{\tabcolsep}{2.2pt}
\begin{tabular}{@{}llrrrrl@{}}
\toprule
Model & Judgment & $n$ & Blind & Indexed & $\Delta$ ($95\%$ CI) & Clustered CI \\
\midrule
mini & NOK & $661$ & $0.4946$ & $0.7339$ & --- & --- \\
mini & FOK & $307$ & $0.4966$ & $0.7003$ & $+0.2037$ $[+0.0935,+0.3191]$ & $[+0.1259,+0.2767]$ \\
GPT-5.6 Luna & NOK & $661$ & $0.4495$ & $0.8182$ & $+0.3687$ $[+0.2932,+0.4358]$ & $[+0.2656,+0.4666]$ \\
GPT-5.6 Luna & FOK & $307$ & $0.5720$ & $0.7605$ & $+0.1885$ $[+0.0439,+0.3352]$ & $[+0.0770,+0.3071]$ \\
Opus 5 & NOK & $661$ & $0.6429$ & $0.8327$ & $+0.1898$ $[+0.1437,+0.2411]$ & $[+0.1241,+0.2523]$ \\
Opus 5 & FOK & $307$ & $0.6793$ & $0.8099$ & $+0.1306$ $[+0.0557,+0.2148]$ & $[+0.0523,+0.2026]$ \\
Haiku 4.5 & NOK & $661$ & $0.4150$ & $0.8210$ & $+0.4060$ $[+0.3283,+0.4791]$ & $[+0.3063,+0.5026]$ \\
Haiku 4.5 & FOK & $307$ & $0.3781$ & $0.7465$ & $+0.3684$ $[+0.2629,+0.4755]$ & $[+0.2700,+0.4747]$ \\
\bottomrule
\end{tabular}
\end{table}

Table~\ref{tab:e1}, from Section~\ref{sec:populations}, gives the two-decimal reading of the obligation rows beside the name lookup's level on the same questions, and the coverage judgment by backend for GPT-4.1 mini.

\begin{table}[H]
\caption{Blind and indexed judgments with the paired gain and both intervals: coverage by backend for GPT-4.1 mini (positives non-\absent{}) and the obligation judgment on FR-Bank by model (positives \pn{}). On the hard tier every coverage positive is an obligation, so that FOK row scores the same positives as the NOK row.}
\label{tab:e1}
\centering
\scriptsize
\setlength{\tabcolsep}{2pt}
\renewcommand{\arraystretch}{0.9}%
\begin{tabular}{@{}lrrrrll@{}}
\toprule
\multicolumn{7}{@{}l}{\textbf{FOK by backend, GPT-4.1 mini}}\\
Backend & $n$ / pos. & Blind & \hd{Indexed}{(ground-truth; constrained on Mem0, A-MEM)} & $\Delta$ & item interval & persona interval \\
\midrule
FR-Bank hard tier & $307$ / $35$ & $0.50$ & $0.70$ & $+0.20$ & $[+0.09,+0.32]$ & $[+0.13,+0.28]$ \\
Mem0 & $238$ / $94$ & $0.57$ & $0.48$ & $-0.10$ & $[-0.17,-0.03]$ & $[-0.16,-0.03]$ \\
A-MEM & $284$ / $126$ & $0.56$ & $0.45$ & $-0.11$ & $[-0.17,-0.04]$ & $[-0.17,-0.04]$ \\
\midrule
\multicolumn{7}{@{}l}{\textbf{NOK by model, FR-Bank}; the name lookup scores $0.75$ (rounded from $0.7484$) on the same $661$ questions}\\
Model & $n$ / pos. & Blind & \hd{Indexed}{(ground-truth summary)} & $\Delta$ & item interval & persona interval \\
\midrule
Haiku 4.5 & $661$ / $104$ & $0.42$ & $0.82$ & $+0.41$ & $[+0.33,+0.48]$ & $[+0.31,+0.50]$ \\
Opus 5 & $661$ / $104$ & $0.64$ & $0.83$ & $+0.19$ & $[+0.14,+0.24]$ & $[+0.12,+0.25]$ \\
GPT-5.6 Luna & $661$ / $104$ & $0.45$ & $0.82$ & $+0.37$ & $[+0.29,+0.44]$ & $[+0.27,+0.47]$ \\
\bottomrule
\end{tabular}
\end{table}

\begin{table}[h]
\caption{Five capability items were registered before the cross-model run. The first four held, while the sol threshold failed, revising the earlier one-call account. The first, third, and fourth items are scored on the $84$-item cross-model set; the second and fifth on the original per-model populations, the shared $150$-item stratified set, with mini's value from its recorded one-call main-study elicitation and luna's and sol's from the five-call average.}
\label{tab:ladder-predictions}
\centering
\scriptsize
\begin{tabular}{@{}p{0.26\linewidth}p{0.38\linewidth}p{0.19\linewidth}p{0.07\linewidth}@{}}
\toprule
Registered item & Observation & Population & Outcome \\
\midrule
P1: luna blind NOK includes chance & $0.5585$, CI $[0.4299,0.6869]$ includes $0.50$ & cross-model set$^{h}$, $n=84$ & held \\
P2: blind FOK increases with scale & $0.5267<0.6745<0.8131$ & stratified set$^{h}$, $n=150$ per model & held \\
P3: indexed FOK gain decreases with scale & $+0.0364<+0.1682<+0.2602$ & cross-model set$^{h}$, $n=84$ & held \\
P4: luna has a positive NOK gain & $+0.2940$, CI $[+0.1696,+0.4172]$ & cross-model set$^{h}$, $n=84$ & held \\
P5: Sol blind FOK remains below $0.75$ & $0.8131>0.75$ & stratified set$^{h}$, $n=150$ & failed \\
\bottomrule
\multicolumn{4}{@{}p{0.96\linewidth}@{}}{\footnotesize $^{h}$Rows elicited through the vendor's agent harness (Appendix~\ref{app:laddertable}).}\\
\end{tabular}
\end{table}

\section{Cross-vendor replication record}\label{app:crossvendor}

This appendix carries the five items registered for the three models from vendors other than OpenAI, and one follow-up on the authorship-matched set.
The item-level interval is the registered primary endpoint and the persona-clustered interval is the sensitivity, because that is the convention everywhere in this paper.

\noindent\textbf{Reading the Outcome column.} \textit{held}: every prediction the registration carried was borne out. \textit{failed}: it was not, and the registration's frozen failure text is reported as written. \textit{held / failed}: a registration carrying several predictions, split as the middle column describes. \textit{no prediction}: a method freeze, or a set of branches enumerated in advance with no pass bar, so nothing could hold or fail. \textit{held / undefined}: one prediction held and one named a quantity its population does not define, which the registration had anticipated.\\

\begin{table}[H]
\caption{Five items were registered for the three new models. The indexed separation and FOK band held, while the universal blind-NOK baseline failed in opposite directions.}
\label{tab:crossvendor-predictions}
\centering
\scriptsize
\begin{tabular}{@{}p{0.15\linewidth}p{0.62\linewidth}p{0.12\linewidth}@{}}
\toprule
Item & Registered reading and observation & Outcome \\
\midrule
XV-1 & Registered: every new blind-NOK interval includes chance. Observed: Opus 5 is above, Haiku 4.5 below, and K3 includes chance & failed \\
XV-2 & Positive indexed-minus-blind NOK delta with interval excluding zero on Opus 5 and K3 & held \\
XV-3 & Opus 5 blind FOK exceeds Haiku 4.5 blind FOK; cross-generation price/capability ordering & held \\
XV-4 & Indexed FOK lies within the registered OpenAI-centered band on all three new models & held \\
XV-5 & K3 blind NOK includes chance under its always-reasoning configuration & reports \\
\bottomrule
\end{tabular}
\end{table}

\noindent\textbf{Per-provider disclosures.} The seven run disclosures for the three providers are in the supplement (Section~S.4), with the correction of one frozen failure explanation whose gate and verdict are unchanged. They are there because they bear on reproduction rather than on any result.

\noindent\textbf{Haiku 4.5 on authorship-matched questions.}
A registered follow-up asked whether Haiku 4.5's below-chance blind obligation score survives when the questions are rewritten in one voice for every label, because question authorship could carry a direction the model does not.
On the authorship-matched cell the Haiku 4.5 blind NOK point stays below chance while its interval includes chance, so the matched cell is indeterminate at $n=36$ on whether the inverted prior of XV-1 reflects a property of the model or of question authorship; XV-1's registered verdict is unchanged.
The point falls $0.0073$ below the registered decision boundary of $0.45$, and the values appear in Table~\ref{tab:matched-cell}.

\section{Question-signal audit}\label{app:questionprior}

This appendix records that the question text alone predicts the labels.
That is why no absolute blind level in this paper is a finding, and why the paired gains and the counterfactual set carry the evidence.
Supervised text probes separate \pn{} from non-\pn{} far above chance on every question set, while a zero-shot GPT-4.1 mini reader does not.
The larger $162$-question comparison stays separable (Table~\ref{tab:questionprior}).
The records we deleted were removed by the store's own operation, yet the questions written about them still read differently from the rest.
The structure runs deeper than authorship, because ranking by the recipe used to build each item alone scores $0.9736$ on the cross-model set, above the probe's $0.7972$ (Table~\ref{tab:av-census}).
Six of the eight recipes carry a single class, so rewording cannot make blind levels clean.
Separating construction from authorship needs comparison within a recipe, and only one is large enough ($n=16$).
The same holds for coverage: on hard questions the FOK-positive class coincides with \pn{}, and the coverage-targeted probe scores $0.9662$ pooled, so blind coverage levels carry the same question signal.
Two quantities are free of this signal, because neither depends on the wording of a question.
The paired gains compare identical questions and still measure the summary's added information, and the counterfactual set fixes the question and varies the store (Table~\ref{tab:counterfactual}).
The balanced authorship-matched set ($36$) holds every slot at the same one-third \pn{} rate, and blind NOK is at chance on it for GPT-4.1 mini and Opus 5.

\begin{table}[h]
\caption{Question-only probes measure how much PN label signal is available without a store. Label-supervised term-frequency--inverse-document-frequency (TF--IDF) features expose generator-specific template signal rather than monitor cue use, whereas the zero-shot mini probe on the cross-model set remains at chance.}
\label{tab:questionprior}
\centering
\scriptsize
\setlength{\tabcolsep}{3pt}
\begin{tabular}{@{}p{0.34\linewidth}rrrp{0.22\linewidth}@{}}
\toprule
Probe and population & $n$ & PN & AUROC ($95\%$ CI) & Reading \\
\midrule
Supervised TF--IDF, cross-model set & $84$ & $40$ & $0.7972$ $[0.6949,0.8927]$ & text carries label signal \\
Supervised TF--IDF, hard row (FOK target $\equiv$ PN)$^{\ddagger}$ & $307$ & $35$ & $0.9199$ $[0.8869,0.9472]$ & text carries label signal \\
Zero-shot mini, cross-model set & $84$ & $40$ & $0.4804$ $[0.3587,0.5935]$ & includes chance \\
Supervised TF--IDF, powered matched set & $162$ & $35$ & $0.8436$ $[0.7813,0.8990]$ & not controlled for the question signal \\
Supervised TF--IDF, FOK target, pooled hard & $588$ & $35$ & $0.9662$ $[0.9510,0.9788]$ & text carries coverage-label signal \\
\bottomrule
\multicolumn{5}{@{}p{0.96\linewidth}@{}}{\footnotesize $^{\ddagger}$The hard tier contains no \present{} items, so the FOK target (non-\absent{}) and the PN target are label-identical on this row; the registered coverage-target probe re-derives the same point estimate, and the recorded PN-probe interval is the one printed.}\\
\end{tabular}
\end{table}

\subsection{Authorship-matched set}\label{app:matched}

The authorship-matched set rewrites every question in one generator voice from each item's entity and slot, so no label is distinguishable by who wrote its question.
It also balances the set so that each slot token carries the same one-third \pn{} rate (Table~\ref{tab:matched-cell}).
Coefficient inspection of the supervised probe first suggested an authorship mismatch, with \pn{} questions bespoke and many negatives reusing generator templates.
The census of Table~\ref{tab:av-census} shows the structure runs deeper than authorship, because six of its eight recipes carry a single class.

\begin{table}[h]
\caption{The authorship-matched set ($n=36$; $12$ \pn{}, $29$ non-\absent{}; one generator voice for every label; label transfer $36/36$). Blind NOK includes chance on all three models. The paired NOK gain is positive on all three and excludes zero on Opus 5 only, the resolution an $n=36$ set affords. FOK rows are descriptive; their blind intervals are not reported. The probe block is the recorded supervised probe of Table~\ref{tab:questionprior}, run unchanged on these same $36$ questions.}
\label{tab:matched-cell}
\centering
\scriptsize
\setlength{\tabcolsep}{3pt}
\begin{tabular}{@{}llllr@{}}
\toprule
Model & Judgment & Blind ($95\%$ CI) & Indexed & $\Delta$ ($95\%$ CI) \\
\midrule
\multicolumn{5}{@{}l}{\textit{Monitors}}\\
mini & NOK & $0.4774$ $[0.2991,0.6623]$ & $0.6771$ & $+0.2028$ $[-0.1121,+0.4984]$ \\
Opus 5 & NOK & $0.3993$ $[0.2008,0.6232]$ & $0.6962$ & $+0.2978$ $[+0.0752,+0.5292]$ \\
Haiku 4.5 & NOK & $0.4427$ $[0.2582,0.6522]$ & $0.6562$ & $+0.2208$ $[-0.0522,+0.4809]$ \\
\addlinespace
mini & FOK & $0.5788$ & $0.8867$ & $+0.3110$ $[+0.1133,+0.5222]$ \\
Opus 5 & FOK & $0.6675$ & $1.0000$ & $+0.3372$ $[+0.1399,+0.5709]$ \\
Haiku 4.5 & FOK & $0.5419$ & $0.9901$ & $+0.4470$ $[+0.2148,+0.6723]$ \\
\addlinespace
\multicolumn{5}{@{}l}{\textit{Question-only probe on the same $36$ questions}}\\
Supervised TF--IDF & \pn{} & $0.3472$ $[0.1673,0.5424]$ & --- & --- \\
\bottomrule
\multicolumn{5}{@{}p{0.96\linewidth}@{}}{\footnotesize The probe interval is wide at $n=36$ with $12$ positives, and the row is a construction check on this set rather than a measurement of the FR-Bank populations; it is not comparable with the Table~\ref{tab:questionprior} rows, which sit on different populations. The zero-shot reader was not re-run here, because no cached responses exist for these newly rendered questions.}\\
\end{tabular}
\end{table}

This set is not a store counterfactual, and what it holds fixed is stated here because the two designs are easy to confuse.
Within a slot the question is near-identical across labels, differing only in the entity named, and the slot's \pn{} rate is one third by construction.
That is what makes it an authorship and blind-score control.
Each item in a slot belongs to a different persona, so the label co-varies with the entire store rather than with the target's lifecycle state alone.
A counterfactual in the strict sense scores one query against versions of the same store differing only in the target's lifecycle state, which is the design of Appendix~\ref{app:counterfactual}.

\begin{table}[h]
\caption{Label structure by item-construction category on the cross-model set ($n=84$, $40$ \pn{}). Ranking by category alone attains AUROC $0.9736$, above the supervised probe's $0.7972$ on the same set. This is a deterministic census on recorded labels (no model is fit and nothing is resampled), so no interval is defined, the same reporting convention as the one-bit is-FR-Bank rule in Appendix~\ref{app:baselines}. Six of the eight categories carry a single class, and only expired logistics carries both at usable size.}
\label{tab:av-census}
\centering
\scriptsize
\setlength{\tabcolsep}{3pt}
\begin{tabular}{@{}lrrrl@{}}
\toprule
Item-construction category & $n$ & \pn{} & $P(\pn{})$ & Classes \\
\midrule
Stable identity & $21$ & $0$ & $0.000$ & single \\
Expired logistics & $16$ & $8$ & $0.500$ & both \\
Superseded preference & $14$ & $13$ & $0.929$ & both ($1$ negative) \\
Selective forgetting & $13$ & $13$ & $1.000$ & single \\
Natural absence & $8$ & $0$ & $0.000$ & single \\
Multi-version fact & $6$ & $6$ & $1.000$ & single \\
Soft supersession & $4$ & $0$ & $0.000$ & single \\
Cross-session contradiction & $2$ & $0$ & $0.000$ & single \\
\bottomrule
\end{tabular}
\end{table}

\subsection{Store-counterfactual set}\label{app:counterfactual}

Opus 5's first draw alone keeps its margin over the name lookup on the whole set.
Scored post hoc against the PRESENT versions alone, which also hold relevant content, every reader moves by at most $0.03$ from its whole-set value, so the comparison stands where coverage is equal.

\begin{table}[H]
\caption{The supersession triple of the counterfactual set that Figure~\ref{fig:example} leaves out: the question is fixed, the three store versions differ in one attribute, and only the summary lines that differ are shown, with the name lookup's score under each. The obligation is named, and the lookup separates the versions.}
\label{tab:sup-triple}
\centering
\footnotesize
\setlength{\tabcolsep}{5pt}
\begin{tabular}{@{}>{\raggedright\arraybackslash}p{0.27\linewidth}>{\raggedright\arraybackslash}p{0.21\linewidth}>{\raggedright\arraybackslash}p{0.23\linewidth}>{\raggedright\arraybackslash}p{0.21\linewidth}@{}}
\multicolumn{4}{@{}l}{Question: \textit{What is my current work style or preference?}}\\
\toprule
summary line & version 1: \present{} & version 2: \pn{} & version 3: \absent{}\\
\midrule
lifecycle total, superseded & \texttt{0} & \texttt{1} & \texttt{0}\\
\texttt{PREFERENCES\_HABITS}: entries, active, superseded & \texttt{46, 46, 0} & \texttt{47, 46, 1} & \texttt{45, 45, 0}\\
names carrying superseded content & \texttt{(none)} & \texttt{work\_preference} & \texttt{(none)}\\
the name lookup's score & $0.00$ & $0.73$ & $0.00$\\
the label & a current value, licensed & the old value must not be used & no such record\\
\bottomrule
\end{tabular}
\end{table}

This appendix carries the counterfactual set in full.
It gives the whole-set and per-state rows of every reader under the three summaries, the set's earlier thirty-question draw (labelled first form in Table~\ref{tab:counterfactual}), the adverse rows, the coverage rows, the lexical lookup, and the expiry pairs the main text refers to.
The set asks $148$ questions, each of three versions of one store that differ in a single attribute, so $n=444$ with $148$ positives: $88$ supersession, $30$ retraction and $30$ expiry pairs.
The pairs were drawn under predicates fixed before any verdict, and every question was written before any version of the store existed.
A question-only reader scores $0.5000$ on this set as an identity, because the same question is asked of all three versions.
Every gain below is measured against it, with the item-level paired interval first and the interval with questions grouped by persona beside it; where the two disagree the wider one decides.

Under the summary that names only superseded content, Haiku 4.5's judgment on the named state exceeds its judgment on the two counted states by $+0.3237$, interval $[+0.2319,+0.4011]$.
Grouped by persona that interval is $[+0.2612,+0.3851]$.
Opus 5 gives $+0.3295$ with intervals $[+0.2420,+0.4181]$ and $[+0.2556,+0.3875]$, and GPT-4.1 mini's $+0.2948$ (interval $[+0.1875,+0.3905]$) is descriptive.
On this cell the superseded-names lookup reaches $0.7663$ against the smallest reader's $0.7167$ under the superseded-only summary, and the all-states lookup reaches $0.8756$ against Haiku 4.5's $0.8773$ under the all-three-named summary, within tolerance.
Under the summary naming all three states the named-minus-counted difference on Haiku 4.5 falls to $+0.0203$ (interval $[-0.0570,+0.0907]$), so the difference belongs to what the summary names.
The totals-randomized summary permutes the retracted and expired lifecycle totals across the $444$ store versions under a recorded seed, so that no version keeps its own pair and every value keeps its multiplicity; the per-category counts are not touched.
Elsewhere it is byte-identical to the superseded-only summary.
It moves GPT-4.1 mini's whole-set judgment by $+0.0011$, Haiku 4.5's by $+0.0072$ and Opus 5's by $-0.0052$, every interval inside the tolerance band, so no model takes anything from the totals.

\paragraph{Count readers on this set.}
Two readers with no generative-model call read the count table alone, so that what a count carries can be measured beside what the models read from it.
The category-aware count reader knows the question's category from the manifest and reads that category's retracted-plus-expired count, a count.
The category-guessing reader assigns the category from the question by the recorded rule of the question-only baseline suite and reads a different quantity at that category, the share of its entries that are superseded, retracted or expired.
A state-aware reader, which knows the state as well as the category, reads the state's own cell.
Under the superseded-only summary the category-aware reader separates the retracted versions at $0.7306$ (item interval $[0.6194,0.8228]$, persona $[0.6900,0.8090]$) and the expired versions at $0.7778$ ($[0.6793,0.8625]$, $[0.7153,0.8680]$), the state-aware reader at $0.8944$ and $0.7778$, and the models on the same items at $0.5283$ to $0.6150$ (Table~\ref{tab:cf-cr}).
The count identifies the version because $26$ of the $30$ retraction pairs and $15$ of the $30$ expiry pairs have a category base count of zero, so the present-negative version's increment reads against zero within its triple.
Within a supersession triple the count table differs only in the superseded column of the target's category, which the category-aware reader does not read, so it ties all $88$ supersession triples; the category-guessing reader's share includes that column, so it orders the $34$ triples whose assigned category is the target's and ties the other $54$.
Read at the manifest category, that share scores $0.8473$ on the whole set ($0.8981$ on supersession, $0.7336$ on retraction, $0.8528$ on expiry): a reader given the question's category, computed after the fact, above GPT-4.1 mini, Haiku 4.5 and Opus 5 on the whole set (Table~\ref{tab:cf-cr}).
A reader of the two lifecycle totals, the lines the totals-randomized summary permutes, reads retraction at $0.6347$ and expiry at $0.6922$ under the superseded-only summary and $0.4044$ and $0.3894$ under the recorded permutation, at chance in expectation over $200$ permutation seeds ($0.4900$ and $0.5065$); the per-category counts are untouched by that permutation, so the category-aware reader is identical under both.

\begin{table}[H]
\caption{The count readers on the counterfactual set beside the models, under the superseded-only summary: AUROC on the whole set and per state, the persona interval of the whole-set AUROC, and each model minus the category-aware reader on the two counted states. The category-aware reader reads the retracted-plus-expired count at the question's category, the state-aware reader the state's own cell at that category, the totals reader the two lifecycle totals, and the category-guessing reader the share of superseded, retracted and expired entries at the category the question-only suite's rule assigns. The last reader row reads that share at the manifest category: privileged, since the category is given, and added after the fact. The category-aware reader separates the versions on the counted states; every model is below it on both. Descriptive; computed after the fact on the recorded texts and rows.}
\label{tab:cf-cr}
\centering
\tiny
\setlength{\tabcolsep}{2pt}
\begin{tabular}{@{}lrrrrl@{}}
\toprule
Reader & whole & supersession & retraction & expiry & persona interval (whole) \\
\midrule
Category-aware reader & $0.6043$ & $0.5000$ & $0.7306$ & $0.7778$ & $[0.5753,0.6389]$ \\
State-aware reader & $0.8580$ & $0.9006$ & $0.8944$ & $0.7778$ & $[0.8150,0.8998]$ \\
Totals reader & $0.5603$ & $0.5028$ & $0.6347$ & $0.6922$ & $[0.5452,0.5882]$ \\
Category-guessing reader & $0.6623$ & $0.6573$ & $0.6669$ & $0.6872$ & $[0.6273,0.6991]$ \\
\addlinespace
\multicolumn{6}{@{}l}{\textit{Post hoc and privileged: the category-guessing reader's share read at the manifest category}}\\
Category-given share reader & $0.8473$ & $0.8981$ & $0.7336$ & $0.8528$ & $[0.8071,0.8873]$ \\
\addlinespace
\multicolumn{6}{@{}l}{\textit{Model minus the category-aware reader: whole set (difference; item interval; persona interval), and on the counted states (model level; difference)}}\\
GPT-4.1 mini & $+0.1124$ & & $0.5283$; $-0.2022$ & $0.5489$; $-0.2289$ & $[+0.0272,+0.1944]$; $[+0.0400,+0.1756]$ \\
Haiku 4.5 & $+0.1523$ & & $0.5433$; $-0.1872$ & $0.5794$; $-0.1983$ & $[+0.0706,+0.2319]$; $[+0.0843,+0.2168]$ \\
Opus 5 & $+0.2362$ & & $0.5872$; $-0.1433$ & $0.6150$; $-0.1628$ & $[+0.1663,+0.3114]$; $[+0.1745,+0.2938]$ \\
\bottomrule
\end{tabular}
\end{table}

\paragraph{The moved-count control.}
A second count check starts from the totals-randomized text and, for each of the $60$ counted-state pairs, moves the obligation's own increment in the category row's retracted or expired column from the present-negative version to one non-present-negative sibling under a recorded seed, so that the version carrying the obligation reads the base count and a sibling reads the increment; every other line is unchanged, every name section equals the superseded-only summary's, the renderer lock holds on all $120$ changed texts, and the leakage screen returns $0$ of $30$.
A reader that follows the moved cell falls below chance on that state, the state-aware reader to $0.3028$ on retraction and $0.3611$ on expiry (Table~\ref{tab:v2rcat}).
The registered reading asked whether a model stays within the $0.05$ band of its superseded-only row on both counted states or moves beyond it reliably, and fired as neither for both models, so each row prints as it came out.
Read after the fact against the direction a count reader falls, and labelled so: GPT-4.1 mini moves by $+0.0269$ on retraction and $+0.0611$ on expiry, upward, with both intervals including zero, and Haiku 4.5 by $-0.0072$ and $-0.0453$, so neither registered run established that it followed the count. Opus 5, added as an optional reader, was run once per item ($k=1$, the registered default) against its recorded five-draw superseded-only row. It moves by $-0.1011$ on retraction and $-0.1236$ on expiry, downward, the direction a count reader falls, with each interval excluding zero but reaching the band. The unchanged supersession texts move by $-0.0022$ under the same single draw, which carries draw-count noise against the five-draw row that the two registered runs did not.
GPT-5.6 Luna and GPT-5.6 Sol, run at five draws on the same texts through the raw API against their recorded five-draw rows under the second amendment, fall reliably on expiry, by $-0.1158$ and $-0.1914$, and on retraction by $-0.0664$ and $-0.0322$ with both intervals including zero.
The amendment's one prediction, that each falls reliably on both counted states, failed for both models.

\begin{table}[H]
\caption{The moved-count control: each model under the moved text against its recorded superseded-only row, whole set and per state, with the paired difference and both intervals, beside the reference readers on the same text. Descriptive; no prediction was registered on the first three readers' rows. Opus 5's five-draw row was registered before its four added draws (Amendment 1), which predicted that the one-draw movement on retraction and expiry would replicate at five draws with both intervals excluding zero; it did. GPT-5.6 Luna and GPT-5.6 Sol ran at five draws under the second amendment, whose one prediction, that each falls reliably on both counted states, failed on retraction for both.}
\label{tab:v2rcat}
\centering
\tiny
\setlength{\tabcolsep}{2pt}
\begin{tabular}{@{}llrrl@{}}
\toprule
Reader & row & sup.\ only & moved-count control & difference (item; persona) \\
\midrule
GPT-4.1 mini ($k=1$) & whole set & $0.7167$ & $0.7410$ & $+0.0242$ $[-0.0047,+0.0523]$; $[-0.0032,+0.0530]$ \\
GPT-4.1 mini & retraction ($30$ pairs) & $0.5283$ & $0.5553$ & $+0.0269$ $[-0.0491,+0.1080]$; $[-0.0692,+0.1331]$ \\
GPT-4.1 mini & expiry ($30$ pairs) & $0.5489$ & $0.6100$ & $+0.0611$ $[-0.0404,+0.1641]$; $[-0.0217,+0.1417]$ \\
Haiku 4.5 ($k=5$) & whole set & $0.7566$ & $0.7525$ & $-0.0041$ $[-0.0220,+0.0133]$; $[-0.0233,+0.0146]$ \\
Haiku 4.5 & retraction ($30$ pairs) & $0.5433$ & $0.5361$ & $-0.0072$ $[-0.0925,+0.0781]$; $[-0.0981,+0.0795]$ \\
Haiku 4.5 & expiry ($30$ pairs) & $0.5794$ & $0.5342$ & $-0.0453$ $[-0.0958,+0.0014]$; $[-0.0781,-0.0169]$ \\
Opus 5 ($k=5$) & whole set & $0.8405$ & $0.8075$ & $-0.0330$ $[-0.0496,-0.0196]$; $[-0.0592,-0.0142]$ \\
Opus 5 & retraction ($30$ pairs) & $0.5872$ & $0.4869$ & $-0.1003$ $[-0.1612,-0.0472]$; $[-0.2079,-0.0317]$ \\
Opus 5 & expiry ($30$ pairs) & $0.6150$ & $0.4869$ & $-0.1281$ $[-0.2127,-0.0517]$; $[-0.2651,-0.0095]$ \\
Opus 5, draw 1 ($k=1$)$^{d}$ & whole set & $0.8405$ & $0.8049$ & $-0.0357$ $[-0.0543,-0.0206]$; $[-0.0617,-0.0169]$ \\
Opus 5, draw 1 & retraction ($30$ pairs) & $0.5872$ & $0.4861$ & $-0.1011$ $[-0.1700,-0.0435]$; $[-0.2145,-0.0211]$ \\
Opus 5, draw 1 & expiry ($30$ pairs) & $0.6150$ & $0.4914$ & $-0.1236$ $[-0.2165,-0.0374]$; $[-0.2598,-0.0118]$ \\
GPT-5.6 Luna ($k=5$) & whole set & $0.8417$ & $0.8121$ & $-0.0297$ $[-0.0499,-0.0089]$; $[-0.0583,-0.0081]$ \\
GPT-5.6 Luna & retraction ($30$ pairs) & $0.5458$ & $0.4794$ & $-0.0664$ $[-0.1505,+0.0156]$; $[-0.1903,+0.0294]$ \\
GPT-5.6 Luna & expiry ($30$ pairs) & $0.6167$ & $0.5008$ & $-0.1158$ $[-0.2162,-0.0210]$; $[-0.2551,-0.0121]$ \\
GPT-5.6 Sol ($k=5$) & whole set & $0.8500$ & $0.8172$ & $-0.0327$ $[-0.0521,-0.0151]$; $[-0.0621,-0.0102]$ \\
GPT-5.6 Sol & retraction ($30$ pairs) & $0.5322$ & $0.5000$ & $-0.0322$ $[-0.1121,+0.0440]$; $[-0.1228,+0.0636]$ \\
GPT-5.6 Sol & expiry ($30$ pairs) & $0.6403$ & $0.4489$ & $-0.1914$ $[-0.2862,-0.1074]$; $[-0.3232,-0.0816]$ \\
\addlinespace
State-aware count reader & retraction; expiry & $0.8944$; $0.7778$ & $0.3028$; $0.3611$ & \\
Totals reader & retraction; expiry & $0.6347$; $0.6922$ & $0.4044$; $0.3894$ & \\
Name lookup & whole set & $0.7663$ & $0.7663$ & the name sections are unchanged \\
\bottomrule
\end{tabular}
\par\smallskip\begin{minipage}{0.97\linewidth}\tiny
$^{d}$The draw-1 rows are the registered comparison of the single-draw moved-count run with the recorded five-draw row ($0.8405$); against Opus 5's first draw alone ($0.8400$, Appendix~\ref{app:counterfactual}) the whole-set difference would read $-0.0351$, a comparison the registration did not make. Amendment 1's $k=5$ rows above remove the mismatch.
\end{minipage}
\end{table}

\begin{table}[H]
\caption{Names against counts on the counterfactual set (E2; registered, ADDENDUM\_CF2\_NAMES\_COUNTS). The superseded-only summary with every count line removed (names only), with the two name blocks removed (counts only), and, for GPT-4.1 mini, with every count replaced by a dash (the committed names-only rule of the pooled hard tier). The recorded intact row is the comparator, GPT-4.1 mini's at one draw (the five-draw row of Table~\ref{tab:cf-main} is not used here); differences carry the item-level and the persona-grouped interval. Both readers stay within tolerance of the intact summary without the counts and fall reliably without the names (P-N1 and P-N2 held).}
\label{tab:cf-names-counts}
\centering\tiny\setlength{\tabcolsep}{1.5pt}
\begin{tabular}{@{}llrlrrrrr@{}}
\toprule
Reader & summary & AUROC & persona interval & minus intact & item interval & persona interval & retraction & expiry \\
\midrule
Haiku 4.5 & intact (recorded) & $0.7566$ & & & & & & \\
 & counts only & $0.5945$ & $[0.5514,0.6460]$ & $-0.1621$ & $[-0.2187,-0.1024]$ & $[-0.1941,-0.1278]$ & $0.5553$ & $0.5189$ \\
 & names only & $0.7507$ & $[0.7114,0.7891]$ & $-0.0059$ & $[-0.0370,+0.0237]$ & $[-0.0365,+0.0216]$ & $0.5631$ & $0.5642$ \\
 & counts only minus names only & & & $-0.1562$ & $[-0.2175,-0.0889]$ & $[-0.2007,-0.1126]$ & & \\
\addlinespace[1pt]
GPT-4.1 mini & intact (recorded) & $0.7167$ & & & & & & \\
 & counts only & $0.5862$ & $[0.5416,0.6322]$ & $-0.1306$ & $[-0.2025,-0.0524]$ & $[-0.1835,-0.0801]$ & $0.6053$ & $0.4933$ \\
 & names only & $0.7103$ & $[0.6644,0.7524]$ & $-0.0064$ & $[-0.0484,+0.0338]$ & $[-0.0439,+0.0290]$ & $0.5344$ & $0.5006$ \\
 & names only, dash rule & $0.6891$ & $[0.6371,0.7356]$ & $-0.0276$ & $[-0.0679,+0.0086]$ & $[-0.0694,+0.0120]$ & $0.4500$ & $0.4764$ \\
 & counts only minus names only & & & $-0.1242$ & $[-0.1981,-0.0525]$ & $[-0.1814,-0.0698]$ & & \\
\addlinespace[1pt]
\bottomrule
\end{tabular}
\end{table}

\paragraph{Names against counts on the three stronger models.}
The same two renderings were run on the three stronger models after the matched metric had been read (Table~\ref{tab:cf-nc-strong}).
Removing every count costs GPT-5.6 Luna $-0.0598$ and Opus 5 $-0.0426$ on the pooled metric, each with both intervals clear of zero, so the prediction that names alone would sit within tolerance of the intact row came out otherwise for Luna and held for Opus 5 at the point.
Counts alone still score reliably below names alone for both, by $-0.0930$ and $-0.1488$, as predicted.
Under the matched metric counts alone read $0.4833$ on retraction for Opus 5 and $0.6667$ on expiry for Luna with the persona interval reaching below $0.50$, so the third prediction came out otherwise for both.
GPT-5.6 Sol was not run: after the sibling-name control's lanes the re-projected cost of its names-and-counts lane would have taken the vendor's total over the cap, and the registration drops the lane in that case.

\begin{table}[H]
\caption{Names against counts on the three stronger models (E2, Amendment 1; registered, ADDENDUM\_R125\_E2\_AMENDMENT\_1). E2's two renderings, byte-identical to Table~\ref{tab:cf-names-counts}'s, on Opus 5 (one draw, Batch), GPT-5.6 Luna (five draws through the raw API; its one-draw reading beside) and GPT-5.6 Sol (one draw; not run, see below): the pooled AUROC (its persona interval is in the ledger), the paired difference from the recorded intact row with both intervals ($^{*}$ both exclude zero), counts only minus names only, and the renderings under the matched metric, the whole set and the two counted states (ret., exp.; their persona intervals are in the ledger, P-N5's two failing lower bounds below). Three predictions: P-N3, names only within tolerance of the intact row at the point, failed (on the models run) (GPT-5.6 Luna $-0.0598$; Opus 5 $-0.0426$, within the band); P-N4, counts only reliably below names only, held (on the models run); P-N5, counts only above $0.50$ on both counted states under the matched metric with the persona interval excluding $0.50$, failed (on the models run) (Opus 5 reads $0.4833$ on retraction; GPT-5.6 Luna's expiry lower bound is $0.4615$). GPT-5.6 Sol's lane was dropped by the registration's own rule when the vendor's projection for this task and the sibling-name control together exceeded the cap, so its rows print as not run.}
\label{tab:cf-nc-strong}
\centering\tiny\setlength{\tabcolsep}{1.5pt}
\begin{tabular}{@{}llrrllrll@{}}
\toprule
Reader & summary & AUROC & minus intact & item interval & persona interval & matched & ret. & exp. \\
\midrule
GPT-5.6 Luna, draw 1 & intact (recorded) & $0.8265$ & & & & & & \\
 & names only & $0.7534$ & $-0.0732$$^{*}$ & $[-0.1147,-0.0327]$ & $[-0.1108,-0.0367]$ & $0.7162$ & $0.3667$ & $0.4333$ \\
 & counts only & $0.6803$ & $-0.1462$$^{*}$ & $[-0.1997,-0.0877]$ & $[-0.1884,-0.1009]$ & $0.7297$ & $0.5833$ & $0.7500$ \\
 & counts minus names & & $-0.0730$$^{*}$ & $[-0.1440,-0.0010]$ & $[-0.1274,-0.0170]$ & & & \\
\addlinespace[1pt]
GPT-5.6 Luna, five draws & intact (recorded) & $0.8417$ & & & & & & \\
 & names only & $0.7819$ & $-0.0598$$^{*}$ & $[-0.0967,-0.0270]$ & $[-0.0965,-0.0295]$ & $0.7939$ & $0.5333$ & $0.5167$ \\
 & counts only & $0.6889$ & $-0.1528$$^{*}$ & $[-0.1991,-0.1033]$ & $[-0.1844,-0.1211]$ & $0.8108$ & $0.7000$ & $0.6667$ \\
 & counts minus names & & $-0.0930$$^{*}$ & $[-0.1536,-0.0264]$ & $[-0.1424,-0.0422]$ & & & \\
\addlinespace[1pt]
GPT-5.6 Sol, draw 1 & not run & \multicolumn{7}{l}{dropped by the registration's budget rule (the caption); its moved-count response is in Table~\ref{tab:v2rcat}} \\
Opus 5, draw 1 & intact (recorded) & $0.8400$ & & & & & & \\
 & names only & $0.7974$ & $-0.0426$$^{*}$ & $[-0.0692,-0.0168]$ & $[-0.0705,-0.0171]$ & $0.7635$ & $0.4333$ & $0.4833$ \\
 & counts only & $0.6486$ & $-0.1914$$^{*}$ & $[-0.2347,-0.1453]$ & $[-0.2317,-0.1513]$ & $0.7162$ & $0.4833$ & $0.7667$ \\
 & counts minus names & & $-0.1488$$^{*}$ & $[-0.2023,-0.0929]$ & $[-0.2015,-0.0942]$ & & & \\
\addlinespace[1pt]
\bottomrule
\end{tabular}
\end{table}

\paragraph{PN versus PRESENT, and the ordering within triples (post hoc, specified before recomputation).}
The pooled obligation AUROC of Table~\ref{tab:cf-main} scores the \pn{} versions against both other versions, so it mixes two distinctions: \pn{} against \present{}, where both versions hold relevant content, and \pn{} against \absent{}.
A reader that only knows whether the store covers the question scores $1$ on \pn{} and \present{} and $0$ on \absent{}, which gives $0.75$ pooled on this set and $0.5000$ on the \pn{}-versus-\present{} comparison, because it ties every triple.
Table~\ref{tab:pnpresent} separates the two for every reader, summary and state, with the pooled value equal to the mean of the two (asserted to four decimals, the negative groups being equal in size), and counts, per triple, whether the reader scores the \pn{} version above, equal to, or below the \present{} version.
The largest move from the pooled value on the whole set, any reader, is $0.0230$, so the model-versus-lookup comparison of Section~\ref{sec:lookup} is the same on the harder comparison: the name lookup reads $0.7663$, GPT-4.1 mini $0.7056$, Haiku 4.5 $0.7493$ and Opus 5 $0.8370$.
The strict-win fraction counts ties as failures and is not a tie-adjusted ranking accuracy; GPT-4.1 mini's single call returns coarse probabilities, so its tie count is large ($57$ of $148$ triples on the whole set), and the interval to read is the persona-grouped one, since triples share personas.
Where the summary only counts a state, the two lookups tie every retraction and expiry triple under the superseded-only summary (their lists do not differ), the count readers win most of them, and the models split between ties and small numbers of wins and losses; naming the states turns most of those ties into wins for every reader.
The expected outcomes written before the run held for the name lookup and GPT-4.1 mini (supersession above $0.85$ and $0.70$; retraction and expiry within tolerance of $0.50$ under the superseded-only summary) and for the all-states lookup (above $0.70$ on both counted states); they did not hold for Opus 5, which reads $0.5689$ on retraction and $0.6128$ on expiry against the \present{} versions, nor for Haiku 4.5 on expiry ($0.5544$), nor for the count readers, which the pooled result had already shown to separate the counted states.

\begin{table}[H]
\caption{PN-versus-PRESENT discrimination and within-triple ordering on the counterfactual set, per reader, summary and state (post hoc, specified before recomputation). PN vs PRESENT is the AUROC of the PN versions against the PRESENT versions alone (the ABSENT versions excluded), with the persona-grouped interval; the pooled AUROC of Table~\ref{tab:cf-main} is the mean of this column and the PN-vs-ABSENT AUROC, which the artifact's ledger carries beside the Wilson intervals. W/T/L counts the triples in which the reader scores the PN version above, equal to, or below the PRESENT version; the strict-win fraction counts ties as failures, is not a tie-adjusted ranking accuracy, and its chance reference is not $0.50$ when ties occur. A reader that only knows whether the store covers the question scores $0.75$ pooled and $0.50$ PN vs PRESENT (every triple a tie).}
\label{tab:pnpresent}
\centering\scriptsize\setlength{\tabcolsep}{1.5pt}
\begin{tabular}{@{}llrlr@{\hspace{3pt}}rrl@{}}
\toprule
Reader, summary & scope & PN vs PRESENT & persona interval & pooled & W/T/L & strict-win & persona interval \\
\midrule
GPT-4.1 mini, sup.\ only & whole & $0.7056$ & $[0.6555,0.7575]$ & $0.7167$ & $81/57/10$ & $0.5473$ & $[0.4625,0.6334]$ \\
 & sup. & $0.8265$ & $[0.7607,0.8931]$ & $0.8308$ & $68/19/1$ & $0.7727$ & $[0.6781,0.8734]$ \\
 & ret. & $0.5044$ & $[0.4165,0.5975]$ & $0.5283$ & $9/15/6$ & $0.3000$ & $[0.1665,0.4516]$ \\
 & exp. & $0.5250$ & $[0.4280,0.6450]$ & $0.5489$ & $4/23/3$ & $0.1333$ & $[0.0000,0.3333]$ \\
\addlinespace[1pt]
GPT-4.1 mini, all three & whole & $0.7840$ & $[0.7240,0.8407]$ & $0.7967$ & $105/32/11$ & $0.7095$ & $[0.6063,0.7961]$ \\
 & sup. & $0.8428$ & $[0.7664,0.9130]$ & $0.8444$ & $71/12/5$ & $0.8068$ & $[0.7096,0.9000]$ \\
 & ret. & $0.8433$ & $[0.7641,0.9096]$ & $0.8725$ & $27/2/1$ & $0.9000$ & $[0.7586,1.0000]$ \\
 & exp. & $0.5494$ & $[0.4429,0.6578]$ & $0.5869$ & $7/18/5$ & $0.2333$ & $[0.0606,0.4444]$ \\
\addlinespace[1pt]
Haiku 4.5, sup.\ only & whole & $0.7493$ & $[0.7006,0.7976]$ & $0.7566$ & $89/52/7$ & $0.6014$ & $[0.5034,0.6968]$ \\
 & sup. & $0.8852$ & $[0.8348,0.9356]$ & $0.8876$ & $76/9/3$ & $0.8636$ & $[0.7816,0.9348]$ \\
 & ret. & $0.5072$ & $[0.4521,0.5684]$ & $0.5433$ & $6/22/2$ & $0.2000$ & $[0.0667,0.3574]$ \\
 & exp. & $0.5544$ & $[0.5047,0.6139]$ & $0.5794$ & $7/21/2$ & $0.2333$ & $[0.0625,0.4688]$ \\
\addlinespace[1pt]
Haiku 4.5, all three & whole & $0.8703$ & $[0.8273,0.9083]$ & $0.8773$ & $125/18/5$ & $0.8446$ & $[0.7822,0.9006]$ \\
 & sup. & $0.8857$ & $[0.8337,0.9362]$ & $0.8852$ & $73/11/4$ & $0.8295$ & $[0.7449,0.9059]$ \\
 & ret. & $0.8933$ & $[0.8264,0.9556]$ & $0.9042$ & $27/2/1$ & $0.9000$ & $[0.8000,1.0000]$ \\
 & exp. & $0.8039$ & $[0.7119,0.9022]$ & $0.8269$ & $25/5/0$ & $0.8333$ & $[0.6774,0.9655]$ \\
\addlinespace[1pt]
Opus 5, sup.\ only & whole & $0.8370$ & $[0.7959,0.8772]$ & $0.8405$ & $128/3/17$ & $0.8649$ & $[0.8027,0.9257]$ \\
 & sup. & $0.9331$ & $[0.8952,0.9682]$ & $0.9325$ & $86/0/2$ & $0.9773$ & $[0.9432,1.0000]$ \\
 & ret. & $0.5689$ & $[0.5244,0.6415]$ & $0.5872$ & $22/2/6$ & $0.7333$ & $[0.5556,0.9000]$ \\
 & exp. & $0.6128$ & $[0.5428,0.7028]$ & $0.6150$ & $20/1/9$ & $0.6667$ & $[0.4643,0.8696]$ \\
\addlinespace[1pt]
Opus 5, all three & whole & $0.9070$ & $[0.8758,0.9370]$ & $0.9079$ & $139/0/9$ & $0.9392$ & $[0.8944,0.9750]$ \\
 & sup. & $0.9332$ & $[0.8936,0.9697]$ & $0.9319$ & $85/0/3$ & $0.9659$ & $[0.9140,1.0000]$ \\
 & ret. & $0.7922$ & $[0.7098,0.8814]$ & $0.8100$ & $25/0/5$ & $0.8333$ & $[0.7000,0.9432]$ \\
 & exp. & $0.9300$ & $[0.8623,0.9789]$ & $0.9267$ & $29/0/1$ & $0.9667$ & $[0.8929,1.0000]$ \\
\addlinespace[1pt]
GPT-5.6 Sol, sup.\\ only & whole & $0.8522$ & $[0.8155,0.8872]$ & $0.8500$ & $126/2/20$ & $0.8514$ & $[0.8038,0.8951]$ \\
 & sup. & $0.9426$ & $[0.9110,0.9738]$ & $0.9383$ & $84/1/3$ & $0.9545$ & $[0.9140,0.9889]$ \\
 & ret. & $0.5206$ & $[0.4483,0.5928]$ & $0.5322$ & $17/1/12$ & $0.5667$ & $[0.3871,0.7333]$ \\
 & exp. & $0.6572$ & $[0.5789,0.7511]$ & $0.6403$ & $25/0/5$ & $0.8333$ & $[0.7000,0.9394]$ \\
\addlinespace[1pt]
GPT-5.6 Luna, sup.\\ only & whole & $0.8544$ & $[0.8151,0.8958]$ & $0.8417$ & $129/2/17$ & $0.8716$ & $[0.8027,0.9324]$ \\
 & sup. & $0.9474$ & $[0.9116,0.9782]$ & $0.9426$ & $84/1/3$ & $0.9545$ & $[0.9069,0.9894]$ \\
 & ret. & $0.5550$ & $[0.4885,0.6314]$ & $0.5458$ & $19/1/10$ & $0.6333$ & $[0.4286,0.8125]$ \\
 & exp. & $0.6633$ & $[0.6067,0.7634]$ & $0.6167$ & $26/0/4$ & $0.8667$ & $[0.7586,0.9677]$ \\
\addlinespace[1pt]
Name lookup, sup.\ only & whole & $0.7663$ & $[0.7226,0.8051]$ & $0.7663$ & $82/66/0$ & $0.5541$ & $[0.4691,0.6377]$ \\
 & sup. & $0.9223$ & $[0.8810,0.9576]$ & $0.9223$ & $82/6/0$ & $0.9318$ & $[0.8571,0.9888]$ \\
 & ret. & $0.5000$ & $[0.5000,0.5000]$ & $0.5000$ & $0/30/0$ & $0.0000$ & $[0.0000,0.0000]$ \\
 & exp. & $0.5000$ & $[0.5000,0.5000]$ & $0.5000$ & $0/30/0$ & $0.0000$ & $[0.0000,0.0000]$ \\
\addlinespace[1pt]
All-states lookup, all three & whole & $0.8733$ & $[0.8350,0.9097]$ & $0.8756$ & $121/27/0$ & $0.8176$ & $[0.7569,0.8757]$ \\
 & sup. & $0.8994$ & $[0.8532,0.9412]$ & $0.8995$ & $78/10/0$ & $0.8864$ & $[0.8043,0.9620]$ \\
 & ret. & $0.7817$ & $[0.6950,0.8734]$ & $0.7975$ & $19/11/0$ & $0.6333$ & $[0.4643,0.7880]$ \\
 & exp. & $0.8711$ & $[0.7864,0.9424]$ & $0.8711$ & $24/6/0$ & $0.8000$ & $[0.6667,0.9091]$ \\
\addlinespace[1pt]
Category-guessing count reader, sup.\ only & whole & $0.6393$ & $[0.6033,0.6782]$ & $0.6623$ & $65/83/0$ & $0.4392$ & $[0.3576,0.5133]$ \\
 & sup. & $0.6475$ & $[0.5984,0.7000]$ & $0.6573$ & $34/54/0$ & $0.3864$ & $[0.2577,0.5132]$ \\
 & ret. & $0.6050$ & $[0.5552,0.6739]$ & $0.6669$ & $16/14/0$ & $0.5333$ & $[0.3793,0.7037]$ \\
 & exp. & $0.6683$ & $[0.5844,0.7700]$ & $0.6872$ & $15/15/0$ & $0.5000$ & $[0.3226,0.6522]$ \\
\addlinespace[1pt]
Category-aware count reader, sup.\ only & whole & $0.5941$ & $[0.5651,0.6292]$ & $0.6043$ & $60/88/0$ & $0.4054$ & $[0.3191,0.4895]$ \\
 & sup. & $0.5000$ & $[0.5000,0.5000]$ & $0.5000$ & $0/88/0$ & $0.0000$ & $[0.0000,0.0000]$ \\
 & ret. & $0.6706$ & $[0.6450,0.7585]$ & $0.7306$ & $30/0/0$ & $1.0000$ & $[1.0000,1.0000]$ \\
 & exp. & $0.7778$ & $[0.7153,0.8680]$ & $0.7778$ & $30/0/0$ & $1.0000$ & $[1.0000,1.0000]$ \\
\addlinespace[1pt]
\bottomrule
\end{tabular}
\end{table}

The same store judged under two intents cannot be read from the recorded rows: each question carries one constructed intent and the \textsc{change} intent is never \pn{}, so any comparison across intents would vary intent with the label; the crossover below builds that test with new questions.
Two things can be read instead (post hoc, specified before recomputation).
An intent-aware name lookup, whose score is set to zero on the $105$ \textsc{change}-intent questions of the full obligation set (superseded content is licensed under that intent), reads $0.7957$ against the plain lookup's $0.7484$, $+0.0473$ with intervals $[+0.0331,+0.0654]$ and $[+0.0284,+0.0689]$; the all-states lookup moves from $0.7806$ to $0.8255$ under the same rule plus a zero on the $9$ past-intent questions whose best match is an expired name.
The items on which intent alone decides the label, \textsc{change}-intent licensed questions whose own slot holds a superseded record, number $4$ of the $45$ on the full obligation set, below the floor of five the specification set for a reading, so no monitor comparison is printed on them.

\begin{figure}[H]
\centering
\begin{tikzpicture}
\begin{groupplot}[group style={group size=2 by 1, horizontal sep=0.9cm, ylabels at=edge left, yticklabels at=edge left},
    width=0.41\linewidth, height=5.0cm, xmin=-0.5, xmax=2.5, ymin=0.45, ymax=1.0,
    xtick={0,1,2}, xticklabels={Supersession,Retraction,Expiry}, ytick={0.5,0.6,0.7,0.8,0.9,1.0},
    ylabel={pooled obligation AUROC}, ylabel style={font=\scriptsize}, tick label style={font=\scriptsize}, title style={font=\scriptsize, yshift=-3pt},
    ymajorgrids, grid style={black!12}, axis line style={black!60}, clip=false,
    legend style={at={(1.04,0.5)}, anchor=west, font=\scriptsize, draw=none, fill=none, row sep=-1pt, nodes={anchor=west}}]
  \nextgroupplot[title={only supersession named}]
      \draw[dashed, black!60] (axis cs:-0.5,0.5) -- (axis cs:2.5,0.5);
      \addplot[only marks, mark=*, mark size=2.0pt, black, xshift=-9pt] table[x=x, y=mini] {\figthreeSupPooled};
      \addplot[only marks, mark=square*, mark size=1.9pt, black, xshift=-6pt] table[x=x, y=haiku] {\figthreeSupPooled};
      \addplot[only marks, mark=triangle*, mark size=2.3pt, black, mark options={fill=white}, xshift=-3pt] table[x=x, y=luna] {\figthreeSupPooled};
      \addplot[only marks, mark=diamond*, mark size=2.3pt, black, mark options={fill=white}, xshift=0pt] table[x=x, y=sol] {\figthreeSupPooled};
      \addplot[only marks, mark=pentagon*, mark size=2.2pt, black, xshift=3pt] table[x=x, y=opus] {\figthreeSupPooled};
      \addplot[only marks, mark=x, mark size=2.6pt, black, thick, xshift=6pt] table[x=x, y=lookup] {\figthreeSupPooled};
      \addplot[only marks, mark=+, mark size=2.6pt, black!60, thick, xshift=9pt] table[x=x, y=count] {\figthreeSupPooled};
  \nextgroupplot[title={all states named}]
      \draw[dashed, black!60] (axis cs:-0.5,0.5) -- (axis cs:2.5,0.5);
      \addplot[only marks, mark=*, mark size=2.0pt, black, xshift=-9pt] table[x=x, y=mini] {\figthreeAllPooled};
      \addplot[only marks, mark=square*, mark size=1.9pt, black, xshift=-6pt] table[x=x, y=haiku] {\figthreeAllPooled};
      \addlegendimage{only marks, mark=triangle*, mark size=2.3pt, black, mark options={fill=white}}
      \addlegendimage{only marks, mark=diamond*, mark size=2.3pt, black, mark options={fill=white}}
      \addplot[only marks, mark=pentagon*, mark size=2.2pt, black, xshift=3pt] table[x=x, y=opus] {\figthreeAllPooled};
      \addplot[only marks, mark=x, mark size=2.6pt, black, thick, xshift=6pt] table[x=x, y=lookup] {\figthreeAllPooled};
      \addlegendimage{only marks, mark=+, mark size=2.6pt, black!60, thick}
      \addlegendimage{dashed, black!60}
      \legend{mini (five draws), Haiku, Luna (left only), Sol (left only), Opus, name lookup, count reader (left only), chance}
\end{groupplot}
\end{tikzpicture}
\caption{The pooled version of Figure~\ref{fig:perstate}: per-state pooled obligation AUROC on the counterfactual set, each state's positives against all of its items' negatives, under the summary naming supersession only (left) and the summary naming all states (right); the dashed line is chance. Mini uses five draws, matching Table~\ref{tab:cf-main}; the count reader guesses the category; Luna and Sol have rows under the default summary only. Every value is in Table~\ref{tab:cf-matched}.}
\label{fig:perstate-pooled}
\end{figure}
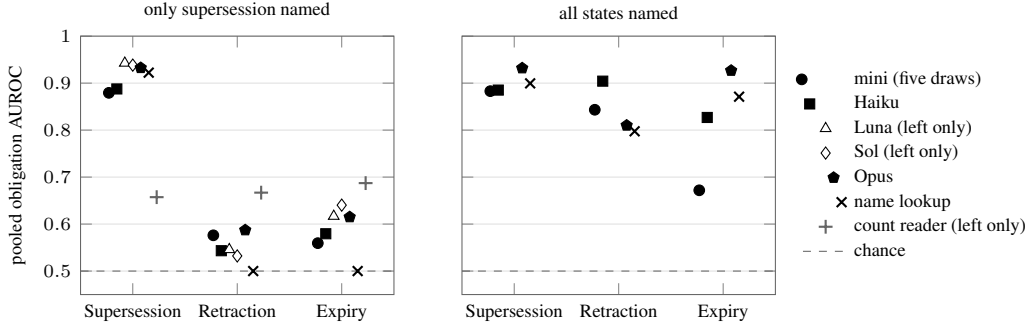

\paragraph{The matched metric.}
The counterfactual set fixes the question and varies the store, so the comparison the design supports is within a triple; the pooled AUROC of Table~\ref{tab:cf-main} also orders each positive against the negatives of every other question.
The matched AUROC of Section~\ref{sec:setup} is $(W+T/2)/n$ over the triples, the \pn{} version against the \present{} version, with a tie counting half; it was defined after every pooled row was known, and every value here is post hoc.
Table~\ref{tab:cf-matched} gives it for every reader of Table~\ref{tab:cf-main} with the triple-level and the persona-grouped interval, whole set and per state; Table~\ref{tab:cf-classes} decomposes the pooled AUROC by pair class; Table~\ref{tab:cf-matched-vs} gives the paired differences against the lookups and the count readers; Table~\ref{tab:cf-matched-ablations} the ablations; Figure~\ref{fig:perstate-pooled} is the pooled version of Figure~\ref{fig:perstate}.
Under this metric a tie is half a win, so the readers that never lose a triple sit on their tie counts: the name lookup ties $66$ of $148$ triples and reads $0.7770$, and the guessing count reader ties $83$ and reads $0.7196$, while a model with fine-grained scores rarely ties and pays for every loss (GPT-4.1 mini's five-draw row loses $33$ triples against its one-draw row's $10$, so it reads $0.7264$ against $0.7399$).
The three claims of Section~\ref{sec:counterfactual} whose reading changes under it are the count readers on the counted states, where GPT-5.6 Luna, GPT-5.6 Sol and Opus 5 are no longer reliably below the guessing reader (Table~\ref{tab:cf-matched-vs}); the moved count, which Luna and Sol follow reliably as well as Opus 5 (Table~\ref{tab:cf-matched-ablations}); and the hybrid reader (Table~\ref{tab:cf-hybrid}).
One reliability changes: Opus 5 under the all-states summary exceeds the all-states lookup by $+0.0304$ with intervals $[-0.0068,+0.0643]$ and $[-0.0038,+0.0648]$, not reliable, where the pooled difference of Table~\ref{tab:cf-main} is; its first draw alone reads $+0.0372$ with both intervals clear of zero.
What the matched comparison does not remove is list length: the random-token lookup reads $0.6453$ on it ($0.7443$ on supersession) because the \pn{} version's list is one name longer than the \present{} version's, and Table~\ref{tab:cf-emptiness} measures that component; the sibling-name control below is built to remove it.

\begin{table}[H]
\caption{Every reader of Table~\ref{tab:cf-main} under both metrics (post hoc; the matched metric was defined after the pooled rows were known). Pooled: the recorded AUROC over the $444$ items, and the mean of the three per-state pooled values (state-balanced). Matched: W/T/L counts the triples in which the reader scores the \pn{} version above, equal to, or below the \present{} version (ties exact equality of the collapsed scores), and the matched AUROC is $(W+T/2)/148$, with the triple-level and the persona-grouped bootstrap interval ($1{,}000$ draws) and the per-state values over each state's own triples ($88$ / $30$ / $30$); the last column is the same comparison against the \absent{} version. The lookups and the count readers never lose a triple, so their matched values are half their tie counts plus their wins; GPT-4.1 mini's five-draw row loses $33$ triples where its one-draw row loses $10$, so it sits lower on this metric and higher on the pooled one.}
\label{tab:cf-matched}
\centering\tiny\setlength{\tabcolsep}{1pt}
\begin{tabular}{@{}lrrrrllcr@{}}
\toprule
Reader & pooled & state-balanced & W/T/L & matched & triple interval & persona interval & matched sup.\ / ret.\ / exp. & vs \absent{} \\
\midrule
\multicolumn{9}{@{}l}{\textit{The default summary (superseded content named)}}\\
Name lookup & $0.7663$ & $0.6408$ & $82/66/0$ & $0.7770$ & $[0.7365,0.8176]$ & $[0.7346,0.8188]$ & $0.9659$ / $0.5000$ / $0.5000$ & $0.7770$ \\
GPT-4.1 mini, one draw & $0.7167$ & $0.6360$ & $81/57/10$ & $0.7399$ & $[0.6858,0.7872]$ & $[0.6916,0.7911]$ & $0.8807$ / $0.5500$ / $0.5167$ & $0.7399$ \\
GPT-4.1 mini, five draws & $0.7648$ & $0.6716$ & $100/15/33$ & $0.7264$ & $[0.6520,0.7973]$ & $[0.6646,0.7871]$ & $0.8920$ / $0.5667$ / $0.4000$ & $0.8514$ \\
Haiku 4.5 & $0.7566$ & $0.6701$ & $89/52/7$ & $0.7770$ & $[0.7264,0.8243]$ & $[0.7175,0.8321]$ & $0.9148$ / $0.5667$ / $0.5833$ & $0.8176$ \\
GPT-5.6 Luna & $0.8417$ & $0.7017$ & $129/2/17$ & $0.8784$ & $[0.8209,0.9291]$ & $[0.8134,0.9363]$ & $0.9602$ / $0.6500$ / $0.8667$ & $0.8176$ \\
GPT-5.6 Sol & $0.8500$ & $0.7036$ & $126/2/20$ & $0.8581$ & $[0.8040,0.9122]$ & $[0.8109,0.9008]$ & $0.9602$ / $0.5833$ / $0.8333$ & $0.8311$ \\
Opus 5 & $0.8405$ & $0.7116$ & $128/3/17$ & $0.8750$ & $[0.8209,0.9257]$ & $[0.8176,0.9317]$ & $0.9773$ / $0.7667$ / $0.6833$ & $0.8919$ \\
Count reader, guessed category & $0.6623$ & $0.6705$ & $65/83/0$ & $0.7196$ & $[0.6791,0.7635]$ & $[0.6788,0.7567]$ & $0.6932$ / $0.7667$ / $0.7500$ & $0.6892$ \\
Count reader, given category & $0.6043$ & $0.6695$ & $60/88/0$ & $0.7027$ & $[0.6622,0.7432]$ & $[0.6596,0.7448]$ & $0.5000$ / $1.0000$ / $1.0000$ & $0.7027$ \\
State-aware count reader & $0.8580$ & $0.8576$ & $148/0/0$ & $1.0000$ & $[1.0000,1.0000]$ & $[1.0000,1.0000]$ & $1.0000$ / $1.0000$ / $1.0000$ & $1.0000$ \\
Category-given share reader & $0.8473$ & $0.8282$ & $148/0/0$ & $1.0000$ & $[1.0000,1.0000]$ & $[1.0000,1.0000]$ & $1.0000$ / $1.0000$ / $1.0000$ & $0.9865$ \\
Lookup with one count feature & $0.7865$ & $0.6967$ & $114/34/0$ & $0.8851$ & $[0.8514,0.9189]$ & $[0.8504,0.9167]$ & $0.9716$ / $0.7667$ / $0.7500$ & $0.8514$ \\
Random-token lookup & $0.5954$ & $0.5550$ & $45/101/2$ & $0.6453$ & $[0.6081,0.6858]$ & $[0.6014,0.6926]$ & $0.7443$ / $0.5000$ / $0.5000$ & $0.6453$ \\
Whole name-map lookup & $0.5776$ & $0.5802$ & $0/148/0$ & $0.5000$ & $[0.5000,0.5000]$ & $[0.5000,0.5000]$ & $0.5000$ / $0.5000$ / $0.5000$ & $0.7061$ \\
Coverage-only reader & $0.7500$ & $0.7500$ & $0/148/0$ & $0.5000$ & -- & -- & -- & $1.0000$ \\
\addlinespace[2pt]\multicolumn{9}{@{}l}{\textit{The summary naming all states}}\\
Name lookup, all states & $0.8756$ & $0.8560$ & $121/27/0$ & $0.9088$ & $[0.8783,0.9392]$ & $[0.8785,0.9379]$ & $0.9432$ / $0.8167$ / $0.9000$ & $0.9223$ \\
GPT-4.1 mini, one draw & $0.7967$ & $0.7679$ & $105/32/11$ & $0.8176$ & $[0.7669,0.8649]$ & $[0.7516,0.8767]$ & $0.8750$ / $0.9333$ / $0.5333$ & $0.8446$ \\
GPT-4.1 mini, five draws & $0.8323$ & $0.7994$ & $130/5/13$ & $0.8953$ & $[0.8446,0.9358]$ & $[0.8449,0.9403]$ & $0.9205$ / $0.9333$ / $0.7833$ & $0.8784$ \\
Haiku 4.5 & $0.8773$ & $0.8721$ & $125/18/5$ & $0.9054$ & $[0.8682,0.9393]$ & $[0.8636,0.9399]$ & $0.8920$ / $0.9333$ / $0.9167$ & $0.9155$ \\
Opus 5 & $0.9079$ & $0.8895$ & $139/0/9$ & $0.9392$ & $[0.8919,0.9730]$ & $[0.8944,0.9750]$ & $0.9659$ / $0.8333$ / $0.9667$ & $0.9595$ \\
Opus 5, draw 1 & $0.9029$ & $0.8789$ & $138/4/6$ & $0.9459$ & $[0.9088,0.9764]$ & $[0.9130,0.9755]$ & $0.9830$ / $0.8167$ / $0.9667$ & $0.9595$ \\
Random-token lookup & $0.5479$ & $0.5592$ & $31/117/0$ & $0.6047$ & $[0.5777,0.6385]$ & $[0.5680,0.6460]$ & $0.5966$ / $0.5500$ / $0.6833$ & $0.6081$ \\
\bottomrule
\end{tabular}
\end{table}

\begin{table}[H]
\caption{The pooled AUROC decomposed by pair class (post hoc). Of the $43,808$ positive-negative pairs the pooled AUROC averages over, $296$ are within a triple, $18,792$ pair a positive with a negative of the same state from another question, and $24,720$ cross states; the pooled value is the pair-count-weighted mean of the three class AUROCs (asserted for every reader). The within-triple class is the matched metric's two comparisons pooled. Readers as in Table~\ref{tab:cf-matched}.}
\label{tab:cf-classes}
\centering\scriptsize\setlength{\tabcolsep}{3pt}
\begin{tabular}{@{}lrrrr@{}}
\toprule
Reader & pooled & within a triple ($296$) & same state, across questions ($18,792$) & across states ($24,720$) \\
\midrule
\multicolumn{5}{@{}l}{\textit{The default summary}}\\
Name lookup & $0.7663$ & $0.7770$ & $0.8437$ & $0.7074$ \\
GPT-4.1 mini, one draw & $0.7167$ & $0.7399$ & $0.7763$ & $0.6712$ \\
GPT-4.1 mini, five draws & $0.7648$ & $0.7889$ & $0.8211$ & $0.7217$ \\
Haiku 4.5 & $0.7566$ & $0.7973$ & $0.8265$ & $0.7029$ \\
GPT-5.6 Luna & $0.8417$ & $0.8480$ & $0.8749$ & $0.8165$ \\
GPT-5.6 Sol & $0.8500$ & $0.8446$ & $0.8724$ & $0.8330$ \\
Opus 5 & $0.8405$ & $0.8834$ & $0.8698$ & $0.8178$ \\
Count reader, guessed category & $0.6623$ & $0.7044$ & $0.6603$ & $0.6633$ \\
Count reader, given category & $0.6043$ & $0.7027$ & $0.5455$ & $0.6478$ \\
State-aware count reader & $0.8580$ & $1.0000$ & $0.8867$ & $0.8344$ \\
Category-given share reader & $0.8473$ & $0.9932$ & $0.8765$ & $0.8234$ \\
Lookup with one count feature & $0.7865$ & $0.8682$ & $0.8471$ & $0.7394$ \\
Random-token lookup & $0.5954$ & $0.6453$ & $0.6338$ & $0.5657$ \\
Whole name-map lookup & $0.5776$ & $0.6030$ & $0.5744$ & $0.5797$ \\
\addlinespace[2pt]\multicolumn{5}{@{}l}{\textit{The summary naming all states}}\\
Name lookup, all states & $0.8756$ & $0.9155$ & $0.8868$ & $0.8667$ \\
GPT-4.1 mini, one draw & $0.7967$ & $0.8311$ & $0.8227$ & $0.7766$ \\
GPT-4.1 mini, five draws & $0.8323$ & $0.8868$ & $0.8589$ & $0.8115$ \\
Haiku 4.5 & $0.8773$ & $0.9105$ & $0.8810$ & $0.8742$ \\
Opus 5 & $0.9079$ & $0.9493$ & $0.9195$ & $0.8986$ \\
Opus 5, draw 1 & $0.9029$ & $0.9527$ & $0.9134$ & $0.8943$ \\
Random-token lookup & $0.5479$ & $0.6064$ & $0.5434$ & $0.5506$ \\
\bottomrule
\end{tabular}
\end{table}

\begin{table}[H]
\caption{Paired differences under the matched metric (post hoc): the mean over triples of the two readers' per-triple outcomes (a win $1$, a tie $\tfrac12$, a loss $0$), with the triple-level and the persona-grouped interval; $^{*}$ marks a difference with both intervals clear of zero. First panel: each reader minus its summary's lookup on the whole set, with the point differences on the two counted states (the lookup ties every counted-state triple under the default summary); second panel: the same on supersession. Under the all-states summary Opus 5's lead over its lookup is not reliable at five draws and is at its first draw alone. Third and fourth panels: each reader minus the guessed-category and the given-category count reader on the counted states; a difference of $-0.5000$ against the given-category reader means that reader won every triple the other tied.}
\label{tab:cf-matched-vs}
\centering\tiny\setlength{\tabcolsep}{2pt}
\begin{tabular}{@{}lrllrr@{}}
\toprule
Reader minus its lookup, whole set & difference & triple interval & persona interval & retraction & expiry \\
\midrule
GPT-4.1 mini, one draw & $-0.0372$ & $[-0.0811,+0.0068]$ & $[-0.0877,+0.0146]$ & $+0.0500$ & $+0.0167$ \\
GPT-4.1 mini, five draws & $-0.0507$ & $[-0.1115,+0.0101]$ & $[-0.1069,+0.0000]$ & $+0.0667$ & $-0.1000$ \\
Haiku 4.5 & $+0.0000$ & $[-0.0439,+0.0372]$ & $[-0.0528,+0.0483]$ & $+0.0667$ & $+0.0833$ \\
GPT-5.6 Luna & $+0.1014$$^{*}$ & $[+0.0473,+0.1554]$ & $[+0.0442,+0.1533]$ & $+0.1500$ & $+0.3667$ \\
GPT-5.6 Sol & $+0.0811$$^{*}$ & $[+0.0270,+0.1351]$ & $[+0.0283,+0.1319]$ & $+0.0833$ & $+0.3333$ \\
Opus 5 & $+0.0980$$^{*}$ & $[+0.0472,+0.1454]$ & $[+0.0435,+0.1503]$ & $+0.2667$ & $+0.1833$ \\
Lookup with one count feature & $+0.1081$$^{*}$ & $[+0.0777,+0.1419]$ & $[+0.0741,+0.1419]$ & $+0.2667$ & $+0.2500$ \\
Count reader, guessed category & $-0.0574$ & $[-0.1149,+0.0034]$ & $[-0.1186,+0.0001]$ & $+0.2667$ & $+0.2500$ \\
Random-token lookup & $-0.1318$$^{*}$ & $[-0.1724,-0.0912]$ & $[-0.1786,-0.0854]$ & $+0.0000$ & $+0.0000$ \\
GPT-4.1 mini, five draws, all states & $-0.0135$ & $[-0.0642,+0.0338]$ & $[-0.0698,+0.0375]$ & $+0.1167$ & $-0.1167$ \\
Haiku 4.5, all states & $-0.0034$ & $[-0.0439,+0.0372]$ & $[-0.0476,+0.0333]$ & $+0.1167$ & $+0.0167$ \\
Opus 5, all states & $+0.0304$ & $[-0.0068,+0.0643]$ & $[-0.0038,+0.0648]$ & $+0.0167$ & $+0.0667$ \\
Opus 5, draw 1, all states & $+0.0372$$^{*}$ & $[+0.0033,+0.0709]$ & $[+0.0076,+0.0662]$ & $+0.0000$ & $+0.0667$ \\
\bottomrule
\end{tabular}\\[4pt]
\begin{tabular}{@{}lrll@{}}
\toprule
Reader minus its lookup, supersession & difference & triple interval & persona interval \\
\midrule
GPT-4.1 mini, one draw & $-0.0852$$^{*}$ & $[-0.1307,-0.0398]$ & $[-0.1420,-0.0294]$ \\
GPT-4.1 mini, five draws & $-0.0739$$^{*}$ & $[-0.1307,-0.0227]$ & $[-0.1393,-0.0214]$ \\
Haiku 4.5 & $-0.0511$ & $[-0.1136,+0.0000]$ & $[-0.1084,+0.0000]$ \\
GPT-5.6 Luna & $-0.0057$ & $[-0.0341,+0.0227]$ & $[-0.0321,+0.0174]$ \\
GPT-5.6 Sol & $-0.0057$ & $[-0.0511,+0.0341]$ & $[-0.0417,+0.0287]$ \\
Opus 5 & $+0.0114$ & $[-0.0227,+0.0455]$ & $[-0.0284,+0.0579]$ \\
Lookup with one count feature & $+0.0057$ & $[+0.0000,+0.0170]$ & $[+0.0000,+0.0179]$ \\
Count reader, guessed category & $-0.2727$$^{*}$ & $[-0.3240,-0.2159]$ & $[-0.3350,-0.2088]$ \\
Random-token lookup & $-0.2216$$^{*}$ & $[-0.2841,-0.1646]$ & $[-0.2889,-0.1506]$ \\
GPT-4.1 mini, five draws, all states & $-0.0227$ & $[-0.0795,+0.0342]$ & $[-0.0886,+0.0337]$ \\
Haiku 4.5, all states & $-0.0511$ & $[-0.1080,+0.0000]$ & $[-0.1123,+0.0054]$ \\
Opus 5, all states & $+0.0227$ & $[-0.0114,+0.0568]$ & $[-0.0105,+0.0550]$ \\
Opus 5, draw 1, all states & $+0.0398$$^{*}$ & $[+0.0170,+0.0682]$ & $[+0.0118,+0.0714]$ \\
\bottomrule
\end{tabular}\\[4pt]
\begin{tabular}{@{}llrll@{}}
\toprule
Reader minus the guessed-category count reader & state & difference & triple interval & persona interval \\
\midrule
Name lookup & retraction & $-0.2667$$^{*}$ & $[-0.3500,-0.1833]$ & $[-0.3519,-0.1897]$ \\
 & expiry & $-0.2500$$^{*}$ & $[-0.3333,-0.1667]$ & $[-0.3261,-0.1613]$ \\
GPT-4.1 mini, one draw & retraction & $-0.2167$$^{*}$ & $[-0.3833,-0.0500]$ & $[-0.3750,-0.0625]$ \\
 & expiry & $-0.2333$$^{*}$ & $[-0.3504,-0.1167]$ & $[-0.3462,-0.1199]$ \\
GPT-4.1 mini, five draws & retraction & $-0.2000$$^{*}$ & $[-0.3671,-0.0167]$ & $[-0.4194,-0.0139]$ \\
 & expiry & $-0.3500$$^{*}$ & $[-0.5167,-0.1833]$ & $[-0.5167,-0.2069]$ \\
Haiku 4.5 & retraction & $-0.2000$$^{*}$ & $[-0.3167,-0.0667]$ & $[-0.3089,-0.0769]$ \\
 & expiry & $-0.1667$$^{*}$ & $[-0.3000,-0.0500]$ & $[-0.2904,-0.0151]$ \\
GPT-5.6 Luna & retraction & $-0.1167$ & $[-0.3333,+0.1167]$ & $[-0.3336,+0.0770]$ \\
 & expiry & $+0.1167$ & $[-0.0333,+0.2500]$ & $[-0.0156,+0.2586]$ \\
GPT-5.6 Sol & retraction & $-0.1833$ & $[-0.3500,+0.0000]$ & $[-0.3519,-0.0284]$ \\
 & expiry & $+0.0833$ & $[-0.0504,+0.2167]$ & $[-0.0653,+0.2143]$ \\
Opus 5 & retraction & $+0.0000$ & $[-0.1833,+0.1833]$ & $[-0.1966,+0.1757]$ \\
 & expiry & $-0.0667$ & $[-0.2504,+0.1167]$ & $[-0.2834,+0.1379]$ \\
Lookup with one count feature & retraction & $+0.0000$ & $[+0.0000,+0.0000]$ & $[+0.0000,+0.0000]$ \\
 & expiry & $+0.0000$ & $[+0.0000,+0.0000]$ & $[+0.0000,+0.0000]$ \\
\bottomrule
\end{tabular}\\[4pt]
\begin{tabular}{@{}llrll@{}}
\toprule
Reader minus the given-category count reader & state & difference & triple interval & persona interval \\
\midrule
Name lookup & retraction & $-0.5000$$^{*}$ & $[-0.5000,-0.5000]$ & $[-0.5000,-0.5000]$ \\
 & expiry & $-0.5000$$^{*}$ & $[-0.5000,-0.5000]$ & $[-0.5000,-0.5000]$ \\
GPT-4.1 mini, one draw & retraction & $-0.4500$$^{*}$ & $[-0.5667,-0.3333]$ & $[-0.5600,-0.3448]$ \\
 & expiry & $-0.4833$$^{*}$ & $[-0.5667,-0.4000]$ & $[-0.5893,-0.3548]$ \\
GPT-4.1 mini, five draws & retraction & $-0.4333$$^{*}$ & $[-0.6000,-0.2667]$ & $[-0.6094,-0.2586]$ \\
 & expiry & $-0.6000$$^{*}$ & $[-0.7500,-0.4500]$ & $[-0.7143,-0.4857]$ \\
Haiku 4.5 & retraction & $-0.4333$$^{*}$ & $[-0.5167,-0.3500]$ & $[-0.5175,-0.3393]$ \\
 & expiry & $-0.4167$$^{*}$ & $[-0.5004,-0.3167]$ & $[-0.5312,-0.2884]$ \\
GPT-5.6 Luna & retraction & $-0.3500$$^{*}$ & $[-0.5167,-0.1833]$ & $[-0.5485,-0.1786]$ \\
 & expiry & $-0.1333$$^{*}$ & $[-0.2667,-0.0333]$ & $[-0.2414,-0.0323]$ \\
GPT-5.6 Sol & retraction & $-0.4167$$^{*}$ & $[-0.5833,-0.2500]$ & $[-0.5893,-0.2500]$ \\
 & expiry & $-0.1667$$^{*}$ & $[-0.3000,-0.0333]$ & $[-0.3000,-0.0606]$ \\
Opus 5 & retraction & $-0.2333$$^{*}$ & $[-0.3833,-0.1000]$ & $[-0.3889,-0.0882]$ \\
 & expiry & $-0.3167$$^{*}$ & $[-0.4833,-0.1667]$ & $[-0.5000,-0.1250]$ \\
Lookup with one count feature & retraction & $-0.2333$$^{*}$ & $[-0.3167,-0.1500]$ & $[-0.3103,-0.1481]$ \\
 & expiry & $-0.2500$$^{*}$ & $[-0.3333,-0.1667]$ & $[-0.3387,-0.1739]$ \\
\bottomrule
\end{tabular}
\end{table}

\begin{table}[H]
\caption{The ablations under the matched metric (post hoc). Upper panel: the names-only, counts-only, totals-randomized and moved-count rows, W/T/L against the \present{} version and the matched AUROC, whole set and per state (Tables~\ref{tab:cf-names-counts} and~\ref{tab:v2rcat} carry the pooled rows). Middle panels: the paired contrasts, one row per scope, with both intervals, $^{*}$ where both exclude zero; the counted states are the $60$ retraction and expiry triples. Names alone leave both smaller models where the intact summary left them and counts alone cost each about $0.14$, all of it on supersession; the moved count moves the three stronger models reliably on the counted states and neither smaller model. Lower panel: Opus 5 under the all-states summary minus the all-states lookup, five draws and its first draw alone.}
\label{tab:cf-matched-ablations}
\centering\tiny\setlength{\tabcolsep}{2pt}
\begin{tabular}{@{}lrrrrr@{}}
\toprule
Reader, summary & W/T/L & matched & supersession & retraction & expiry \\
\midrule
GPT-4.1 mini, names only & $75/69/4$ & $0.7399$ & $0.8864$ & $0.5500$ & $0.5000$ \\
GPT-4.1 mini, names only, dash rule & $69/66/13$ & $0.6892$ & $0.8523$ & $0.4500$ & $0.4500$ \\
GPT-4.1 mini, counts only & $48/83/17$ & $0.6047$ & $0.6534$ & $0.6167$ & $0.4500$ \\
Haiku 4.5, names only & $85/59/4$ & $0.7736$ & $0.9318$ & $0.5167$ & $0.5667$ \\
Haiku 4.5, counts only & $61/66/21$ & $0.6351$ & $0.7159$ & $0.5833$ & $0.4500$ \\
GPT-4.1 mini, totals randomized & $78/50/20$ & $0.6959$ & $0.8750$ & $0.4000$ & $0.4667$ \\
Haiku 4.5, totals randomized & $89/50/9$ & $0.7703$ & $0.8977$ & $0.6167$ & $0.5500$ \\
Opus 5, totals randomized & $128/0/20$ & $0.8649$ & $0.9659$ & $0.6667$ & $0.7667$ \\
GPT-4.1 mini, moved count & $80/58/10$ & $0.7365$ & $0.8636$ & $0.5500$ & $0.5500$ \\
Haiku 4.5, moved count & $84/53/11$ & $0.7466$ & $0.8977$ & $0.5667$ & $0.4833$ \\
Opus 5, moved count (five draws) & $108/3/37$ & $0.7399$ & $0.9432$ & $0.4000$ & $0.4833$ \\
GPT-5.6 Luna, moved count (five draws) & $113/1/34$ & $0.7669$ & $0.9716$ & $0.3333$ & $0.6000$ \\
GPT-5.6 Sol, moved count (five draws) & $107/1/40$ & $0.7264$ & $0.9602$ & $0.4000$ & $0.3667$ \\
State-aware count reader, moved count & $88/33/27$ & $0.7061$ & $1.0000$ & $0.2667$ & $0.2833$ \\
\bottomrule
\end{tabular}\\[4pt]
\begin{tabular}{@{}llrll@{}}
\toprule
Contrast (names, counts, totals) & scope & difference & triple interval & persona interval \\
\midrule
GPT-4.1 mini, names only minus intact & whole set & $+0.0000$ & $[-0.0439,+0.0473]$ & $[-0.0440,+0.0426]$ \\
 & supersession & $+0.0057$ & $[-0.0398,+0.0511]$ & $[-0.0349,+0.0533]$ \\
 & counted states & $-0.0083$ & $[-0.1083,+0.0750]$ & $[-0.1059,+0.0819]$ \\
GPT-4.1 mini, counts only minus names only & whole set & $-0.1351$$^{*}$ & $[-0.2027,-0.0743]$ & $[-0.1943,-0.0827]$ \\
 & supersession & $-0.2330$$^{*}$ & $[-0.3125,-0.1533]$ & $[-0.3133,-0.1522]$ \\
 & counted states & $+0.0083$ & $[-0.0833,+0.1000]$ & $[-0.0970,+0.1226]$ \\
Haiku 4.5, names only minus intact & whole set & $-0.0034$ & $[-0.0507,+0.0473]$ & $[-0.0671,+0.0578]$ \\
 & supersession & $+0.0170$ & $[-0.0341,+0.0739]$ & $[-0.0353,+0.0787]$ \\
 & counted states & $-0.0333$ & $[-0.1333,+0.0500]$ & $[-0.1392,+0.0741]$ \\
Haiku 4.5, counts only minus names only & whole set & $-0.1385$$^{*}$ & $[-0.2095,-0.0776]$ & $[-0.2113,-0.0640]$ \\
 & supersession & $-0.2159$$^{*}$ & $[-0.3068,-0.1364]$ & $[-0.3193,-0.1111]$ \\
 & counted states & $-0.0250$ & $[-0.1167,+0.0669]$ & $[-0.1215,+0.0715]$ \\
GPT-4.1 mini, totals randomized minus intact & whole set & $-0.0439$ & $[-0.1014,+0.0135]$ & $[-0.0893,+0.0000]$ \\
Haiku 4.5, totals randomized minus intact & whole set & $-0.0068$ & $[-0.0508,+0.0338]$ & $[-0.0580,+0.0520]$ \\
Opus 5, totals randomized minus intact & whole set & $-0.0101$ & $[-0.0777,+0.0575]$ & $[-0.0725,+0.0548]$ \\
\bottomrule
\end{tabular}\\[4pt]
\begin{tabular}{@{}llrll@{}}
\toprule
Contrast (the moved count; all states named) & scope & difference & triple interval & persona interval \\
\midrule
GPT-4.1 mini, moved count minus intact & whole set & $-0.0034$ & $[-0.0541,+0.0473]$ & $[-0.0537,+0.0472]$ \\
 & retraction & $+0.0000$ & $[-0.1500,+0.1337]$ & $[-0.1452,+0.1452]$ \\
 & expiry & $+0.0333$ & $[-0.1000,+0.1833]$ & $[-0.0962,+0.1667]$ \\
 & counted states & $+0.0167$ & $[-0.0750,+0.1083]$ & $[-0.0896,+0.1334]$ \\
Haiku 4.5, moved count minus intact & whole set & $-0.0304$ & $[-0.0743,+0.0101]$ & $[-0.0756,+0.0245]$ \\
 & retraction & $+0.0000$ & $[-0.1333,+0.1333]$ & $[-0.1613,+0.1481]$ \\
 & expiry & $-0.1000$ & $[-0.2000,+0.0000]$ & $[-0.2000,+0.0000]$ \\
 & counted states & $-0.0500$ & $[-0.1417,+0.0250]$ & $[-0.1455,+0.0704]$ \\
Opus 5, moved count minus intact (five against five draws) & whole set & $-0.1351$$^{*}$ & $[-0.2162,-0.0608]$ & $[-0.2169,-0.0577]$ \\
 & retraction & $-0.3667$$^{*}$ & $[-0.6167,-0.1167]$ & $[-0.5962,-0.1296]$ \\
 & expiry & $-0.2000$ & $[-0.4171,+0.0500]$ & $[-0.4501,+0.0371]$ \\
 & counted states & $-0.2833$$^{*}$ & $[-0.4417,-0.0915]$ & $[-0.4818,-0.0895]$ \\
GPT-5.6 Luna, moved count minus intact (five against five) & whole set & $-0.1115$$^{*}$ & $[-0.1824,-0.0338]$ & $[-0.1793,-0.0392]$ \\
 & retraction & $-0.3167$$^{*}$ & $[-0.5667,-0.0667]$ & $[-0.5758,-0.0333]$ \\
 & expiry & $-0.2667$$^{*}$ & $[-0.4667,-0.0667]$ & $[-0.4571,-0.0714]$ \\
 & counted states & $-0.2917$$^{*}$ & $[-0.4500,-0.1250]$ & $[-0.4630,-0.1231]$ \\
GPT-5.6 Sol, moved count minus intact (five against five) & whole set & $-0.1318$$^{*}$ & $[-0.2163,-0.0507]$ & $[-0.2169,-0.0419]$ \\
 & retraction & $-0.1833$ & $[-0.4500,+0.1000]$ & $[-0.4483,+0.1347]$ \\
 & expiry & $-0.4667$$^{*}$ & $[-0.7000,-0.2000]$ & $[-0.6452,-0.2496]$ \\
 & counted states & $-0.3250$$^{*}$ & $[-0.5000,-0.1333]$ & $[-0.4825,-0.1344]$ \\
GPT-4.1 mini, five draws, all states minus default & whole set & $+0.1689$$^{*}$ & $[+0.0946,+0.2432]$ & $[+0.1114,+0.2293]$ \\
 & counted states & $+0.3750$$^{*}$ & $[+0.2500,+0.5002]$ & $[+0.2755,+0.4828]$ \\
Haiku 4.5, all states minus default & whole set & $+0.1284$$^{*}$ & $[+0.0743,+0.1824]$ & $[+0.0625,+0.1930]$ \\
 & counted states & $+0.3500$$^{*}$ & $[+0.2667,+0.4333]$ & $[+0.2659,+0.4310]$ \\
Opus 5, all states minus default & whole set & $+0.0642$ & $[+0.0068,+0.1182]$ & $[-0.0037,+0.1234]$ \\
 & counted states & $+0.1750$$^{*}$ & $[+0.0667,+0.3000]$ & $[+0.0614,+0.2924]$ \\
Name lookup, all states minus default & whole set & $+0.1318$$^{*}$ & $[+0.0945,+0.1724]$ & $[+0.0906,+0.1690]$ \\
 & counted states & $+0.3583$$^{*}$ & $[+0.3000,+0.4167]$ & $[+0.3071,+0.4105]$ \\
\bottomrule
\end{tabular}\\[4pt]
\begin{tabular}{@{}llrll@{}}
\toprule
Opus 5 minus the all-states lookup & scope & difference & triple interval & persona interval \\
\midrule
Opus 5, five draws & whole set & $+0.0304$ & $[-0.0068,+0.0643]$ & $[-0.0038,+0.0648]$ \\
 & supersession & $+0.0227$ & $[-0.0114,+0.0568]$ & $[-0.0105,+0.0550]$ \\
 & retraction & $+0.0167$ & $[-0.0833,+0.1333]$ & $[-0.0800,+0.1207]$ \\
 & expiry & $+0.0667$ & $[-0.0333,+0.1667]$ & $[+0.0000,+0.1407]$ \\
Opus 5, draw 1 & whole set & $+0.0372$$^{*}$ & $[+0.0033,+0.0709]$ & $[+0.0076,+0.0662]$ \\
 & supersession & $+0.0398$$^{*}$ & $[+0.0170,+0.0682]$ & $[+0.0118,+0.0714]$ \\
 & retraction & $+0.0000$ & $[-0.1000,+0.1004]$ & $[-0.0962,+0.0968]$ \\
 & expiry & $+0.0667$ & $[-0.0333,+0.1667]$ & $[+0.0000,+0.1407]$ \\
\bottomrule
\end{tabular}
\end{table}

\pgfplotstableread[col sep=comma]{
x,state,mini,haiku,luna,sol,opus,lookup,count
0,Supersession,0.8793,0.8876,0.9426,0.9383,0.9325,0.9223,0.6573
1,Retraction,0.5761,0.5433,0.5458,0.5322,0.5872,0.5000,0.6669
2,Expiry,0.5594,0.5794,0.6167,0.6403,0.6150,0.5000,0.6872
}\figthreeSupPooled
\pgfplotstableread[col sep=comma]{
x,state,mini,haiku,opus,lookup
0,Supersession,0.8829,0.8852,0.9319,0.8995
1,Retraction,0.8433,0.9042,0.8100,0.7975
2,Expiry,0.6719,0.8269,0.9267,0.8711
}\figthreeAllPooled

\begin{table}[H]
\caption{How much of the lookup is list length and emptiness (post hoc). Upper panel: the superseded-names list of the $88$ supersession pairs per store version under the default summary, the count of empty lists, the mean length and the length distribution; the \pn{} version's list is longer than the \present{} version's by one name on $81$ pairs, two on $6$ and three on $1$, and on the $24$ pairs whose \present{} list is empty emptiness alone identifies the \pn{} version. Middle panel: the random-token control (every name replaced by a random token, the list's length and emptiness kept) on the counterfactual set under both summaries, pooled and matched, and on the full obligation set (no triples, so no matched row; each state's positives against all $557$ negatives). Lower panel: the name lookup minus the random-token control, pooled with the item-level and persona-grouped intervals and matched with the triple-level and persona-grouped intervals ($^{*}$ both exclude zero). Under the default summary $0.60$ of the lookup's $0.77$ pooled and $0.65$ of its $0.78$ matched is reachable with no name meaning anything; the sibling-name control (Table~\ref{tab:cf-sibling}) removes that component.}
\label{tab:cf-emptiness}
\centering\tiny\setlength{\tabcolsep}{2pt}
\begin{tabular}{@{}lrrl@{}}
\toprule
Store version & empty lists & mean length & length: pairs \\
\midrule
\pn{} & $0$ & $2.6250$ & 1: 23, 2: 17, 3: 25, 4: 17, 5: 5, 6: 1 \\
\present{} & $24$ & $1.5341$ & 0: 24, 1: 16, 2: 31, 3: 12, 4: 4, 5: 1 \\
\absent{} & $24$ & $1.5341$ & 0: 24, 1: 16, 2: 31, 3: 12, 4: 4, 5: 1 \\
\bottomrule
\end{tabular}\\[4pt]
\begin{tabular}{@{}lrlrrr@{}}
\toprule
Random-token lookup & pooled & sup.\ / ret.\ / exp. & W/T/L & matched & matched, supersession \\
\midrule
$444$, default summary & $0.5954$ & $0.6651$ / $0.5000$ / $0.5000$ & $45/101/2$ & $0.6453$ & $0.7443$ \\
$444$, all states named & $0.5479$ & $0.5385$ / $0.5697$ / $0.5694$ & $31/117/0$ & $0.6047$ & $0.5966$ \\
$661$, as committed & $0.6654$ & $0.7389$ / $0.6032$ / $0.4647$ & -- & -- & -- \\
$661$, marker-free & $0.6627$ & $0.7764$ / $0.6134$ / $0.2706$ & -- & -- & -- \\
\bottomrule
\end{tabular}\\[4pt]
\begin{tabular}{@{}lrll@{}}
\toprule
Name lookup minus random token & difference & item or triple interval & persona interval \\
\midrule
$444$, default summary, pooled & $+0.1709$$^{*}$ & $[+0.1218,+0.2218]$ & $[+0.1284,+0.2132]$ \\
$444$, default summary, matched & $+0.1318$$^{*}$ & $[+0.0912,+0.1724]$ & $[+0.0854,+0.1786]$ \\
$444$, all states named, pooled & $+0.3277$$^{*}$ & $[+0.2576,+0.3925]$ & $[+0.2928,+0.3631]$ \\
$444$, all states named, matched & $+0.3041$$^{*}$ & $[+0.2568,+0.3447]$ & $[+0.2516,+0.3493]$ \\
$661$, as committed, pooled & $+0.0830$ & $[+0.0076,+0.1654]$ & $[-0.0097,+0.1826]$ \\
$661$, marker-free, pooled & $+0.0815$$^{*}$ & $[+0.0270,+0.1438]$ & $[+0.0164,+0.1603]$ \\
\bottomrule
\end{tabular}
\end{table}

\paragraph{The intent crossover.}
The crossover asks each in-scope supersession pair of the set its recorded present-tense question and a second question about how the attribute has changed, and each expiry pair a question about the past event, of the same two store versions, so that the same stale record is forbidden under the original question and licensed under the crossed one (Table~\ref{tab:cf-intent}).
The crossed questions were drafted by GPT-4.1 mini from the attribute path and the time sense alone, never from any stored value, screened for value leakage, and read by us before any monitor call; five pairs were dropped and two took a second drafting seed.
Two rubrics were used: the main-study rubric, and one that states that the obligation depends on the question's intent.
The name lookup, a keyword-intent lookup over a fixed cue list, and an oracle-intent lookup that reads the constructed intent are the readers with no generative-model call.
Three predictions were written before any call.
The first, that every model would sit within tolerance of $0.50$ on the same-store contrast under the main-study rubric, came out otherwise: four models sit below it and Opus 5 above it.
The second, that Opus 5 and Haiku 4.5 would separate the intents under the intent-explicit rubric, came out otherwise for Haiku 4.5.
The third, that the name lookup would sit within tolerance of $0.50$, came out as written.
GPT-5.6 Sol's prediction, made after an interim read of three models' rows, that it separates the intents under the intent-explicit rubric, came out as written.
Under the intent-explicit rubric the counterfactual set itself was rerun for GPT-5.6 Luna at five draws and GPT-4.1 mini at one (Table~\ref{tab:cf-p1}).
Each reads within $0.02$ of its main-rubric row ($0.8531$ against $0.8417$; $0.7253$ against $0.7167$).
Luna keeps its per-state order and a reliable margin over the name lookup ($+0.0867$, intervals $[+0.0447,+0.1302]$ and $[+0.0562,+0.1219]$); GPT-4.1 mini's two counted states swap order (expiry $0.5489$ to $0.4758$, retraction $0.5283$ to $0.5456$), neither move reliable, which is a failed prediction.

\begin{table}[H]
\caption{The intent crossover on the counterfactual set (registered; ADDENDUM\_CF2\_INTENT\_CROSSOVER and its Amendment 1). Each supersession pair of the set is asked its recorded present-tense question and a second question about how the attribute has changed, of the same two store versions (\pn{} and \present{}) under the default summary (population A, $87$ of the $88$ pairs); each expiry pair is asked its recorded upcoming-intent question and a question about the past event under the all-states summary (population B, $26$ of the $30$ pairs); our read of the crossed questions dropped five pairs before any call. The crossed question makes the same stale record licensed, so a reader that conditions on intent scores the \pn{} version of the original question above the same version under the crossed question. This table: that same-store AUROC, chance $0.50$, with the persona-grouped and the pair-clustered interval and the within-pair W/T/L. Table~\ref{tab:cf-intent-b}: the interaction, the mean of the two-by-two contrast, with both intervals, and the pooled AUROC of the original question's \pn{} version against the other three cells. Each reader ran under the main-study rubric and under a rubric that states the intent rule. GPT-5.6 Luna and Haiku 4.5 ran five draws; their draw-1 rows are the column comparable with the one-draw readers, and the body reads that column. GPT-5.6 Sol was added after an interim read of three models' rows, and its prediction is labelled so in Appendix~\ref{app:registration}; under the intent-explicit rubric it counts $76/0/11$ within pairs on population A, and in $9$ of its $11$ losses both questions score $0.96$ or higher, so the rule was not applied on those pairs. Under the intent-explicit rubric no model's discrimination on its original questions falls by more than $0.05$ (the largest fall $-0.0458$, Haiku 4.5 on population A; the largest rise $+0.1716$, GPT-4.1 mini on population B).}
\label{tab:cf-intent}
\centering\tiny\setlength{\tabcolsep}{1.5pt}
\begin{tabular}{@{}lllrllr@{}}
\toprule
Reader & pop. & rubric & same-store AUROC & persona interval & pair interval & W/T/L \\
\midrule
GPT-4.1 mini ($k=1$) & A & main-study & $0.3973$ & $[0.3447,0.4455]$ & $[0.3460,0.4426]$ & 16/26/45 \\
 & A & intent-explicit & $0.4789$ & $[0.4169,0.5413]$ & $[0.4261,0.5279]$ & 22/34/31 \\
 & B & main-study & $0.3388$ & $[0.2385,0.4297]$ & $[0.2411,0.4157]$ & 0/13/13 \\
 & B & intent-explicit & $0.4238$ & $[0.3264,0.5236]$ & $[0.3262,0.5297]$ & 7/5/14 \\
\addlinespace[1pt]
Haiku 4.5 ($k=5$) & A & main-study & $0.3765$ & $[0.3183,0.4257]$ & $[0.3239,0.4203]$ & 18/16/53 \\
\quad draw 1 only & A & main-study & $0.3872$ & $[0.3302,0.4367]$ & $[0.3350,0.4318]$ & 16/22/49 \\
 & A & intent-explicit & $0.5483$ & $[0.4793,0.6135]$ & $[0.4819,0.6152]$ & 42/10/35 \\
\quad draw 1 only & A & intent-explicit & $0.5464$ & $[0.4770,0.6073]$ & $[0.4792,0.6115]$ & 37/16/34 \\
 & B & main-study & $0.4238$ & $[0.3179,0.5136]$ & $[0.3461,0.5022]$ & 8/2/16 \\
\quad draw 1 only & B & main-study & $0.4120$ & $[0.3202,0.4863]$ & $[0.3395,0.4712]$ & 4/6/16 \\
 & B & intent-explicit & $0.5333$ & $[0.4437,0.6206]$ & $[0.4378,0.6331]$ & 13/5/8 \\
\quad draw 1 only & B & intent-explicit & $0.5377$ & $[0.4442,0.6278]$ & $[0.4393,0.6442]$ & 12/6/8 \\
\addlinespace[1pt]
GPT-5.6 Luna ($k=5$) & A & main-study & $0.3467$ & $[0.2944,0.3847]$ & $[0.2994,0.3827]$ & 10/1/76 \\
\quad draw 1 only & A & main-study & $0.3484$ & $[0.3121,0.3773]$ & $[0.3027,0.3876]$ & 15/8/64 \\
 & A & intent-explicit & $0.9836$ & $[0.9558,1.0000]$ & $[0.9564,1.0000]$ & 85/0/2 \\
\quad draw 1 only & A & intent-explicit & $0.9640$ & $[0.9238,0.9935]$ & $[0.9252,0.9940]$ & 82/2/3 \\
 & B & main-study & $0.4172$ & $[0.3497,0.4664]$ & $[0.3328,0.4786]$ & 8/0/18 \\
\quad draw 1 only & B & main-study & $0.4800$ & $[0.4058,0.5608]$ & $[0.3972,0.5703]$ & 13/2/11 \\
 & B & intent-explicit & $0.9793$ & $[0.9401,1.0000]$ & $[0.9356,1.0000]$ & 26/0/0 \\
\quad draw 1 only & B & intent-explicit & $0.9808$ & $[0.9424,1.0000]$ & $[0.9423,1.0000]$ & 26/0/0 \\
\addlinespace[1pt]
Opus 5 ($k=1$) & A & main-study & $0.8966$ & $[0.8471,0.9477]$ & $[0.8428,0.9453]$ & 78/3/6 \\
 & A & intent-explicit & $0.9885$ & $[0.9634,1.0000]$ & $[0.9655,1.0000]$ & 86/0/1 \\
 & B & main-study & $0.8484$ & $[0.7434,0.9407]$ & $[0.7559,0.9334]$ & 23/0/3 \\
 & B & intent-explicit & $1.0000$ & $[1.0000,1.0000]$ & $[1.0000,1.0000]$ & 26/0/0 \\
\addlinespace[1pt]
GPT-5.6 Sol ($k=1$; Amendment 1) & A & main-study & $0.4406$ & $[0.3798,0.5042]$ & $[0.3853,0.4934]$ & 15/32/40 \\
 & A & intent-explicit & $0.8815$ & $[0.8200,0.9349]$ & $[0.8131,0.9449]$ & 76/0/11 \\
 & B & main-study & $0.6354$ & $[0.5261,0.7378]$ & $[0.5318,0.7478]$ & 15/4/7 \\
 & B & intent-explicit & $0.9179$ & $[0.7917,1.0000]$ & $[0.8077,1.0000]$ & 23/0/3 \\
\addlinespace[1pt]
\multicolumn{7}{@{}l}{\textit{Readers with no generative-model call}}\\
Name lookup & A & -- & $0.4739$ & $[0.4413,0.5079]$ & $[0.4336,0.5143]$ & 46/0/41 \\
 & B & -- & $0.4527$ & $[0.3888,0.4912]$ & $[0.3905,0.5015]$ & 10/0/16 \\
Keyword-intent lookup (fixed cue list) & A & -- & $0.9943$ & $[0.9817,1.0000]$ & $[0.9828,1.0000]$ & 86/1/0 \\
 & B & -- & $0.9460$ & $[0.8535,1.0000]$ & $[0.8654,1.0000]$ & 25/1/0 \\
Oracle-intent lookup & A & -- & $1.0000$ & $[1.0000,1.0000]$ & $[1.0000,1.0000]$ & 87/0/0 \\
 & B & -- & $1.0000$ & $[1.0000,1.0000]$ & $[1.0000,1.0000]$ & 26/0/0 \\
\bottomrule
\end{tabular}
\end{table}

\begin{table}[H]
\caption{The intent crossover, continued from Table~\ref{tab:cf-intent}: the interaction, the mean of the two-by-two contrast (the original question's \pn{} version minus its \present{} version, minus the same difference under the crossed question), with the persona-grouped and the pair-clustered interval, and the pooled AUROC of the original question's \pn{} version against the other three cells, with the same readers, populations and rubrics.}
\label{tab:cf-intent-b}
\centering\tiny\setlength{\tabcolsep}{1.5pt}
\begin{tabular}{@{}lllrllr@{}}
\toprule
Reader & pop. & rubric & interaction & persona interval & pair interval & pooled AUROC \\
\midrule
GPT-4.1 mini ($k=1$) & A & main-study & $-0.0695$ & $[-0.1149,-0.0268]$ & $[-0.1161,-0.0276]$ & $0.6821$ \\
 & A & intent-explicit & $-0.0259$ & $[-0.0899,+0.0300]$ & $[-0.0822,+0.0242]$ & $0.7275$ \\
 & B & main-study & $-0.1058$ & $[-0.1655,-0.0444]$ & $[-0.1731,-0.0365]$ & $0.4497$ \\
 & B & intent-explicit & $-0.0538$ & $[-0.1340,+0.0310]$ & $[-0.1635,+0.0520]$ & $0.6146$ \\
\addlinespace[1pt]
Haiku 4.5 ($k=5$) & A & main-study & $-0.0218$ & $[-0.0683,+0.0204]$ & $[-0.0609,+0.0160]$ & $0.7035$ \\
\quad draw 1 only & A & main-study & $-0.0195$ & $[-0.0634,+0.0223]$ & $[-0.0609,+0.0209]$ & $0.7028$ \\
 & A & intent-explicit & $-0.0456$ & $[-0.1046,+0.0202]$ & $[-0.1005,+0.0130]$ & $0.7572$ \\
\quad draw 1 only & A & intent-explicit & $-0.0438$ & $[-0.1043,+0.0252]$ & $[-0.1047,+0.0185]$ & $0.7549$ \\
 & B & main-study & $-0.0592$ & $[-0.1406,+0.0427]$ & $[-0.1252,+0.0197]$ & $0.7064$ \\
\quad draw 1 only & B & main-study & $-0.0504$ & $[-0.1284,+0.0447]$ & $[-0.1289,+0.0392]$ & $0.6970$ \\
 & B & intent-explicit & $-0.0255$ & $[-0.1114,+0.0539]$ & $[-0.1024,+0.0403]$ & $0.7470$ \\
\quad draw 1 only & B & intent-explicit & $-0.0250$ & $[-0.1229,+0.0622]$ & $[-0.1073,+0.0439]$ & $0.7465$ \\
\addlinespace[1pt]
GPT-5.6 Luna ($k=5$) & A & main-study & $-0.0626$ & $[-0.0957,-0.0290]$ & $[-0.1014,-0.0211]$ & $0.7460$ \\
\quad draw 1 only & A & main-study & $-0.0963$ & $[-0.1389,-0.0544]$ & $[-0.1509,-0.0454]$ & $0.7317$ \\
 & A & intent-explicit & $+0.6098$ & $[+0.5405,+0.6829]$ & $[+0.5413,+0.6778]$ & $0.9714$ \\
\quad draw 1 only & A & intent-explicit & $+0.5997$ & $[+0.5233,+0.6751]$ & $[+0.5291,+0.6704]$ & $0.9523$ \\
 & B & main-study & $-0.0825$ & $[-0.1464,-0.0281]$ & $[-0.1309,-0.0369]$ & $0.7690$ \\
\quad draw 1 only & B & main-study & $-0.0500$ & $[-0.1297,+0.0208]$ & $[-0.1165,+0.0142]$ & $0.7840$ \\
 & B & intent-explicit & $+0.4975$ & $[+0.3965,+0.6067]$ & $[+0.3810,+0.6120]$ & $0.9588$ \\
\quad draw 1 only & B & intent-explicit & $+0.5192$ & $[+0.4327,+0.6150]$ & $[+0.4023,+0.6282]$ & $0.9608$ \\
\addlinespace[1pt]
Opus 5 ($k=1$) & A & main-study & $+0.2208$ & $[+0.1588,+0.2880]$ & $[+0.1648,+0.2784]$ & $0.9342$ \\
 & A & intent-explicit & $+0.4437$ & $[+0.3834,+0.5009]$ & $[+0.3934,+0.4976]$ & $0.9672$ \\
 & B & main-study & $-0.0127$ & $[-0.0743,+0.0540]$ & $[-0.0762,+0.0515]$ & $0.9241$ \\
 & B & intent-explicit & $+0.2508$ & $[+0.1770,+0.3271]$ & $[+0.1892,+0.3108]$ & $0.9753$ \\
\addlinespace[1pt]
GPT-5.6 Sol ($k=1$; Amendment 1) & A & main-study & $-0.1197$ & $[-0.1828,-0.0559]$ & $[-0.1910,-0.0473]$ & $0.7731$ \\
 & A & intent-explicit & $+0.5205$ & $[+0.4153,+0.6366]$ & $[+0.4146,+0.6375]$ & $0.9226$ \\
 & B & main-study & $+0.0465$ & $[-0.1434,+0.2313]$ & $[-0.1231,+0.2185]$ & $0.8378$ \\
 & B & intent-explicit & $+0.4050$ & $[+0.2065,+0.5793]$ & $[+0.2500,+0.5477]$ & $0.9475$ \\
\addlinespace[1pt]
\multicolumn{7}{@{}l}{\textit{Readers with no generative-model call}}\\
Name lookup & A & -- & $-0.0435$ & $[-0.0650,-0.0204]$ & $[-0.0737,-0.0148]$ & $0.7770$ \\
 & B & -- & $-0.0528$ & $[-0.0822,-0.0290]$ & $[-0.0854,-0.0241]$ & $0.7567$ \\
Keyword-intent lookup (fixed cue list) & A & -- & $+0.4041$ & $[+0.3405,+0.4736]$ & $[+0.3476,+0.4642]$ & $0.9679$ \\
 & B & -- & $+0.2561$ & $[+0.1848,+0.3352]$ & $[+0.1806,+0.3348]$ & $0.9290$ \\
Oracle-intent lookup & A & -- & $+0.4047$ & $[+0.3406,+0.4742]$ & $[+0.3486,+0.4645]$ & $0.9735$ \\
 & B & -- & $+0.2973$ & $[+0.2312,+0.3820]$ & $[+0.2184,+0.3761]$ & $0.9662$ \\
\bottomrule
\end{tabular}
\end{table}

\begin{table}[H]
\caption{The counterfactual set under the intent-explicit rubric (registered; ADDENDUM\_R122\_CF2\_P1): GPT-5.6 Luna at five draws and GPT-4.1 mini at one, each against its recorded row under the main-study rubric. Upper panel: whole set with the persona interval, and the margin over the name lookup with both intervals. Lower panel: the per-state values and the post hoc columns. Predicted before the run: each reader within $0.05$ of its recorded row (both readers, as predicted), the same per-state order (GPT-5.6 Luna as predicted; GPT-4.1 mini's two counted states swap, neither move reliable, a failed prediction), and Luna's margin over the lookup reliable (as predicted).}
\label{tab:cf-p1}
\centering\tiny\setlength{\tabcolsep}{1pt}
\begin{tabular}{@{}llrlrll@{}}
\toprule
Reader & row & whole set & persona interval & minus the lookup & item interval & persona interval \\
\midrule
GPT-5.6 Luna ($k=5$) & main-study (recorded) & $0.8417$ & $[0.7999,0.8835]$ & $+0.0754$ & $[+0.0339,+0.1216]$ & $[+0.0450,+0.1090]$ \\
 & intent-explicit & $0.8531$ & $[0.8108,0.8950]$ & $+0.0867$ & $[+0.0447,+0.1302]$ & $[+0.0562,+0.1219]$ \\
 & intent-explicit minus main-study & $+0.0113$ & & & $[-0.0052,+0.0277]$ & $[-0.0010,+0.0227]$ \\
\addlinespace[1pt]
GPT-4.1 mini ($k=1$) & main-study (recorded) & $0.7167$ & $[0.6663,0.7649]$ & $-0.0496$ & $[-0.1012,+0.0013]$ & $[-0.0933,-0.0054]$ \\
 & intent-explicit & $0.7253$ & $[0.6738,0.7731]$ & $-0.0411$ & $[-0.0878,+0.0034]$ & $[-0.0817,-0.0036]$ \\
 & intent-explicit minus main-study & $+0.0085$ & & & $[-0.0260,+0.0430]$ & $[-0.0290,+0.0417]$ \\
\addlinespace[1pt]
\bottomrule
\end{tabular}\\[4pt]
\begin{tabular}{@{}llrrrrrrr@{}}
\toprule
Reader & row & supersession & retraction & expiry & PN vs PRESENT & W/T/L & distinct values & tied triples \\
\midrule
GPT-5.6 Luna ($k=5$) & main-study (recorded) & $0.9426$ & $0.5458$ & $0.6167$ & $0.8544$ & 129/2/17 & 239 & 0 \\
 & intent-explicit & $0.9470$ & $0.5997$ & $0.6347$ & $0.8566$ & 125/1/22 & 228 & 0 \\
 & intent-explicit minus main-study & $+0.0043$ & $+0.0539$ & $+0.0181$ & & & & \\
\addlinespace[1pt]
GPT-4.1 mini ($k=1$) & main-study (recorded) & $0.8308$ & $0.5283$ & $0.5489$ & $0.7056$ & 81/57/10 & 12 & 51 \\
 & intent-explicit & $0.8591$ & $0.5456$ & $0.4758$ & $0.7120$ & 80/52/16 & 10 & 45 \\
 & intent-explicit minus main-study & $+0.0282$ & $+0.0172$ & $-0.0731$ & & & & \\
\addlinespace[1pt]
\bottomrule
\end{tabular}
\end{table}

\paragraph{Score resolution.}
The models scored from five calls carry finer score resolution than GPT-4.1 mini's single call, and the margin over the lookup does not depend on it.
Opus 5's first draw alone reads $0.8400$ on this set ($53$ distinct values against $264$ after averaging), $+0.0736$ above the name lookup with intervals $[+0.0298,+0.1223]$ and $[+0.0388,+0.1129]$, and its five draws span $0.0059$; Haiku 4.5's first draw reads $0.7555$ ($17$ values), $-0.0108$ against the lookup with both intervals including zero; GPT-4.1 mini's single draw takes $12$ values.
At five draws with the provider's default temperature GPT-4.1 mini reads $0.76$ under the superseded-only summary and $0.83$ under the all-states summary, within $0.05$ of its one-draw rows, with the tied triples falling from $51$ to $6$ (Table~\ref{tab:mini-k5}).
Table~\ref{tab:cf-main} prints those rows.
At five draws the model sits within tolerance of the name lookup on the whole set ($-0.0016$, both intervals including zero) and on supersession ($-0.0429$, intervals $[-0.0898,+0.0006]$ and $[-0.1001,+0.0035]$), where its one-draw row is reliably below (Table~\ref{tab:sup-margins}).
On the full obligation set the first draws read $0.8156$ (Haiku 4.5), $0.8335$ (Opus 5) and $0.8141$ (GPT-5.6 Luna) against averaged rows of $0.8210$, $0.8327$ and $0.8182$.
GPT-4.1 mini's repeated calls return identical scores on $46$ of $50$ questions.

\begin{table}[H]
\caption{GPT-4.1 mini at five draws on the counterfactual set (registered; ADDENDUM\_R122\_MINI\_K5), under the default and the all-states summary, beside its recorded one-draw rows at temperature $0$. Upper panel: whole set with the persona interval, and the margin over the same summary's lookup with both intervals. Lower panel: per state, and the post hoc columns. The five-draw run used the provider's default temperature, so two things changed between the rows. Predicted before the run: the five-draw row within $0.05$ of the one-draw row and fewer tied triples, under both summaries (as predicted). Table~\ref{tab:cf-main} prints the five-draw rows; the names-against-counts comparison of Table~\ref{tab:cf-names-counts} keeps the one-draw row as its comparator.}
\label{tab:mini-k5}
\centering\tiny\setlength{\tabcolsep}{1pt}
\begin{tabular}{@{}llrlrll@{}}
\toprule
Reader & row & whole set & persona interval & minus the lookup & item interval & persona interval \\
\midrule
default (superseded-only) & one draw, temperature $0$ (recorded) & $0.7167$ & $[0.6663,0.7649]$ & $-0.0496$ & $[-0.1012,+0.0013]$ & $[-0.0933,-0.0054]$ \\
 & five draws, provider default & $0.7648$ & $[0.7218,0.8047]$ & $-0.0016$ & $[-0.0426,+0.0393]$ & $[-0.0415,+0.0328]$ \\
 & five minus one & $+0.0481$ & & & $[+0.0119,+0.0857]$ & $[+0.0171,+0.0827]$ \\
\addlinespace[1pt]
all-states & one draw, temperature $0$ (recorded) & $0.7967$ & $[0.7438,0.8478]$ & $-0.0789$ & $[-0.1229,-0.0344]$ & $[-0.1283,-0.0334]$ \\
 & five draws, provider default & $0.8323$ & $[0.7889,0.8725]$ & $-0.0433$ & $[-0.0879,-0.0035]$ & $[-0.0838,-0.0042]$ \\
 & five minus one & $+0.0356$ & & & $[+0.0094,+0.0619]$ & $[+0.0092,+0.0688]$ \\
\addlinespace[1pt]
\bottomrule
\end{tabular}\\[4pt]
\begin{tabular}{@{}llrrrrrrr@{}}
\toprule
Reader & row & supersession & retraction & expiry & PN vs PRESENT & W/T/L & distinct values & tied triples \\
\midrule
default (superseded-only) & one draw, temperature $0$ (recorded) & $0.8308$ & $0.5283$ & $0.5489$ & $0.7056$ & 81/57/10 & 12 & 51 \\
 & five draws, provider default & $0.8793$ & $0.5761$ & $0.5594$ & $0.7394$ & 100/15/33 & 74 & 6 \\
 & five minus one & $+0.0485$ & $+0.0478$ & $+0.0106$ & & & & \\
\addlinespace[1pt]
all-states & one draw, temperature $0$ (recorded) & $0.8444$ & $0.8725$ & $0.5869$ & $0.7840$ & 105/32/11 & 12 & 26 \\
 & five draws, provider default & $0.8829$ & $0.8433$ & $0.6719$ & $0.8307$ & 130/5/13 & 78 & 1 \\
 & five minus one & $+0.0385$ & $-0.0292$ & $+0.0850$ & & & & \\
\addlinespace[1pt]
\bottomrule
\end{tabular}
\end{table}

The registered texts for this set (Appendix~\ref{app:registration}) read as follows.
On $148$ questions written before any version of the store existed, and asked of three versions of the same store that differ in one attribute each, the smallest model's obligation judgment reaches $0.7167$ against the $0.5$ a question-only reader attains here, with an interval of $[+0.1494,+0.2846]$. The gain survives where the question cannot carry the label.
Naming retracted and expired content lifts Haiku 4.5's judgment on those states from $0.5638$ to $0.8649$ on the store-counterfactual set ($60$ questions), a paired gain of $+0.3010$ with interval $[+0.2268,+0.3721]$.

The adverse rows are reported in full, because a row that runs against the paper's reading is evidence too.
The all-states lookup exceeds GPT-4.1 mini on the whole set under every summary ($+0.1589$, $+0.0789$, $+0.1577$), each beyond the $\pm0.05$ tolerance.
It exceeds Haiku 4.5 under the two summaries that do not name the counted states ($+0.1190$, $+0.1118$), also beyond tolerance, and Opus 5 under those two summaries within it ($+0.0351$, $+0.0403$).
The name lookup exceeds GPT-4.1 mini and Haiku 4.5 under the superseded-only summary within tolerance ($+0.0496$, $+0.0097$).
By state, the name lookup ($0.9223$ on supersession) exceeds GPT-4.1 mini and Haiku 4.5 under every summary and trails Opus 5.
The all-states lookup ($0.7975$ on retraction, $0.8711$ on expiry) exceeds every reader on the two counted states under the summaries that only count them.
The whole name-map lookup ($0.5758$ on retraction, $0.5919$ on expiry) exceeds GPT-4.1 mini on both counted states under every summary except retraction under the all-three summary.
That is what a reader of the whole map should do when the counted names sit in that map.
Table~\ref{tab:sup-margins} gives each model's supersession-only margin over the name lookup with both intervals.
Under the all-states summary Opus 5's margin over the all-states lookup is reliable on the whole set ($+0.0323$) but on no single state: supersession $+0.0324$, retraction $+0.0125$ and expiry $+0.0556$, each with an interval that includes zero.

\begin{table}[H]
\caption{Each model's supersession-only margin over the name lookup on the counterfactual set under the default summary ($88$ pairs, $264$ items; the lookup reads $0.9223$ there): the model's whole-set and supersession AUROC, the paired difference from the lookup with the item-level and the persona-grouped interval, and whether both intervals exclude zero. Descriptive; computed on the recorded rows after the fact. At one draw GPT-4.1 mini is reliably below the lookup; at five draws its interval includes zero; no model is reliably above it.}
\label{tab:sup-margins}
\centering\tiny\setlength{\tabcolsep}{1pt}
\begin{tabular}{@{}lrrrlll@{}}
\toprule
Model & whole set & supersession & minus the lookup & item interval & persona interval & both exclude zero \\
\midrule
GPT-4.1 mini, one draw (recorded) & $0.7167$ & $0.8308$ & $-0.0915$ & $[-0.1399,-0.0478]$ & $[-0.1436,-0.0425]$ & yes \\
GPT-4.1 mini, five draws (provider default) & $0.7648$ & $0.8793$ & $-0.0429$ & $[-0.0898,+0.0006]$ & $[-0.1001,+0.0035]$ & no \\
Haiku 4.5 (five draws) & $0.7566$ & $0.8876$ & $-0.0347$ & $[-0.0813,+0.0086]$ & $[-0.0809,+0.0066]$ & no \\
Opus 5 (five draws) & $0.8405$ & $0.9325$ & $+0.0102$ & $[-0.0183,+0.0392]$ & $[-0.0259,+0.0432]$ & no \\
GPT-5.6 Sol (five draws) & $0.8500$ & $0.9383$ & $+0.0161$ & $[-0.0130,+0.0473]$ & $[-0.0202,+0.0478]$ & no \\
GPT-5.6 Luna (five draws) & $0.8417$ & $0.9426$ & $+0.0204$ & $[-0.0053,+0.0473]$ & $[-0.0056,+0.0451]$ & no \\
\bottomrule
\end{tabular}
\end{table}

The earlier draw of this set asked thirty questions ($n=90$, ten positives per state) and reproduced the naming boundary.
Under the summary that names only superseded content the named state scored $0.6550$ to $0.8275$ across the three models, while the two counted states scored $0.5225$ to $0.6225$.
At that size GPT-4.1 mini reached $0.5939$ against $0.5000$, with an item-level interval of $[-0.0345,+0.2315]$ that includes chance and a question-clustered interval of $[+0.0325,+0.1625]$ that excludes it.
The set could therefore neither establish the aggregate effect nor establish contamination, and the $444$-item form was built with the power the first lacked.
On that form the all-states lookup exceeded three monitor rows by more than the tolerance: GPT-4.1 mini by $+0.1947$ under the first summary and $+0.1467$ under the second, and Haiku 4.5 by $+0.1480$ under the first.
The name lookup exceeded GPT-4.1 mini by $+0.0322$, inside it.
The struck clause of that form's failure text is in the supplement (Section~S.6).
Of the $204$ questions drafted for the set and its replacements, $111$ name the plan or value their attribute is about, against $3$ of $30$ in the earlier draw.
A question is identical across the three versions of its store, so what it names cannot carry the label, and Appendix~\ref{app:posthoc-flags} splits the set by that flag.
The summary-leakage screen returned $0$ of $30$ on the all-states summary and $0$ of $30$ on the totals-randomized one.
The draw replaced $45$ pairs whose successor record failed our read and closed at $88$ supersession pairs (Table~\ref{tab:registrations-b}).

\begin{table}[h]
\caption{The store-counterfactual set: $148$ questions, each asked of three versions of one store differing in a single attribute, so $n=444$ with $148$ positives ($88$ supersession, $30$ retraction, $30$ expiry). The question is fixed and the store varies, so a question-only reader scores $0.5000$ by construction and every gain is measured against that. Three summaries: naming superseded content only, naming all three non-active states, and the first with its two lifecycle totals, retracted and expired, permuted across store versions (the per-category counts untouched). Intervals are the item-level paired bootstrap and the bootstrap with questions grouped by persona, both at $95\%$; a prediction holds only when both agree; the first form's gains are against the same $0.5000$. The first form of the set ($30$ questions, $n=90$) is reported beside the new values as a first-form column or row; it had no totals-randomized summary and its counted-state block used the $45$ items whose attribute name identified the pair, a predicate the new draw applies before any pair is drawn. A dash marks an entry the design does not define: a lookup reads one summary and has no paired interval, and the first form has no totals-randomized value. Adverse rows are stated in the text above.}
\label{tab:counterfactual}
\centering
\scriptsize
\setlength{\tabcolsep}{1.5pt}
\begin{tabular}{@{}lrrrrr@{}}
\toprule
\multicolumn{6}{@{}l}{\textit{Whole set, $n=444$; blind $0.5000$ by construction; gain is AUROC less $0.5000$}}\\
Reader, summary & AUROC & item interval & persona interval & first draw & first-draw gain \\
\midrule
GPT-4.1 mini, sup.\ only & $0.7167$ & $[+0.1494,+0.2846]$ & $[+0.1663,+0.2649]$ & $0.5939$ & $+0.0939$ \\
GPT-4.1 mini, all three & $0.7967$ & $[+0.2314,+0.3568]$ & $[+0.2438,+0.3478]$ & $0.6419$ & $+0.1419$ \\
GPT-4.1 mini, totals rand.\ & $0.7179$ & $[+0.1479,+0.2857]$ & $[+0.1691,+0.2638]$ & --- & \\
Haiku 4.5, sup.\ only & $0.7566$ & $[+0.1942,+0.3189]$ & $[+0.2102,+0.3035]$ & $0.6406$ & $+0.1406$ \\
Haiku 4.5, all three & $0.8773$ & $[+0.3174,+0.4395]$ & $[+0.3362,+0.4124]$ & $0.7656$ & $+0.2656$ \\
Haiku 4.5, totals rand.\ & $0.7638$ & $[+0.2003,+0.3254]$ & $[+0.2193,+0.3071]$ & --- & \\
Opus 5, sup.\ only & $0.8405$ & $[+0.2882,+0.3913]$ & $[+0.2996,+0.3817]$ & $0.7122$ & $+0.2122$ \\
Opus 5, all three & $0.9079$ & $[+0.3508,+0.4628]$ & $[+0.3769,+0.4377]$ & $0.8172$ & $+0.3172$ \\
Opus 5, totals rand.\ & $0.8353$ & $[+0.2825,+0.3870]$ & $[+0.2937,+0.3768]$ & --- & \\
\addlinespace
\multicolumn{6}{@{}l}{\textit{Lookups, whole set, each on its own summary; persona interval of the AUROC}}\\
Name lookup (sup.\ only) & $0.7663$ & --- & $[0.7226,0.8051]$ & $0.6261$ & \\
All-states lookup (all three) & $\mathbf{0.8756}$ & --- & $[0.8391,0.9111]$ & $0.7886$ & \\
Whole name-map lookup (sup.\ only) & $0.5776$ & --- & $[0.5587,0.5981]$ & $0.5347$ & \\
\addlinespace
\multicolumn{6}{@{}l}{\textit{By lifecycle state ($88$ / $30$ / $30$ positives; first form ten each)}}\\
Reader, state & superseded only & all three & totals rand. & first: sup. & first: all \\
GPT-4.1 mini, supersession & $0.8308$ & $0.8444$ & $0.8500$ & $0.6550$ & $0.6950$ \\
GPT-4.1 mini, retraction & $0.5283$ & $0.8725$ & $0.4950$ & $0.5700$ & $0.7500$ \\
GPT-4.1 mini, expiry & $0.5489$ & $0.5869$ & $0.5144$ & $0.5825$ & $0.4800$ \\
Haiku 4.5, supersession & $0.8876$ & $0.8852$ & $0.8874$ & $0.7825$ & $0.7450$ \\
Haiku 4.5, retraction & $0.5433$ & $0.9042$ & $0.5844$ & $0.6225$ & $0.7375$ \\
Haiku 4.5, expiry & $0.5794$ & $0.8269$ & $0.5847$ & $0.5225$ & $0.8575$ \\
Opus 5, supersession & $0.9325$ & $0.9319$ & $0.9322$ & $0.8275$ & $0.8350$ \\
Opus 5, retraction & $0.5872$ & $0.8100$ & $0.5681$ & $0.5975$ & $0.7300$ \\
Opus 5, expiry & $0.6150$ & $0.9267$ & $0.6092$ & $0.6150$ & $0.9450$ \\
Lookup, own summary & supersession & retraction & expiry & & \\
Name lookup & $0.9223$ & $0.5000$ & $0.5000$ & --- & --- \\
All-states names lookup & $0.8995$ & $0.7975$ & $0.8711$ & --- & --- \\
Whole name-map lookup & $0.5728$ & $0.5758$ & $0.5919$ & --- & --- \\
\addlinespace
\multicolumn{6}{@{}l}{\textit{Counted states pooled (retraction, expiry): $60$ questions, $n=180$; first form $15$, $n=45$}}\\
Reader & paired gain & item interval & persona interval & superseded only & all three \\
GPT-4.1 mini & $+0.1854$ & $[+0.1007,+0.2689]$ & $[+0.1000,+0.2734]$ & $0.5360$ & $0.7214$ \\
Haiku 4.5 & $+0.3010$ & $[+0.2268,+0.3721]$ & $[+0.2247,+0.3591]$ & $0.5638$ & $\mathbf{0.8649}$ \\
Opus 5 & $+0.2506$ & $[+0.1784,+0.3349]$ & $[+0.1941,+0.3019]$ & $0.6030$ & $0.8536$ \\
GPT-4.1 mini, first form & --- & --- & --- & $0.5578$ & $0.6600$ \\
Haiku 4.5, first form & --- & $[+0.0772,+0.4945]$ & --- & $0.5656$ & $0.8489$ \\
Opus 5, first form & --- & --- & --- & $0.6122$ & $0.9144$ \\
\addlinespace
\multicolumn{6}{@{}l}{\textit{Named minus counted, superseded-only summary (two-subset bootstrap)}}\\
Reader & difference & two-subset interval & persona interval & named & counted \\
GPT-4.1 mini & $+0.2948$ & $[+0.1875,+0.3905]$ & $[+0.2099,+0.3838]$ & $0.8308$ & $0.5360$ \\
Haiku 4.5 & $\mathbf{+0.3237}$ & $[+0.2319,+0.4011]$ & $[+0.2612,+0.3851]$ & $0.8876$ & $0.5638$ \\
Opus 5 & $+0.3295$ & $[+0.2420,+0.4181]$ & $[+0.2556,+0.3875]$ & $0.9325$ & $0.6030$ \\
\addlinespace
\multicolumn{6}{@{}l}{\textit{Totals randomized minus superseded only, whole set (band $\pm0.05$)}}\\
Reader & difference & item interval & persona interval & & \\
GPT-4.1 mini & $+0.0011$ & $[-0.0282,+0.0297]$ & $[-0.0210,+0.0230]$ & & \\
Haiku 4.5 & $+0.0072$ & $[-0.0079,+0.0229]$ & $[-0.0103,+0.0248]$ & & \\
Opus 5 & $-0.0052$ & $[-0.0130,+0.0026]$ & $[-0.0122,+0.0020]$ & & \\
\addlinespace
\multicolumn{6}{@{}l}{\textit{Model minus lookup on the same summary, whole set (paired bootstrap; computed after the fact)}}\\
Reader, summary & difference & item interval & persona interval & model & lookup \\
GPT-4.1 mini, superseded only & $-0.0496$ & $[-0.1012,+0.0013]$ & $[-0.0933,-0.0054]$ & $0.7167$ & $0.7663$ \\
GPT-4.1 mini, all three & $-0.0789$ & $[-0.1229,-0.0344]$ & $[-0.1283,-0.0334]$ & $0.7967$ & $0.8756$ \\
Haiku 4.5, superseded only & $-0.0098$ & $[-0.0640,+0.0398]$ & $[-0.0494,+0.0336]$ & $0.7566$ & $0.7663$ \\
Haiku 4.5, all three & $+0.0017$ & $[-0.0326,+0.0361]$ & $[-0.0312,+0.0320]$ & $0.8773$ & $0.8756$ \\
Opus 5, superseded only & $+0.0742$ & $[+0.0315,+0.1203]$ & $[+0.0414,+0.1126]$ & $0.8405$ & $0.7663$ \\
Opus 5, all three & $+0.0323$ & $[+0.0060,+0.0617]$ & $[+0.0077,+0.0608]$ & $0.9079$ & $0.8756$ \\
GPT-5.6 Sol, superseded only & $+0.0836$ & $[+0.0401,+0.1315]$ & $[+0.0544,+0.1163]$ & $0.8500$ & $0.7663$ \\
GPT-5.6 Luna, superseded only & $+0.0754$ & $[+0.0339,+0.1216]$ & $[+0.0450,+0.1090]$ & $0.8417$ & $0.7663$ \\
\bottomrule
\end{tabular}
\end{table}
\FloatBarrier

\paragraph{Coverage on the counterfactual set.}
The coverage score was recorded in the same calls, so the set also measures coverage under the same identity.
One blind score per question, reused across its three versions, gives blind FOK $0.5000$ by construction, and the positives are the \present{} and \pn{} versions ($296$) against the \absent{} version ($148$).
Table~\ref{tab:cf-fok} gives each model's coverage AUROC under the three summaries with the interval grouped by persona, and every interval excludes $0.50$.
Opus 5, whose scores take about $200$ distinct values, separates the versions more than the two smaller models, whose scores crowd a dozen values.

\begin{table}[H]
\caption{Coverage AUROC on the counterfactual set ($n=444$; positives the \present{} and \pn{} versions, $296$) from the recorded coverage scores, by model and summary, with the interval grouped by persona and the mean score by store version. Blind coverage is $0.5000$ by construction, so the gain is the AUROC minus $0.5000$. Descriptive; no prediction was registered on these rows.}
\label{tab:cf-fok}
\centering
\scriptsize
\setlength{\tabcolsep}{3pt}
\begin{tabular}{@{}llrlrrr@{}}
\toprule
Model & Summary & AUROC & persona interval & mean, \present{} & mean, \absent{} & mean, \pn{} \\
\midrule
GPT-4.1 mini & sup.\ only & $0.6357$ & $[0.5972,0.6770]$ & $0.4517$ & $0.3530$ & $0.4966$ \\
GPT-4.1 mini & all three & $0.6436$ & $[0.6120,0.6802]$ & $0.4926$ & $0.3875$ & $0.5439$ \\
GPT-4.1 mini & totals rand. & $0.6371$ & $[0.6071,0.6709]$ & $0.4551$ & $0.3503$ & $0.4882$ \\
Haiku 4.5 & sup.\ only & $0.6168$ & $[0.5828,0.6537]$ & $0.5130$ & $0.4178$ & $0.5263$ \\
Haiku 4.5 & all three & $0.6133$ & $[0.5816,0.6522]$ & $0.5055$ & $0.4163$ & $0.5180$ \\
Haiku 4.5 & totals rand. & $0.6050$ & $[0.5713,0.6416]$ & $0.5072$ & $0.4231$ & $0.5180$ \\
Opus 5 & sup.\ only & $0.6922$ & $[0.6576,0.7353]$ & $0.7746$ & $0.6604$ & $0.8369$ \\
Opus 5 & all three & $0.7098$ & $[0.6763,0.7535]$ & $0.7818$ & $0.6637$ & $0.8567$ \\
Opus 5 & totals rand. & $0.6894$ & $[0.6572,0.7299]$ & $0.7741$ & $0.6633$ & $0.8386$ \\
\bottomrule
\end{tabular}
\end{table}

\paragraph{The lexical lookup on this set.}
The similarity lookup of the main text is the embedding variant.
The lexical variant, token overlap between the question and the names, reads the same lists (Table~\ref{tab:lex-444}).
It reaches $0.6698$ on the whole set under the superseded-only summary and $0.8085$ under the all-states summary, below the embedding variant by $0.10$ and $0.07$.
Under the superseded-only summary every model exceeds it with intervals that exclude zero, so the lexical reader is the weaker of the two, and the main text's comparisons use the stronger.

\begin{table}[H]
\caption{The lexical lookup on the counterfactual set beside the embedding lookup, whole set and by state, and each model minus the lexical lookup on the same summary with both intervals. The lexical reader trails the embedding reader on every row. Exploratory, computed after the fact on the recorded rows.}
\label{tab:lex-444}
\centering
\scriptsize
\setlength{\tabcolsep}{3pt}
\begin{tabular}{@{}lrrrr@{}}
\toprule
Lookup, summary & whole set & supersession & retraction & expiry \\
\midrule
Lexical, sup.\ only & $0.6698$ & $0.7829$ & $0.5000$ & $0.5000$ \\
Lexical, all three & $0.8085$ & $0.7829$ & $0.8617$ & $0.8244$ \\
Embedding, sup.\ only & $0.7663$ & $0.9223$ & $0.5000$ & $0.5000$ \\
Embedding, all three & $0.8756$ & $0.8995$ & $0.7975$ & $0.8711$ \\
\addlinespace
\multicolumn{5}{@{}l}{\textit{Model minus the lexical lookup on the same summary (paired bootstrap): difference, item interval, persona interval}}\\
GPT-4.1 mini, sup.\ only & $+0.0469$ & $[+0.0037,+0.0925]$ & $[+0.0037,+0.0895]$ & \\
GPT-4.1 mini, all three & $-0.0117$ & $[-0.0591,+0.0360]$ & $[-0.0566,+0.0294]$ & \\
Haiku 4.5, sup.\ only & $+0.0868$ & $[+0.0405,+0.1299]$ & $[+0.0476,+0.1274]$ & \\
Haiku 4.5, all three & $+0.0689$ & $[+0.0283,+0.1087]$ & $[+0.0321,+0.1046]$ & \\
Opus 5, sup.\ only & $+0.1707$ & $[+0.1314,+0.2102]$ & $[+0.1317,+0.2118]$ & \\
Opus 5, all three & $+0.0994$ & $[+0.0602,+0.1409]$ & $[+0.0545,+0.1451]$ & \\
\bottomrule
\end{tabular}
\end{table}

\paragraph{The lookup under other encoders.}
Two further encoders, \texttt{all-mpnet-base-v2} and \texttt{bge-small-en-v1.5}, score the same lookup at $0.76$ and $0.76$ on the counterfactual set, within $0.01$ of MiniLM; replacing every name by a random token scores $0.60$ (Table~\ref{tab:enc-lookup}).

\begin{table}[H]
\caption{The name lookup under two further sentence encoders, and the random-token control (registered; ADDENDUM\_R122\_ENCODERS): the same lookup code with the encoder swapped, on the counterfactual set under the default and the all-states summary and on the full obligation set, AUROC with the persona interval where recorded and the difference from the recorded \texttt{all-MiniLM-L6-v2} lookup with both intervals. The full-obligation-set rows as committed read the summaries where an empty section's \texttt{(none)} marker is encoded as a name; the marker-free rows drop the marker and are the fair cross-encoder comparison there. The control replaces every name by a random token, so what it scores comes from the list's length and from whether it is empty. Predicted before the run: both encoders within $0.05$ of MiniLM on the counterfactual set under the default summary (as predicted).}
\label{tab:enc-lookup}
\centering\tiny\setlength{\tabcolsep}{1pt}
\begin{tabular}{@{}llrlrll@{}}
\toprule
Encoder & population, summary & AUROC & persona interval & minus MiniLM & item interval & persona interval \\
\midrule
\texttt{all-MiniLM-L6-v2} (recorded) & $444$, default & $0.7663$ & $[0.7226,0.8051]$ & -- & -- & -- \\
 & $444$, all-states & $0.8756$ & $[0.8391,0.9111]$ & -- & -- & -- \\
 & $661$, committed & $0.7484$ & $[0.6636,0.8298]$ & -- & -- & -- \\
 & $661$, marker-free & $0.7442$ & $[0.6513,0.8336]$ & -- & -- & -- \\
\addlinespace[1pt]
\texttt{all-mpnet-base-v2} & $444$, default & $0.7605$ & $[0.7140,0.8024]$ & $-0.0059$ & $[-0.0240,+0.0126]$ & $[-0.0224,+0.0093]$ \\
 & $444$, all-states & $0.8646$ & $[0.8296,0.9018]$ & $-0.0110$ & $[-0.0296,+0.0063]$ & $[-0.0257,+0.0034]$ \\
 & $661$, committed & $0.7642$ & $[0.6705,0.8573]$ & $+0.0158$ & $[-0.0275,+0.0585]$ & $[-0.0469,+0.0686]$ \\
 & $661$, marker-free & $0.7642$ & $[0.6574,0.8619]$ & $+0.0200$ & $[-0.0037,+0.0426]$ & $[-0.0095,+0.0507]$ \\
\addlinespace[1pt]
\texttt{bge-small-en-v1.5} & $444$, default & $0.7584$ & $[0.7166,0.7989]$ & $-0.0080$ & $[-0.0298,+0.0143]$ & $[-0.0244,+0.0098]$ \\
 & $444$, all-states & $0.8690$ & $[0.8326,0.9076]$ & $-0.0066$ & $[-0.0254,+0.0101]$ & $[-0.0259,+0.0134]$ \\
 & $661$, committed & $0.8133$ & $[0.7452,0.8856]$ & $+0.0649$ & $[+0.0292,+0.1026]$ & $[+0.0101,+0.1297]$ \\
 & $661$, marker-free & $0.7620$ & $[0.6595,0.8519]$ & $+0.0178$ & $[-0.0014,+0.0392]$ & $[-0.0170,+0.0530]$ \\
\addlinespace[1pt]
\multicolumn{7}{@{}l}{\textit{Random tokens in place of every name}}\\
\texttt{all-MiniLM-L6-v2} & $444$, default & $0.5954$ & $[0.5601,0.6383]$ & -- & -- & -- \\
 & $444$, all-states & $0.5479$ & $[0.5260,0.5740]$ & -- & -- & -- \\
 & $661$, committed & $0.6654$ & $[0.5874,0.7417]$ & -- & -- & -- \\
 & $661$, marker-free & $0.6627$ & $[0.5798,0.7393]$ & -- & -- & -- \\
\addlinespace[1pt]
\texttt{all-mpnet-base-v2} & $444$, default & $0.5977$ & $[0.5620,0.6365]$ & -- & -- & -- \\
 & $444$, all-states & $0.5363$ & $[0.5203,0.5555]$ & -- & -- & -- \\
 & $661$, committed & $0.7150$ & $[0.6475,0.7854]$ & -- & -- & -- \\
 & $661$, marker-free & $0.7003$ & $[0.6225,0.7763]$ & -- & -- & -- \\
\addlinespace[1pt]
\texttt{bge-small-en-v1.5} & $444$, default & $0.5948$ & $[0.5580,0.6396]$ & -- & -- & -- \\
 & $444$, all-states & $0.5452$ & $[0.5262,0.5701]$ & -- & -- & -- \\
 & $661$, committed & $0.6989$ & $[0.6228,0.7666]$ & -- & -- & -- \\
 & $661$, marker-free & $0.6437$ & $[0.5681,0.7174]$ & -- & -- & -- \\
\addlinespace[1pt]
\bottomrule
\end{tabular}
\end{table}

\paragraph{A hybrid reader with no model in it.}
A logistic regression of the label on the name lookup's cosine, the category-guessing reader's share and an indicator that a category was assigned, fitted under leave-persona-out and scored out of fold, reads $0.7865$ on the whole set under the superseded-only summary.
That is $+0.0201$ above the name lookup, with the item-level interval $[-0.0016,+0.0410]$ including zero and the persona interval $[+0.0039,+0.0409]$ excluding it, so the registered prediction that the combination exceeds the lookup reliably failed.
GPT-5.6 Sol stays above it by $+0.0635$ with both intervals excluding zero, as predicted; Opus 5 and GPT-5.6 Luna are reliably above it as well, Haiku 4.5 sits within tolerance of it, and GPT-4.1 mini is reliably below (Table~\ref{tab:cf-hybrid}).
The combination gains on the two counted states ($+0.0889$ on retraction and $+0.0942$ on expiry over the lookup's $0.5000$) and reads $-0.0153$ against the lookup on supersession, where the lookup alone scores $0.9223$.
Under the matched metric the same reader reads $0.8851$ ($114/34/0$), $+0.1081$ above the name lookup with intervals $[+0.0777,+0.1419]$ and $[+0.0741,+0.1419]$, and the three stronger models are level with it at the point: GPT-5.6 Sol $-0.0270$, GPT-5.6 Luna $-0.0068$ and Opus 5 $-0.0101$, each with both intervals including zero, while Haiku 4.5 ($-0.1081$) and GPT-4.1 mini at five draws ($-0.1588$) are reliably below it.
So the two registered predictions, read under the pooled metric, reverse under the matched one, which was defined afterwards: the combination is reliably above the lookup within triples, and Sol is not reliably above the combination.
Table~\ref{tab:cf-classes} shows where the pooled metric's ordering comes from: the hybrid's within-triple and same-state values are close to the three models', and its across-state value is not.

\begin{table}[H]
\caption{A hybrid reader with no generative-model call on the counterfactual set (registered; ADDENDUM\_CF2\_HYBRID\_READER): a logistic regression of the label on the name lookup's cosine, the category-guessing reader's share and an indicator that a category was assigned, fitted under leave-persona-out over the $40$ personas and scored out of fold, under the superseded-only summary. Upper panel: whole set and per state with the persona interval, and the paired differences against the name lookup and the category-guessing reader with both intervals. Lower panels: each model's recorded row minus the hybrid, and GPT-4.1 mini's five-draw row (Table~\ref{tab:cf-main}) beside its one-draw row; then the same reader under the matched metric (post hoc), whole set and per state, minus the lookup and against each model. Predicted before the run: that the hybrid exceeds the name lookup reliably (it does not: the item-level interval includes zero, a failed prediction) and that GPT-5.6 Sol stays reliably above it (as predicted). The fitted coefficients on the standardized features range over the folds from $+1.13$ to $+1.29$ on the lookup and $+0.39$ to $+0.53$ on the share.}
\label{tab:cf-hybrid}
\centering\tiny\setlength{\tabcolsep}{2pt}
\begin{tabular}{@{}lrlrll@{}}
\toprule
Scope & hybrid & persona interval & minus the lookup & item interval & persona interval \\
\midrule
whole set & $0.7865$ & $[0.7488,0.8234]$ & $+0.0201$ & $[-0.0016,+0.0410]$ & $[+0.0039,+0.0409]$ \\
supersession & $0.9070$ & -- & $-0.0153$ & $[-0.0293,-0.0029]$ & $[-0.0306,-0.0018]$ \\
retraction & $0.5889$ & -- & $+0.0889$ & $[+0.0120,+0.1673]$ & $[+0.0477,+0.1483]$ \\
expiry & $0.5942$ & -- & $+0.0942$ & $[+0.0165,+0.1717]$ & $[+0.0517,+0.1635]$ \\
whole set, minus the category-guessing reader & & & $+0.1242$ & $[+0.0668,+0.1816]$ & $[+0.0825,+0.1656]$ \\
\bottomrule
\end{tabular}\\[4pt]
\begin{tabular}{@{}lrrll@{}}
\toprule
Model (recorded row) & model & minus the hybrid & item interval & persona interval \\
\midrule
GPT-4.1 mini & $0.7167$ & $-0.0698$ & $[-0.1187,-0.0192]$ & $[-0.1094,-0.0316]$ \\
GPT-4.1 mini, five draws & $0.7648$ & $-0.0217$ & $[-0.0621,+0.0186]$ & $[-0.0587,+0.0067]$ \\
Haiku 4.5 & $0.7566$ & $-0.0299$ & $[-0.0862,+0.0210]$ & $[-0.0694,+0.0093]$ \\
Opus 5 & $0.8405$ & $+0.0540$ & $[+0.0061,+0.1004]$ & $[+0.0188,+0.0888]$ \\
GPT-5.6 Sol & $0.8500$ & $+0.0635$ & $[+0.0214,+0.1108]$ & $[+0.0346,+0.0913]$ \\
GPT-5.6 Luna & $0.8417$ & $+0.0553$ & $[+0.0169,+0.0987]$ & $[+0.0245,+0.0864]$ \\
\bottomrule
\end{tabular}%
\\[4pt]
\begin{tabular}{@{}lrrll@{}}
\toprule
Under the matched metric (post hoc) & W/T/L & matched & triple interval & persona interval \\
\midrule
whole set & $114/34/0$ & $0.8851$ & $[0.8514,0.9189]$ & $[0.8504,0.9167]$ \\
supersession & $83/5/0$ & $0.9716$ & $[0.9489,0.9943]$ & $[0.9323,1.0000]$ \\
retraction & $16/14/0$ & $0.7667$ & $[0.6833,0.8500]$ & $[0.6897,0.8519]$ \\
expiry & $15/15/0$ & $0.7500$ & $[0.6667,0.8333]$ & $[0.6613,0.8261]$ \\
\bottomrule
\end{tabular}\\[4pt]
\begin{tabular}{@{}lrll@{}}
\toprule
Hybrid minus the lookup, matched & difference & triple interval & persona interval \\
\midrule
whole set & $+0.1081$$^{*}$ & $[+0.0777,+0.1419]$ & $[+0.0741,+0.1419]$ \\
supersession & $+0.0057$ & $[+0.0000,+0.0170]$ & $[+0.0000,+0.0179]$ \\
retraction & $+0.2667$$^{*}$ & $[+0.1833,+0.3500]$ & $[+0.1897,+0.3519]$ \\
expiry & $+0.2500$$^{*}$ & $[+0.1667,+0.3333]$ & $[+0.1613,+0.3261]$ \\
\bottomrule
\end{tabular}\\[4pt]
\begin{tabular}{@{}lrrll@{}}
\toprule
Model minus the hybrid, matched & model & difference & triple interval & persona interval \\
\midrule
GPT-4.1 mini, one draw & $0.7399$ & $-0.1453$$^{*}$ & $[-0.1993,-0.0946]$ & $[-0.1985,-0.0871]$ \\
GPT-4.1 mini, five draws & $0.7264$ & $-0.1588$$^{*}$ & $[-0.2264,-0.0979]$ & $[-0.2244,-0.0965]$ \\
Haiku 4.5 & $0.7770$ & $-0.1081$$^{*}$ & $[-0.1555,-0.0642]$ & $[-0.1644,-0.0479]$ \\
Opus 5 & $0.8750$ & $-0.0101$ & $[-0.0676,+0.0439]$ & $[-0.0733,+0.0563]$ \\
GPT-5.6 Sol & $0.8581$ & $-0.0270$ & $[-0.0812,+0.0203]$ & $[-0.0737,+0.0175]$ \\
GPT-5.6 Luna & $0.8784$ & $-0.0068$ & $[-0.0642,+0.0507]$ & $[-0.0659,+0.0536]$ \\
\bottomrule
\end{tabular}
\end{table}

\paragraph{The sibling-name control.}
The matched comparison still lets a reader profit from list length, since the \pn{} version's superseded-names list is one name longer than its siblings' and $24$ of the $88$ \present{} lists are empty (Table~\ref{tab:cf-emptiness}).
The sibling-name control removes that: a fourth store version per supersession pair carries the \pn{} version's text with the target's name in the superseded-names list replaced by another attribute of the same category, so the two versions differ only in which name is listed (Table~\ref{tab:cf-sibling}).
The name lookup separates the \pn{} version from it at $0.9489$ ($82/3/3$), and the five models at $0.8807$ to $0.9716$, none reliably above the lookup (Opus 5 $+0.0227$ with intervals $[-0.0170,+0.0682]$ and $[-0.0114,+0.0632]$); the random-token lookup falls to $0.4716$ and the count reader ties every pair.
An unrelated name in the list does raise the scores: the sibling version scores above the \present{} version for every reader but Opus 5 (the sibling-minus-\present{} mean for Haiku 4.5 $+0.0859$, both intervals clear of zero; Opus 5 $+0.0376$, intervals including zero), which is the list-length component the primary comparison removes.

\begin{table}[H]
\caption{The sibling-name control (registered; ADDENDUM\_R125\_SIBLING): a fourth version of each of the $88$ supersession pairs, rendered from the \pn{} version's text with the target's name in the superseded-names list replaced by a same-category sibling under HV-B's pool and neighbourhood rule (no content token shared with the question or the target; $88$ of $88$ pairs kept; the leak screen $0$ of $30$), so counts, list length and every other line are identical to the \pn{} version's. Upper panel: W/T/L and the matched AUROC of the \pn{} version against the sibling version with both intervals; the \pn{}-against-\present{} value on the same pairs; and the sibling version against the \present{} version (how far an unrelated name in the list raises the score, the list-length component; descriptive). Lower panels: the sibling-minus-\present{} mean score with both intervals, and each reader minus the name lookup on the primary comparison, $^{*}$ where both intervals exclude zero. Four predictions, all as written: P-SB1, the lookup at or above $0.90$ ($0.9489$); P-SB2, every model at or above $0.80$ with the persona interval above $0.50$ (lower bounds $0.8409$, $0.8351$, $0.8681$, $0.8274$, $0.9387$); P-SB3, the random-token lookup and the count reader within $0.05$ of $0.50$ ($0.4716$, $0.5000$); P-SB4, no model reliably above the lookup. Disclosure: the scorer was dry-run once before the freeze to check its code path and computed the three no-model readers on the sibling texts, so P-SB1 and P-SB3 are recorded values rather than predictions and are not counted in Appendix~\ref{app:registration}; no model row existed at the dry run. GPT-4.1 mini, Haiku 4.5 and GPT-5.6 Luna are five-draw means as in Table~\ref{tab:cf-main}; Opus 5 and GPT-5.6 Sol one draw.}
\label{tab:cf-sibling}
\centering\tiny\setlength{\tabcolsep}{2pt}
\begin{tabular}{@{}lrrllrr@{}}
\toprule
Reader & W/T/L & \pn{} vs sibling & triple interval & persona interval & \pn{} vs \present{} & sibling vs \present{} \\
\midrule
Name lookup & $82/3/3$ & $0.9489$ & $[0.9034,0.9830]$ & $[0.9070,0.9831]$ & $0.9659$ & $0.7557$ \\
GPT-4.1 mini (five draws) & $78/3/7$ & $0.9034$ & $[0.8409,0.9602]$ & $[0.8409,0.9551]$ & $0.8920$ & $0.6932$ \\
Haiku 4.5 (five draws) & $73/9/6$ & $0.8807$ & $[0.8182,0.9375]$ & $[0.8274,0.9318]$ & $0.9148$ & $0.7045$ \\
GPT-5.6 Luna (five draws) & $80/1/7$ & $0.9148$ & $[0.8523,0.9659]$ & $[0.8351,0.9767]$ & $0.9602$ & $0.6989$ \\
GPT-5.6 Sol (draw 1) & $78/5/5$ & $0.9148$ & $[0.8636,0.9602]$ & $[0.8681,0.9620]$ & $0.9489$ & $0.6080$ \\
Opus 5 (draw 1) & $85/1/2$ & $0.9716$ & $[0.9375,1.0000]$ & $[0.9387,1.0000]$ & $0.9716$ & $0.5284$ \\
Random-token lookup & $25/33/30$ & $0.4716$ & $[0.3864,0.5511]$ & $[0.4011,0.5402]$ & $0.7443$ & $0.7500$ \\
Count reader, guessed category & $0/88/0$ & $0.5000$ & $[0.5000,0.5000]$ & $[0.5000,0.5000]$ & $0.6932$ & $0.6932$ \\
\bottomrule
\end{tabular}\\[4pt]
\begin{tabular}{@{}lrll@{}}
\toprule
Reader: sibling minus \present{} mean score & difference & triple interval & persona interval \\
\midrule
Name lookup & $+0.0689$$^{*}$ & $[+0.0500,+0.0901]$ & $[+0.0438,+0.0988]$ \\
GPT-4.1 mini (five draws) & $+0.0759$$^{*}$ & $[+0.0469,+0.1043]$ & $[+0.0410,+0.1073]$ \\
Haiku 4.5 (five draws) & $+0.0859$$^{*}$ & $[+0.0556,+0.1167]$ & $[+0.0499,+0.1201]$ \\
GPT-5.6 Luna (five draws) & $+0.0644$$^{*}$ & $[+0.0295,+0.1057]$ & $[+0.0238,+0.1055]$ \\
GPT-5.6 Sol (draw 1) & $+0.0656$ & $[-0.0024,+0.1331]$ & $[-0.0018,+0.1373]$ \\
Opus 5 (draw 1) & $+0.0376$ & $[-0.0007,+0.0812]$ & $[-0.0183,+0.0844]$ \\
Random-token lookup & $+0.0334$$^{*}$ & $[+0.0242,+0.0437]$ & $[+0.0218,+0.0459]$ \\
Count reader, guessed category & $+0.0149$$^{*}$ & $[+0.0101,+0.0200]$ & $[+0.0092,+0.0213]$ \\
\bottomrule
\end{tabular}\\[4pt]
\begin{tabular}{@{}lrll@{}}
\toprule
Reader minus the name lookup, \pn{} vs sibling & difference & triple interval & persona interval \\
\midrule
GPT-4.1 mini (five draws) & $-0.0455$ & $[-0.1136,+0.0170]$ & $[-0.1112,+0.0115]$ \\
Haiku 4.5 (five draws) & $-0.0682$ & $[-0.1364,+0.0000]$ & $[-0.1266,+0.0000]$ \\
GPT-5.6 Luna (five draws) & $-0.0341$ & $[-0.0854,+0.0227]$ & $[-0.1075,+0.0294]$ \\
GPT-5.6 Sol (draw 1) & $-0.0341$ & $[-0.0966,+0.0286]$ & $[-0.0905,+0.0253]$ \\
Opus 5 (draw 1) & $+0.0227$ & $[-0.0170,+0.0682]$ & $[-0.0114,+0.0632]$ \\
Random-token lookup & $-0.4773$$^{*}$ & $[-0.5682,-0.3919]$ & $[-0.5476,-0.4107]$ \\
Count reader, guessed category & $-0.4489$$^{*}$ & $[-0.4830,-0.4034]$ & $[-0.4831,-0.4070]$ \\
\bottomrule
\end{tabular}
\end{table}

\paragraph{GPT-4.1 mini on the expiry pairs under the all-states summary.}
Under the all-states summary GPT-4.1 mini reads retraction at $0.8725$ and expiry at $0.5869$ on this set, although both states are named in the same way.
The $30$ expiry triples show why, because the three versions of each triple can be read side by side.
The model scores its \pn{} version at or below $0.05$ on $26$ of the $30$ pairs, and the \present{} and \absent{} versions score the same low values, so the three versions are indistinguishable to it.
Only four pairs draw a score of $0.40$ or more on the \pn{} version, and in two of those the \present{} version draws the same score.
Every expiry question carries the upcoming intent, and every \pn{} version names the expired attribute in the expired-names section.
The model therefore does not take a name listed under the expired heading as evidence against a question about an upcoming event, whereas a name under the retracted heading moves it.
Appendix~\ref{app:resp} records the same asymmetry on the full population, where the expiry row falls under the all-states summary.

\paragraph{Three registered checks on this set.}
The list-length check pads the superseded-names list of each of the $661$ main-study summaries with correct attribute names that no scored question of the persona targets, to $15$, $30$ and $60$ names, and scores both lookups and GPT-4.1 mini at each length against the $1.75$-name row (Table~\ref{tab:pad}).
The embedding lookup falls from $0.7484$ to $0.6696$ at $60$ names, a difference whose item interval excludes zero and whose persona interval includes it; the lexical lookup is flat at every length; GPT-4.1 mini degrades first at $15$ names, reliably below $-0.05$ at $15$ and $30$.
With every untargeted name the store holds on the list, a mean of $99$ names per summary and at most $129$, the embedding lookup reads $0.6531$; continued past that pool with other personas' names to $100$ and $200$ it reads $0.6558$ and $0.5960$ (Table~\ref{tab:pad}), so a longer list of the store's own names is untested.
Beyond the store's own name pool, a mean of $99$ names, the lookup is tested only with other personas' names.
Its amendment injects wrong names into the counterfactual set's superseded-only lists at FR-Bank's measured rate of $0.232$ and at $0.40$: the lookup falls from $0.7663$ to $0.6740$ and $0.6001$, and GPT-4.1 mini and Haiku 4.5 fall with it and stay within $0.02$ of it on the same lists, every weaker interval including zero.
The paraphrase check asks each of the $148$ questions again in a wording that shares no content token with the attribute's slot path or with any attribute name the summary lists; it is reported after Table~\ref{tab:pad}.
The long-list check pads the superseded-names lists of this set's own $444$ summaries to $60$ names for the three lookups and all five models, with four predictions; it is reported after the paraphrase check.
In Table~\ref{tab:cf-main}, the lookup with one count feature is a logistic reader over the name lookup's similarity and the guessed category's non-active share, fit with personas left out (Table~\ref{tab:cf-hybrid}); the two count readers read different quantities: the given-category reader ties every supersession triple and is perfect on the counted states by construction.
Table~\ref{tab:robust} reads the pooled AUROC on the counterfactual set unless noted: rows 2 and 5 matched and post hoc (marked $^{p}$), row 5 on the $60$ counted-state pairs; row 3 the \pn{} version against its sibling, matched, on the $88$ supersession pairs; row 7 the $115$ reworded pairs; row 9 the $88$ supersession triples, the lists padded to $60$ names, Opus at one draw against its five-draw row; row 10 the full obligation set.

\begin{table}[H]
\caption{The list-length check and its amendment. Upper block: the superseded-names list of each of the $661$ main-study summaries padded with correct attribute names that no scored question of the persona targets, to $15$, $30$ and $60$ names, and for the two lookups to $100$ and $200$ names under the registered draw order, which continues into other personas' names once the store's own pool is exhausted ($79$ of the $242$ summaries at $100$, every summary at $200$); obligation AUROC of each reader against its recorded $1.75$-name row, with the paired difference and both intervals. Lower block: the counterfactual set's $444$ superseded-only summaries with wrong names injected into the negatives' lists at FR-Bank's measured rate ($0.232$) and at $0.40$; AUROC against the recorded list, the paired difference and both intervals, and for the models the difference from the embedding lookup on the same lists (the weaker of the two intervals). Descriptive; no prediction was registered, and every branch prints as it came out.}
\label{tab:pad}
\centering
\scriptsize
\setlength{\tabcolsep}{3pt}
\begin{tabular}{@{}llrrll@{}}
\toprule
Reader & list & AUROC & $\Delta$ vs recorded & item interval & persona interval \\
\midrule
\multicolumn{6}{@{}l}{\textit{Padding with correct names, the $661$ questions, ground-truth summary}}\\
Name lookup & $1.75$ names & $0.7484$ & -- & -- & -- \\
Name lookup & $15$ names & $0.7298$ & $-0.0186$ & $[-0.0777,+0.0400]$ & $[-0.0917,+0.0522]$ \\
Name lookup & $30$ names & $0.7191$ & $-0.0293$ & $[-0.0968,+0.0392]$ & $[-0.1156,+0.0525]$ \\
Name lookup & $60$ names & $0.6696$ & $-0.0788$ & $[-0.1541,-0.0096]$ & $[-0.1794,+0.0284]$ \\
Name lookup & $100$ names & $0.6558$ & $-0.0926$ & $[-0.1718,-0.0221]$ & $[-0.1953,+0.0199]$ \\
Name lookup & $200$ names & $0.5960$ & $-0.1525$ & $[-0.2357,-0.0773]$ & $[-0.2595,-0.0399]$ \\
Lexical lookup & $1.75$ names & $0.5738$ & -- & -- & -- \\
Lexical lookup & $15$ names & $0.5772$ & $+0.0034$ & $[-0.0296,+0.0398]$ & $[-0.0367,+0.0501]$ \\
Lexical lookup & $30$ names & $0.5817$ & $+0.0079$ & $[-0.0302,+0.0546]$ & $[-0.0425,+0.0672]$ \\
Lexical lookup & $60$ names & $0.5825$ & $+0.0087$ & $[-0.0434,+0.0592]$ & $[-0.0662,+0.0865]$ \\
Lexical lookup & $100$ names & $0.5601$ & $-0.0137$ & $[-0.0672,+0.0369]$ & $[-0.0926,+0.0716]$ \\
Lexical lookup & $200$ names & $0.5383$ & $-0.0355$ & $[-0.0903,+0.0189]$ & $[-0.1125,+0.0455]$ \\
GPT-4.1 mini & $1.75$ names & $0.7339$ & -- & -- & -- \\
GPT-4.1 mini & $15$ names & $0.6675$ & $-0.0664$ & $[-0.1132,-0.0219]$ & $[-0.1291,-0.0128]$ \\
GPT-4.1 mini & $30$ names & $0.6595$ & $-0.0744$ & $[-0.1112,-0.0388]$ & $[-0.1231,-0.0301]$ \\
GPT-4.1 mini & $60$ names & $0.6944$ & $-0.0395$ & $[-0.0840,+0.0043]$ & $[-0.1021,+0.0274]$ \\
\bottomrule
\end{tabular}\\[6pt]
\setlength{\tabcolsep}{2pt}
\begin{tabular}{@{}llrrll@{}}
\toprule
Reader & list & AUROC & $\Delta$ & intervals (item; persona) & minus lookup (weaker) \\
\midrule
\multicolumn{6}{@{}l}{\textit{Wrong names injected, the counterfactual set ($444$), superseded-only summary}}\\
Emb.\ lookup & recorded & $0.7663$ & -- & -- & -- \\
Emb.\ lookup & $c=0.232$ & $0.6740$ & $-0.0923$ & $[-0.1157,-0.0696]$; $[-0.1177,-0.0695]$ & -- \\
Emb.\ lookup & $c=0.40$ & $0.6001$ & $-0.1662$ & $[-0.1995,-0.1351]$; $[-0.2020,-0.1314]$ & -- \\
Lex.\ lookup & recorded & $0.6698$ & -- & -- & -- \\
Lex.\ lookup & $c=0.232$ & $0.5961$ & $-0.0737$ & $[-0.0941,-0.0549]$; $[-0.0960,-0.0521]$ & -- \\
Lex.\ lookup & $c=0.40$ & $0.5497$ & $-0.1201$ & $[-0.1458,-0.0962]$; $[-0.1493,-0.0918]$ & -- \\
GPT-4.1 mini & recorded & $0.7167$ & -- & -- & -- \\
GPT-4.1 mini & $c=0.232$ & $0.6809$ & $-0.0358$ & $[-0.0712,-0.0029]$; $[-0.0771,-0.0006]$ & $+0.0069$ $[-0.0362,+0.0498]$ \\
GPT-4.1 mini & $c=0.40$ & $0.6123$ & $-0.1045$ & $[-0.1365,-0.0708]$; $[-0.1473,-0.0638]$ & $+0.0122$ $[-0.0291,+0.0528]$ \\
Haiku 4.5 & recorded & $0.7566$ & -- & -- & -- \\
Haiku 4.5 & $c=0.232$ & $0.6823$ & $-0.0742$ & $[-0.0983,-0.0495]$; $[-0.1027,-0.0476]$ & $+0.0083$ $[-0.0405,+0.0527]$ \\
Haiku 4.5 & $c=0.40$ & $0.6192$ & $-0.1373$ & $[-0.1660,-0.1078]$; $[-0.1709,-0.1044]$ & $+0.0192$ $[-0.0227,+0.0602]$ \\
\bottomrule
\end{tabular}
\end{table}

\paragraph{The paraphrase check.}
Each of the $148$ questions was rewritten by GPT-4.1 mini under a deterministic screen that forbids the attribute's own words and every name the summary prints; $24$ pairs had no rewrite that passed the screen after three drafts.
An author read the remaining $124$ against two questions, whether the rewrite asks the same thing under the same intent and whether it names the plan or value, and excluded $9$ on the first, leaving $115$ ($76$ supersession, $17$ retraction, $22$ expiry) whose labels transfer from the original triples because the store versions are unchanged.
The registered floor of $120$ accepted pairs was amended to the $115$ before any scoring call, nothing was revisited, and the $124$ are reported as a sensitivity analysis, not registered.
The registration is descriptive: it carries no prediction, only the branch readings of Table~\ref{tab:para-pooled} at a $\pm0.05$ tolerance.

The read was assisted.
An AI assistant pre-read the exported sheet and flagged rows with a one-line reason, and a coding session printed persona evidence for nine flagged rows; neither saw a reader's output or a score, and every verdict is an author's.
Against those verdicts, the rows the assistant flagged as ones it would have rejected were excluded $5$ of $5$, the rows it flagged as depending on the persona $3$ of $4$, the rows it flagged as vaguer but passable $1$ of $12$, and the unflagged rows $0$ of $103$.
The sheet, the flags, the evidence and every verdict are in the artifact.

\begin{table}[H]
\caption{The paraphrase check on the $115$ accepted pairs ($345$ items, $115$ obligations; registered, descriptive): obligation AUROC on the original questions and on the paraphrases under the same store versions, the paired difference paraphrase minus original with the item-level and the persona-clustered interval, the registered reading at the $\pm0.05$ tolerance (a reader \emph{holds} inside it and \emph{collapses} below $-0.05$ with both intervals clear of zero), and the paraphrase-set AUROC per state (superseded / retracted / expired). $^{*}$: both intervals exclude zero. The lexical lookup reads $0.5000$ on every paraphrase set by construction: no content token is shared with any printed name.}
\label{tab:para-pooled}
\centering
\tiny
\setlength{\tabcolsep}{1.5pt}
\begin{tabular}{@{}llrrrllll@{}}
\toprule
Reader & States & original & paraphrase & difference & item interval & persona interval & reading & per state, paraphrase \\
\midrule
Name lookup & superseded & $0.7921$ & $0.7705$ & $-0.0216$ & $[-0.0463,+0.0010]$ & $[-0.0476,+0.0055]$ & holds & $0.8937$ / $0.5000$ / $0.5000$ \\
Lexical lookup & superseded & $0.6860$ & $0.5000$ & $-0.1860$$^{*}$ & $[-0.2327,-0.1374]$ & $[-0.2355,-0.1343]$ & collapses & $0.5000$ / $0.5000$ / $0.5000$ \\
Name lookup & all & $0.8785$ & $0.8227$ & $-0.0559$$^{*}$ & $[-0.0908,-0.0196]$ & $[-0.0896,-0.0194]$ & collapses & $0.8553$ / $0.6548$ / $0.8264$ \\
Lexical lookup & all & $0.8053$ & $0.5000$ & $-0.3053$$^{*}$ & $[-0.3505,-0.2623]$ & $[-0.3492,-0.2570]$ & collapses & $0.5000$ / $0.5000$ / $0.5000$ \\
GPT-4.1 mini & superseded & $0.7127$ & $0.6801$ & $-0.0326$ & $[-0.0879,+0.0181]$ & $[-0.0816,+0.0116]$ & holds & $0.7742$ / $0.4896$ / $0.5207$ \\
GPT-4.1 mini & all & $0.7957$ & $0.7186$ & $-0.0771$$^{*}$ & $[-0.1363,-0.0170]$ & $[-0.1259,-0.0307]$ & collapses & $0.7684$ / $0.7154$ / $0.6043$ \\
Haiku 4.5 & superseded & $0.7779$ & $0.7137$ & $-0.0642$$^{*}$ & $[-0.1104,-0.0208]$ & $[-0.0998,-0.0343]$ & collapses & $0.8194$ / $0.5112$ / $0.5320$ \\
Haiku 4.5 & all & $0.8779$ & $0.7522$ & $-0.1257$$^{*}$ & $[-0.1781,-0.0765]$ & $[-0.1659,-0.0828]$ & collapses & $0.7674$ / $0.6349$ / $0.7805$ \\
Opus 5 (draw 1)$^{a}$ & superseded & $0.8666$ & $0.8335$ & $-0.0331$$^{*}$ & $[-0.0669,-0.0037]$ & $[-0.0571,-0.0108]$ & holds & $0.9162$ / $0.5779$ / $0.5723$ \\
Opus 5 (draw 1)$^{a}$ & all & $0.9172$ & $0.8770$ & $-0.0402$$^{*}$ & $[-0.0676,-0.0167]$ & $[-0.0659,-0.0199]$ & holds & $0.9159$ / $0.6721$ / $0.8161$ \\
\midrule
\multicolumn{9}{@{}l}{\textit{Added after the registration, not registered}}\\
GPT-5.6 Luna$^{b}$ & superseded & $0.8684$ & $0.8650$ & $-0.0034$ & $[-0.0312,+0.0236]$ & $[-0.0256,+0.0215]$ & holds & $0.9382$ / $0.6168$ / $0.6798$ \\
GPT-5.6 Sol$^{b}$ & superseded & $0.8745$ & $0.8596$ & $-0.0149$ & $[-0.0410,+0.0120]$ & $[-0.0407,+0.0101]$ & holds & $0.9333$ / $0.5415$ / $0.6612$ \\
\bottomrule
\end{tabular}
\par\smallskip
\begin{minipage}{0.97\linewidth}\scriptsize
$^{a}$Opus 5 ran the paraphrases once, against its recorded five-draw rows on the original questions, so its paired row carries a draw-count difference as well; $^{b}$added after the registration, not registered, no prediction: GPT-5.6 Luna and GPT-5.6 Sol at five draws through the raw API under the summary naming superseded records only, because both summaries together exceeded that lane's budget.
\end{minipage}
\end{table}
\begin{table}[H]
\caption{The same $115$ pairs under the matched metric (post hoc): within each triple the \pn{} version against the \present{} version, W/T/L and $(W+T/2)/115$, on the original questions and on the paraphrases, with the paired difference (the mean over triples of the per-triple outcome difference) and the triple-level and persona-grouped intervals. $^{*}$: both intervals exclude zero.}
\label{tab:para-matched}
\centering
\tiny
\setlength{\tabcolsep}{1.5pt}
\begin{tabular}{@{}llrrrrrll@{}}
\toprule
Reader & States & original W/T/L & matched & paraphrase W/T/L & matched & difference & triple interval & persona interval \\
\midrule
Name lookup & superseded & $72/43/0$ & $0.8130$ & $69/43/3$ & $0.7870$ & $-0.0261$ & $[-0.0522,-0.0043]$ & $[-0.0636,+0.0000]$ \\
Lexical lookup & superseded & $43/72/0$ & $0.6870$ & $0/115/0$ & $0.5000$ & $-0.1870$$^{*}$ & $[-0.2261,-0.1435]$ & $[-0.2328,-0.1381]$ \\
Name lookup & all & $98/15/2$ & $0.9174$ & $93/21/1$ & $0.9000$ & $-0.0174$ & $[-0.0522,+0.0217]$ & $[-0.0572,+0.0192]$ \\
Lexical lookup & all & $69/46/0$ & $0.8000$ & $0/115/0$ & $0.5000$ & $-0.3000$$^{*}$ & $[-0.3435,-0.2564]$ & $[-0.3440,-0.2521]$ \\
GPT-4.1 mini & superseded & $63/46/6$ & $0.7478$ & $50/58/7$ & $0.6870$ & $-0.0609$$^{*}$ & $[-0.1174,-0.0087]$ & $[-0.1191,-0.0039]$ \\
GPT-4.1 mini & all & $80/25/10$ & $0.8043$ & $69/36/10$ & $0.7565$ & $-0.0478$ & $[-0.1174,+0.0217]$ & $[-0.1355,+0.0345]$ \\
Haiku 4.5 & superseded & $77/33/5$ & $0.8130$ & $66/41/8$ & $0.7522$ & $-0.0609$ & $[-0.1130,+0.0000]$ & $[-0.1179,+0.0000]$ \\
Haiku 4.5 & all & $96/15/4$ & $0.9000$ & $75/31/9$ & $0.7870$ & $-0.1130$$^{*}$ & $[-0.1739,-0.0565]$ & $[-0.1709,-0.0564]$ \\
Opus 5 (draw 1)$^{a}$ & superseded & $104/2/9$ & $0.9130$ & $95/6/14$ & $0.8522$ & $-0.0609$ & $[-0.1348,+0.0087]$ & $[-0.1296,+0.0042]$ \\
Opus 5 (draw 1)$^{a}$ & all & $110/0/5$ & $0.9565$ & $103/2/10$ & $0.9043$ & $-0.0522$$^{*}$ & $[-0.1043,-0.0087]$ & $[-0.0958,-0.0094]$ \\
\midrule
\multicolumn{9}{@{}l}{\textit{Added after the registration, not registered}}\\
GPT-5.6 Luna$^{b}$ & superseded & $102/1/12$ & $0.8913$ & $105/1/9$ & $0.9174$ & $+0.0261$ & $[-0.0435,+0.0870]$ & $[-0.0414,+0.0909]$ \\
GPT-5.6 Sol$^{b}$ & superseded & $99/1/15$ & $0.8652$ & $98/1/16$ & $0.8565$ & $-0.0087$ & $[-0.0697,+0.0565]$ & $[-0.0764,+0.0591]$ \\
\bottomrule
\end{tabular}
\par\smallskip
\begin{minipage}{0.97\linewidth}\scriptsize
$^{a}$Opus 5 ran the paraphrases once, against its recorded five-draw rows on the original questions, so its paired row carries a draw-count difference as well; $^{b}$added after the registration, not registered, no prediction: GPT-5.6 Luna and GPT-5.6 Sol at five draws through the raw API under the summary naming superseded records only, because both summaries together exceeded that lane's budget.
\end{minipage}
\end{table}
\begin{table}[H]
\caption{Each model minus the name lookup under the same summary on the whole set of $115$ pairs, pooled AUROC with the item-level and the persona-clustered paired interval: on the paraphrase set (the registered row) and on the original questions of the same pairs beside it (post hoc), so that a lead can be read against the one it replaces. $^{*}$: both intervals exclude zero.}
\label{tab:para-minus-lookup}
\centering
\tiny
\setlength{\tabcolsep}{1.5pt}
\begin{tabular}{@{}llrllrll@{}}
\toprule
Model & States & paraphrases & item & persona & originals & item & persona \\
\midrule
GPT-4.1 mini & superseded & $-0.0904$$^{*}$ & $[-0.1453,-0.0352]$ & $[-0.1517,-0.0337]$ & $-0.0794$$^{*}$ & $[-0.1303,-0.0290]$ & $[-0.1236,-0.0348]$ \\
GPT-4.1 mini & all & $-0.1040$$^{*}$ & $[-0.1775,-0.0351]$ & $[-0.1649,-0.0512]$ & $-0.0828$$^{*}$ & $[-0.1323,-0.0357]$ & $[-0.1319,-0.0356]$ \\
Haiku 4.5 & superseded & $-0.0568$ & $[-0.1170,+0.0040]$ & $[-0.1066,-0.0074]$ & $-0.0142$ & $[-0.0675,+0.0365]$ & $[-0.0557,+0.0346]$ \\
Haiku 4.5 & all & $-0.0704$$^{*}$ & $[-0.1300,-0.0130]$ & $[-0.1230,-0.0149]$ & $-0.0006$ & $[-0.0363,+0.0378]$ & $[-0.0426,+0.0385]$ \\
Opus 5 (draw 1)$^{a}$ & superseded & $+0.0630$$^{*}$ & $[+0.0078,+0.1134]$ & $[+0.0101,+0.1139]$ & $+0.0745$$^{*}$ & $[+0.0230,+0.1252]$ & $[+0.0357,+0.1147]$ \\
Opus 5 (draw 1)$^{a}$ & all & $+0.0544$$^{*}$ & $[+0.0013,+0.0983]$ & $[+0.0060,+0.1003]$ & $+0.0387$$^{*}$ & $[+0.0053,+0.0718]$ & $[+0.0071,+0.0705]$ \\
\midrule
\multicolumn{8}{@{}l}{\textit{Added after the registration, not registered}}\\
GPT-5.6 Luna$^{b}$ & superseded & $+0.0945$$^{*}$ & $[+0.0485,+0.1401]$ & $[+0.0465,+0.1431]$ & $+0.0763$$^{*}$ & $[+0.0305,+0.1192]$ & $[+0.0400,+0.1123]$ \\
GPT-5.6 Sol$^{b}$ & superseded & $+0.0891$$^{*}$ & $[+0.0370,+0.1384]$ & $[+0.0374,+0.1380]$ & $+0.0824$$^{*}$ & $[+0.0379,+0.1300]$ & $[+0.0443,+0.1234]$ \\
\bottomrule
\end{tabular}
\par\smallskip
\begin{minipage}{0.97\linewidth}\scriptsize
$^{a}$Opus 5 ran the paraphrases once, against its recorded five-draw rows on the original questions, so its paired row carries a draw-count difference as well; $^{b}$added after the registration, not registered, no prediction: GPT-5.6 Luna and GPT-5.6 Sol at five draws through the raw API under the summary naming superseded records only, because both summaries together exceeded that lane's budget.
\end{minipage}
\end{table}
\begin{table}[H]
\caption{Supersession only (post hoc): each model minus the name lookup under the same summary on the $76$ supersession pairs ($228$ items), on the paraphrases and on the original questions of the same pairs; upper panel the pooled AUROC with the item-level and the persona-clustered paired intervals, lower panel the matched metric with the triple-level and the persona-grouped intervals. $^{*}$: both intervals clear of zero.}
\label{tab:para-sup}
\centering
\tiny
\setlength{\tabcolsep}{2pt}
\begin{tabular}{@{}lllrll@{}}
\toprule
Model & States & questions & Pooled AUROC & item interval & persona interval \\
\midrule
GPT-4.1 mini & superseded & paraphrases & $-0.1195$$^{*}$ & $[-0.1842,-0.0572]$ & $[-0.1959,-0.0512]$ \\
 &  & originals & $-0.1044$$^{*}$ & $[-0.1593,-0.0532]$ & $[-0.1665,-0.0456]$ \\
GPT-4.1 mini & all & paraphrases & $-0.0868$$^{*}$ & $[-0.1583,-0.0115]$ & $[-0.1572,-0.0143]$ \\
 &  & originals & $-0.0703$$^{*}$ & $[-0.1177,-0.0197]$ & $[-0.1320,-0.0099]$ \\
Haiku 4.5 & superseded & paraphrases & $-0.0743$$^{*}$ & $[-0.1371,-0.0202]$ & $[-0.1355,-0.0116]$ \\
 &  & originals & $-0.0304$ & $[-0.0746,+0.0152]$ & $[-0.0833,+0.0222]$ \\
Haiku 4.5 & all & paraphrases & $-0.0879$$^{*}$ & $[-0.1577,-0.0148]$ & $[-0.1488,-0.0241]$ \\
 &  & originals & $-0.0187$ & $[-0.0572,+0.0290]$ & $[-0.0678,+0.0265]$ \\
Opus 5 (draw 1)$^{a}$ & superseded & paraphrases & $+0.0226$ & $[-0.0241,+0.0698]$ & $[-0.0357,+0.0806]$ \\
 &  & originals & $+0.0136$ & $[-0.0218,+0.0464]$ & $[-0.0249,+0.0513]$ \\
Opus 5 (draw 1)$^{a}$ & all & paraphrases & $+0.0607$$^{*}$ & $[+0.0114,+0.1156]$ & $[+0.0009,+0.1256]$ \\
 &  & originals & $+0.0309$ & $[-0.0027,+0.0662]$ & $[-0.0090,+0.0723]$ \\
\midrule
\multicolumn{6}{@{}l}{\textit{Added after the registration, not registered}}\\
GPT-5.6 Luna$^{b}$ & superseded & paraphrases & $+0.0445$$^{*}$ & $[+0.0093,+0.0845]$ & $[+0.0042,+0.0895]$ \\
 &  & originals & $+0.0255$ & $[-0.0003,+0.0520]$ & $[-0.0040,+0.0514]$ \\
GPT-5.6 Sol$^{b}$ & superseded & paraphrases & $+0.0396$ & $[-0.0063,+0.0845]$ & $[-0.0176,+0.0929]$ \\
 &  & originals & $+0.0172$ & $[-0.0169,+0.0501]$ & $[-0.0228,+0.0517]$ \\
\bottomrule
\end{tabular}
\par\vspace{3pt}
\begin{tabular}{@{}lllrll@{}}
\toprule
Model & States & questions & Matched metric & triple interval & persona interval \\
\midrule
GPT-4.1 mini & superseded & paraphrases & $-0.1513$$^{*}$ & $[-0.2171,-0.0789]$ & $[-0.2501,-0.0548]$ \\
 &  & originals & $-0.1053$$^{*}$ & $[-0.1581,-0.0526]$ & $[-0.1667,-0.0417]$ \\
GPT-4.1 mini & all & paraphrases & $-0.1118$$^{*}$ & $[-0.1908,-0.0395]$ & $[-0.1875,-0.0384]$ \\
 &  & originals & $-0.0724$ & $[-0.1316,-0.0132]$ & $[-0.1583,+0.0000]$ \\
Haiku 4.5 & superseded & paraphrases & $-0.0526$ & $[-0.1250,+0.0197]$ & $[-0.1205,+0.0236]$ \\
 &  & originals & $-0.0592$ & $[-0.1250,+0.0000]$ & $[-0.1181,-0.0068]$ \\
Haiku 4.5 & all & paraphrases & $-0.0921$$^{*}$ & $[-0.1645,-0.0132]$ & $[-0.1603,-0.0241]$ \\
 &  & originals & $-0.0395$ & $[-0.1053,+0.0263]$ & $[-0.1081,+0.0260]$ \\
Opus 5 (draw 1)$^{a}$ & superseded & paraphrases & $+0.0263$ & $[-0.0329,+0.0855]$ & $[-0.0375,+0.1000]$ \\
 &  & originals & $+0.0132$ & $[-0.0066,+0.0395]$ & $[-0.0141,+0.0435]$ \\
Opus 5 (draw 1)$^{a}$ & all & paraphrases & $+0.0526$ & $[+0.0000,+0.1118]$ & $[-0.0068,+0.1071]$ \\
 &  & originals & $+0.0461$ & $[+0.0000,+0.0987]$ & $[+0.0061,+0.0890]$ \\
\midrule
\multicolumn{6}{@{}l}{\textit{Added after the registration, not registered}}\\
GPT-5.6 Luna$^{b}$ & superseded & paraphrases & $+0.0395$ & $[-0.0066,+0.0921]$ & $[-0.0071,+0.1053]$ \\
 &  & originals & $-0.0197$ & $[-0.0658,+0.0132]$ & $[-0.0556,+0.0072]$ \\
GPT-5.6 Sol$^{b}$ & superseded & paraphrases & $+0.0263$ & $[-0.0395,+0.0855]$ & $[-0.0390,+0.0974]$ \\
 &  & originals & $-0.0197$ & $[-0.0658,+0.0197]$ & $[-0.0596,+0.0130]$ \\
\bottomrule
\end{tabular}
\par\smallskip
\begin{minipage}{0.97\linewidth}\scriptsize
$^{a}$Opus 5 ran the paraphrases once, against its recorded five-draw rows on the original questions, so its paired row carries a draw-count difference as well; $^{b}$added after the registration, not registered, no prediction: GPT-5.6 Luna and GPT-5.6 Sol at five draws through the raw API under the summary naming superseded records only, because both summaries together exceeded that lane's budget.
\end{minipage}
\end{table}
On the named state, reworded, GPT-5.6 Luna is the only reader above the lookup (Table~\ref{tab:para-sup}), by $0.04$ on the pooled metric.
It is the reader added after the registration; its matched interval includes zero; its pooled lower bounds are $0.009$ and $0.004$; this is one of five comparisons and no multiplicity correction is applied; and Haiku 4.5's pooled and matched readings also disagree on this state.
On the original questions of the same pairs no reader is reliably above the lookup.

\begin{table}[H]
\caption{Sensitivity, not registered: the same scoring on all $124$ screen-passing pairs ($372$ items), the $115$ accepted and the $9$ an author's read excluded; original, paraphrase, the paired difference with both intervals and the reading at the same tolerance. $^{*}$: both intervals exclude zero.}
\label{tab:para-sens}
\centering
\tiny
\setlength{\tabcolsep}{1.5pt}
\begin{tabular}{@{}llrrrlll@{}}
\toprule
Reader & States & original & paraphrase & difference & item interval & persona interval & reading \\
\midrule
Name lookup & superseded & $0.7901$ & $0.7643$ & $-0.0258$ & $[-0.0494,-0.0008]$ & $[-0.0542,+0.0017]$ & holds \\
Lexical lookup & superseded & $0.6847$ & $0.5000$ & $-0.1847$$^{*}$ & $[-0.2286,-0.1421]$ & $[-0.2317,-0.1356]$ & collapses \\
Name lookup & all & $0.8741$ & $0.8104$ & $-0.0637$$^{*}$ & $[-0.0969,-0.0243]$ & $[-0.0977,-0.0254]$ & collapses \\
Lexical lookup & all & $0.8058$ & $0.5000$ & $-0.3058$$^{*}$ & $[-0.3467,-0.2633]$ & $[-0.3504,-0.2606]$ & collapses \\
GPT-4.1 mini & superseded & $0.7252$ & $0.6721$ & $-0.0531$$^{*}$ & $[-0.1060,-0.0025]$ & $[-0.1020,-0.0075]$ & collapses \\
GPT-4.1 mini & all & $0.7979$ & $0.7034$ & $-0.0945$$^{*}$ & $[-0.1540,-0.0339]$ & $[-0.1417,-0.0510]$ & collapses \\
Haiku 4.5 & superseded & $0.7782$ & $0.7064$ & $-0.0718$$^{*}$ & $[-0.1201,-0.0267]$ & $[-0.1077,-0.0347]$ & collapses \\
Haiku 4.5 & all & $0.8791$ & $0.7387$ & $-0.1404$$^{*}$ & $[-0.1913,-0.0923]$ & $[-0.1810,-0.0994]$ & collapses \\
Opus 5 (draw 1)$^{a}$ & superseded & $0.8655$ & $0.8228$ & $-0.0428$$^{*}$ & $[-0.0815,-0.0102]$ & $[-0.0662,-0.0203]$ & holds \\
Opus 5 (draw 1)$^{a}$ & all & $0.9171$ & $0.8677$ & $-0.0494$$^{*}$ & $[-0.0826,-0.0224]$ & $[-0.0767,-0.0253]$ & holds \\
\midrule
\multicolumn{8}{@{}l}{\textit{Added after the registration, not registered}}\\
GPT-5.6 Luna$^{b}$ & superseded & $0.8692$ & $0.8638$ & $-0.0054$ & $[-0.0344,+0.0222]$ & $[-0.0299,+0.0202]$ & holds \\
GPT-5.6 Sol$^{b}$ & superseded & $0.8720$ & $0.8512$ & $-0.0208$ & $[-0.0487,+0.0054]$ & $[-0.0443,+0.0037]$ & holds \\
\bottomrule
\end{tabular}
\par\smallskip
\begin{minipage}{0.97\linewidth}\scriptsize
$^{a}$Opus 5 ran the paraphrases once, against its recorded five-draw rows on the original questions, so its paired row carries a draw-count difference as well; $^{b}$added after the registration, not registered, no prediction: GPT-5.6 Luna and GPT-5.6 Sol at five draws through the raw API under the summary naming superseded records only, because both summaries together exceeded that lane's budget.
\end{minipage}
\end{table}
The picture is the same with the nine excluded pairs put back, slightly lower for every reader (Table~\ref{tab:para-sens}).

The registered readings, per summary, at the $\pm0.05$ tolerance on the point estimate: under the summary naming superseded records only, the lexical lookup collapses and the embedding lookup holds; GPT-4.1 mini and Opus 5 hold where the lexical lookup collapses, Haiku 4.5 collapses, and the two added readers hold.
Under the summary naming all states both lookups collapse and only Opus 5 holds; the embedding lookup's pooled loss there is beyond the registered tolerance, while its within-triple loss is not reliable (Table~\ref{tab:para-matched}).

\begin{table}[H]
\caption{The attribute-naming split of Appendix~\ref{app:posthoc-flags} recomputed on the paraphrase set (a disclosure; it excludes nothing), under two rules for the sheet's second column: an author's read to that appendix's sentence, whether the rewrite names the plan or value its attribute is about ($61$ flagged of the $115$, $54$ not), and the token rule of the original read, whether the rewrite carries a content token of the value part of the slot path ($32$ flagged, $83$ not). Pooled AUROC on each half; under the first rule the original questions' values on the same halves beside.}
\label{tab:para-flags}
\centering
\tiny
\setlength{\tabcolsep}{1.5pt}
\begin{tabular}{@{}llrrrrrr@{}}
\toprule
Reader & States & rule 1, flagged & unflagged & originals, flagged & unflagged & rule 2, flagged & unflagged \\
\midrule
Name lookup & superseded & $0.7163$ & $0.8443$ & $0.7291$ & $0.8795$ & $0.6641$ & $0.8080$ \\
Lexical lookup & superseded & $0.5000$ & $0.5000$ & $0.5567$ & $0.8354$ & $0.5000$ & $0.5000$ \\
Name lookup & all & $0.7762$ & $0.9124$ & $0.8170$ & $0.9769$ & $0.7385$ & $0.8566$ \\
Lexical lookup & all & $0.5000$ & $0.5000$ & $0.6672$ & $0.9588$ & $0.5000$ & $0.5000$ \\
GPT-4.1 mini & superseded & $0.6280$ & $0.7647$ & $0.6303$ & $0.8384$ & $0.5527$ & $0.7374$ \\
GPT-4.1 mini & all & $0.6694$ & $0.7943$ & $0.7020$ & $0.9240$ & $0.6182$ & $0.7699$ \\
Haiku 4.5 & superseded & $0.6485$ & $0.7971$ & $0.6905$ & $0.8798$ & $0.5527$ & $0.7792$ \\
Haiku 4.5 & all & $0.7032$ & $0.8171$ & $0.7759$ & $0.9870$ & $0.6436$ & $0.8003$ \\
Opus 5 (draw 1)$^{a}$ & superseded & $0.7984$ & $0.8911$ & $0.8394$ & $0.9180$ & $0.7283$ & $0.8740$ \\
Opus 5 (draw 1)$^{a}$ & all & $0.8363$ & $0.9408$ & $0.8665$ & $0.9799$ & $0.7859$ & $0.9095$ \\
\midrule
\multicolumn{8}{@{}l}{\textit{Added after the registration, not registered}}\\
GPT-5.6 Luna$^{b}$ & superseded & $0.8236$ & $0.9339$ & $0.8138$ & $0.9255$ & $0.7393$ & $0.9136$ \\
GPT-5.6 Sol$^{b}$ & superseded & $0.8255$ & $0.9102$ & $0.8189$ & $0.9444$ & $0.7544$ & $0.8959$ \\
\bottomrule
\end{tabular}
\par\smallskip
\begin{minipage}{0.97\linewidth}\scriptsize
$^{a}$Opus 5 ran the paraphrases once, against its recorded five-draw rows on the original questions, so its paired row carries a draw-count difference as well; $^{b}$added after the registration, not registered, no prediction: GPT-5.6 Luna and GPT-5.6 Sol at five draws through the raw API under the summary naming superseded records only, because both summaries together exceeded that lane's budget.
\end{minipage}
\end{table}
The rewrites are systematically more generic than the originals.
Because the screen forbids the attribute's own words and every printed name, specific nouns give way to generic ones: ``What materials do I need to prepare for my upcoming meeting?'' became ``Which items should I get ready for the coming discussion?''; ``What type of headlamps am I currently carrying?'' became ``Can you tell me the model of the lights I currently possess?''; ``At this moment, how much am I paying per bag through the cooperative?'' became ``At present, what is the charge for every portion I get from the association?''.
Over the $115$ pairs the mean cosine of the question to its target name falls from $0.486$ to $0.317$ and the mean content-token overlap with it from $0.484$ to $0.000$, so every change above reflects both the loss of shared words and vaguer wording.

Draws: GPT-4.1 mini $1$, Haiku 4.5 $5$, Opus 5 $1$ against its recorded $5$, GPT-5.6 Luna and GPT-5.6 Sol $5$; the GPT-4.1 mini lane synchronous at list price, the Haiku 4.5 and Opus 5 lanes through the batch interface, the two added readers through the raw interface with the settings of their Table~\ref{tab:cf-main} rows.
Every served model identifier was checked before its lane, every lock of the registration held, and the lookup gate reproduced the recorded values on the original questions at four decimals.
An interim aggregate read was taken with the Opus 5 and Sol lanes then pending; no lane or analysis changed after it.

\paragraph{The long-list check.}
The superseded-names list of each of this set's $444$ summaries was padded with attribute names drawn from the persona's own printed name map, the pool being the names printed in all three versions of the triple less every name in any version's superseded section, every name a scored question of the persona targets in either population, and the change set; the pool holds $71$ to $121$ names per triple (median $102$), so no pad draws on other personas' names.
One pad is drawn per triple under a recorded seed and the triple's three versions receive the identical pad, so the \pn{} version's list has exactly $60$ names and its siblings' lists are shorter by the committed section alone ($57$ to $60$ names); the counts above the list are left as written, so the superseded total disagrees with the list length, as in the list-length check above.
The target is never in its own pad; its alphabetical rank in the \pn{} list is never first and last on $4$ of the $88$ supersession triples (minimum $2$, median $41.5$, maximum $60$; by tenths of the list, $5$, $3$, $2$, $5$, $8$, $12$, $12$, $9$, $9$, $23$).
The renderer reproduces every committed text byte for byte over the committed names alone ($444$ of $444$), the leak scan reports $0$ failures, the lookups reproduce their recorded values on the committed lists at four decimals, and the registration was frozen and committed before the first call, with both endpoints, the pooled AUROC and the matched score on the supersession triples, fixed in it.
Caps of \$$12$ (Anthropic) and \$$20$ (OpenAI) were metered from the rows at the page prices; the run spent \$$8.11$ and \$$15.54$, no cap was reached, no draw was dropped, and no deviation was recorded.

The four predictions, each read on the pooled AUROC of the supersession slice with the matched reading beside it:
P-L1, that each of GPT-5.6 Luna, GPT-5.6 Sol and Opus 5 reliably exceeds the name lookup on the $60$-name list, fails on Opus 5 (Luna $+0.1375$ and Sol $+0.1274$ above it with both intervals, Opus 5 $-0.0336$ with both intervals including zero; Table~\ref{tab:cfpad-minus-lookup}); P-L2, that neither GPT-4.1 mini nor Haiku 4.5 reliably exceeds it, holds ($-0.0525$ and $-0.0062$); P-L3, that the lookup loses more than $0.05$ against its unpadded row with both intervals below zero, holds ($-0.1807$; Table~\ref{tab:cfpad-sup}); P-L4, that each of the three stronger models loses less than the lookup at the point estimate, fails on Opus 5 (Luna $-0.0636$ and Sol $-0.0693$ lose less, Opus 5 $-0.2245$ loses more).
Luna and Sol are above the lookup on the matched metric as well, both intervals clear of zero (Table~\ref{tab:cfpad-minus-lookup}).

\begin{table}[H]
\caption{The long-list check, supersession (registered): the $264$ items of the $88$ supersession triples with the superseded-names list padded to $60$ names, every reader's pooled AUROC (upper panel) and matched score (lower panel, with W/T/L) beside its recorded unpadded value on the same items, and the paired difference padded minus unpadded with the item-level and the persona-grouped interval (pooled) or the triple-level and the persona-grouped interval (matched). The three lookups read the padded lists as printed; the models ran the padded summaries at the settings of their Table~\ref{tab:cf-main} rows. $^{*}$: both intervals exclude zero.}
\label{tab:cfpad-sup}
\centering
\tiny
\setlength{\tabcolsep}{1.5pt}
\begin{tabular}{@{}lrrrll@{}}
\toprule
\multicolumn{6}{@{}l}{\textit{Pooled AUROC}}\\
Reader & unpadded & padded & difference & item interval & persona interval \\
\midrule
Name lookup & $0.9223$ & $0.7416$ & $-0.1807$$^{*}$ & $[-0.2420,-0.1272]$ & $[-0.2335,-0.1311]$ \\
Lexical lookup & $0.7829$ & $0.6892$ & $-0.0937$$^{*}$ & $[-0.1435,-0.0501]$ & $[-0.1320,-0.0599]$ \\
Random-token lookup & $0.6651$ & $0.5008$ & $-0.1643$$^{*}$ & $[-0.2325,-0.0957]$ & $[-0.2302,-0.1106]$ \\
\midrule
GPT-4.1 mini & $0.8793$ & $0.6891$ & $-0.1902$$^{*}$ & $[-0.2451,-0.1369]$ & $[-0.2407,-0.1354]$ \\
Haiku 4.5 & $0.8876$ & $0.7354$ & $-0.1522$$^{*}$ & $[-0.2106,-0.1001]$ & $[-0.2017,-0.0997]$ \\
GPT-5.6 Luna & $0.9426$ & $0.8791$ & $-0.0636$$^{*}$ & $[-0.0978,-0.0293]$ & $[-0.0989,-0.0307]$ \\
GPT-5.6 Sol & $0.9383$ & $0.8690$ & $-0.0693$$^{*}$ & $[-0.1058,-0.0352]$ & $[-0.1175,-0.0221]$ \\
Opus 5$^{a}$ & $0.9325$ & $0.7080$ & $-0.2245$$^{*}$ & $[-0.2792,-0.1705]$ & $[-0.3449,-0.1294]$ \\
\bottomrule
\end{tabular}
\par\vspace{3pt}
\begin{tabular}{@{}lrrrll@{}}
\toprule
\multicolumn{6}{@{}l}{\textit{Matched score, \pn{} against \present{}}}\\
Reader & unpadded (W/T/L) & padded (W/T/L) & difference & triple interval & persona interval \\
\midrule
Name lookup & $0.9659$ (82/6/0) & $0.8295$ (58/30/0) & $-0.1364$$^{*}$ & $[-0.1875,-0.0909]$ & $[-0.1883,-0.0823]$ \\
Lexical lookup & $0.7841$ (50/38/0) & $0.7443$ (43/45/0) & $-0.0398$$^{*}$ & $[-0.0739,-0.0170]$ & $[-0.0745,-0.0120]$ \\
Random-token lookup & $0.7443$ (45/41/2) & $0.5057$ (1/87/0) & $-0.2386$$^{*}$ & $[-0.2955,-0.1818]$ & $[-0.3056,-0.1772]$ \\
\midrule
GPT-4.1 mini & $0.8920$ (76/5/7) & $0.7841$ (67/4/17) & $-0.1080$$^{*}$ & $[-0.2159,-0.0055]$ & $[-0.2143,-0.0055]$ \\
Haiku 4.5 & $0.9148$ (76/9/3) & $0.8011$ (59/23/6) & $-0.1136$$^{*}$ & $[-0.1706,-0.0511]$ & $[-0.1786,-0.0455]$ \\
GPT-5.6 Luna & $0.9602$ (84/1/3) & $0.9318$ (82/0/6) & $-0.0284$ & $[-0.0795,+0.0170]$ & $[-0.0787,+0.0188]$ \\
GPT-5.6 Sol & $0.9602$ (84/1/3) & $0.9205$ (81/0/7) & $-0.0398$ & $[-0.1193,+0.0284]$ & $[-0.1222,+0.0319]$ \\
Opus 5$^{a}$ & $0.9773$ (86/0/2) & $0.7841$ (67/4/17) & $-0.1932$$^{*}$ & $[-0.2841,-0.1136]$ & $[-0.3148,-0.0841]$ \\
\bottomrule
\end{tabular}
\par\smallskip
\begin{minipage}{0.97\linewidth}\scriptsize
$^{a}$Opus 5: one draw; recorded row five draws.
\end{minipage}
\end{table}
\begin{table}[H]
\caption{The long-list check, supersession (registered): each model minus the name lookup on the same lists, pooled (item-level and persona-grouped intervals) and matched (triple-level and persona-grouped), on the padded lists (upper panel) and on the unpadded lists (lower panel: the recorded model minus the recorded lookup). A difference in bold is reliable (both intervals exclude zero, $^{*}$).}
\label{tab:cfpad-minus-lookup}
\centering
\tiny
\setlength{\tabcolsep}{1.5pt}
\begin{tabular}{@{}lrllrll@{}}
\toprule
\multicolumn{7}{@{}l}{\textit{Padded lists}}\\
Model & pooled & item interval & persona interval & matched & triple interval & persona interval \\
\midrule
GPT-4.1 mini & $-0.0525$ & $[-0.1065,+0.0041]$ & $[-0.1092,+0.0058]$ & $-0.0455$ & $[-0.1307,+0.0455]$ & $[-0.1461,+0.0450]$ \\
Haiku 4.5 & $-0.0062$ & $[-0.0652,+0.0515]$ & $[-0.0558,+0.0464]$ & $-0.0284$ & $[-0.0852,+0.0341]$ & $[-0.0805,+0.0294]$ \\
GPT-5.6 Luna & $\mathbf{+0.1375}$$^{*}$ & $[+0.0883,+0.1934]$ & $[+0.0971,+0.1777]$ & $\mathbf{+0.1023}$$^{*}$ & $[+0.0398,+0.1648]$ & $[+0.0482,+0.1552]$ \\
GPT-5.6 Sol & $\mathbf{+0.1274}$$^{*}$ & $[+0.0726,+0.1837]$ & $[+0.0802,+0.1726]$ & $\mathbf{+0.0909}$$^{*}$ & $[+0.0227,+0.1477]$ & $[+0.0341,+0.1519]$ \\
Opus 5$^{a}$ & $-0.0336$ & $[-0.1070,+0.0448]$ & $[-0.1597,+0.0748]$ & $-0.0455$ & $[-0.1420,+0.0398]$ & $[-0.1845,+0.0756]$ \\
\bottomrule
\end{tabular}
\par\vspace{3pt}
\begin{tabular}{@{}lrllrll@{}}
\toprule
\multicolumn{7}{@{}l}{\textit{Unpadded lists (the recorded rows)}}\\
Model & pooled & item interval & persona interval & matched & triple interval & persona interval \\
\midrule
GPT-4.1 mini & $-0.0429$ & $[-0.0898,+0.0006]$ & $[-0.1001,+0.0035]$ & $\mathbf{-0.0739}$$^{*}$ & $[-0.1307,-0.0227]$ & $[-0.1393,-0.0214]$ \\
Haiku 4.5 & $-0.0347$ & $[-0.0813,+0.0086]$ & $[-0.0809,+0.0066]$ & $-0.0511$ & $[-0.1136,+0.0000]$ & $[-0.1084,+0.0000]$ \\
GPT-5.6 Luna & $+0.0204$ & $[-0.0053,+0.0473]$ & $[-0.0056,+0.0451]$ & $-0.0057$ & $[-0.0341,+0.0227]$ & $[-0.0321,+0.0174]$ \\
GPT-5.6 Sol & $+0.0161$ & $[-0.0130,+0.0473]$ & $[-0.0202,+0.0478]$ & $-0.0057$ & $[-0.0511,+0.0341]$ & $[-0.0417,+0.0287]$ \\
Opus 5$^{a}$ & $+0.0102$ & $[-0.0183,+0.0392]$ & $[-0.0259,+0.0432]$ & $+0.0114$ & $[-0.0227,+0.0455]$ & $[-0.0284,+0.0579]$ \\
\bottomrule
\end{tabular}
\par\smallskip
\begin{minipage}{0.97\linewidth}\scriptsize
$^{a}$Opus 5: one draw; recorded row five draws.
\end{minipage}
\end{table}
\begin{table}[H]
\caption{The long-list check on the whole counterfactual set, $444$ items and $148$ triples (descriptive; the registered comparisons are the supersession slice): every reader's padded and unpadded scores as in Table~\ref{tab:cfpad-sup} (the first two panels), then each model minus the name lookup as in Table~\ref{tab:cfpad-minus-lookup} (the last two). $^{*}$: both intervals exclude zero; bold: reliable.}
\label{tab:cfpad-whole}
\centering
\tiny
\setlength{\tabcolsep}{1.5pt}
\begin{tabular}{@{}lrrrll@{}}
\toprule
\multicolumn{6}{@{}l}{\textit{Pooled AUROC}}\\
Reader & unpadded & padded & difference & item interval & persona interval \\
\midrule
Name lookup & $0.7663$ & $0.6470$ & $-0.1194$$^{*}$ & $[-0.1805,-0.0571]$ & $[-0.1551,-0.0857]$ \\
Lexical lookup & $0.6698$ & $0.6153$ & $-0.0544$$^{*}$ & $[-0.0960,-0.0109]$ & $[-0.0791,-0.0298]$ \\
Random-token lookup & $0.5954$ & $0.5004$ & $-0.0951$$^{*}$ & $[-0.1567,-0.0316]$ & $[-0.1377,-0.0601]$ \\
\midrule
GPT-4.1 mini & $0.7648$ & $0.6304$ & $-0.1344$$^{*}$ & $[-0.1794,-0.0936]$ & $[-0.1702,-0.0992]$ \\
Haiku 4.5 & $0.7566$ & $0.6525$ & $-0.1041$$^{*}$ & $[-0.1475,-0.0636]$ & $[-0.1453,-0.0631]$ \\
GPT-5.6 Luna & $0.8417$ & $0.7390$ & $-0.1027$$^{*}$ & $[-0.1388,-0.0677]$ & $[-0.1327,-0.0762]$ \\
GPT-5.6 Sol & $0.8500$ & $0.7477$ & $-0.1023$$^{*}$ & $[-0.1367,-0.0707]$ & $[-0.1390,-0.0627]$ \\
Opus 5$^{a}$ & $0.8405$ & $0.6527$ & $-0.1879$$^{*}$ & $[-0.2268,-0.1480]$ & $[-0.2595,-0.1289]$ \\
\bottomrule
\end{tabular}
\par\vspace{3pt}
\begin{tabular}{@{}lrrrll@{}}
\toprule
\multicolumn{6}{@{}l}{\textit{Matched score, \pn{} against \present{}}}\\
Reader & unpadded (W/T/L) & padded (W/T/L) & difference & triple interval & persona interval \\
\midrule
Name lookup & $0.7770$ (82/66/0) & $0.6959$ (58/90/0) & $-0.0811$$^{*}$ & $[-0.1115,-0.0541]$ & $[-0.1157,-0.0521]$ \\
Lexical lookup & $0.6689$ (50/98/0) & $0.6453$ (43/105/0) & $-0.0236$$^{*}$ & $[-0.0405,-0.0101]$ & $[-0.0426,-0.0074]$ \\
Random-token lookup & $0.6453$ (45/101/2) & $0.5034$ (1/147/0) & $-0.1419$$^{*}$ & $[-0.1824,-0.1047]$ & $[-0.1871,-0.1007]$ \\
\midrule
GPT-4.1 mini & $0.7264$ (100/15/33) & $0.7128$ (102/7/39) & $-0.0135$ & $[-0.1115,+0.0845]$ & $[-0.0988,+0.0621]$ \\
Haiku 4.5 & $0.7770$ (89/52/7) & $0.6520$ (74/45/29) & $-0.1250$$^{*}$ & $[-0.1858,-0.0642]$ & $[-0.1948,-0.0508]$ \\
GPT-5.6 Luna & $0.8784$ (129/2/17) & $0.8649$ (128/0/20) & $-0.0135$ & $[-0.0811,+0.0574]$ & $[-0.0966,+0.0669]$ \\
GPT-5.6 Sol & $0.8581$ (126/2/20) & $0.7635$ (113/0/35) & $-0.0946$$^{*}$ & $[-0.1655,-0.0236]$ & $[-0.1633,-0.0280]$ \\
Opus 5$^{a}$ & $0.8750$ (128/3/17) & $0.7095$ (98/14/36) & $-0.1655$$^{*}$ & $[-0.2433,-0.0811]$ & $[-0.2568,-0.0810]$ \\
\bottomrule
\end{tabular}
\par\vspace{3pt}
\begin{tabular}{@{}lrllrll@{}}
\toprule
\multicolumn{7}{@{}l}{\textit{Padded lists}}\\
Model & pooled & item interval & persona interval & matched & triple interval & persona interval \\
\midrule
GPT-4.1 mini & $-0.0165$ & $[-0.0699,+0.0359]$ & $[-0.0541,+0.0207]$ & $+0.0169$ & $[-0.0541,+0.0878]$ & $[-0.0578,+0.0743]$ \\
Haiku 4.5 & $+0.0055$ & $[-0.0462,+0.0538]$ & $[-0.0278,+0.0447]$ & $-0.0439$ & $[-0.0981,+0.0068]$ & $[-0.0896,+0.0031]$ \\
GPT-5.6 Luna & $\mathbf{+0.0921}$$^{*}$ & $[+0.0450,+0.1438]$ & $[+0.0572,+0.1277]$ & $\mathbf{+0.1689}$$^{*}$ & $[+0.1115,+0.2297]$ & $[+0.1102,+0.2245]$ \\
GPT-5.6 Sol & $\mathbf{+0.1007}$$^{*}$ & $[+0.0516,+0.1531]$ & $[+0.0629,+0.1413]$ & $+0.0676$ & $[+0.0000,+0.1318]$ & $[+0.0096,+0.1241]$ \\
Opus 5$^{a}$ & $+0.0057$ & $[-0.0538,+0.0711]$ & $[-0.0780,+0.0847]$ & $+0.0135$ & $[-0.0574,+0.0845]$ & $[-0.0868,+0.1084]$ \\
\bottomrule
\end{tabular}
\par\vspace{3pt}
\begin{tabular}{@{}lrllrll@{}}
\toprule
\multicolumn{7}{@{}l}{\textit{Unpadded lists (the recorded rows)}}\\
Model & pooled & item interval & persona interval & matched & triple interval & persona interval \\
\midrule
GPT-4.1 mini & $-0.0016$ & $[-0.0426,+0.0393]$ & $[-0.0415,+0.0328]$ & $-0.0507$ & $[-0.1115,+0.0101]$ & $[-0.1069,+0.0000]$ \\
Haiku 4.5 & $-0.0098$ & $[-0.0640,+0.0398]$ & $[-0.0494,+0.0336]$ & $+0.0000$ & $[-0.0439,+0.0372]$ & $[-0.0528,+0.0483]$ \\
GPT-5.6 Luna & $\mathbf{+0.0754}$$^{*}$ & $[+0.0339,+0.1216]$ & $[+0.0450,+0.1090]$ & $\mathbf{+0.1014}$$^{*}$ & $[+0.0473,+0.1554]$ & $[+0.0442,+0.1533]$ \\
GPT-5.6 Sol & $\mathbf{+0.0836}$$^{*}$ & $[+0.0401,+0.1315]$ & $[+0.0544,+0.1163]$ & $\mathbf{+0.0811}$$^{*}$ & $[+0.0270,+0.1351]$ & $[+0.0283,+0.1319]$ \\
Opus 5$^{a}$ & $\mathbf{+0.0742}$$^{*}$ & $[+0.0315,+0.1203]$ & $[+0.0414,+0.1126]$ & $\mathbf{+0.0980}$$^{*}$ & $[+0.0472,+0.1454]$ & $[+0.0435,+0.1503]$ \\
\bottomrule
\end{tabular}
\par\smallskip
\begin{minipage}{0.97\linewidth}\scriptsize
$^{a}$Opus 5: one draw; recorded row five draws.
\end{minipage}
\end{table}
\begin{table}[H]
\caption{The long-list check on retraction and expiry, $90$ items and $30$ triples each (descriptive): unpadded/padded/the paired difference, pooled and matched, per state. The padded list names neither state and a triple's three versions carry identical padded lists, so the lookups tie every triple on both states by construction. $^{*}$: both intervals exclude zero.}
\label{tab:cfpad-counted}
\centering
\tiny
\setlength{\tabcolsep}{1pt}
\begin{tabular}{@{}lllll@{}}
\toprule
Reader & retraction, pooled & retraction, matched & expiry, pooled & expiry, matched \\
\midrule
Name lookup & $0.5000$/$0.5000$/$+0.0000$ & $0.5000$/$0.5000$/$+0.0000$ & $0.5000$/$0.5000$/$+0.0000$ & $0.5000$/$0.5000$/$+0.0000$ \\
Lexical lookup & $0.5000$/$0.5000$/$+0.0000$ & $0.5000$/$0.5000$/$+0.0000$ & $0.5000$/$0.5000$/$+0.0000$ & $0.5000$/$0.5000$/$+0.0000$ \\
Random-token lookup & $0.5000$/$0.5000$/$+0.0000$ & $0.5000$/$0.5000$/$+0.0000$ & $0.5000$/$0.5000$/$+0.0000$ & $0.5000$/$0.5000$/$+0.0000$ \\
\midrule
GPT-4.1 mini & $0.5761$/$0.5597$/$-0.0164$ & $0.5667$/$0.6667$/$+0.1000$ & $0.5594$/$0.5147$/$-0.0447$ & $0.4000$/$0.5500$/$+0.1500$ \\
Haiku 4.5 & $0.5433$/$0.5014$/$-0.0419$ & $0.5667$/$0.3833$/$-0.1833$$^{*}$ & $0.5794$/$0.5606$/$-0.0189$ & $0.5833$/$0.4833$/$-0.1000$ \\
GPT-5.6 Luna & $0.5458$/$0.4956$/$-0.0503$ & $0.6500$/$0.7333$/$+0.0833$ & $0.6167$/$0.5700$/$-0.0467$ & $0.8667$/$0.8000$/$-0.0667$ \\
GPT-5.6 Sol & $0.5322$/$0.4636$/$-0.0686$ & $0.5833$/$0.4000$/$-0.1833$ & $0.6403$/$0.5728$/$-0.0675$ & $0.8333$/$0.6667$/$-0.1667$ \\
Opus 5$^{a}$ & $0.5872$/$0.5206$/$-0.0667$$^{*}$ & $0.7667$/$0.5667$/$-0.2000$ & $0.6150$/$0.6031$/$-0.0119$ & $0.6833$/$0.6333$/$-0.0500$ \\
\bottomrule
\end{tabular}
\par\smallskip
\begin{minipage}{0.97\linewidth}\scriptsize
$^{a}$Opus 5: one draw; recorded row five draws.
\end{minipage}
\end{table}
The padded list names neither retraction nor expiry and a triple's three versions carry identical padded lists, so no reader's change on those states is reliable except Haiku 4.5's matched fall on retraction and Opus 5's pooled fall on retraction (Table~\ref{tab:cfpad-counted}).

\begin{table}[H]
\caption{The long-list check, supersession (descriptive): each model minus the lexical lookup and minus the random-token lookup on the same lists, padded and unpadded, pooled and matched; point estimates, $^{*}$ where both intervals exclude zero (the intervals are in the artifact's ledger).}
\label{tab:cfpad-other-lookups}
\centering
\tiny
\setlength{\tabcolsep}{1.5pt}
\begin{tabular}{@{}lrrrrrrrr@{}}
\toprule
& \multicolumn{4}{c}{minus the lexical lookup} & \multicolumn{4}{c@{}}{minus the random-token lookup} \\
Model & padded, pooled & matched & unpadded, pooled & matched & padded, pooled & matched & unpadded, pooled & matched \\
\midrule
GPT-4.1 mini & $-0.0001$ & $+0.0398$ & $+0.0964$$^{*}$ & $+0.1080$$^{*}$ & $+0.1883$$^{*}$ & $+0.2784$$^{*}$ & $+0.2142$$^{*}$ & $+0.1477$$^{*}$ \\
Haiku 4.5 & $+0.0461$ & $+0.0568$ & $+0.1046$$^{*}$ & $+0.1307$$^{*}$ & $+0.2345$$^{*}$ & $+0.2955$$^{*}$ & $+0.2225$$^{*}$ & $+0.1705$$^{*}$ \\
GPT-5.6 Luna & $+0.1898$$^{*}$ & $+0.1875$$^{*}$ & $+0.1597$$^{*}$ & $+0.1761$$^{*}$ & $+0.3782$$^{*}$ & $+0.4261$$^{*}$ & $+0.2775$$^{*}$ & $+0.2159$$^{*}$ \\
GPT-5.6 Sol & $+0.1798$$^{*}$ & $+0.1761$$^{*}$ & $+0.1554$$^{*}$ & $+0.1761$$^{*}$ & $+0.3682$$^{*}$ & $+0.4148$$^{*}$ & $+0.2732$$^{*}$ & $+0.2159$$^{*}$ \\
Opus 5$^{a}$ & $+0.0188$ & $+0.0398$ & $+0.1496$$^{*}$ & $+0.1932$$^{*}$ & $+0.2072$$^{*}$ & $+0.2784$$^{*}$ & $+0.2674$$^{*}$ & $+0.2330$$^{*}$ \\
\bottomrule
\end{tabular}
\par\smallskip
\begin{minipage}{0.97\linewidth}\scriptsize
$^{a}$Opus 5: one draw; recorded row five draws.
\end{minipage}
\end{table}
\begin{table}[H]
\caption{The target's alphabetical rank in the \pn{} version's $60$-name list over the $88$ supersession triples against each reader (descriptive): Spearman correlation of the rank with the reader's \pn{} score and with its \pn{}-against-\present{} outcome.}
\label{tab:cfpad-position}
\centering
\scriptsize
\setlength{\tabcolsep}{4pt}
\begin{tabular}{@{}lrr@{}}
\toprule
Reader & rank against \pn{} score & rank against outcome \\
\midrule
Name lookup & $+0.3264$ & $+0.2040$ \\
Lexical lookup & $+0.3316$ & $+0.1612$ \\
Random-token lookup & $-0.0496$ & $-0.0613$ \\
\midrule
GPT-4.1 mini & $+0.1434$ & $+0.2762$ \\
Haiku 4.5 & $+0.1647$ & $+0.2717$ \\
GPT-5.6 Luna & $+0.2634$ & $+0.1199$ \\
GPT-5.6 Sol & $+0.2388$ & $-0.0546$ \\
Opus 5$^{a}$ & $+0.0467$ & $+0.0172$ \\
\bottomrule
\end{tabular}
\par\smallskip
\begin{minipage}{0.97\linewidth}\scriptsize
$^{a}$Opus 5: one draw; recorded row five draws.
\end{minipage}
\end{table}
Opus 5 ran the padded summaries once against a recorded five-draw row; its recorded draw 1 alone reads $0.9317$ pooled and $0.9716$ matched on supersession ($0.8400$ and $0.8682$ on the whole set), and the padded draw minus that draw reads $-0.2236$ ($[-0.2776,-0.1690]$; $[-0.3416,-0.1294]$) on supersession and $-0.1873$ ($[-0.2277,-0.1476]$; $[-0.2569,-0.1278]$) on the whole set, so the fall is not the draw count.
Under seeds $43$ to $46$ every registered difference keeps its reading; one interval of one difference, GPT-4.1 mini minus the lookup at seed $45$, moves across zero on the persona side without changing the reading.

Draws: GPT-4.1 mini, Haiku 4.5, GPT-5.6 Luna and GPT-5.6 Sol $5$, Opus 5 $1$ against its recorded $5$, at the settings of their Table~\ref{tab:cf-main} rows; the GPT-4.1 mini, Luna and Sol runs through the raw interface, draw by draw, and the Haiku 4.5 and Opus 5 runs through the batch interface; $2220$ rows per five-draw run and $444$ for Opus 5, none unparseable.
Every served model identifier was checked on a probe call before its run and on every row at scoring; no run reached its cap.

\paragraph{Equivalence within the tolerance, post hoc.}\label{app:equiv}
A reliable difference is one shown to differ from zero; a difference that is not reliable is not thereby shown to be small. The mirror test is two one-sided tests at $5\%$: a difference is equivalent within $0.05$ when both of its $90\%$ intervals, the item-level (triple-level on the matched metric) and the persona-grouped one, lie inside $[-0.05,+0.05]$. The $90\%$ intervals come from the same $1{,}000$ draws that give the printed $95\%$ ones. Tables~\ref{tab:equiv} and~\ref{tab:equiv-para} apply the test to every model-minus-lookup row printed for the default summary, the sibling-name control, the all-states summary, the padded set and the paraphrase set, and to the lookup with one count feature against the three stronger readers. A row can be neither reliable nor equivalent; its difference is then neither shown nor excluded. The test is post hoc and changes no reliable verdict.
On the short list's supersession triples, on the matched metric, Luna, Sol and Opus are equivalent to the lookup, mini is reliably below it and Haiku is neither; pooled, mini is neither. Under the sibling-name control no reader is equivalent to the lookup, and the lookup with one count feature is neither reliably different from nor equivalent to any of the three stronger readers.

\begin{table}[H]
\caption{Equivalence within $0.05$ (post hoc): each model minus the name lookup under the same summary, every row printed for the default summary (Tables~\ref{tab:sup-margins}, \ref{tab:cfpad-whole} and~\ref{tab:cf-matched-vs}), the sibling-name control (Table~\ref{tab:cf-sibling}), the all-states summary (Tables~\ref{tab:cf-main} and~\ref{tab:cf-matched-vs}) and the padded set (Tables~\ref{tab:cfpad-minus-lookup} and~\ref{tab:cfpad-whole}), and the lookup with one count feature minus each stronger reader (matched, $148$ triples). Pooled rows: item-level and persona-grouped intervals; matched rows: triple-level and persona-grouped. Equivalent: both $90\%$ intervals inside $[-0.05,+0.05]$, the two one-sided tests at $5\%$; reliable: both $95\%$ intervals exclude zero; a row that is neither is neither shown nor excluded.}
\label{tab:equiv}
\centering\tiny\setlength{\tabcolsep}{1.5pt}
\resizebox{\textwidth}{!}{\begin{tabular}{@{}llrllllll@{}}
\toprule
Reader & scope, metric & difference & 95\% item/triple & 95\% persona & 90\% item/triple & 90\% persona & reliable & equivalent \\
\midrule
\multicolumn{9}{@{}l}{\textit{default summary}}\\
GPT-4.1 mini, one draw & supersession, matched & $-0.0852$ & $[-0.1307,-0.0398]$ & $[-0.1420,-0.0294]$ & $[-0.1250,-0.0509]$ & $[-0.1330,-0.0372]$ & yes & no \\
GPT-4.1 mini, five draws & supersession, matched & $-0.0739$ & $[-0.1307,-0.0227]$ & $[-0.1393,-0.0214]$ & $[-0.1193,-0.0284]$ & $[-0.1304,-0.0294]$ & yes & no \\
Haiku 4.5 & supersession, matched & $-0.0511$ & $[-0.1136,+0.0000]$ & $[-0.1084,+0.0000]$ & $[-0.0966,-0.0114]$ & $[-0.0988,-0.0066]$ & no & no \\
GPT-5.6 Luna & supersession, matched & $-0.0057$ & $[-0.0341,+0.0227]$ & $[-0.0321,+0.0174]$ & $[-0.0284,+0.0227]$ & $[-0.0291,+0.0150]$ & no & yes \\
GPT-5.6 Sol & supersession, matched & $-0.0057$ & $[-0.0511,+0.0341]$ & $[-0.0417,+0.0287]$ & $[-0.0455,+0.0284]$ & $[-0.0372,+0.0230]$ & no & yes \\
Opus 5 & supersession, matched & $+0.0114$ & $[-0.0227,+0.0455]$ & $[-0.0284,+0.0579]$ & $[-0.0170,+0.0398]$ & $[-0.0222,+0.0495]$ & no & yes \\
GPT-4.1 mini, one draw & supersession, pooled & $-0.0915$ & $[-0.1399,-0.0478]$ & $[-0.1436,-0.0425]$ & $[-0.1310,-0.0550]$ & $[-0.1381,-0.0487]$ & yes & no \\
GPT-4.1 mini, five draws & supersession, pooled & $-0.0429$ & $[-0.0898,+0.0006]$ & $[-0.1001,+0.0035]$ & $[-0.0818,-0.0074]$ & $[-0.0896,-0.0038]$ & no & no \\
Haiku 4.5 & supersession, pooled & $-0.0347$ & $[-0.0813,+0.0086]$ & $[-0.0809,+0.0066]$ & $[-0.0727,+0.0014]$ & $[-0.0706,-0.0017]$ & no & no \\
GPT-5.6 Luna & supersession, pooled & $+0.0204$ & $[-0.0053,+0.0473]$ & $[-0.0056,+0.0451]$ & $[-0.0003,+0.0429]$ & $[-0.0021,+0.0403]$ & no & yes \\
GPT-5.6 Sol & supersession, pooled & $+0.0161$ & $[-0.0130,+0.0473]$ & $[-0.0202,+0.0478]$ & $[-0.0082,+0.0424]$ & $[-0.0109,+0.0411]$ & no & yes \\
Opus 5 & supersession, pooled & $+0.0102$ & $[-0.0183,+0.0392]$ & $[-0.0259,+0.0432]$ & $[-0.0143,+0.0352]$ & $[-0.0190,+0.0382]$ & no & yes \\
GPT-4.1 mini, one draw & whole, matched & $-0.0372$ & $[-0.0811,+0.0068]$ & $[-0.0877,+0.0146]$ & $[-0.0743,-0.0034]$ & $[-0.0800,+0.0070]$ & no & no \\
GPT-4.1 mini, five draws & whole, matched & $-0.0507$ & $[-0.1115,+0.0101]$ & $[-0.1069,+0.0000]$ & $[-0.1014,-0.0034]$ & $[-0.0992,-0.0063]$ & no & no \\
Haiku 4.5 & whole, matched & $+0.0000$ & $[-0.0439,+0.0372]$ & $[-0.0528,+0.0483]$ & $[-0.0372,+0.0338]$ & $[-0.0437,+0.0420]$ & no & yes \\
GPT-5.6 Luna & whole, matched & $+0.1014$ & $[+0.0473,+0.1554]$ & $[+0.0442,+0.1533]$ & $[+0.0574,+0.1453]$ & $[+0.0530,+0.1449]$ & yes & no \\
GPT-5.6 Sol & whole, matched & $+0.0811$ & $[+0.0270,+0.1351]$ & $[+0.0283,+0.1319]$ & $[+0.0372,+0.1284]$ & $[+0.0362,+0.1218]$ & yes & no \\
Opus 5 & whole, matched & $+0.0980$ & $[+0.0472,+0.1454]$ & $[+0.0435,+0.1503]$ & $[+0.0541,+0.1387]$ & $[+0.0526,+0.1444]$ & yes & no \\
GPT-4.1 mini, five draws & whole, pooled & $-0.0016$ & $[-0.0426,+0.0393]$ & $[-0.0415,+0.0328]$ & $[-0.0371,+0.0316]$ & $[-0.0343,+0.0278]$ & no & yes \\
Haiku 4.5 & whole, pooled & $-0.0098$ & $[-0.0640,+0.0398]$ & $[-0.0494,+0.0336]$ & $[-0.0527,+0.0320]$ & $[-0.0448,+0.0262]$ & no & no \\
GPT-5.6 Luna & whole, pooled & $+0.0754$ & $[+0.0339,+0.1216]$ & $[+0.0450,+0.1090]$ & $[+0.0407,+0.1127]$ & $[+0.0505,+0.1038]$ & yes & no \\
GPT-5.6 Sol & whole, pooled & $+0.0836$ & $[+0.0401,+0.1315]$ & $[+0.0544,+0.1163]$ & $[+0.0487,+0.1243]$ & $[+0.0578,+0.1110]$ & yes & no \\
Opus 5 & whole, pooled & $+0.0742$ & $[+0.0315,+0.1203]$ & $[+0.0414,+0.1126]$ & $[+0.0378,+0.1139]$ & $[+0.0461,+0.1063]$ & yes & no \\
\multicolumn{9}{@{}l}{\textit{sibling-name control}}\\
GPT-4.1 mini (five draws) & supersession (PN vs sibling), matched & $-0.0455$ & $[-0.1136,+0.0170]$ & $[-0.1112,+0.0115]$ & $[-0.1023,+0.0057]$ & $[-0.1001,+0.0051]$ & no & no \\
Haiku 4.5 (five draws) & supersession (PN vs sibling), matched & $-0.0682$ & $[-0.1364,+0.0000]$ & $[-0.1266,+0.0000]$ & $[-0.1250,-0.0114]$ & $[-0.1176,-0.0118]$ & no & no \\
GPT-5.6 Luna (five draws) & supersession (PN vs sibling), matched & $-0.0341$ & $[-0.0854,+0.0227]$ & $[-0.1075,+0.0294]$ & $[-0.0795,+0.0114]$ & $[-0.0958,+0.0215]$ & no & no \\
GPT-5.6 Sol (draw 1) & supersession (PN vs sibling), matched & $-0.0341$ & $[-0.0966,+0.0286]$ & $[-0.0905,+0.0253]$ & $[-0.0909,+0.0170]$ & $[-0.0825,+0.0163]$ & no & no \\
Opus 5 (draw 1) & supersession (PN vs sibling), matched & $+0.0227$ & $[-0.0170,+0.0682]$ & $[-0.0114,+0.0632]$ & $[-0.0114,+0.0625]$ & $[-0.0059,+0.0568]$ & no & no \\
\multicolumn{9}{@{}l}{\textit{all-states summary}}\\
GPT-4.1 mini, five draws & supersession, matched & $-0.0227$ & $[-0.0795,+0.0342]$ & $[-0.0886,+0.0337]$ & $[-0.0685,+0.0284]$ & $[-0.0778,+0.0266]$ & no & no \\
Haiku 4.5 & supersession, matched & $-0.0511$ & $[-0.1080,+0.0000]$ & $[-0.1123,+0.0054]$ & $[-0.0966,-0.0057]$ & $[-0.1001,-0.0057]$ & no & no \\
Opus 5 & supersession, matched & $+0.0227$ & $[-0.0114,+0.0568]$ & $[-0.0105,+0.0550]$ & $[-0.0057,+0.0511]$ & $[-0.0056,+0.0500]$ & no & no \\
Opus 5, draw 1 & supersession, matched & $+0.0398$ & $[+0.0170,+0.0682]$ & $[+0.0118,+0.0714]$ & $[+0.0170,+0.0625]$ & $[+0.0167,+0.0650]$ & yes & no \\
GPT-4.1 mini, five draws & whole, matched & $-0.0135$ & $[-0.0642,+0.0338]$ & $[-0.0698,+0.0375]$ & $[-0.0541,+0.0270]$ & $[-0.0577,+0.0289]$ & no & no \\
Haiku 4.5 & whole, matched & $-0.0034$ & $[-0.0439,+0.0372]$ & $[-0.0476,+0.0333]$ & $[-0.0372,+0.0304]$ & $[-0.0402,+0.0283]$ & no & yes \\
Opus 5 & whole, matched & $+0.0304$ & $[-0.0068,+0.0643]$ & $[-0.0038,+0.0648]$ & $[+0.0000,+0.0608]$ & $[+0.0000,+0.0599]$ & no & no \\
Opus 5, draw 1 & whole, matched & $+0.0372$ & $[+0.0033,+0.0709]$ & $[+0.0076,+0.0662]$ & $[+0.0068,+0.0642]$ & $[+0.0133,+0.0621]$ & yes & no \\
Opus 5 & whole, pooled & $+0.0323$ & $[+0.0060,+0.0617]$ & $[+0.0077,+0.0608]$ & $[+0.0087,+0.0569]$ & $[+0.0118,+0.0554]$ & yes & no \\
\multicolumn{9}{@{}l}{\textit{padded set}}\\
GPT-4.1 mini, five draws & supersession, matched & $-0.0455$ & $[-0.1307,+0.0455]$ & $[-0.1461,+0.0450]$ & $[-0.1193,+0.0284]$ & $[-0.1329,+0.0345]$ & no & no \\
Haiku 4.5 & supersession, matched & $-0.0284$ & $[-0.0852,+0.0341]$ & $[-0.0805,+0.0294]$ & $[-0.0795,+0.0227]$ & $[-0.0722,+0.0195]$ & no & no \\
GPT-5.6 Luna & supersession, matched & $+0.1023$ & $[+0.0398,+0.1648]$ & $[+0.0482,+0.1552]$ & $[+0.0511,+0.1534]$ & $[+0.0561,+0.1468]$ & yes & no \\
GPT-5.6 Sol & supersession, matched & $+0.0909$ & $[+0.0227,+0.1477]$ & $[+0.0341,+0.1519]$ & $[+0.0398,+0.1420]$ & $[+0.0426,+0.1453]$ & yes & no \\
Opus 5 & supersession, matched & $-0.0455$ & $[-0.1420,+0.0398]$ & $[-0.1845,+0.0756]$ & $[-0.1253,+0.0227]$ & $[-0.1596,+0.0585]$ & no & no \\
GPT-4.1 mini, five draws & supersession, pooled & $-0.0525$ & $[-0.1065,+0.0041]$ & $[-0.1092,+0.0058]$ & $[-0.0981,-0.0031]$ & $[-0.1000,-0.0064]$ & no & no \\
Haiku 4.5 & supersession, pooled & $-0.0062$ & $[-0.0652,+0.0515]$ & $[-0.0558,+0.0464]$ & $[-0.0550,+0.0437]$ & $[-0.0497,+0.0378]$ & no & no \\
GPT-5.6 Luna & supersession, pooled & $+0.1375$ & $[+0.0883,+0.1934]$ & $[+0.0971,+0.1777]$ & $[+0.0980,+0.1838]$ & $[+0.1016,+0.1722]$ & yes & no \\
GPT-5.6 Sol & supersession, pooled & $+0.1274$ & $[+0.0726,+0.1837]$ & $[+0.0802,+0.1726]$ & $[+0.0821,+0.1763]$ & $[+0.0883,+0.1653]$ & yes & no \\
Opus 5 & supersession, pooled & $-0.0336$ & $[-0.1070,+0.0448]$ & $[-0.1597,+0.0748]$ & $[-0.0930,+0.0315]$ & $[-0.1444,+0.0614]$ & no & no \\
GPT-4.1 mini, five draws & whole, matched & $+0.0169$ & $[-0.0541,+0.0878]$ & $[-0.0578,+0.0743]$ & $[-0.0439,+0.0745]$ & $[-0.0455,+0.0684]$ & no & no \\
Haiku 4.5 & whole, matched & $-0.0439$ & $[-0.0981,+0.0068]$ & $[-0.0896,+0.0031]$ & $[-0.0880,+0.0000]$ & $[-0.0820,-0.0068]$ & no & no \\
GPT-5.6 Luna & whole, matched & $+0.1689$ & $[+0.1115,+0.2297]$ & $[+0.1102,+0.2245]$ & $[+0.1182,+0.2196]$ & $[+0.1172,+0.2177]$ & yes & no \\
GPT-5.6 Sol & whole, matched & $+0.0676$ & $[+0.0000,+0.1318]$ & $[+0.0096,+0.1241]$ & $[+0.0135,+0.1184]$ & $[+0.0170,+0.1151]$ & no & no \\
Opus 5 & whole, matched & $+0.0135$ & $[-0.0574,+0.0845]$ & $[-0.0868,+0.1084]$ & $[-0.0439,+0.0743]$ & $[-0.0684,+0.0946]$ & no & no \\
GPT-4.1 mini, five draws & whole, pooled & $-0.0165$ & $[-0.0699,+0.0359]$ & $[-0.0541,+0.0207]$ & $[-0.0615,+0.0295]$ & $[-0.0486,+0.0136]$ & no & no \\
Haiku 4.5 & whole, pooled & $+0.0055$ & $[-0.0462,+0.0538]$ & $[-0.0278,+0.0447]$ & $[-0.0395,+0.0453]$ & $[-0.0232,+0.0390]$ & no & yes \\
GPT-5.6 Luna & whole, pooled & $+0.0921$ & $[+0.0450,+0.1438]$ & $[+0.0572,+0.1277]$ & $[+0.0510,+0.1348]$ & $[+0.0640,+0.1237]$ & yes & no \\
GPT-5.6 Sol & whole, pooled & $+0.1007$ & $[+0.0516,+0.1531]$ & $[+0.0629,+0.1413]$ & $[+0.0583,+0.1443]$ & $[+0.0696,+0.1338]$ & yes & no \\
Opus 5 & whole, pooled & $+0.0057$ & $[-0.0538,+0.0711]$ & $[-0.0780,+0.0847]$ & $[-0.0455,+0.0582]$ & $[-0.0624,+0.0717]$ & no & no \\
\multicolumn{9}{@{}l}{\textit{B6: lookup with one count feature}}\\
count-feature lookup $-$ GPT-5.6 Luna & whole, matched & $+0.0068$ & $[-0.0507,+0.0642]$ & $[-0.0536,+0.0659]$ & $[-0.0405,+0.0541]$ & $[-0.0431,+0.0568]$ & no & no \\
count-feature lookup $-$ GPT-5.6 Sol & whole, matched & $+0.0270$ & $[-0.0203,+0.0812]$ & $[-0.0175,+0.0737]$ & $[-0.0169,+0.0709]$ & $[-0.0104,+0.0658]$ & no & no \\
count-feature lookup $-$ Opus 5 & whole, matched & $+0.0101$ & $[-0.0439,+0.0676]$ & $[-0.0563,+0.0733]$ & $[-0.0338,+0.0608]$ & $[-0.0438,+0.0642]$ & no & no \\
\bottomrule
\end{tabular}}
\end{table}
\begin{table}[H]
\caption{Equivalence within $0.05$ on the paraphrase set (post hoc): every row of Tables~\ref{tab:para-minus-lookup} and~\ref{tab:para-sup}, on the paraphrases and on the original questions of the same pairs; Luna and Sol were added after the registration. Columns and rule as in Table~\ref{tab:equiv}.}
\label{tab:equiv-para}
\centering\tiny\setlength{\tabcolsep}{1.5pt}
\resizebox{\textwidth}{!}{\begin{tabular}{@{}llrllllll@{}}
\toprule
Reader & scope, metric & difference & 95\% item/triple & 95\% persona & 90\% item/triple & 90\% persona & reliable & equivalent \\
\midrule
\multicolumn{9}{@{}l}{\textit{paraphrase set}}\\
GPT-4.1 mini, paraphrases & whole, pooled & $-0.0904$ & $[-0.1453,-0.0352]$ & $[-0.1517,-0.0337]$ & $[-0.1387,-0.0438]$ & $[-0.1405,-0.0424]$ & yes & no \\
GPT-4.1 mini, originals & whole, pooled & $-0.0794$ & $[-0.1303,-0.0290]$ & $[-0.1236,-0.0348]$ & $[-0.1238,-0.0375]$ & $[-0.1179,-0.0404]$ & yes & no \\
GPT-4.1 mini, paraphrases & supersession, pooled & $-0.1195$ & $[-0.1842,-0.0572]$ & $[-0.1959,-0.0512]$ & $[-0.1687,-0.0666]$ & $[-0.1833,-0.0598]$ & yes & no \\
GPT-4.1 mini, originals & supersession, pooled & $-0.1044$ & $[-0.1593,-0.0532]$ & $[-0.1665,-0.0456]$ & $[-0.1475,-0.0594]$ & $[-0.1558,-0.0550]$ & yes & no \\
GPT-4.1 mini, paraphrases & supersession, matched & $-0.1513$ & $[-0.2171,-0.0789]$ & $[-0.2501,-0.0548]$ & $[-0.2105,-0.0855]$ & $[-0.2344,-0.0730]$ & yes & no \\
GPT-4.1 mini, originals & supersession, matched & $-0.1053$ & $[-0.1581,-0.0526]$ & $[-0.1667,-0.0417]$ & $[-0.1513,-0.0592]$ & $[-0.1594,-0.0533]$ & yes & no \\
GPT-4.1 mini, all states, paraphrases & whole, pooled & $-0.1040$ & $[-0.1775,-0.0351]$ & $[-0.1649,-0.0512]$ & $[-0.1645,-0.0499]$ & $[-0.1557,-0.0612]$ & yes & no \\
GPT-4.1 mini, all states, originals & whole, pooled & $-0.0828$ & $[-0.1323,-0.0357]$ & $[-0.1319,-0.0356]$ & $[-0.1255,-0.0416]$ & $[-0.1235,-0.0438]$ & yes & no \\
GPT-4.1 mini, all states, paraphrases & supersession, pooled & $-0.0868$ & $[-0.1583,-0.0115]$ & $[-0.1572,-0.0143]$ & $[-0.1471,-0.0229]$ & $[-0.1448,-0.0308]$ & yes & no \\
GPT-4.1 mini, all states, originals & supersession, pooled & $-0.0703$ & $[-0.1177,-0.0197]$ & $[-0.1320,-0.0099]$ & $[-0.1122,-0.0271]$ & $[-0.1187,-0.0196]$ & yes & no \\
GPT-4.1 mini, all states, paraphrases & supersession, matched & $-0.1118$ & $[-0.1908,-0.0395]$ & $[-0.1875,-0.0384]$ & $[-0.1776,-0.0461]$ & $[-0.1757,-0.0493]$ & yes & no \\
GPT-4.1 mini, all states, originals & supersession, matched & $-0.0724$ & $[-0.1316,-0.0132]$ & $[-0.1583,+0.0000]$ & $[-0.1250,-0.0197]$ & $[-0.1439,-0.0128]$ & no & no \\
Haiku 4.5, paraphrases & whole, pooled & $-0.0568$ & $[-0.1170,+0.0040]$ & $[-0.1066,-0.0074]$ & $[-0.1030,-0.0080]$ & $[-0.0960,-0.0158]$ & no & no \\
Haiku 4.5, originals & whole, pooled & $-0.0142$ & $[-0.0675,+0.0365]$ & $[-0.0557,+0.0346]$ & $[-0.0593,+0.0307]$ & $[-0.0504,+0.0253]$ & no & no \\
Haiku 4.5, paraphrases & supersession, pooled & $-0.0743$ & $[-0.1371,-0.0202]$ & $[-0.1355,-0.0116]$ & $[-0.1256,-0.0274]$ & $[-0.1285,-0.0206]$ & yes & no \\
Haiku 4.5, originals & supersession, pooled & $-0.0304$ & $[-0.0746,+0.0152]$ & $[-0.0833,+0.0222]$ & $[-0.0660,+0.0066]$ & $[-0.0730,+0.0119]$ & no & no \\
Haiku 4.5, paraphrases & supersession, matched & $-0.0526$ & $[-0.1250,+0.0197]$ & $[-0.1205,+0.0236]$ & $[-0.1118,+0.0066]$ & $[-0.1111,+0.0070]$ & no & no \\
Haiku 4.5, originals & supersession, matched & $-0.0592$ & $[-0.1250,+0.0000]$ & $[-0.1181,-0.0068]$ & $[-0.1118,-0.0132]$ & $[-0.1053,-0.0130]$ & no & no \\
Haiku 4.5, all states, paraphrases & whole, pooled & $-0.0704$ & $[-0.1300,-0.0130]$ & $[-0.1230,-0.0149]$ & $[-0.1204,-0.0217]$ & $[-0.1150,-0.0241]$ & yes & no \\
Haiku 4.5, all states, originals & whole, pooled & $-0.0006$ & $[-0.0363,+0.0378]$ & $[-0.0426,+0.0385]$ & $[-0.0317,+0.0320]$ & $[-0.0342,+0.0310]$ & no & yes \\
Haiku 4.5, all states, paraphrases & supersession, pooled & $-0.0879$ & $[-0.1577,-0.0148]$ & $[-0.1488,-0.0241]$ & $[-0.1449,-0.0284]$ & $[-0.1390,-0.0338]$ & yes & no \\
Haiku 4.5, all states, originals & supersession, pooled & $-0.0187$ & $[-0.0572,+0.0290]$ & $[-0.0678,+0.0265]$ & $[-0.0508,+0.0205]$ & $[-0.0593,+0.0185]$ & no & no \\
Haiku 4.5, all states, paraphrases & supersession, matched & $-0.0921$ & $[-0.1645,-0.0132]$ & $[-0.1603,-0.0241]$ & $[-0.1513,-0.0197]$ & $[-0.1500,-0.0337]$ & yes & no \\
Haiku 4.5, all states, originals & supersession, matched & $-0.0395$ & $[-0.1053,+0.0263]$ & $[-0.1081,+0.0260]$ & $[-0.0921,+0.0197]$ & $[-0.0987,+0.0143]$ & no & no \\
Opus 5, draw 1, paraphrases & whole, pooled & $+0.0630$ & $[+0.0078,+0.1134]$ & $[+0.0101,+0.1139]$ & $[+0.0162,+0.1052]$ & $[+0.0188,+0.1066]$ & yes & no \\
Opus 5, draw 1, originals & whole, pooled & $+0.0745$ & $[+0.0230,+0.1252]$ & $[+0.0357,+0.1147]$ & $[+0.0288,+0.1160]$ & $[+0.0414,+0.1075]$ & yes & no \\
Opus 5, draw 1, paraphrases & supersession, pooled & $+0.0226$ & $[-0.0241,+0.0698]$ & $[-0.0357,+0.0806]$ & $[-0.0164,+0.0603]$ & $[-0.0239,+0.0722]$ & no & no \\
Opus 5, draw 1, originals & supersession, pooled & $+0.0136$ & $[-0.0218,+0.0464]$ & $[-0.0249,+0.0513]$ & $[-0.0148,+0.0411]$ & $[-0.0190,+0.0438]$ & no & yes \\
Opus 5, draw 1, paraphrases & supersession, matched & $+0.0263$ & $[-0.0329,+0.0855]$ & $[-0.0375,+0.1000]$ & $[-0.0263,+0.0789]$ & $[-0.0286,+0.0833]$ & no & no \\
Opus 5, draw 1, originals & supersession, matched & $+0.0132$ & $[-0.0066,+0.0395]$ & $[-0.0141,+0.0435]$ & $[-0.0066,+0.0329]$ & $[-0.0081,+0.0398]$ & no & yes \\
Opus 5, draw 1, all states, paraphrases & whole, pooled & $+0.0544$ & $[+0.0013,+0.0983]$ & $[+0.0060,+0.1003]$ & $[+0.0129,+0.0919]$ & $[+0.0148,+0.0921]$ & yes & no \\
Opus 5, draw 1, all states, originals & whole, pooled & $+0.0387$ & $[+0.0053,+0.0718]$ & $[+0.0071,+0.0705]$ & $[+0.0104,+0.0665]$ & $[+0.0123,+0.0661]$ & yes & no \\
Opus 5, draw 1, all states, paraphrases & supersession, pooled & $+0.0607$ & $[+0.0114,+0.1156]$ & $[+0.0009,+0.1256]$ & $[+0.0192,+0.1049]$ & $[+0.0102,+0.1159]$ & yes & no \\
Opus 5, draw 1, all states, originals & supersession, pooled & $+0.0309$ & $[-0.0027,+0.0662]$ & $[-0.0090,+0.0723]$ & $[+0.0022,+0.0606]$ & $[-0.0023,+0.0641]$ & no & no \\
Opus 5, draw 1, all states, paraphrases & supersession, matched & $+0.0526$ & $[+0.0000,+0.1118]$ & $[-0.0068,+0.1071]$ & $[+0.0066,+0.0987]$ & $[+0.0000,+0.0964]$ & no & no \\
Opus 5, draw 1, all states, originals & supersession, matched & $+0.0461$ & $[+0.0000,+0.0987]$ & $[+0.0061,+0.0890]$ & $[+0.0132,+0.0921]$ & $[+0.0071,+0.0833]$ & no & no \\
GPT-5.6 Luna, paraphrases & whole, pooled & $+0.0945$ & $[+0.0485,+0.1401]$ & $[+0.0465,+0.1431]$ & $[+0.0556,+0.1330]$ & $[+0.0544,+0.1336]$ & yes & no \\
GPT-5.6 Luna, originals & whole, pooled & $+0.0763$ & $[+0.0305,+0.1192]$ & $[+0.0400,+0.1123]$ & $[+0.0381,+0.1141]$ & $[+0.0467,+0.1072]$ & yes & no \\
GPT-5.6 Luna, paraphrases & supersession, pooled & $+0.0445$ & $[+0.0093,+0.0845]$ & $[+0.0042,+0.0895]$ & $[+0.0152,+0.0785]$ & $[+0.0096,+0.0804]$ & yes & no \\
GPT-5.6 Luna, originals & supersession, pooled & $+0.0255$ & $[-0.0003,+0.0520]$ & $[-0.0040,+0.0514]$ & $[+0.0044,+0.0483]$ & $[+0.0004,+0.0466]$ & no & yes \\
GPT-5.6 Luna, paraphrases & supersession, matched & $+0.0395$ & $[-0.0066,+0.0921]$ & $[-0.0071,+0.1053]$ & $[+0.0000,+0.0789]$ & $[-0.0061,+0.0903]$ & no & no \\
GPT-5.6 Luna, originals & supersession, matched & $-0.0197$ & $[-0.0658,+0.0132]$ & $[-0.0556,+0.0072]$ & $[-0.0526,+0.0066]$ & $[-0.0507,+0.0063]$ & no & no \\
GPT-5.6 Sol, paraphrases & whole, pooled & $+0.0891$ & $[+0.0370,+0.1384]$ & $[+0.0374,+0.1380]$ & $[+0.0458,+0.1312]$ & $[+0.0458,+0.1290]$ & yes & no \\
GPT-5.6 Sol, originals & whole, pooled & $+0.0824$ & $[+0.0379,+0.1300]$ & $[+0.0443,+0.1234]$ & $[+0.0442,+0.1219]$ & $[+0.0493,+0.1147]$ & yes & no \\
GPT-5.6 Sol, paraphrases & supersession, pooled & $+0.0396$ & $[-0.0063,+0.0845]$ & $[-0.0176,+0.0929]$ & $[+0.0002,+0.0774]$ & $[-0.0075,+0.0850]$ & no & no \\
GPT-5.6 Sol, originals & supersession, pooled & $+0.0172$ & $[-0.0169,+0.0501]$ & $[-0.0228,+0.0517]$ & $[-0.0103,+0.0457]$ & $[-0.0160,+0.0470]$ & no & yes \\
GPT-5.6 Sol, paraphrases & supersession, matched & $+0.0263$ & $[-0.0395,+0.0855]$ & $[-0.0390,+0.0974]$ & $[-0.0263,+0.0789]$ & $[-0.0308,+0.0833]$ & no & no \\
GPT-5.6 Sol, originals & supersession, matched & $-0.0197$ & $[-0.0658,+0.0197]$ & $[-0.0596,+0.0130]$ & $[-0.0592,+0.0132]$ & $[-0.0530,+0.0074]$ & no & no \\
\bottomrule
\end{tabular}}
\end{table}

\paragraph{The length-only reader, post hoc.}\label{app:length}
The default lists are short: the \pn{} version's superseded list holds $2.6$ names on average and at most $6$ on the supersession triples, and $24$ of the $88$ \present{} lists are empty; at these lengths the lookup is close to a string match against one or two candidates (Table~\ref{tab:cf-emptiness}; FR-Bank's own lists in Appendix~\ref{app:census}).
A reader whose score is the number of names in the superseded-names section of the summary it is given, and nothing else, is scored as the count readers are (Table~\ref{tab:length}). On the supersession triples the \pn{} version's list is longer than each sibling's in every triple, on the default and the padded lists alike (by one name on $81$ triples, two on $6$ and three on $1$), so the reader wins all $88$ triples on both; on the padded lists, where every \pn{} list holds $60$ names and every sibling's fewer, it is also perfect pooled. On the sibling-name control the lists have equal length by construction and the reader ties every triple. The pad is identical across a triple, so the padded set carries the length difference the default lists already carry; the sibling padded to equal length is the control of the next paragraph (Table~\ref{tab:sb60}).

\begin{table}[H]
\caption{The length-only reader (post hoc): its score is the number of names in the superseded-names section of the summary it is given, and nothing else, scored as the count readers are. Pooled AUROC with the persona-grouped interval; matched W/T/L and score with the triple-level and persona-grouped intervals. The sibling-name control compares the \pn{} version with its sibling version. \pn{} list length minus each sibling's, per triple and sibling: padded lists, supersession: $1$ on $162$, $2$ on $12$, $3$ on $2$; padded lists, whole: $0$ on $120$, $1$ on $162$, $2$ on $12$, $3$ on $2$; unpadded lists, supersession: $1$ on $162$, $2$ on $12$, $3$ on $2$; unpadded lists, whole: $0$ on $120$, $1$ on $162$, $2$ on $12$, $3$ on $2$; sibling lists, supersession: $0$ on $88$.}
\label{tab:length}
\centering\tiny\setlength{\tabcolsep}{2pt}
\begin{tabular}{@{}llrrrlrrll@{}}
\toprule
Summary & scope & items & triples & pooled & persona interval & W/T/L & matched & triple interval & persona interval \\
\midrule
padded to $60$ & whole & $444$ & $148$ & $0.7973$ & $[0.7552,0.8404]$ & $88/60/0$ & $0.7973$ & $[0.7568,0.8378]$ & $[0.7552,0.8404]$ \\
padded to $60$ & supersession & $264$ & $88$ & $1.0000$ & $[1.0000,1.0000]$ & $88/0/0$ & $1.0000$ & $[1.0000,1.0000]$ & $[1.0000,1.0000]$ \\
padded to $60$ & retraction & $90$ & $30$ & $0.5000$ & $[0.5000,0.5000]$ & $0/30/0$ & $0.5000$ & $[0.5000,0.5000]$ & $[0.5000,0.5000]$ \\
padded to $60$ & expiry & $90$ & $30$ & $0.5000$ & $[0.5000,0.5000]$ & $0/30/0$ & $0.5000$ & $[0.5000,0.5000]$ & $[0.5000,0.5000]$ \\
default (short list) & whole & $444$ & $148$ & $0.6295$ & $[0.6078,0.6654]$ & $88/60/0$ & $0.7973$ & $[0.7568,0.8378]$ & $[0.7552,0.8404]$ \\
default (short list) & supersession & $264$ & $88$ & $0.7199$ & $[0.6949,0.7716]$ & $88/0/0$ & $1.0000$ & $[1.0000,1.0000]$ & $[1.0000,1.0000]$ \\
default (short list) & retraction & $90$ & $30$ & $0.5000$ & $[0.5000,0.5000]$ & $0/30/0$ & $0.5000$ & $[0.5000,0.5000]$ & $[0.5000,0.5000]$ \\
default (short list) & expiry & $90$ & $30$ & $0.5000$ & $[0.5000,0.5000]$ & $0/30/0$ & $0.5000$ & $[0.5000,0.5000]$ & $[0.5000,0.5000]$ \\
sibling-name control & supersession & $176$ & $88$ & $0.5000$ & $[0.5000,0.5000]$ & $0/88/0$ & $0.5000$ & $[0.5000,0.5000]$ & $[0.5000,0.5000]$ \\
\bottomrule
\end{tabular}
\end{table}

\paragraph{The sibling-name control at $60$ names.}\label{app:sb60}
A fourth version of each of the $88$ supersession pairs takes the padded \pn{} text and swaps the target's name for a same-category sibling, so the two lists hold the same $60$ names but one, under an identical pad and an identical count table (Table~\ref{tab:sb60}). The sibling is the pair's sibling from the short-list control; where that name was already in the pad, which happened on $35$ pairs, a substitute was drawn from the pair's remaining eligible names under a recorded seed. The renderer reproduces every text byte for byte, every list holds $60$ distinct names, and the leak screen found $0$ of $30$ answers correct. Before the freeze the name lookup was recorded at $0.8239$ matched and the length-only reader, the count reader and the random-token lookup at $0.5000$. Of the four predictions, Luna reliably above the lookup came out otherwise (matched $+0.0398$, the persona interval reaching below zero; pooled $+0.1122$, reliable); Sol reliably above it held ($+0.0852$ matched, $+0.1238$ pooled); GPT-4.1 mini not reliably above it held ($-0.0909$ matched); and Luna and Sol at or above $0.80$ with the persona interval above $0.50$ held ($0.8636$ and $0.9091$). Seeds $43$ to $46$ change no reading. Luna, Sol and mini ran five draws each through the raw interface; Haiku 4.5 and Opus 5 were not run. Section~\ref{sec:lookup} words Luna's result as holding on the pooled metric only, a wording deviation from the registered branch clause (\emph{Luna's lead is not separated from the cue}): the registration promised the pooled reading beside the matched one, and the paper names the pooled metric wherever the two disagree.

\begin{table}[H]
\caption{The sibling-name control at $60$ names (registered; ADDENDUM\_R141\_SB60): the \pn{} version with its list padded to $60$ names against a fourth version with one name swapped for a same-category sibling, identical pad and counts, on the $88$ supersession pairs. Upper panel: each reader's matched score (W + T/2)/88 with the triple-level and persona-grouped intervals, the pooled AUROC over the $176$ items, and the sibling against the padded \present{} version (descriptive). The no-model rows were recorded before the freeze. Lower panel: each model minus the name lookup, paired; $^{*}$: both $95\%$ intervals exclude zero; equivalent: both $90\%$ intervals inside $[-0.05,+0.05]$. Haiku 4.5 and Opus 5 were not run on this control.}
\label{tab:sb60}
\centering\tiny\setlength{\tabcolsep}{2pt}
\begin{tabular}{@{}lrrllrr@{}}
\toprule
Reader & W/T/L & matched & triple interval & persona interval & pooled & sibling vs \present{} \\
\midrule
Name lookup (embedding) & $57/31/0$ & $0.8239$ & $[0.7727,0.8750]$ & $[0.7670,0.8816]$ & $0.7395$ & $0.5227$ \\
Lexical lookup & $42/46/0$ & $0.7386$ & $[0.6875,0.7898]$ & $[0.6875,0.7927]$ & $0.6815$ & $0.5114$ \\
Random-token lookup & $1/86/1$ & $0.5000$ & $[0.4830,0.5170]$ & $[0.4837,0.5170]$ & $0.5006$ & $0.5057$ \\
Length-only reader & $0/88/0$ & $0.5000$ & $[0.5000,0.5000]$ & $[0.5000,0.5000]$ & $0.5000$ & $1.0000$ \\
Count reader, guessed category (share) & $0/88/0$ & $0.5000$ & $[0.5000,0.5000]$ & $[0.5000,0.5000]$ & $0.5000$ & $0.6932$ \\
GPT-5.6 Luna (five draws) & $76/0/12$ & $0.8636$ & $[0.7955,0.9318]$ & $[0.8043,0.9259]$ & $0.8517$ & $0.6818$ \\
GPT-5.6 Sol (five draws) & $80/0/8$ & $0.9091$ & $[0.8523,0.9659]$ & $[0.8506,0.9620]$ & $0.8633$ & $0.5114$ \\
GPT-4.1 mini (five draws) & $62/5/21$ & $0.7330$ & $[0.6420,0.8239]$ & $[0.6455,0.8193]$ & $0.5931$ & $0.6307$ \\
\bottomrule
\end{tabular}\\[4pt]
\begin{tabular}{@{}llrllr@{}}
\toprule
Model minus the name lookup & metric & difference & triple / item interval & persona interval & equivalent \\
\midrule
GPT-5.6 Luna & matched & $+0.0398$ & $[-0.0284,+0.1080]$ & $[-0.0112,+0.0987]$ & no \\
GPT-5.6 Luna & pooled & $+0.1122$$^{*}$ & $[+0.0521,+0.1732]$ & $[+0.0688,+0.1575]$ & no \\
GPT-5.6 Sol & matched & $+0.0852$$^{*}$ & $[+0.0284,+0.1420]$ & $[+0.0322,+0.1398]$ & no \\
GPT-5.6 Sol & pooled & $+0.1238$$^{*}$ & $[+0.0615,+0.1911]$ & $[+0.0812,+0.1645]$ & no \\
GPT-4.1 mini & matched & $-0.0909$ & $[-0.1818,+0.0001]$ & $[-0.1889,+0.0108]$ & no \\
GPT-4.1 mini & pooled & $-0.1464$$^{*}$ & $[-0.2110,-0.0778]$ & $[-0.1920,-0.1006]$ & no \\
\bottomrule
\end{tabular}
\end{table}

\paragraph{The expiry state extended by a third descriptive draw.}
The candidate census behind the counterfactual draw (in the artifact) found $19$ expiry candidates that our read had passed and the draw had left undrawn, $14$ of them within the cap of four pairs per persona and the one-pair-per-name rule.
A third descriptive registration draws those $14$, builds their three store versions by the same construction, drafts their questions through the same renderer before any version exists, and scores the same readers under the same summaries; its questions and summaries pass the same leakage screen ($0$ of $14$ on every summary).
Table~\ref{tab:cf3-expiry} prints the extension beside the $444$'s expiry pairs and pools the two, $44$ pairs, with the pooling stated in the caption.
The counted-state group stays small with it, $44$ expiry pairs against $88$ supersession pairs.
On the extension the all-states lookup reads expiry at $0.97$ and GPT-4.1 mini at $0.68$ under the all-states summary; retraction stays at $30$ pairs, because one candidate in ten passes our read for a cancellable plan ($36$ of $370$ read), and the candidate census in the artifact records every candidate with the rule that removed it.

\begin{table}[H]
\caption{The expiry state of the counterfactual set extended by a third descriptive draw: the $444$'s $30$ expiry pairs ($90$ items), the $14$ eligible pairs the first draw left undrawn ($42$ items, built by the same construction with questions written before any store version existed), and the two pooled ($44$ pairs, $132$ items). Obligation AUROC per reader and summary; question only (blind) is $0.5000$ by construction on every set, so each value is also a gain. Intervals are the weaker of the item-level and persona-grouped paired intervals for the models and the persona interval of the AUROC for the lookups. The pooled row joins two cells scored on different dates under two registrations with the same prompts, models and construction; Opus 5's rows are five-call averages on the $444$ and single calls on the extension, so its pooled row mixes the two; the rows below them read Opus 5 at its first draw on every pair, so the pooled value there mixes nothing. Descriptive; no prediction was registered.}
\label{tab:cf3-expiry}
\centering
\scriptsize
\setlength{\tabcolsep}{2.5pt}
\begin{tabular}{@{}llrrr@{}}
\toprule
Reader & summary & $444$ expiry ($30$ pairs) & extension ($14$ pairs) & pooled ($44$ pairs) \\
\midrule
Name lookup & sup.\ only & $0.5000$ $[0.5000,0.5000]$ & $0.5000$ $[0.5000,0.5000]$ & $0.5000$ $[0.5000,0.5000]$ \\
All-states lookup & all three & $0.8711$ $[0.7864,0.9424]$ & $0.9694$ $[0.9475,1.0000]$ & $0.8977$ $[0.8376,0.9477]$ \\
GPT-4.1 mini & sup.\ only & $0.5489$ $[-0.1022,+0.1893]$ & $0.5421$ $[-0.1446,+0.2222]$ & $0.5461$ $[-0.0628,+0.1585]$ \\
GPT-4.1 mini & all three & $0.5869$ $[-0.0524,+0.2230]$ & $0.6811$ $[-0.0308,+0.3901]$ & $0.6139$ $[-0.0042,+0.2198]$ \\
GPT-4.1 mini & totals rand. & $0.5144$ $[-0.1317,+0.1534]$ & $0.5651$ $[-0.1021,+0.2373]$ & $0.5302$ $[-0.0797,+0.1325]$ \\
Haiku 4.5 & sup.\ only & $0.5794$ $[-0.0626,+0.2182]$ & $0.6059$ $[-0.0441,+0.2595]$ & $0.5872$ $[-0.0307,+0.1932]$ \\
Haiku 4.5 & all three & $0.8269$ $[+0.1935,+0.4675]$ & $0.9630$ $[+0.2917,+0.6377]$ & $0.8644$ $[+0.2552,+0.4676]$ \\
Haiku 4.5 & totals rand. & $0.5847$ $[-0.0594,+0.2188]$ & $0.6212$ $[-0.0321,+0.2577]$ & $0.5980$ $[-0.0186,+0.2056]$ \\
Opus 5 & sup.\ only & $0.6150$ $[+0.0077,+0.2282]$ & $0.5638$ $[-0.0867,+0.2840]$ & $0.5914$ $[-0.0061,+0.1930]$ \\
Opus 5 & all three & $0.9267$ $[+0.2974,+0.5637]$ & $0.9452$ $[+0.2593,+0.6318]$ & $0.9232$ $[+0.3246,+0.5218]$ \\
Opus 5 & totals rand. & $0.6092$ $[+0.0008,+0.2140]$ & $0.6084$ $[-0.0830,+0.2792]$ & $0.6068$ $[+0.0115,+0.1974]$ \\
Opus 5, draw 1 & sup.\ only & $0.6169$ $[+0.0059,+0.2313]$ & $0.5638$ $[-0.0867,+0.2840]$ & $0.5917$ $[-0.0053,+0.1869]$ \\
Opus 5, draw 1 & all three & $0.9231$ $[+0.3006,+0.5502]$ & $0.9452$ $[+0.2593,+0.6318]$ & $0.9247$ $[+0.3291,+0.5149]$ \\
Opus 5, draw 1 & totals rand. & $0.6047$ $[-0.0081,+0.2187]$ & $0.6084$ $[-0.0830,+0.2792]$ & $0.6043$ $[+0.0081,+0.1967]$ \\
\bottomrule
\end{tabular}
\end{table}

\section{Monitoring-scoped instrument ledger}\label{app:ledger}

Every failure in the ledger is the same kind, a semantic assumption hidden inside a step that looked mechanical, so the ledger is kept as one list.
The following ledger contains only measurement steps germane to E0, E1, the cross-model set, the question-signal audit, and their economics; measurement steps outside this paper's monitoring scope are excluded.
The lesson: printing a quantity beside the rule that constrains it makes the redundancy a detector for exactly this kind of failure.
The twelve entries of the ledger, each stating what went wrong and what changed, are in the supplement (Section~S.1), because each is a record rather than a result.
The supplement also carries the account of the four failures of the same kind in this manuscript's own verification battery.

\section{Experimental and reproducibility detail}\label{app:repro}

\subsection{Notation and study glossary}\label{app:glossary}

The paper uses one vocabulary and the artifact records another.
Three pairs of terms recur in the appendix prose before the table at the end of this section defines them: \emph{blind} and \emph{indexed} are the main text's question only and with the summary; \emph{route A} and \emph{route B} are the summary route and the tag route; and \emph{the monitor} is the reader, a name lookup or a model.
The two blocks below define every term the main text uses, and a third block in the supplement (Section~S.5) maps the internal names a reader meets in the registrations, ledgers and row files onto the paper's term.
The notation box and the definition of the lifecycle summary follow the two blocks, so that every symbol the appendices use is defined here.
Two labeled queries from the hiking persona illustrate Definition~\ref{def:coverage}.
Under the current intent ``What's David's family background?'' is \present{}.
Under the upcoming intent ``Does David have any upcoming exams to prepare students for?'' is \pn{}, because the store holds an exam record whose date has passed.

\begin{center}
\fbox{\begin{minipage}{0.97\linewidth}
\ttfamily\tiny
MEMORY STORE INDEX  (metadata only; no fact content)\\
entries in this store: 575\\[1mm]
CATEGORY\hspace{18mm} ENTRIES\quad ACTIVE\quad SUPRSD\quad RETRACT\quad EXPIRED\\
HOBBIES\_RECREATION\hspace{10mm}29\hspace{8mm}18\hspace{8mm}11\hspace{10mm}0\hspace{10mm}2\\
PREFERENCES\_HABITS\hspace{8mm}46\hspace{8mm}37\hspace{9mm}9\hspace{10mm}0\hspace{10mm}1\\
PROJECTS\_ENDEAVORS\hspace{8mm}61\hspace{8mm}54\hspace{9mm}7\hspace{10mm}0\hspace{9mm}18\\[1mm]
AGES (days)\quad newest 1\quad oldest 518\quad median 258\quad bands 0-30: 47, 31-90: 54, 91-180: 75, 181-365: 229, 365+: 170\\[1mm]
ATTRIBUTE NAMES BY CATEGORY (value-scrubbed; abridged)\\
HOBBIES\_RECREATION: hiking\_experience, hiking\_preference, hiking\_preference\_scenery, ...\\
PROJECTS\_ENDEAVORS: camp\_stove\_status, project\_status, trip, trip\_plan, ...\\
ATTRIBUTE NAMES CARRYING SUPERSEDED CONTENT (abridged): hiking\_preference, ...
\end{minipage}}
\end{center}
\noindent\textit{This is what the indexed monitor reads}, an abridged, real FR-Bank summary that exposes counts and attribute names but no stored value.
The EXPIRED column is a flag derived from each entry's event time and the ACTIVE column counts the entries that are neither superseded nor retracted, so an expired entry can also be counted ACTIVE and the four lifecycle columns overlap; a record carrying more than one mark is labelled by the first of retracted, superseded, expired (Section~\ref{sec:framework}).

\begin{center}
\fbox{\begin{minipage}{0.97\linewidth}
\scriptsize
\textbf{The two judgments, and the label they are scored against.}\par\smallskip
\begin{tabular}{@{}p{.20\linewidth}p{.74\linewidth}@{}}
FOK & feeling of knowing: the score elicited as a probability that the store covers the query under its intent \\
NOK & negative-obligation knowledge: the score elicited as a probability that the store holds content whose lifecycle state forbids use \\
\pn{} & present-negative: the label for a query whose relevant content is present and would make the answer wrong if used \\
intent & the time frame a query asks about: \textsc{current}, \textsc{upcoming}, \textsc{past}, or \textsc{change} \\
lifecycle state & \textsc{active}, \textsc{superseded}, \textsc{expired}, or \textsc{retracted} \\
AUROC & the probability that a randomly chosen positive scores above a randomly chosen negative \\
ECE & expected calibration error: the gap between stated probabilities and observed frequencies \\
\end{tabular}
\end{minipage}}
\end{center}

\begin{center}
\fbox{\begin{minipage}{0.97\linewidth}
\scriptsize
\textbf{Conditions, stores, and populations.}\par\smallskip
\begin{tabular}{@{}p{.20\linewidth}p{.74\linewidth}@{}}
blind / indexed & the monitor sees the question only / also sees the metadata summary \\
the gain ($\Delta$) & the indexed score minus the blind score on the same questions \\
ground-truth / store-written summary & built from ground truth / from the fields the backend itself exposes \\
easy / hard tier & a whole category removed / a target missing inside a populated category \\
monitor call & one pre-retrieval request returning FOK and NOK \\
FR-Bank & the lifecycle-tracking store of our earlier system \citep{fr}, the one evaluated store that records a lifecycle state on every entry \\
Mem0 & a production memory layer whose \texttt{UPDATE} overwrites history \\
A-MEM & an agentic memory system organized around linked notes \\
pilot / main study & the feasibility probe / the registered study \\
mini / luna / sol & \texttt{gpt-4.1-mini} / \texttt{gpt-5.6-luna} / \texttt{gpt-5.6-sol}, in ascending size and price \\
\end{tabular}
\end{minipage}}
\end{center}

\begin{center}
\setlength{\fboxsep}{1.5pt}%
\fbox{\begin{minipage}{0.97\linewidth}
\scriptsize
\setlength{\tabcolsep}{1.5pt}
\renewcommand{\arraystretch}{0.92}
\noindent\textbf{Notation.} The terms below are used verbatim throughout.\par
\noindent\begin{tabular}[t]{@{}p{0.115\linewidth}p{0.36\linewidth}@{}}
\midrule
lifecycle state; non-active & the tag a record carries, \textsc{active}, \textsc{superseded}, \textsc{retracted} or \textsc{expired}; the last three are the non-active states \\
forbidding record & the non-active record that would make an answer wrong if used \\
intent; \pn{} & the time frame a question asks about (\textsc{current}, \textsc{upcoming}, \textsc{past}, \textsc{change}); present-negative: the store holds relevant content and using it would make the answer wrong. An \emph{obligation} is a question whose label is \pn{} \\
FOK; NOK & the coverage judgment, is the question covered; the obligation judgment, does the store hold anything not to use \\
summary & the count table (records per category and lifecycle state) and the name map (the attribute names each category exposes, values removed); the superseded-names list is the part of the name map that carries superseded content \\
\end{tabular}\hfill\begin{tabular}[t]{@{}p{0.115\linewidth}p{0.36\linewidth}@{}}
\midrule
names, counts & a summary \emph{names} a state when it lists the attribute names in that state and \emph{counts} it when it gives only a count of it \\
blind; indexed & the monitor sees the question alone; the monitor sees the question and the summary \\
ground-truth summary; store-written summary & the summary built from ground truth; the summary built from the fields the backend itself exposes \\
name lookup & the similarity lookup over the summary's superseded-names list, the baseline with no generative-model call that every model is measured against \\
padded list & a superseded-names list lengthened to a fixed size ($60$ names) with correct attribute names of the same store that no scored question targets, the counts left as written \\
\end{tabular}
\end{minipage}}
\end{center}
\begin{definition}[Lifecycle summary]\label{def:index}
A summary $M(S)$ is maintained outside query time. Ours pairs a count table with an attribute-name map from which every stored value is removed. The count table holds each category--lifecycle count and, per category, the count of updates that superseded an entry. The name map records which names each category exposes and which carry superseded content; the names that carry superseded content form the superseded-names list. The size $|M(S)|$ is $O(|C|\cdot|\Xi|+|N|)$ in the number $N$ of distinct names exposed, and for a fixed name vocabulary it is independent of the number of records $|S|$.
\end{definition}
\noindent A summary \emph{names} a state when it lists the attribute names in that state and \emph{counts} it when it gives only a count of it; a reader \emph{reads} a state when its score separates that state's items from the rest.

\paragraph{The artifact's names in the registered texts, and the appendix prose's terms in the main text.}
The registered texts quoted in Appendices~\ref{app:counterfactual}, \ref{app:baselines}, \ref{app:census}, \ref{app:resp} and~\ref{app:registration} are reproduced byte for byte, so they keep the artifact's names.
Six of those names differ from this paper's terms, and eight terms of the appendix prose differ from the main text's; the table gives both.
\begin{center}
\scriptsize
\begin{tabular}{@{}ll@{}}
\toprule
In the registered texts or the appendix prose & In this paper's main text \\
\midrule
\textsc{true} (the truthful summary) & the ground-truth summary; on Mem0 and A-MEM, the constrained rendering \\
\textsc{believed} (the believed summary) & the store-written summary \\
the join & ground-truth tags \\
the matcher & the name lookup \\
the smallest model & GPT-4.1 mini \\
user profile & persona \\
\midrule
blind & question only (the reader sees the question alone) \\
indexed & with the summary (the reader sees the question and the summary) \\
route A & the summary route \\
route B & the tag route \\
NOK & the obligation judgment \\
FOK & the coverage judgment \\
the monitor & the reader (a name lookup or a model) \\
the structureless stores & Mem0 and A-MEM, the two backends that record no lifecycle state \\
\bottomrule
\end{tabular}
\end{center}

\subsection{Pilot estimands}\label{app:pilot}

The pilot preceded the main study with a different prompt, so its values support the design lesson and cannot be compared numerically with the main study.
Its estimands and their table are in the supplement (Section~S.7).

\subsection{Main-study construction and checks}\label{app:construction}

This appendix carries the two user prompts, the obligation half of the rubric, and the construction record of the main study, because every reported estimate is bounded by one of them.
Listing~\ref{lst:summary} is the abridged summary the indexed monitor sees, rendered from the store of one persona.

\paragraph{Provenance of the personas, the stores and the questions.}
The persona bank, the three main stores and the questions of the main study come from the FR-Bank release \citep{fr}; the question sets this study added were drafted by language models.
The release holds $40$ personas, each with a human-written manifest and $35$ language-model-generated sessions ($1{,}400$ in all), the FR-Bank, Mem0 and A-MEM stores built from those sessions, and $516$ questions.
This study asks $187$ of those questions of the full FR-Bank store and its two ablated variants, the category-removed and the slot-removed store ($178$, $176$ and $162$ questions; $516$ of the $661$ items of the full obligation set).
It adds the $145$ natural-absent probes, drafted by a language model and verified against the manifests.
GPT-4.1 mini drafted, through frozen renderers, the counterfactual set's $148$ questions with their three store versions ($444$ items) and the expiry extension ($14$ pairs, $42$ items).
It also drafted the $113$ crossed intent questions asked of $118$ drafted and the $148$ paraphrases.
The cross-vendor set is $84$ of the $661$ items ($76$ with a released question, $8$ probes) and the question-only probes are classifiers over the existing question strings, so neither adds a question.
The Graphiti graphs, one per persona, were built by the unmodified release from $1{,}365$ episodes rendered from the FR-Bank store's records, one per session, with every lifecycle field removed (Appendix~\ref{app:posthoc-graphiti}).

Summary size grows with categories, states and distinct names; at a fixed vocabulary, it does not grow with record count. Counts remain because the design fixed them before the study.

\begin{listing}[h]
\caption{An abridged real summary: the count table (lines two and three), the name map (line four) and the superseded-names list (line five).}
\label{lst:summary}
\centering
\fbox{\begin{minipage}{0.97\linewidth}
\ttfamily\scriptsize
MEMORY STORE INDEX  (metadata only; no fact content)\\
CATEGORY\hspace{16mm} ENTRIES\quad ACTIVE\quad SUPRSD\quad RETRACT\quad EXPIRED\\
HOBBIES\_RECREATION\hspace{8mm}29\hspace{8mm}18\hspace{8mm}11\hspace{10mm}0\hspace{10mm}2\\
ATTRIBUTE NAMES BY CATEGORY (value-scrubbed; abridged)\quad HOBBIES\_RECREATION: hiking\_experience, hiking\_preference, ...\\
ATTRIBUTE NAMES CARRYING SUPERSEDED CONTENT (abridged): hiking\_preference, ...
\end{minipage}}
\end{listing}

\begin{center}
\fbox{\begin{minipage}{0.97\linewidth}
\scriptsize
\begin{minipage}[t]{0.28\linewidth}
\textbf{Blind user prompt}\par\smallskip
\ttfamily Question: \{question\}
\end{minipage}\hfill
\begin{minipage}[t]{0.68\linewidth}
\textbf{Indexed user prompt}\par\smallskip
\ttfamily Metadata summary of this user's memory store (structure and counts only -- no fact content). Fields this memory system cannot record are marked as such; a field marked ``not expressible by this backend'' means the system does not track it, which is different from tracking it and finding none.\par\smallskip
\char0\char0\char0\{summary\}\char0\char0\char0\par\smallskip
Question: \{question\}
\end{minipage}
\end{minipage}}
\end{center}
\noindent\textit{The two user prompts.} The indexed condition inserts the serialized summary, and the blind condition supplies only the question.

One call elicits both judgments. The coverage half of the rubric read in full: ``FOK (feeling of knowing): the probability, 0 to 100, that the store contains content relevant to this question -- anything that should shape the answer.'' The reply is one JSON object, \texttt{\{"fok": <0-100>, "nok\_candidates\_found": <true|false>, "nok": <0-100>, "would\_search": <true|false>\}}, and the system text adds that the two judgments are independent (``A store can hold plenty of relevant content with nothing stale in it, and it can hold stale content on a topic it otherwise covers poorly.'') and that the model must not attempt to answer the question; \texttt{would\_search} is not scored.
The obligation half of the elicitation rubric, which Section~\ref{sec:setup} points to, read in full:
``NEGATIVE OBLIGATION: content the store still holds whose correct influence on the answer is SUPPRESSION rather than use, because it has been superseded, retracted, or has expired. Judge this in two steps. Step 1, nok\_candidates\_found: is there content on this topic in the store that has been superseded by a later value, retracted by the user, or has passed its date? true or false. Step 2, nok: the probability, 0 to 100, that such content is present.'' Step 2 is the scored quantity.
For four of the five models the main-study rubric does not elicit the intent condition; Opus 5 applies it unprompted.
In the main populations every \pn{} question has current or upcoming intent by construction (Table~\ref{tab:e3-intent}), but a staleness judgment and an obligation judgment do not coincide even there: $44$ of the $174$ \present{} questions hold a non-active record of their own attribute, $33$ under current or upcoming intent, where the label follows the record the question targets, and $11$ under change intent, where intent licenses the old content; the crossover of Appendix~\ref{app:counterfactual} is where the two judgments are separated on one store.
\begin{table}[H]
\caption{The $661$ questions of the full obligation set by intent, label and the state of the question's own attribute. For a \pn{} question the state is the forbidding record's, from the manifest; for a \present{} question it is the first of retracted, superseded and expired that the ground-truth join puts on any record of the question's own attribute in its store version, else active only (unaligned: no gold record is aligned, the ten past-event questions); an \absent{} question has no such record. Every \pn{} question has current or upcoming intent, and $44$ \present{} questions hold a non-active record of their own attribute: $33$ under current or upcoming intent, where the label follows the record the question targets, and $11$ under change intent, where intent licenses the old content (the $4$ with a superseded record are the items Appendix~\ref{app:counterfactual} counts).}
\label{tab:e3-intent}
\centering\scriptsize\setlength{\tabcolsep}{4pt}
\begin{tabular}{@{}llrrrrrrr@{}}
\toprule
Intent & Label & superseded & retracted & expired & active only & unaligned & none & total \\
\midrule
\textsc{current} & \pn{} & $63$ & $26$ & $0$ & $0$ & $0$ & $0$ & $89$ \\
\textsc{current} & \present{} & $1$ & $1$ & $15$ & $68$ & $0$ & $0$ & $85$ \\
\textsc{current} & \absent{} & $0$ & $0$ & $0$ & $0$ & $0$ & $274$ & $274$ \\
\textsc{upcoming} & \pn{} & $0$ & $0$ & $15$ & $0$ & $0$ & $0$ & $15$ \\
\textsc{upcoming} & \present{} & $5$ & $3$ & $8$ & $18$ & $0$ & $0$ & $34$ \\
\textsc{upcoming} & \absent{} & $0$ & $0$ & $0$ & $0$ & $0$ & $49$ & $49$ \\
\textsc{past} & \present{} & $0$ & $0$ & $0$ & $0$ & $10$ & $0$ & $10$ \\
\textsc{change} & \present{} & $3$ & $3$ & $5$ & $34$ & $0$ & $0$ & $45$ \\
\textsc{change} & \absent{} & $0$ & $0$ & $0$ & $0$ & $0$ & $60$ & $60$ \\
\bottomrule
\end{tabular}
\end{table}

Table~\ref{tab:e3-groups} scores the $104$ obligations against each of the three \present{} groups.
Group (a)'s non-active record is an unrelated fact under the same broad key, so the labels stand: for ``Does Bruna have any upcoming events or deadlines?'' the expired record under \texttt{event} is a marathon she had already run, while the conference she is to present at is active, and for ``What's Elena's family background?'' the expired record under \texttt{family\_member} is a sister's birthday reminder, while the family facts are active.
Every model separates the obligations from group (a) less well than from group (c), $0.57$ to $0.68$ against $0.69$ to $0.83$ with the summary.
Question only, the three models that score anything without a summary already rank group (a) above the obligations ($0.29$ to $0.40$, below chance), and the summary reverses that reliably without closing the gap; the name lookup, which reads no question, separates the obligations from each group about equally ($0.74$, $0.69$ and $0.74$).

\begin{table}[H]
\caption{The $104$ obligations against the three \present{} groups of Table~\ref{tab:e3-intent}: (a) the $33$ whose own attribute holds a non-active record under current or upcoming intent, (b) the $11$ under change intent, (c) the remaining $130$. Upper panel: the mean score question only and with the summary, and the AUROC of the obligations against the group with the summary, with the item-level and persona-grouped intervals ($1{,}000$ draws). Lower panel: the question-only AUROC and the paired gain from adding the summary, with both intervals. GPT-4.1 mini's rows are one call; Haiku 4.5, Opus 5 and GPT-5.6 Luna the five-call collapse; the name lookup reads the superseded section and has no question-only form. Descriptive; read after the results were known.}
\label{tab:e3-groups}
\centering\tiny\setlength{\tabcolsep}{3pt}
\begin{tabular}{@{}llrrrrll@{}}
\toprule
Reader & group & $n$ & mean, question only & mean, with summary & AUROC, with summary & item interval & persona interval \\
\midrule
GPT-4.1 mini & obligations & $104$ & $0.0000$ & $0.2591$ & & & \\
 & (a) & $33$ & $0.0000$ & $0.0773$ & $0.6801$ & $[0.5924,0.7619]$ & $[0.5649,0.7791]$ \\
 & (b) & $11$ & $0.0000$ & $0.0455$ & $0.6875$ & $[0.5809,0.7815]$ & $[0.5574,0.7973]$ \\
 & (c) & $130$ & $0.0015$ & $0.0296$ & $0.6947$ & $[0.6261,0.7659]$ & $[0.6051,0.7868]$ \\
Haiku 4.5 & obligations & $104$ & $0.1201$ & $0.3399$ & & & \\
 & (a) & $33$ & $0.3559$ & $0.2177$ & $0.6716$ & $[0.5738,0.7598]$ & $[0.5623,0.7620]$ \\
 & (b) & $11$ & $0.1340$ & $0.1800$ & $0.7019$ & $[0.5759,0.8049]$ & $[0.5549,0.8349]$ \\
 & (c) & $130$ & $0.1787$ & $0.1216$ & $0.7889$ & $[0.7238,0.8436]$ & $[0.7237,0.8564]$ \\
GPT-5.6 Luna & obligations & $104$ & $0.0651$ & $0.4720$ & & & \\
 & (a) & $33$ & $0.1175$ & $0.3322$ & $0.6173$ & $[0.5177,0.7166]$ & $[0.4752,0.7310]$ \\
 & (b) & $11$ & $0.0880$ & $0.1589$ & $0.7906$ & $[0.6609,0.9077]$ & $[0.6210,0.9202]$ \\
 & (c) & $130$ & $0.0858$ & $0.1308$ & $0.8311$ & $[0.7735,0.8833]$ & $[0.7507,0.8990]$ \\
Opus 5 & obligations & $104$ & $0.3906$ & $0.5171$ & & & \\
 & (a) & $33$ & $0.4587$ & $0.4495$ & $0.5747$ & $[0.4462,0.6920]$ & $[0.4189,0.7047]$ \\
 & (b) & $11$ & $0.3905$ & $0.2667$ & $0.7841$ & $[0.6529,0.9110]$ & $[0.6360,0.9074]$ \\
 & (c) & $130$ & $0.3394$ & $0.2336$ & $0.8160$ & $[0.7606,0.8681]$ & $[0.7552,0.8797]$ \\
Name lookup & obligations & $104$ & -- & $0.2636$ & & & \\
 & (a) & $33$ & -- & $0.1279$ & $0.7395$ & $[0.6484,0.8213]$ & $[0.6299,0.8277]$ \\
 & (b) & $11$ & -- & $0.1583$ & $0.6932$ & $[0.5518,0.8165]$ & $[0.5451,0.8092]$ \\
 & (c) & $130$ & -- & $0.1226$ & $0.7428$ & $[0.6692,0.8112]$ & $[0.6548,0.8276]$ \\
\bottomrule
\end{tabular}\\[4pt]
\begin{tabular}{@{}llrrll@{}}
\toprule
Reader & group & AUROC, question only & gain from the summary & item interval & persona interval \\
\midrule
GPT-4.1 mini & (a) & $0.5000$ & $+0.1801$ & $[+0.0924,+0.2619]$ & $[+0.0649,+0.2791]$ \\
 & (b) & $0.5000$ & $+0.1875$ & $[+0.0809,+0.2815]$ & $[+0.0574,+0.2973]$ \\
 & (c) & $0.4923$ & $+0.2024$ & $[+0.1362,+0.2742]$ & $[+0.1100,+0.2943]$ \\
Haiku 4.5 & (a) & $0.2902$ & $+0.3814$ & $[+0.2577,+0.5018]$ & $[+0.2671,+0.5158]$ \\
 & (b) & $0.4637$ & $+0.2382$ & $[-0.0092,+0.4692]$ & $[-0.0743,+0.5068]$ \\
 & (c) & $0.4315$ & $+0.3574$ & $[+0.2704,+0.4448]$ & $[+0.2450,+0.4719]$ \\
GPT-5.6 Luna & (a) & $0.3550$ & $+0.2622$ & $[+0.1315,+0.3839]$ & $[+0.1341,+0.3717]$ \\
 & (b) & $0.3951$ & $+0.3955$ & $[+0.1343,+0.6512]$ & $[+0.0888,+0.6503]$ \\
 & (c) & $0.4098$ & $+0.4213$ & $[+0.3389,+0.5097]$ & $[+0.3135,+0.5207]$ \\
Opus 5 & (a) & $0.3977$ & $+0.1770$ & $[+0.0924,+0.2693]$ & $[+0.0616,+0.2906]$ \\
 & (b) & $0.5109$ & $+0.2732$ & $[+0.0395,+0.4901]$ & $[+0.0076,+0.5129]$ \\
 & (c) & $0.6139$ & $+0.2021$ & $[+0.1392,+0.2671]$ & $[+0.1300,+0.2810]$ \\
\bottomrule
\end{tabular}
\end{table}

\begin{table}[H]
\caption{The main-study record separates frozen design choices from measured checks in pipeline order. It shows how the labels, presence control, leakage screen, and querying apparatus constrain the reported estimates.}
\label{tab:e1-construction}
\centering
\scriptsize
\setlength{\tabcolsep}{3pt}
\begin{tabular}{@{}p{0.25\linewidth}p{0.12\linewidth}lp{0.26\linewidth}p{0.13\linewidth}@{}}
\toprule
Record & $n$ & $95\%$ CI & Point or composition & Constrains \\
\midrule
\multicolumn{5}{@{}l}{\textit{Design records}}\\
Scoreable corpus & $1{,}183$ & n/a & $40$ personas across three backends, with preregistered Mem0 exclusions & design record \\
Hard tier & $588$ & n/a & $35$ FOK-positive items; external negatives: $145$ A-MEM and $136$ Mem0 & design record \\
FR-Bank hard tier & $307$ & n/a & $145$ natural-absent; $127$ slot-absent; $35$ PN & design record \\
Intent audit & $413$ & not recorded & route precision $20\%$; intent conditioning makes $110$ items ($26.6\%$) ineligible for PN by definition & design record \\
Inherited retrieval premise\newline\citep{fr} & not recorded & not recorded & retraction-record pass $5\%$ & design record \\
\addlinespace
\multicolumn{5}{@{}l}{\textit{Measured checks}}\\
PN verification & $104$ & n/a & $248$ judgments: $192/179$ membership reads/confirmations; $56/48$ semantic reads/obligations & the \pn{} labels (Section~\ref{sec:setup}) \\
Presence control & reference-derived targets & not recorded & lexical $0.697$; embedding $0.965$; combined $0.995$; $\varepsilon=0.005$ & conservative gains (Section~\ref{sec:setup}) \\
Leakage screen & $180$ calls & n/a & $0$ leaks; frozen threshold $>3/30$ & values removed (Section~\ref{sec:setup}) \\
Monitor querying & $2{,}698$ responses & n/a & mini; temperature $0$; $0$ parse failures & Table~\ref{tab:e1} \\
Blind stability & $50$ & n/a & $k=3$; $46/50$ identical & blind rows, Table~\ref{tab:e1} \\
Response format & $150$ & not recorded & AUROC $0.7516$; $5$ distinct staircase values & five-level scale (below) \\
Believed-summary FOK retention & $661$ & not recorded & FR-Bank FOK retention $101.4\%$ & coverage retention (Section~\ref{sec:backends}) \\
\bottomrule
\end{tabular}
\end{table}

A five-level scale in place of the free number scores within noise of the free-number format (Table~\ref{tab:e1-construction}), a descriptive comparison because we ran no equivalence test.
On FR-Bank the store-written summary keeps $101.4\%$ of the ground-truth summary's coverage benefit despite severe metadata error, and the excess over $100\%$ is within what sampling alone produces.
The $60\%$ retention threshold was fixed in advance at $60\%$ in the main study's preregistration (day $0$ of the registration record, Appendix~\ref{app:registration}) for coverage retention and was inherited unchanged when obligation retention was scored, before that figure existed.

The presence control's residual error is $\varepsilon=0.005$, the rate at which the check misses a target that is present.
Turning that into the chance an \absent{} claim is wrong also needs the base rate, which is $0.917$ present among resolved targets.
At that base rate the chance is $5.2\%$, roughly ten times $\varepsilon$, because the sampled questions are answerability-skewed.
Splitting the hard-question negatives by kind gives the two coverage results the main text rounds.
Against facts the user never had, coverage discrimination improves from $0.5517$ to $0.8782$ under the summary.
Against records the store's own operation deleted it moves from $0.4336$ to $0.4972$ and stays at chance.
A mislabeled \absent{} item enters the negative class in both conditions and is expected to shrink the measured discrimination, so the reported gains are conservative with respect to this error.

The wrong-summary control replaces only the summary text and holds the question, prompts, protocol and populations fixed (Table~\ref{tab:wrongindex}).
Giving the monitor another persona's summary costs it the whole benefit and inverts coverage below chance, from $0.7003$ to $0.3631$, so a mismatched summary actively misleads.
A shuffle of the categories, by contrast, leaves both judgments where they were.
Permuting which counts and attribute-name lists belong to which category changes neither judgment, which bounds the mechanism rather than that specificity: per-category correspondence is not necessary on FR-Bank, consistent with the name map rather than the count structure carrying the effect (Definition~\ref{def:index}), though the shuffle preserves the global name and count multiset and the flat superseded-names section and so bounds correspondence alone.

\begin{table}[h]
\caption{The registered wrong-summary control replaces only the summary text and holds the question, prompts, protocol, and populations fixed; the intact condition is the recorded elicitation, read from the recorded rows. Giving the monitor another persona's summary costs it the whole benefit and inverts FOK below chance. Permuting which counts and which attribute-name lists belong to which category costs it nothing. $\Delta$ is the changed condition minus the original, so a negative number means the change hurt.}
\label{tab:wrongindex}
\centering
\scriptsize
\setlength{\tabcolsep}{3pt}
\begin{tabular}{@{}llrrrrl@{}}
\toprule
Manipulation & Judgment & $n$ & Intact & Manipulated & $\Delta$ & $95\%$ CI \\
\midrule
Another persona's summary & FOK & $307$ & $0.7003$ & $\mathbf{0.3631}$ & $-0.3383$ & $[-0.4506,-0.2169]$ \\
Another persona's summary & NOK & $661$ & $0.7339$ & $\mathbf{0.4659}$ & $-0.2668$ & $[-0.3298,-0.2010]$ \\
\addlinespace
Own summary, categories shuffled & FOK & $307$ & $0.7003$ & $0.7053$ & $+0.0031$ & $[-0.0499,+0.0546]$ \\
Own summary, categories shuffled & NOK & $661$ & $0.7339$ & $0.7116$ & $-0.0217$ & $[-0.0611,+0.0178]$ \\
\bottomrule
\multicolumn{7}{@{}p{0.96\linewidth}@{}}{\footnotesize The shuffle permutes the per-category count rows and the per-category attribute-name lists under a fixed seed, preserving byte length exactly on all $661$ summaries and leaving column alignment intact, so the manipulation is correspondence rather than damaged formatting. It preserves the global multiset of names and counts and leaves the flat superseded-names section untouched, so it bounds per-category correspondence and nothing more. Each $\Delta$ is the recorded paired-bootstrap estimate of manipulated minus intact ($1{,}000$ item-level draws, seed $42$; single-class draws discarded) rather than the difference of the two displayed columns, from which it differs by Monte Carlo error only, at most $0.002$ on every row.}\\
\end{tabular}
\end{table}

\subsection{Statistics and serialization}

FOK AUROC treats non-\absent{} coverage as positive, while NOK AUROC treats \pn{} as positive.
Items are nested within the $40$ personas.
Each registered family of model-minus-lookup comparisons holds five readers (three under the all-states summary).
Reported $95\%$ confidence intervals on paired AUROC differences use the fixed scorers' paired percentile bootstrap \citep{efron1993}, preserved in the artifact.
Within each reported row, $1{,}000$ item-level resamples are drawn with replacement under fixed seed $42$, with paired-condition scores moving together.
AUROC differences are recomputed per draw, single-class draws are discarded, and the $2.5$th and $97.5$th percentiles of retained differences form the interval.
A whole row containing one outcome class has undefined AUROC and no interval, so the mechanism probe reports mean coverage for such external-backend strata.
Ladder one-sample AUROC intervals use the same item-level draw-and-discard rule, so the scored item rather than the persona is the implemented resampling unit.
A persona-clustered sensitivity \citep{cameron2008}, resampling whole personas under the same paired draw-and-discard rule, was computed after the item-level records were frozen.
The registered item-level interval remains the primary endpoint.
The measured intracluster correlation of the paired indexed$-$blind contribution spans $[-0.0544,+0.0870]$, and the clustered-to-jackknife width ratio is $0.95$--$0.99$ on all six calibration rows.
Two independent cluster-level variance estimators therefore agree, and four of six clustered intervals are narrower than their item-level counterparts (Table~\ref{tab:cluster-sensitivity}).
Exact model identifiers were pinned, prompts were hash-verified, and the artifact preserves the exact prompts and serialized summaries used in monitor calls.

\begin{table}[h]
\caption{The clustered-sensitivity table reports the registered item-level interval beside the persona-clustered interval for every calibration row. Every clustered interval excludes zero, and four of six are narrower than their item-level counterparts. The last row is the paired intact-minus-blind ECE difference, whose interval includes zero.}
\label{tab:cluster-sensitivity}
\centering
\scriptsize
\begin{tabular}{@{}lrrll@{}}
\toprule
Row & $n$ & $\Delta$ & Item-level CI & Persona-clustered CI \\
\midrule
FOK, hard headline & $588$ & $+0.2272$ & $[+0.1177,+0.3504]$ & $[+0.1552,+0.2996]$ \\
FOK, FR-Bank hard & $307$ & $+0.2037$ & $[+0.0935,+0.3191]$ & $[+0.1259,+0.2767]$ \\
FOK, easy contrast & $595$ & $+0.1914$ & $[+0.1179,+0.2596]$ & $[+0.1327,+0.2433]$ \\
NOK, FR-Bank & $661$ & $+0.2393$ & $[+0.1801,+0.2949]$ & $[+0.1552,+0.3237]$ \\
FOK, Mem0 & $238$ & $-0.0969$ & $[-0.1686,-0.0250]$ & $[-0.1638,-0.0286]$ \\
FOK, A-MEM & $284$ & $-0.1065$ & $[-0.1726,-0.0403]$ & $[-0.1654,-0.0446]$ \\
ECE, pooled hard (intact $-$ blind) & $588$ & $+0.0145$ & $[-0.0037,+0.0345]$ & not recorded \\
\bottomrule
\end{tabular}
\end{table}

\subsection{Summary-channel probe}

The count-destruction probe compares intact, altered-count and names-only summaries on the pooled hard tier and on FR-Bank alone (Tables~\ref{tab:destruction} and~\ref{tab:destruction-fr}).
The pooled names-only row is driven by the structureless stores.
The FR-Bank row supports only the narrower conclusion that counts are not necessary for coverage, because every manipulated-condition interval on FR-Bank includes zero (Table~\ref{tab:destruction-fr}, with the differences and intervals read from the recorded probe scores).

\begin{table}[h]
\caption{The pooled count-destruction probe compares intact, altered-count, and names-only summaries on the hard tier. It shows flat intact-summary calibration and a different names-only result driven by the structureless external stores. Every positive of the pooled tier is an FR-Bank item and every Mem0 and A-MEM item a negative, so the pooled level reflects cross-store differences; Table~\ref{tab:destruction-fr} is the controlled within-store comparison.}
\label{tab:destruction}
\centering
\scriptsize
\begin{tabular}{@{}lrrrrr@{}}
\toprule
Condition & $n$ & AUROC & $\Delta$ vs. intact & $95\%$ CI & ECE \\
\midrule
Blind & $588$ & $0.5258$ & --- & --- & $0.1596$ \\
Intact summary & $588$ & $0.7530$ & --- & --- & $0.1741$ \\
Counts jittered & $588$ & $0.7697$ & $+0.0167$ & $[-0.0174,+0.0552]$ & $0.1720$ \\
Counts coarsened & $588$ & $0.7625$ & $+0.0095$ & $[-0.0390,+0.0657]$ & $0.1668$ \\
Names only & $588$ & $0.8516$ & $+0.0985$ & $[+0.0426,+0.1577]$ & $0.0583$ \\
\bottomrule
\end{tabular}
\end{table}

\begin{table}[h]
\caption{The FR-Bank count-destruction probe compares each altered summary with the intact summary. Every manipulated-condition interval includes zero; the differences and their item-level bootstrap intervals are read from the recorded probe scores (the artifact's mechanism-score ledger).}
\label{tab:destruction-fr}
\centering
\scriptsize
\begin{tabular}{@{}lrrrr@{}}
\toprule
Condition & $n$ & AUROC & $\Delta$ vs. intact & $95\%$ CI \\
\midrule
Intact summary & $307$ & $0.7003$ & --- & --- \\
Counts jittered & $307$ & $0.7018$ & $+0.0015$ & $[-0.0310,+0.0366]$ \\
Counts coarsened & $307$ & $0.7015$ & $+0.0012$ & $[-0.0472,+0.0509]$ \\
Names only & $307$ & $0.7283$ & $+0.0279$ & $[-0.0231,+0.0801]$ \\
\bottomrule
\end{tabular}
\end{table}

\subsection{Factorial attribution}\label{app:factorial}

The factorial separates the lifecycle-index change from the two prompt changes and checks the main-study reproduction.
The lifecycle-index main effect is $+0.1980$, with intervals $[+0.1522,+0.2423]$ at item level and $[+0.1297,+0.2639]$ grouped by persona; the two prompt-fix effects are $+0.0134$ ($[-0.0105,+0.0354]$; $[-0.0073,+0.0348]$) and $-0.0193$ ($[-0.0451,+0.0034]$; $[-0.0396,+0.0011]$), both including zero.
The matching factorial condition reproduces the main study at $0.7344$ against its indexed NOK of $0.7339$; the intervals are recomputed from the recorded rows of the factorial, every cell rescored on the same bootstrap draw.

After the pilot's null obligation result the index gained its lifecycle columns and the prompt gained worked examples and a detection step, so the NOK redesign changed three things at once.
The attribution comparison estimates the effect of each of the three changes.
The lifecycle-index effect dominates both prompt effects, so the recovery is attributed to the lifecycle information, within the conditions we varied.

\FloatBarrier
\subsection{Structureless-store inversion}\label{app:inversion}

Scope: the negative index effect is measured; the single-channel ablations do not identify its cause.

The association does not identify which rendered feature causes the inversion.
The candidate mechanisms in Section~\ref{sec:inversion} remain unresolved after the registered single-channel probe.

What the summary does to the score is a collapse onto a few low values, with the positives averaging slightly below the negatives (Table~\ref{tab:dist}).
Scores collapse onto a few low values, with positives slightly below negatives, and the trigger remains unresolved.
This is an inversion at every scale, not a below-chance score at every scale.
What triggers the collapse is unresolved, and the ablations that narrowed it follow.
On these two stores the ground-truth summary is the constrained rendering: the manifest's lifecycle totals, with every name section printed as not expressible (the verbatim summaries below).
The store-written summary carries the store's own fields alone, a superseded count on Mem0 and recency on A-MEM.
The loss is measured on the constrained rendering.
The store-written summary is within tolerance of blind: $-0.0005$ on Mem0, intervals $[-0.0663,+0.0644]$ item-level and $[-0.0672,+0.0635]$ persona-grouped, and $-0.0313$ on A-MEM, $[-0.0983,+0.0400]$ and $[-0.0798,+0.0208]$ (Table~\ref{tab:baselines}).
The pooled $66$ external questions of its GPT-5.6 rows are the Mem0 and A-MEM part of a $150$-item stratified subsample of the main study's universe (seed $42$; quota $50$ present, $60$ absent, $40$ obligations, which resolves by backend to FR-Bank $84$, Mem0 $29$, A-MEM $37$): $15$ Mem0 and $21$ A-MEM questions on the full stores, all \present{}, and $14$ and $16$ natural-absent probes, all \absent{}; $36$ positives and $30$ negatives over $29$ personas; no obligation, since the obligation labels are FR-Bank's.

The inversion table collects the backend and model rows behind the cross-scale claim, so that the reading below can be checked row by row.
Section~\ref{sec:backends} reads one generalization from them: the benefit appears on the backend that carries lifecycle structure and inverts on the two that do not.
This law is directional but not mechanistic: the source of the observed below-chance behavior remains open.

\begin{table}[h]
\caption{The table records negative summary effects on structureless stores for every tested model scale. It shows inversion at every scale, while only some scores under the summary (the indexed rows) fall below chance. $^{h}$ marks rows elicited through the vendor's agent harness (Appendix~\ref{app:laddertable}).}
\label{tab:inversion}
\centering
\scriptsize
\begin{tabular}{@{}llrrrr@{}}
\toprule
Model & Backend/population & $n$ & Blind & Indexed & $\Delta$ ($95\%$ CI) \\
\midrule
mini & Mem0 & $238$ & $0.5746$ & $0.4777$ & $-0.0969$ $[-0.1686,-0.0250]$ \\
mini & A-MEM & $284$ & $0.5591$ & $0.4526$ & $-0.1065$ $[-0.1726,-0.0403]$ \\
GPT-5.6 Sol$^{h}$ & external pooled & $66$ & $0.9130$ & $0.8083$ & $-0.1046$ $[-0.2024,-0.0174]$ \\
GPT-5.6 Luna$^{h}$ & external pooled & $66$ & $0.7894$ & $0.4514$ & $-0.3380$ $[-0.4896,-0.1811]$ \\
\bottomrule
\end{tabular}
\end{table}

The registered summary-ablation probe (Table~\ref{tab:registrations}) elicited mini on the recorded external items under single-channel and combined ablations of the serialized summary. The blind and intact conditions are the recorded rows, re-derived exactly.
The names condition has no ablation site (these summaries expose no category names, every name section's body being exactly the inexpressibility marker) and therefore equals the intact summary by construction.

The two structureless summaries as the monitor saw them, verbatim, for the persona of median size (\texttt{leila}), are reproduced below, so that the marker's share of the text can be seen rather than inferred.
Under the constrained rendering the marker occupies $3$ of $31$ lines and $18$ of $216$ tokens ($8.3\%$) on both backends. Under the store-written summary it occupies $7$ lines and $18.1\%$ of tokens on Mem0 and $8$ lines and $20.3\%$ on A-MEM.

\noindent\begin{minipage}{\linewidth}
\scriptsize\noindent\texttt{mem0:leila:FULL:TRUE}
\tiny
\begin{verbatim}
MEMORY STORE INDEX  (metadata only; no fact content)
entries in this store: 259
Ages below are in days before the store's pinned current time.

LIFECYCLE TOTALS
  active                259
  superseded            0
  retracted             0
  expired               1
  supersession events   0

PER-CATEGORY BREAKDOWN
  not expressible by this backend

RECENCY
  newest entry          1
  oldest entry          550
  median entry age      291

AGE DISTRIBUTION  (entries by age band)
  0-30                  10
  31-90                 47
  91-180                32
  181-365               71
  over 365              99

ATTRIBUTE NAMES BY CATEGORY  (value-scrubbed)
  not expressible by this backend

ATTRIBUTE NAMES CARRYING SUPERSEDED CONTENT  (value-scrubbed)
  not expressible by this backend
\end{verbatim}
\end{minipage}

\noindent\begin{minipage}{\linewidth}
\scriptsize\noindent\texttt{amem:leila:FULL:TRUE}
\tiny
\begin{verbatim}
MEMORY STORE INDEX  (metadata only; no fact content)
entries in this store: 272
Ages below are in days before the store's pinned current time.

LIFECYCLE TOTALS
  active                271
  superseded            1
  retracted             0
  expired               2
  supersession events   2

PER-CATEGORY BREAKDOWN
  not expressible by this backend

RECENCY
  newest entry          1
  oldest entry          550
  median entry age      298

AGE DISTRIBUTION  (entries by age band)
  0-30                  14
  31-90                 35
  91-180                24
  181-365               82
  over 365              117

ATTRIBUTE NAMES BY CATEGORY  (value-scrubbed)
  not expressible by this backend

ATTRIBUTE NAMES CARRYING SUPERSEDED CONTENT  (value-scrubbed)
  not expressible by this backend
\end{verbatim}
\end{minipage}

\begin{table}[h]
\caption{The summary-ablation probe removes candidate channels from the external-store summaries. No single channel separates from the intact inversion; removing the global count and the markers together reduces the A-MEM inversion and leaves Mem0 unchanged.}
\label{tab:x2-ablation}
\centering
\scriptsize
\begin{tabular}{@{}llrrrl@{}}
\toprule
Store & Condition & $n$ & Indexed AUROC & $\Delta$ vs blind & $95\%$ CI \\
\midrule
Mem0 & intact (recorded) & $238$ & $0.4777$ & $-0.0969$ & $[-0.1686,-0.0250]$ \\
Mem0 & count removed & $238$ & $0.4802$ & $-0.0943$ & $[-0.1674,-0.0233]$ \\
Mem0 & markers removed & $238$ & $0.4796$ & $-0.0949$ & $[-0.1654,-0.0250]$ \\
Mem0 & count$+$markers removed & $238$ & $0.4664$ & $-0.1082$ & $[-0.1769,-0.0392]$ \\
\addlinespace
A-MEM & intact (recorded) & $284$ & $0.4526$ & $-0.1065$ & $[-0.1726,-0.0403]$ \\
A-MEM & count removed & $284$ & $0.4587$ & $-0.1004$ & $[-0.1699,-0.0309]$ \\
A-MEM & markers removed & $284$ & $0.4705$ & $-0.0886$ & $[-0.1550,-0.0269]$ \\
A-MEM & count$+$markers removed & $284$ & $0.5223$ & $-0.0368$ & $[-0.1041,+0.0268]$ \\
\bottomrule
\end{tabular}
\end{table}

Whether the marker itself carries the inversion is a separate question. A descriptive probe left every summary byte-identical and appended one instruction line inside it: treat any field marked not expressible by this backend as absent, because the backend does not record it.
GPT-4.1 mini re-read the $522$ recorded items under that line, with the blind and intact rows read from the recorded records (Table~\ref{tab:q4-marker}).
Both stores move up to about $0.54$, and the difference from the intact summary excludes zero on A-MEM under both intervals and includes it on Mem0. On neither store does the score then differ from blind by an interval that excludes zero.
Grouped by persona the intervals read $[-0.0149,+0.1292]$ (Mem0) and $[+0.0285,+0.1403]$ (A-MEM) for the difference from the intact summary, and $[-0.1026,+0.0294]$ and $[-0.0739,+0.0327]$ for the difference from blind.
The probe carried no prediction and is reported whichever way it came out. It does not resolve the mechanism, because the score stays below blind at the point estimate on both stores. The instruction's word ``absent'' is also the name of a label, so the probe cannot separate a reading of the marker from a reading of the label.

\begin{table}[H]
\caption{The marker read as absence: GPT-4.1 mini's coverage AUROC on the two structureless stores when one instruction line is appended to the otherwise unchanged constrained rendering, beside the recorded question-only (blind) and intact rows. The condition lifts both stores toward chance without lifting either above question only. Intervals are the item-level paired bootstrap; the persona-clustered intervals are in the text above; descriptive, no prediction.}
\label{tab:q4-marker}
\centering
\scriptsize
\setlength{\tabcolsep}{3pt}
\begin{tabular}{@{}lrrrrll@{}}
\toprule
Store & $n$ & Blind & Intact & Marker as absence & $-$ intact ($95\%$ CI) & $-$ blind ($95\%$ CI) \\
\midrule
Mem0 & $238$ & $0.5746$ & $0.4777$ & $0.5353$ & $+0.0577$ $[-0.0074,+0.1157]$ & $-0.0393$ $[-0.1042,+0.0226]$ \\
A-MEM & $284$ & $0.5591$ & $0.4526$ & $0.5355$ & $+0.0829$ $[+0.0277,+0.1411]$ & $-0.0236$ $[-0.0871,+0.0388]$ \\
\bottomrule
\end{tabular}
\end{table}

A registered follow-up asked whether the inversion is a property of the store or of the summary it was handed.
Each stored entry of the two structureless stores was mapped to the manifest slot it holds, by exact lexical match and by embedding similarity. The map was checked by hand on thirty pairs per store: $24$ of $30$ correct on A-MEM ($0.800$) and $17$ of $30$ on Mem0 ($0.567$).
The Mem0 result fell below the pre-specified bar of $0.80$, so its map was restricted to lexically supported pairs, which cover $4{,}809$ of its $9{,}736$ entries ($49.4\%$), below the registered coverage floor of one half. A-MEM's map covers $10{,}974$ of $11{,}319$ entries ($0.970$).
From each map a name-bearing summary was rendered in the paper's summary format, with the per-category attribute names and the superseded-names section filled from the manifest for the mapped entries, and every stored value removed. The leakage screen scored $0$ of $30$ on both stores.
The recorded coverage items of each store were then scored by the same monitor against that summary, with the blind and ground-truth rows read from the recorded records (Table~\ref{tab:s1-names}).

\begin{table}[h]
\caption{The name-bearing summary on the two structureless stores. Each store's recorded coverage items are scored by GPT-4.1 mini against a summary built from the manifest for the entries the store holds, beside the two lookups on that summary; question only (blind) and constrained are the recorded rows of Table~\ref{tab:inversion}. A-MEM carries the two registered predictions and both hold (Section~\ref{sec:backends}); Mem0's summary covers $49.4\%$ of its entries, below the coverage floor of one half, so its row is descriptive and carries no verdict. Intervals are the item-level paired bootstrap; the persona-clustered intervals are in the text.}
\label{tab:s1-names}
\centering
\scriptsize
\setlength{\tabcolsep}{2.4pt}
\begin{tabular}{@{}lrrrrllrr@{}}
\toprule
Store & $n$ & Blind & Constrained & Names & Names $-$ blind ($95\%$ CI) & Names $-$ constrained ($95\%$ CI) & NB-4 & NB-5 \\
\midrule
A-MEM & $284$ & $0.5591$ & $0.4526$ & $0.8974$ & $+0.3383$ $[+0.2673,+0.4122]$ & $+0.4448$ $[+0.3800,+0.5116]$ & $0.7000$ & $0.8554$ \\
Mem0 & $238$ & $0.5746$ & $0.4777$ & $0.8751$ & $+0.3006$ $[+0.2208,+0.3728]$ & $+0.3975$ $[+0.3218,+0.4747]$ & $0.7104$ & $0.8700$ \\
\bottomrule
\end{tabular}
\end{table}

Under persona clustering the A-MEM intervals read $[+0.2791,+0.3976]$ for the gain over blind and $[+0.3875,+0.5042]$ for the difference from the constrained rendering; Mem0's read $[+0.2429,+0.3669]$ and $[+0.3327,+0.4617]$.
On the paper's own summary of these stores the two lookups sit at $0.5000$ (Table~\ref{tab:baselines}), because that summary names nothing. On the name-bearing summary they reach $0.7000$ and $0.8554$ on A-MEM and $0.7104$ and $0.8700$ on Mem0.
The name lookup was not scored on these summaries, whose superseded-names section is thin by construction.

The two readings written before the run, reported as they were preregistered: Given the manifest's names for what it holds, the store no longer inverts the benefit by the margin it did; the inversion was a property of the index the store was handed, not of the store alone.
The name-bearing summary does not invert on either store.
On Mem0 the name-bearing index could be built for $49.4$ percent of the store's entries under the lexical-required mapping the spot-check forced (two-channel precision $0.567$); below the registered coverage floor of one half, its coverage gain of $+0.3006$ (interval $[+0.2208,+0.3728]$; \trueidx{} $-0.0969$) is reported and carries no verdict.

\FloatBarrier

\paragraph{What the summary does to the score.}
The recorded coverage scores on the two structureless stores show what the inversion consists of (Table~\ref{tab:dist}).
Blind, GPT-4.1 mini's score takes six distinct values with a variance of $0.0245$ on Mem0 and $0.0242$ on A-MEM.
Indexed under the constrained rendering it takes five values with a variance of $0.0060$ and $0.0052$, and the mean falls from $0.22$ to $0.17$ on Mem0 and from $0.22$ to $0.16$ on A-MEM.
The summary lowers the score on $86$ Mem0 items and raises it on $46$, and lowers it on $110$ A-MEM items and raises it on $48$, while the mean on positives and on negatives moves together.
Compression alone cannot lower an AUROC below $0.50$, because AUROC is unchanged by any order-preserving transform; what the rows show is a collapse onto five values in which the positives average slightly below the negatives ($0.1617$ against $0.1688$ on Mem0, $0.1544$ against $0.1658$ on A-MEM), and that ordering is the inversion.
The store-written summaries, which carry no not-expressible line, show no loss (within tolerance of blind, above), and the name-bearing summary rebuilt from ground truth restores coverage (Table~\ref{tab:s1-names}), so the loss belongs to the constrained rendering; which line of it produces the collapse is the question the ablations above leave open, and the marker's wording is not it (Table~\ref{tab:marker}).

\begin{table}[H]
\caption{The distribution of GPT-4.1 mini's coverage score on the two structureless stores, question only (blind) against with the summary (indexed) under the constrained rendering, from the recorded rows: distinct values, variance, mean, and the mean on positive and negative items. The summary collapses the score onto a few low values on both stores, the positives slightly below the negatives. Descriptive.}
\label{tab:dist}
\centering
\scriptsize
\setlength{\tabcolsep}{4pt}
\begin{tabular}{@{}llrrrrr@{}}
\toprule
Store & Condition & distinct values & variance & mean & mean, positives & mean, negatives \\
\midrule
Mem0 ($238$) & blind & $6$ & $0.0245$ & $0.2183$ & $0.2202$ & $0.2170$ \\
Mem0 ($238$) & indexed & $5$ & $0.0060$ & $0.1660$ & $0.1617$ & $0.1688$ \\
A-MEM ($284$) & blind & $6$ & $0.0242$ & $0.2234$ & $0.2230$ & $0.2237$ \\
A-MEM ($284$) & indexed & $5$ & $0.0052$ & $0.1607$ & $0.1544$ & $0.1658$ \\
\bottomrule
\end{tabular}
\end{table}

\begin{table}[H]
\caption{The inexpressibility marker on the two structureless stores (E3; registered, ADDENDUM\_MARKER\_ABLATION): coverage AUROC under the constrained rendering (structural), with the three inexpressible sections omitted (the recorded omitted arm of the main study), and with the marker's three lines reworded to say the store does not record the field (neutral), against question only (blind). GPT-4.1 mini on each store's coverage population at one call; GPT-5.6 Luna on the pooled $66$ questions through the raw API at five calls, with its recorded harness rows beside ($^{h}$: elicited through the vendor's agent harness, Appendix~\ref{app:laddertable}). Differences against question only carry the item-level and the persona-grouped interval. The modal share is the fraction of items at the most common score.}
\label{tab:marker}
\centering\tiny\setlength{\tabcolsep}{2pt}
\begin{tabular}{@{}lllrlrrrr@{}}
\toprule
Reader & store & summary & AUROC & persona interval & minus blind & item interval & persona interval & modal share \\
\midrule
GPT-4.1 mini & Mem0 & blind & $0.5746$ & $[0.5141,0.6327]$ &  & &  & $0.4202$ \\
 &  & structural & $0.4777$ & $[0.4238,0.5378]$ & $-0.0969$ & $[-0.1686,-0.0250]$ & $[-0.1638,-0.0286]$ & $0.4622$ \\
 &  & omitted & $0.4881$ & $[0.4206,0.5475]$ & $-0.0865$ & $[-0.1508,-0.0254]$ & $[-0.1501,-0.0254]$ & $0.6176$ \\
 &  & neutral & $0.5024$ & $[0.4387,0.5590]$ & $-0.0722$ & $[-0.1391,-0.0046]$ & $[-0.1311,-0.0121]$ & $0.4664$ \\
\addlinespace[1pt]
GPT-4.1 mini & A-MEM & blind & $0.5591$ & $[0.5072,0.6136]$ &  & &  & $0.4014$ \\
 &  & structural & $0.4526$ & $[0.3951,0.5083]$ & $-0.1065$ & $[-0.1726,-0.0403]$ & $[-0.1654,-0.0446]$ & $0.4648$ \\
 &  & omitted & $0.5213$ & $[0.4652,0.5773]$ & $-0.0378$ & $[-0.1024,+0.0260]$ & $[-0.0956,+0.0163]$ & $0.6162$ \\
 &  & neutral & $0.5319$ & $[0.4725,0.5925]$ & $-0.0272$ & $[-0.0890,+0.0354]$ & $[-0.0708,+0.0251]$ & $0.5282$ \\
\addlinespace[1pt]
GPT-5.6 Luna, raw API & pooled $66$ & blind & $0.6454$ & $[0.5121,0.7657]$ &  & &  & $0.0909$ \\
 &  & structural & $0.3528$ & $[0.2259,0.4948]$ & $-0.2926$ & $[-0.4261,-0.1590]$ & $[-0.4344,-0.1359]$ & $0.0606$ \\
 &  & omitted & $0.3088$ & $[0.2005,0.4250]$ & $-0.3366$ & $[-0.4859,-0.1824]$ & $[-0.4588,-0.2106]$ & $0.0606$ \\
 &  & neutral & $0.3782$ & $[0.2595,0.5126]$ & $-0.2671$ & $[-0.4127,-0.1232]$ & $[-0.3951,-0.1267]$ & $0.0606$ \\
GPT-5.6 Luna, harness$^{h}$ & pooled $66$ & blind & $0.7894$ & & & & & \\
(recorded) & & structural & $0.4514$ & & $-0.3380$ & & & \\
\bottomrule
\end{tabular}
\end{table}

\FloatBarrier
\subsection{Non-LLM baseline suite}\label{app:baselines}

The non-model readers of the summary carry numbered names in the registrations, and the main text calls the sixth the name lookup.
The registrations number them NB-1 to NB-6.
This appendix carries the whole family, the per-state split of the name lookup against the models, and the lookup on the coverage tiers, because each bounds a claim the main text makes about what a model adds.
Both readings were registered, the whole-population comparison as a sentence pair and the per-state one as a single reading.

Six non-model readers were run on the same serialized summaries the monitor reads, to ask whether a language model is needed at all.
The first is a deterministic lifecycle rule that parses per-category entry and lifecycle counts and assigns the question to a category by content-token overlap with that category's attribute names.
It scores coverage as the assigned category's share of the store's largest category and obligation as its superseded-plus-retracted-plus-expired share.
The second is a logistic model over numeric summary features only, label-supervised and persona-grouped.
The fourth and fifth are unsupervised query-to-summary lookups, lexical and embedding respectively, scored against the whole per-category name map.
The sixth is that same pair of lookups aimed at one section instead, the flat list under \texttt{ATTRIBUTE NAMES CARRYING SUPERSEDED CONTENT}, whose populations, tolerance and outcome sentences were fixed before it was scored.
The third was disqualified before scoring, because its features concatenated the summary with the question and its ablation showed question-only text outscoring the concatenated version on every population.
It read question templates rather than the store (supplement Section~S.1, item 12), and its column below is the summary-only ablation, which is the part that reads the store.

\begin{table}[h]
\caption{Mechanical readers of the same serialized summary. ``Best'' is the strongest baseline on that row; the final column is the monitor minus that best, under the $\pm0.05$ tolerance inherited from the count-destruction probe. The NB-6 column (the name lookup) is filled on the obligation rows only: on the coverage rows NB-6 reads the same name map as NB-4 and NB-5 and reproduces both exactly, so an entry there would repeat a number already in the table. The NB-6 values shown are its embedding variant, the stronger of its two wherever it is scored. NB-5 and NB-6 use a local open-weights encoder and are not the presence control's embedding channel.}
\label{tab:baselines}
\centering
\scriptsize
\setlength{\tabcolsep}{2.2pt}
\begin{tabular}{@{}llrrrrrrrrrl@{}}
\toprule
Population & summary & $n$ & monitor & NB-1 & NB-2 & NB-4 & NB-5 & NB-3\textsuperscript{idx} & NB-6 & best & $\Delta$ \\
\midrule
FR-Bank FOK & ground-truth & $307$ & $0.7003$ & $0.5194$ & $0.5228$ & $0.6080$ & $0.6186$ & $0.5048$ & -- & $0.6186$ & $\mathbf{+0.0817}$ \\
FR-Bank FOK & store-written & $307$ & $0.7005$ & $0.5194$ & $0.4630$ & $0.6080$ & $0.6186$ & $0.5419$ & -- & $0.6186$ & $\mathbf{+0.0819}$ \\
FR-Bank NOK & ground-truth & $661$ & $0.7339$ & $0.5514$ & $0.5834$ & $0.5383$ & $0.5974$ & $0.4780$ & $\mathbf{0.7484}$ & $0.7484$ & $-0.0145$ \\
FR-Bank NOK & store-written & $661$ & $0.5959$ & $0.4987$ & $0.5459$ & $0.5383$ & $0.5974$ & $0.4715$ & $0.5858$ & $0.5974$ & $-0.0015$ \\
\addlinespace
Mem0 FOK & constrained$^{\ddagger}$ & $238$ & $0.4777$ & $0.4818$ & $0.4923$ & $0.5000$ & $0.5000$ & $0.4464$ & -- & $0.5000$ & $-0.0223$ \\
Mem0 FOK & store-written & $238$ & $0.5741$ & $0.4818$ & $0.5372$ & $0.5000$ & $0.5000$ & $0.4621$ & -- & $0.5372$ & $+0.0369$ \\
A-MEM FOK & constrained$^{\ddagger}$ & $284$ & $0.4526$ & $0.5226$ & $0.4184$ & $0.5000$ & $0.5000$ & $0.5163$ & -- & $0.5226$ & $-0.0700$ \\
A-MEM FOK & store-written & $284$ & $0.5278$ & $0.5226$ & $0.4452$ & $0.5000$ & $0.5000$ & $0.5098$ & -- & $0.5226$ & $+0.0052$ \\
\addlinespace
Pooled hard FOK$^{\dagger}$ & ground-truth & $588$ & $0.7530$ & $0.4513$ & $0.7246$ & $0.6983$ & $0.8124$ & $0.6827$ & -- & $0.8124$ & $-0.0594$ \\
Pooled hard FOK$^{\dagger}$ & store-written & $588$ & $0.7955$ & $0.4513$ & $0.7367$ & $0.6983$ & $0.8124$ & $0.6909$ & -- & $0.8124$ & $-0.0169$ \\
\bottomrule
\multicolumn{12}{@{}p{0.96\linewidth}@{}}{\footnotesize $^{\dagger}$Pooled hard-tier absolute levels are not interpretable for any reader, monitor included: all $35$ positives are FR-Bank, so the one-bit is-FR-Bank rule scores $0.7541$, and its Mem0 and A-MEM items are under the constrained rendering. Paired deltas are unaffected. Of the rows where a baseline outranks the monitor, two are of this kind, two are on a structureless backend, and one is the ground-truth obligation row, where NB-6 leads by $0.0145$ inside the tolerance. $^{\ddagger}$Ground truth, rendered under the backend's constraints: the lifecycle totals exact, every name section printed as not expressible (Appendix~\ref{app:inversion}).}\\
\end{tabular}
\end{table}

Three readings carry beyond the table, because each bears on a claim the main text makes.
The first is the store-written obligation row, where the monitor at $0.5959$ is indistinguishable from the embedding lookup at $0.5974$, a difference of $-0.0015$ inside the tolerance: under corrupted self-description the monitor's obligation judgment falls back to what a similarity lookup already achieves, which is the same degradation the $42.3\%$ retention records in Section~\ref{sec:backends}, reached by an independent route.
The second is that the monitor's coverage margin over the whole-map readers, the lookups that score the question against the whole name map, is not uniform across backends.
It exists where lifecycle structure exists and vanishes where it does not, which is the same boundary the inversion result draws.
The third is the structureless pair, where the lexical and embedding lookups sit at $0.5000$ and the other readers scatter around chance ($0.4184$ to $0.5226$).
The monitor's indexed point estimates under the constrained rendering fall below them ($0.4777$, $0.4526$; Section~\ref{sec:backends}).

The name lookup was registered as a direct test of one exposure: whether the obligation judgment under a ground-truth summary is close to reading the answer off the superseded-names section.
The whole-map lookups cannot detect that, because they dilute the section across the whole name map.
Under the ground-truth summary it reaches $0.7484$ against the monitor's $0.7339$, so on this population the monitor holds no margin over that reader.
The difference is inside the $\pm0.05$ tolerance, so it does not establish the lookup as the better judge either; what it removes is the claim of a margin.
Under the store-written summary it reaches $0.5858$ against the monitor's $0.5959$, also inside the tolerance.
The whole-population difference under the ground-truth summary, monitor minus lookup, is $-0.0145$ with item-level interval $[-0.0769,+0.0549]$ and persona-clustered interval $[-0.1080,+0.0765]$ (Table~\ref{tab:nb6-composition}, last row), so the interval and the tolerance reading agree.
The ground-truth superseded list averages $1.75$ names per item and the store-written list averages $15.04$, so the two summaries give the lookup very different amounts of text to work with.
Mem0 and A-MEM were not scored by the lookup, because their superseded-names section carries only the marker the renderer writes where a store exposes no field, and the lookup would measure the marker rather than the store.
On the coverage rows NB-6 scores the whole per-category name map and reproduces NB-4 and NB-5 to four decimals, so the coverage cells print a dash rather than repeating them.
The lookup has one descriptive value on the cross-model set, whole-set and per-state values on the store-counterfactual set, and none on the authorship-matched set. The comparison between model and lookup is therefore complete on the full obligation population ($661$) and on the store-counterfactual set ($444$).
On the counterfactual set every model's coverage interval excludes chance (Table~\ref{tab:cf-fok}, Appendix~\ref{app:counterfactual}).
Table~\ref{tab:cov-lookup} below adds the superseded section scored on the coverage rows, a measurement added after the main study.

The registered texts for these rows (Appendix~\ref{app:registration}) read as follows.
A matcher that scores each question directly against the store's list of names marked superseded reaches $0.7484$ on the same $661$ questions, against the monitor's $0.7339$.
Under a truthful summary the obligation judgment is available to a direct lookup on that one list.
The superseded-names matcher matches the smallest model tested. Larger models score higher on the states the index only counts, an advantage the paired blind comparison does not attribute to the index (Tables~\ref{tab:nb6-attrib} and~\ref{tab:nb6-frontier}).
Under the metadata the backend actually exposes, the monitor's obligation judgment is no better than a matcher that reads the store's list of superseded names directly.
The monitor scores $0.5959$ on that row and the matcher $0.5858$ (Appendix~\ref{app:baselines}).

A census run after that result asks why the lookup works where it does, because the answer says what a reader can take from this summary and what it cannot.
The summary names attributes in exactly one place, the superseded-names section, and expiry and retraction reach the summary as counts and are never named.
Splitting the obligation items by the lifecycle state each obligation arises from reproduces that asymmetry exactly (Table~\ref{tab:nb6-composition}).
The lookup reaches $0.9534$ on supersession-derived items and sits at $0.4823$ on expiry and $0.4053$ on retraction, the last two at or below chance.
The monitor is weaker off supersession as well, at $0.5110$ and $0.5638$, and on retraction it is the higher of the two readers.
A check of whether each obligation's own attribute name appears in the section the lookup reads finds it for $74$ of $104$ under exact slot match: $63$ of $63$ supersession-derived, $11$ of $26$ retraction-derived, and none of the $15$ expiry-derived.
On the state the summary names, the lookup has the item's own name to read; whether the monitor reads it is a separate question that this census does not settle.
Where the self-description names an obligation, a lookup reaches it; where the self-description only counts it, the lookup does not.
Composition follows the same line: obligations on the full population are $63$ of $104$ supersession-derived, against $19$ of $40$ on the cross-model set, where the lookup scores $0.6023$.

\begin{table}[h]
\caption{Post-hoc census of the obligation population under the ground-truth summary, split by the lifecycle state each obligation arises from and scored against the same negatives. The summary names superseded content and only counts the other two states, and discrimination follows that line for both readers. Designed after the baseline result; it carries no registered prediction and is reported whichever way it came out.}
\label{tab:nb6-composition}
\centering
\scriptsize
\begin{tabular}{@{}lrrrrr@{}}
\toprule
Obligation arises from & $n$ & NB-6 ($95\%$ CI) & monitor ($95\%$ CI) & monitor $-$ NB-6 \\
\midrule
Supersession & $63$ & $0.9534$ $[0.9281,0.9760]$ & $0.8572$ $[0.7946,0.9142]$ & $-0.0962$ $[-0.1520,-0.0495]$ \\
Expiry & $15$ & $0.4823$ $[0.3626,0.6110]$ & $0.5110$ $[0.3971,0.6383]$ & $+0.0287$ $[-0.1795,+0.2212]$ \\
Retraction & $26$ & $0.4053$ $[0.2812,0.5320]$ & $0.5638$ $[0.4629,0.6760]$ & $+0.1585$ $[+0.0002,+0.3341]$ \\
\addlinespace
Whole population & $104$ & $0.7484$ $[0.6636,0.8298]$ & $0.7339$ $[0.6502,0.8186]$ & $-0.0145$ $[-0.0769,+0.0549]$ \\
\bottomrule
\end{tabular}
\end{table}

The monitor exceeds the matcher on retraction-derived obligations at $n=26$, by $+0.1585$ $[+0.0002,+0.3341]$.
That interval's lower bound sits at $+0.0002$, so the row is marginal and is read as marginal.
The index counts expired content, and neither reader recovers it.
Under persona clustering the mini per-state differences read Supersession $-0.0962$ item $[-0.1520,-0.0495]$ clustered $[-0.1712,-0.0307]$; Expiry $+0.0287$ item $[-0.1795,+0.2212]$ clustered $[-0.3115,+0.4296]$; Retraction $+0.1585$ item $[+0.0002,+0.3341]$ clustered $[-0.0560,+0.4152]$.
The retraction row is marginal at item level (lower bound $+0.0002$) and not separated under persona clustering.
The same split run on the three models scored on this population under five calls per question separates the two readers in opposite directions.
On supersession, the content the summary names, the lookup is above all three.
On expiry and retraction, the content the summary only counts, all three are above the lookup, and every difference in Table~\ref{tab:nb6-frontier} excludes zero.
Expiry rests on $15$ obligations and retraction on $26$, so what these rows support is the direction of the crossover rather than a calibrated size for it.
The comparison sets the models against an intent-blind reader, so it cannot say whether their advantage on the counted states comes from the store or from the question.
Answering that needs the paired gain on the same rows, which Table~\ref{tab:nb6-attrib} reports per state.
On supersession the summary carries every model, each difference excluding zero.
On expiry one model of three separates from its own blind condition and on retraction one does, so the models' advantage over the lookup on the counted states is not attributable to the summary.
GPT-4.1 mini is absent from that table because its blind condition spans two distinct values on this population.

\begin{table}[h]
\caption{Test of the per-state split, controlled for the question signal: blind and indexed discrimination for the three models scored on the full obligation population, with the paired gain (indexed minus blind) and its interval. The summary carries every model on the state it names. On the two states it only counts, one model of three separates from its own blind condition on each, so the advantage those models hold over the name lookup there is not attributable to the summary. GPT-4.1 mini is absent because its blind condition spans two distinct values on this population.}
\label{tab:nb6-attrib}
\centering
\scriptsize
\setlength{\tabcolsep}{3pt}
\begin{tabular}{@{}llrrrr@{}}
\toprule
Model & Obligation arises from & $n$ & blind & indexed & indexed $-$ blind ($95\%$ CI) \\
\midrule
GPT-5.6 Luna & Supersession & $63$ & $0.3693$ & $0.8933$ & $+0.5240$ $[+0.4544,+0.5970]$ \\
GPT-5.6 Luna & Expiry & $15$ & $0.7197$ & $0.8248$ & $+0.1051$ $[+0.0056,+0.2163]$ \\
GPT-5.6 Luna & Retraction & $26$ & $0.4881$ & $0.6323$ & $+0.1442$ $[-0.0077,+0.3041]$ \\
\addlinespace
Opus 5 & Supersession & $63$ & $0.5387$ & $0.8690$ & $+0.3303$ $[+0.2814,+0.3831]$ \\
Opus 5 & Expiry & $15$ & $0.8709$ & $0.8881$ & $+0.0172$ $[-0.0057,+0.0411]$ \\
Opus 5 & Retraction & $26$ & $0.7641$ & $0.7129$ & $-0.0511$ $[-0.0987,+0.0009]$ \\
\addlinespace
Haiku 4.5 & Supersession & $63$ & $0.3264$ & $0.8897$ & $+0.5634$ $[+0.4936,+0.6292]$ \\
Haiku 4.5 & Expiry & $15$ & $0.6542$ & $0.6874$ & $+0.0332$ $[-0.1263,+0.1773]$ \\
Haiku 4.5 & Retraction & $26$ & $0.4918$ & $0.7317$ & $+0.2399$ $[+0.1301,+0.3577]$ \\
\bottomrule
\end{tabular}
\end{table}

\begin{table}[h]
\caption{Per-state split for the three models scored on the full obligation population under five calls per question, against the name lookup, with the difference and its interval from the stratified paired bootstrap of the registration. The crossover is the finding: the lookup leads on the state the summary names and trails on the two it only counts. The GPT-5.6 Sol and Kimi K3 runs are absent because a budget directive stopped them before they were scored on this population.}
\label{tab:nb6-frontier}
\centering
\scriptsize
\setlength{\tabcolsep}{3pt}
\begin{tabular}{@{}llrrr@{}}
\toprule
Model & Obligation arises from & $n$ & AUROC ($95\%$ CI) & $-$ lookup ($95\%$ CI) \\
\midrule
GPT-5.6 Luna & Supersession & $63$ & $0.8933$ $[0.8495,0.9324]$ & $-0.0601$ $[-0.1021,-0.0246]$ \\
GPT-5.6 Luna & Expiry & $15$ & $0.8248$ $[0.7825,0.8656]$ & $+0.3425$ $[+0.2093,+0.4636]$ \\
GPT-5.6 Luna & Retraction & $26$ & $0.6323$ $[0.5255,0.7361]$ & $+0.2270$ $[+0.0836,+0.3679]$ \\
\addlinespace
Opus 5 & Supersession & $63$ & $0.8690$ $[0.8322,0.9026]$ & $-0.0845$ $[-0.1226,-0.0486]$ \\
Opus 5 & Expiry & $15$ & $0.8881$ $[0.8576,0.9153]$ & $+0.4058$ $[+0.2751,+0.5264]$ \\
Opus 5 & Retraction & $26$ & $0.7129$ $[0.6526,0.7736]$ & $+0.3077$ $[+0.1772,+0.4386]$ \\
\addlinespace
Haiku 4.5 & Supersession & $63$ & $0.8897$ $[0.8382,0.9306]$ & $-0.0637$ $[-0.1166,-0.0159]$ \\
Haiku 4.5 & Expiry & $15$ & $0.6874$ $[0.5720,0.7962]$ & $+0.2051$ $[+0.0002,+0.3893]$ \\
Haiku 4.5 & Retraction & $26$ & $0.7317$ $[0.6406,0.8222]$ & $+0.3264$ $[+0.1956,+0.4663]$ \\
\bottomrule
\end{tabular}
\end{table}

\paragraph{The lookup against every model on the full population, both variants.}
Table~\ref{tab:margin-661} gives each model's whole-population difference from the name lookup in both variants with both intervals.
Against the embedding variant Haiku 4.5, Opus 5 and GPT-5.6 Luna lead by $+0.0698$ to $+0.0843$, with item-level intervals that exclude zero and persona-clustered intervals that include it, so the wider interval leaves the margin unsettled.
Against the lexical variant, which reads $0.5738$ on this population, every model leads reliably.

\begin{table}[H]
\caption{Each model minus the name lookup on the $661$ obligation questions, for the embedding and the lexical variant of the lookup, with the item-level and the persona-clustered interval. Haiku 4.5, Opus 5 and GPT-5.6 Luna lead the embedding lookup at item level and not under clustering; every model leads the lexical lookup. Exploratory, computed after the fact on the recorded rows.}
\label{tab:margin-661}
\centering
\scriptsize
\setlength{\tabcolsep}{3pt}
\begin{tabular}{@{}llrrrll@{}}
\toprule
Model & Lookup variant & model & lookup & difference & item interval & persona interval \\
\midrule
GPT-4.1 mini & embedding & $0.7339$ & $0.7484$ & $-0.0145$ & $[-0.0769,+0.0549]$ & $[-0.1080,+0.0765]$ \\
GPT-4.1 mini & lexical & $0.7339$ & $0.5738$ & $+0.1601$ & $[+0.1086,+0.2153]$ & $[+0.0676,+0.2585]$ \\
GPT-5.6 Luna & embedding & $0.8182$ & $0.7484$ & $+0.0698$ & $[+0.0123,+0.1264]$ & $[-0.0180,+0.1520]$ \\
GPT-5.6 Luna & lexical & $0.8182$ & $0.5738$ & $+0.2444$ & $[+0.1977,+0.2888]$ & $[+0.1655,+0.3197]$ \\
Opus 5 & embedding & $0.8327$ & $0.7484$ & $+0.0843$ & $[+0.0182,+0.1448]$ & $[-0.0091,+0.1759]$ \\
Opus 5 & lexical & $0.8327$ & $0.5738$ & $+0.2589$ & $[+0.2195,+0.2987]$ & $[+0.1935,+0.3199]$ \\
Haiku 4.5 & embedding & $0.8210$ & $0.7484$ & $+0.0726$ & $[+0.0052,+0.1393]$ & $[-0.0143,+0.1634]$ \\
Haiku 4.5 & lexical & $0.8210$ & $0.5738$ & $+0.2473$ & $[+0.1993,+0.2933]$ & $[+0.1747,+0.3178]$ \\
\bottomrule
\end{tabular}
\end{table}

\paragraph{The superseded-section lookup on the coverage tiers.}
The coverage rows of Table~\ref{tab:baselines} score the whole name map.
Table~\ref{tab:cov-lookup} scores the superseded section alone on the same coverage populations, beside GPT-4.1 mini's recorded indexed coverage score.
On the hard tier every coverage positive is an obligation, so the embedding lookup on the superseded section reads $0.7864$ against the model's $0.7003$, and the difference of $-0.0861$ has intervals that include zero.
The FR-Bank easy tier has ordinary present records as its positives, and there the model reads $0.7835$ against the superseded-section lookup's $0.6083$, a reliable margin of $+0.1752$.
A lookup over the whole name map, the reader the coverage rows of Table~\ref{tab:baselines} score, was computed on the easy tier after the fact and specified before recomputation: it reads $0.7469$ (persona interval $[0.6866,0.8028]$), and the model's margin over it is $+0.0367$ with intervals $[-0.0127,+0.0861]$ and $[-0.0036,+0.0776]$, not reliable and within tolerance, so the coverage margin the superseded-section lookup showed does not survive the stronger reader with no generative-model call (Table~\ref{tab:cov-lookup}, last row).
Pooling the easy tier across backends drops the lookup to $0.4021$, because the structureless stores' sections carry only the marker, and the model's margin there is $+0.2315$.

\begin{table}[H]
\caption{The name lookup, lexical and embedding, on the coverage populations, beside GPT-4.1 mini's recorded blind and indexed coverage AUROC, with the model minus the embedding lookup and both intervals; the last row reads the whole name map instead of the superseded section (post hoc, specified before recomputation). Where the coverage positives are obligations the lookup reads them; where they are ordinary records the model leads the superseded-section lookup and not, reliably, the whole-map lookup. Exploratory, computed after the fact on the recorded rows.}
\label{tab:cov-lookup}
\centering
\scriptsize
\setlength{\tabcolsep}{3pt}
\begin{tabular}{@{}lrrrrrrll@{}}
\toprule
Population & $n$ / pos. & blind & indexed & lexical & embedding & model $-$ embedding & item interval & persona interval \\
\midrule
FR-Bank hard tier & $307$ / $35$ & $0.4966$ & $0.7003$ & $0.5970$ & $0.7864$ & $-0.0861$ & $[-0.1879,+0.0305]$ & $[-0.1793,+0.0166]$ \\
FR-Bank easy tier & $354$ / $243$ & $0.4660$ & $0.7835$ & $0.5162$ & $0.6083$ & $+0.1752$ & $[+0.0938,+0.2515]$ & $[+0.0959,+0.2579]$ \\
Easy tier, all backends & $595$ / $463$ & $0.4422$ & $0.6336$ & $0.5071$ & $0.4021$ & $+0.2315$ & $[+0.1628,+0.2979]$ & $[+0.1680,+0.2970]$ \\
\midrule
Easy tier, whole map & $354$ / $243$ & $0.4660$ & $0.7835$ & $0.6311$ & $0.7469$ & $+0.0367$ & $[-0.0127,+0.0861]$ & $[-0.0036,+0.0776]$ \\
\bottomrule
\end{tabular}
\end{table}

\paragraph{The coverage judgment against its floor.}
The coverage margin over the name lookup does not survive a lookup over the whole name map.
On the easy tier ($354$ questions), where the positives are ordinary present records, GPT-4.1 mini scores $0.78$ against the name lookup's $0.61$, a reliable margin.
A lookup over every name in the map, computed post hoc, scores $0.75$ on the same questions, and the model's margin over it is $+0.04$ with interval $[-0.01,+0.09]$, which is not a reliable one (Table~\ref{tab:cov-lookup}).
On the hard tier every coverage positive is an obligation, so the coverage and obligation scores are one row, and the lookup's $0.79$ against the model's $0.70$ is not a reliable difference.
FOK detects a missing topic and not a single deleted record, which is NOK's target (Appendix~\ref{app:construction}).
The score ranks well and calibrates badly (Appendix~\ref{app:posthoc}).

\subsection{Census of the store-written state}\label{app:census}

Table~\ref{tab:census} in Section~\ref{sec:backends} carries the census behind the over-mark, overwrite and unresolved-links reading.
Four counts near one hundred in that census are four different things: the manifests record $109$ supersession events across the forty personas; the ground-truth join explains about $102$ of the store's inactive entries; the manifest's supersessions touch $104$ stored entries, $40$ of them still active; and the $104$ obligations of the full obligation set are questions, $63$ supersession, $26$ retraction and $15$ expiry, not entries; the $72\%$ of Table~\ref{tab:census} is a share of the $109$ events ($78$ of them), not of the $104$ entries.
This appendix records the store-written superseded list, the renderer's cap, and the fourth-backend trial, because each bears on how much of the loss is the store's.

\paragraph{Two census facts the body states in brief.}
The pass rate of $5\%$ that Section~\ref{sec:setup} cites for FR-Bank's mask is the released store's own figure, measured on its $40$ retraction questions with the mask in place \citep{fr}.
The store also conflates retraction with supersession: of $31$ entries marked retracted across the $40$ personas, $22$ share a slot key with another entry and $16$ carry a supersedes pointer, so the write path records retraction through the supersession mechanism and the backend's own metadata cannot separate retracted from superseded content.
The manifests record $40$ retraction events, one per persona, and the ground-truth join resolves $9$ of them to a retracted entry ($23$ entries).
The count of retained active assertions in Table~\ref{tab:census} depends on the stored-versus-obsolete matching threshold of the census, and the artifact's \texttt{repro\_mem0\_erase.py} and \texttt{repro\_amem\_links.py} reproduce the table's Mem0 and A-MEM rows from the frozen banks.
Mem0's per-memory history call, which keeps the prior text under a known identifier, was not consulted, so the overwrite reading is of the record as exposed and says nothing about the history log.

\paragraph{The store-written superseded list, and the renderer's cap.}
A registered follow-up asked whether a filter on the metadata FR-Bank itself exposes could recover the obligation gain the store-written summary loses.
Two candidate rules were fixed before any call, with a precision bar of $0.80$, and a census of the store decided it at no cost.
Two filters on the metadata FR-Bank exposes --- the store's own supersession pointers, alone and restricted to an active successor at least fourteen days later --- leave the superseded list at $15.04$ and $8.96$ names per item against $1.75$ under ground truth, at precision $0.035$ and $0.062$. The store records $5{,}968$ deactivations and the manifest $104$ supersessions, and on the fields the store exposes the two are indistinguishable. Forty of the $104$ are still active in the store. The believed-state loss is not a filtering problem.
The one condition that ran is descriptive: the recorded store-written store versions rendered with the per-category cap of fifteen names lifted.
The superseded-names section then runs to $32.55$ names per item against $15.04$ under the cap and $1.75$ under the ground-truth summary.
Its recall of the ground-truth names rises from $30.0\%$ to $66.6\%$ at precision $0.0348$ under the cap and $0.0357$ without it.
On the $661$ obligation questions the obligation AUROC reads blind $0.4946$, ground-truth $0.7339$, store-written $0.5959$, and store-written with the cap lifted $0.6513$.
The paired gains over blind are $+0.2393$ $[+0.1801,+0.2949]$, $+0.1012$ $[+0.0458,+0.1543]$ (persona-clustered $[+0.0217,+0.1744]$) and $+0.1567$ $[+0.0961,+0.2162]$ (persona-clustered $[+0.0552,+0.2536]$).
The lifted summary retains $65.5\%$ of the ground-truth gain, item-level $[44.6\%,87.2\%]$ and persona-clustered $[31.5\%,93.6\%]$, against the recorded $42.3\%$ under the cap.
The lifted-minus-capped difference is $+0.0554$, item-level $[+0.0110,+0.0992]$ and persona-clustered $[-0.0019,+0.1173]$, so the two intervals disagree and the wider one governs in Section~\ref{sec:backends}.
The three registered texts for this row (Appendix~\ref{app:registration}) read as follows.
Under the \believed{} index the NOK gain is $+0.1012$ ($95\%$ CI $[+0.0217,+0.1744]$, with questions grouped by user profile). That keeps $42.3\%$ of the \trueidx{} gain, short of the $60\%$ we had set as the threshold.
Lifting the cap on the believed list, which raises its recall of ground-truth superseded names from $30.0$ to $66.6$ percent at unchanged precision, retains $65.5$ percent of the \trueidx{} gain, at or above the $60$ percent threshold.
Part of the believed-state loss recorded at $42.3$ percent is an effect of the renderer's cap of fifteen names rather than of the store's metadata.
NOK degrades materially under believed metadata but remains positive.

Haiku 4.5 fares better where FR-Bank keeps the target's name: with the uncapped store-written summary it scores $0.87$ on the supersession obligations whose name survives, against $0.89$ under ground truth, and $0.68$ where the name is missing (Table~\ref{tab:bh-split}). This descriptive split resembles the target-name ablation, but the missing-name subgroup is small and its paired decline is not reliable.

\paragraph{Haiku 4.5 under the store's own metadata.}
A descriptive registration scored Haiku 4.5 at five calls per question on the recorded store-written texts under both renderings, the capped ($15.04$ superseded names per item) and the lifted ($32.55$ names, at most $50$), against its recorded blind ($0.4150$) and ground-truth ($0.8210$) rows on the full obligation set, with the name lookup on the same lists ($0.5858$ capped, $0.6588$ lifted, persona interval $[0.5777,0.7319]$).
Haiku 4.5 reads $0.7069$ capped ($[0.6657,0.7504]$; $[0.6620,0.7541]$) and $0.7364$ lifted ($[0.6911,0.7858]$; $[0.6665,0.8024]$), $-0.1141$ and $-0.0847$ below its ground-truth row, both reliable, and $+0.1212$ ($[+0.0628,+0.1849]$; $[+0.0317,+0.2107]$) and $+0.0776$ ($[+0.0272,+0.1355]$; $[-0.0031,+0.1498]$) above the name lookup.
Its retention, the ratio of the paired resampled differences, is $0.7190$ capped ($[0.6214,0.8497]$; $[0.5935,0.8603]$) and $0.7916$ lifted ($[0.6838,0.9243]$; $[0.6342,0.9425]$), at or above the sixty-percent line under both renderings.
The sixty-percent line was a conventional threshold, fixed before the main study, with no recorded reason for its value.
GPT-4.1 mini's retention figures above divide by an indexed-minus-blind difference that Appendix~\ref{app:laddertable} does not report as a gain, because its blind score is one value on $604$ of the $661$ questions, so Table~\ref{tab:levels} reports retention for Haiku 4.5 only.
Split by whether the obligation's own scrubbed name survives in the store-written list (Table~\ref{tab:bh-split}), Haiku 4.5 reads the $53$ supersession obligations whose name survives under the lifted rendering at $0.8658$ against $0.8855$ under the ground-truth summary ($-0.0197$, $[-0.0690,+0.0303]$; $[-0.0842,+0.0424]$), and the $10$ whose name is gone at $0.6839$ against $0.9122$ ($-0.2283$, $[-0.4288,-0.0430]$; $[-0.6568,+0.0182]$); under the cap the halves are $24$ and $39$, at $0.7993$ and $0.6933$.

\begin{table}[H]
\caption{Haiku 4.5 under the store's own metadata, the $63$ supersession obligations split by whether the item's own scrubbed name survives in the store-written list, each half against the same $557$ negatives: Haiku 4.5 under the store-written and the ground-truth summary, the lookup on the store-written list, and the paired differences, each with its item-level and persona-grouped intervals on the rows beneath. Descriptive; no prediction was registered.}
\label{tab:bh-split}
\centering
\tiny
\setlength{\tabcolsep}{1.5pt}
\begin{tabular}{@{}llrrrrll@{}}
\toprule
Rend. & own name & $n$ & Haiku, store-written & ground-truth & lookup & store-written minus ground-truth & Haiku minus lookup \\
\midrule
capped & present & $24$ & $0.7993$ & $0.8925$ & $0.7292$ & $-0.0932$ & $+0.0701$ \\
 & item & & & & & $[-0.1746,-0.0006]$ & $[-0.0204,+0.1663]$ \\
 & persona & & & & & $[-0.2041,+0.0235]$ & $[-0.0261,+0.1709]$ \\
capped & absent & $39$ & $0.6933$ & $0.8880$ & $0.6146$ & $-0.1947$ & $+0.0787$ \\
 & item & & & & & $[-0.2673,-0.1224]$ & $[-0.0137,+0.1666]$ \\
 & persona & & & & & $[-0.2813,-0.1011]$ & $[-0.0497,+0.2355]$ \\
lifted & present & $53$ & $0.8658$ & $0.8855$ & $0.7593$ & $-0.0197$ & $+0.1065$ \\
 & item & & & & & $[-0.0690,+0.0303]$ & $[+0.0411,+0.1785]$ \\
 & persona & & & & & $[-0.0842,+0.0424]$ & $[+0.0233,+0.1846]$ \\
lifted & absent & $10$ & $0.6839$ & $0.9122$ & $0.7969$ & $-0.2283$ & $-0.1130$ \\
 & item & & & & & $[-0.4288,-0.0430]$ & $[-0.3613,+0.1503]$ \\
 & persona & & & & $[0.6722,0.8992]$ & $[-0.6568,+0.0182]$ & $[-0.5208,+0.4381]$ \\
\bottomrule
\end{tabular}
\end{table}

\begin{table}[H]
\caption{Obligation AUROC on the full obligation set ($661$ questions, $104$ obligations) under the ground-truth summary and under FR-Bank's store-written summary, rendered with the fifteen-name cap and with the cap lifted. Retention is Haiku 4.5's store-written gain as a fraction of its ground-truth gain.}
\label{tab:levels}
\centering
\scriptsize
\setlength{\tabcolsep}{3pt}
\begin{tabular}{@{}lrrr@{}}
\toprule
Reader & Ground-truth summary & Store-written, capped & Store-written, cap lifted \\
\midrule
GPT-4.1 mini & $0.73$ & $0.60$ & $0.65$ \\
Haiku 4.5 & $0.82$ & $0.71$ & $0.74$ \\
\quad retention$^{a}$ & & $0.72$ & $0.79$ \\
Name lookup & $0.75$ & $0.59$ & $0.66$ \\
\bottomrule
\end{tabular}
\par\smallskip
\begin{minipage}{0.97\linewidth}\scriptsize
$^{a}$The ratio of the paired resampled differences, with the item-level and the persona-grouped interval: capped $[0.62,0.85]$ and $[0.59,0.86]$, lifted $[0.68,0.92]$ and $[0.63,0.94]$. Reported for Haiku 4.5 only, because GPT-4.1 mini's blind score is one value on $604$ of the $661$ questions and the lookup has no blind score (Appendix~\ref{app:census}).
\end{minipage}
\end{table}

\paragraph{What the store is in each scored condition, and where the labels come from.}
On FR-Bank the store is its own persona bank, read through the store's own fields (\texttt{is\_active}, \texttt{is\_retracted}, \texttt{supersedes}) by the one renderer into the store-written summary, capped at fifteen names per category or with the cap lifted.
The \pn{} labels come from the manifest, checked by hand, and each obligation's forbidding records are entries of that same bank, so the forbidding content is stored by construction for all $104$ obligations.
What the label does not say is whether the alignment of manifest events to stored records marks one of those records, which is what the ground-truth summary's superseded-names list and the ground-truth tags are built from: it marks $82$ at the $0.80$ bar, and the $22$ it does not mark are all retraction- or expiry-derived.
So on FR-Bank the label is manifest-relative, the ground-truth summary is alignment-relative and the store-written summary is store-relative; the three coincide on the $63$ supersession obligations and diverge on the counted states.
Table~\ref{tab:levels-strata} re-scores the rows of Table~\ref{tab:levels} on each stratum: on the $82$ the alignment marks, the capped store-written summary still costs GPT-4.1 mini $-0.1274$, Haiku 4.5 $-0.1258$ and the lookup $-0.2027$ against the ground-truth summary, each with both intervals clear of zero, so the loss the body reports is not an artifact of the obligations the alignment misses; on the $22$ it does not mark, the ground-truth summary itself reads $0.5210$ for mini, $0.7074$ for Haiku 4.5 and $0.4272$ for the lookup, because its list is built from the same alignment.
On Graphiti the store is one graph per persona built by the backend's own extraction from the same conversations with every lifecycle field removed, so whether it retains the forbidding content is a question of extraction (Appendix~\ref{app:posthoc-graphiti}, Table~\ref{tab:graphiti-strata}).
On Mem0 and A-MEM no obligation is scored: their scored populations hold \present{} and \absent{} items only, so their evaluation is the coverage judgment and the census, and there is no \pn{} label to state relative to what they retain.

\begin{table}[H]
\caption{The rows of Table~\ref{tab:levels} re-scored on strata of the $104$ obligations, the $557$ negatives unchanged (post hoc). The alignment of manifest events to stored records marks a forbidding record for $82$ obligations at the $0.80$ overlap bar ($63$ supersession, $8$ retraction, $11$ expiry), $84$ at $0.75$ and $95$ at $0.50$; the store's own fields mark a member of $73$; $56$ are marked by both, $26$ by the alignment alone, $17$ by the store alone and $5$ by neither. Upper panel: the AUROC per stratum. Lower panel: on the $82$ the alignment marks, each store-written row minus its ground-truth row with both intervals, $^{*}$ where both exclude zero. On the $22$ it does not mark every reader is at or below $0.71$ under every summary, the ground-truth summary included, because the ground-truth name list is built from the same alignment.}
\label{tab:levels-strata}
\centering\tiny\setlength{\tabcolsep}{1.5pt}
\begin{tabular}{@{}lrrrrrrrrrr@{}}
\toprule
Obligations & $n$ & \multicolumn{3}{c}{GPT-4.1 mini} & \multicolumn{3}{c}{Haiku 4.5} & \multicolumn{3}{c}{Name lookup} \\
 & & truth & capped & lifted & truth & capped & lifted & truth & capped & lifted \\
\midrule
all $104$ (the printed rows) & $104$ & $0.7339$ & $0.5959$ & $0.6513$ & $0.8210$ & $0.7069$ & $0.7364$ & $0.7484$ & $0.5858$ & $0.6588$ \\
alignment marks the forbidding record ($0.80$ bar) & $82$ & $0.7910$ & $0.6636$ & $0.7297$ & $0.8515$ & $0.7258$ & $0.7821$ & $0.8346$ & $0.6319$ & $0.7061$ \\
the same at the $0.75$ bar & $84$ & $0.7866$ & $0.6573$ & $0.7180$ & $0.8468$ & $0.7235$ & $0.7738$ & $0.8177$ & $0.6261$ & $0.6986$ \\
the same at the $0.50$ bar & $95$ & $0.7571$ & $0.6193$ & $0.6750$ & $0.8358$ & $0.7156$ & $0.7473$ & $0.7766$ & $0.5942$ & $0.6640$ \\
not marked at $0.80$ (the $22$) & $22$ & $0.5210$ & $0.3433$ & $0.3591$ & $0.7074$ & $0.6368$ & $0.5659$ & $0.4272$ & $0.4138$ & $0.4822$ \\
the store's own mark on a member & $73$ & $0.7366$ & $0.5996$ & $0.6627$ & $0.8303$ & $0.7133$ & $0.7622$ & $0.7977$ & $0.5793$ & $0.6763$ \\
no store mark on any member & $31$ & $0.7276$ & $0.5871$ & $0.6244$ & $0.7992$ & $0.6919$ & $0.6755$ & $0.6324$ & $0.6011$ & $0.6175$ \\
marked by both & $56$ & $0.7883$ & $0.6741$ & $0.7494$ & $0.8664$ & $0.7262$ & $0.8185$ & $0.8975$ & $0.6436$ & $0.7381$ \\
marked by the alignment only & $26$ & $0.7970$ & $0.6411$ & $0.6872$ & $0.8196$ & $0.7247$ & $0.7037$ & $0.6991$ & $0.6068$ & $0.6373$ \\
\bottomrule
\end{tabular}\\[4pt]
\begin{tabular}{@{}ll@{}}
\toprule
On the $82$ the alignment marks: store-written minus ground truth & difference, item interval; persona interval \\
\midrule
GPT-4.1 mini, capped & $-0.1274$ $[-0.1813,-0.0731]$; $[-0.2036,-0.0485]$$^{*}$ \\
GPT-4.1 mini, lifted & $-0.0614$ $[-0.1149,-0.0053]$; $[-0.1425,+0.0099]$ \\
Haiku 4.5, capped & $-0.1258$ $[-0.1795,-0.0701]$; $[-0.1913,-0.0600]$$^{*}$ \\
Haiku 4.5, lifted & $-0.0694$ $[-0.1185,-0.0204]$; $[-0.1401,-0.0060]$$^{*}$ \\
Name lookup, capped & $-0.2027$ $[-0.2721,-0.1319]$; $[-0.2866,-0.1270]$$^{*}$ \\
Name lookup, lifted & $-0.1285$ $[-0.1842,-0.0716]$; $[-0.1993,-0.0615]$$^{*}$ \\
\bottomrule
\end{tabular}
\end{table}

\paragraph{A fourth backend, measured on three personas.}
A trial on an unmodified public release of a fourth backend, Graphiti \citep{zep}, a temporal knowledge-graph store built by another group, asks whether the self-description problem is confined to the three backends above.
On three personas and $137$ episodes it names at least one superseded attribute per persona, so its summary is not degenerate.
Against the persona manifests it recovers $2$ of $10$ supersession events at the $0.80$ overlap bar this study inherited, a recall of $0.20$.
At the $0.75$ and $0.50$ bands the same census recovers $5$ and $6$, and three of the ten events sit between $0.75$ and $0.80$.
The verdict therefore turns on a threshold rather than on a wide margin, and all three bands are reported.
What it does mark is mostly not a manifest event, because precision runs $0.00$ to $0.04$ across the three personas.
Retraction and category are inexpressible under its stock defaults, so the summary emits the marker for both.
Three properties of the store account for most of the misses.
The user is split across three entity nodes in every persona, so a later fact about the same slot does not meet the earlier one.
About a tenth of edges carry no validity time, and invalidation requires one on both sides, and some temporal bookkeeping is incoherent, with one edge invalidated before its own validity begins.
The failure mode is loss of content with over-marking in what survives, the same direction as FR-Bank's at a far smaller absolute scale.
Because it covers three personas it is a trial rather than a census, and no obligation judgment is scored on this backend.
Disclosure, because it bears on independence: our earlier system \citep{fr} ran on a copy of this backend carrying four documented numeric-preservation modifications, and ingested these personas into it, before FR-Bank replaced it.
Re-running the unmodified release on the same personas and episodes changes edge counts by $2.4\%$ and leaves supersession recall identical at the frozen bar, so those modifications are near-inert on this corpus.
Since the trial, the census and both monitoring routes were run on all forty personas (Appendix~\ref{app:posthoc-graphiti}). The name lookup retains useful signal; native tags add no established monitoring gain.

\FloatBarrier
\subsection{Own-name ablation, retrieved-view monitors, the rubric gate, and the downstream experiment}\label{app:resp}

Three mechanism tests and one downstream experiment act on the judgment.
Each was preregistered with its texts written before any call (Table~\ref{tab:registrations-b}), so every outcome below is read against a text that predates it.
Every monitor call is GPT-4.1 mini at temperature $0$.
The blind and ground-truth-summary rows are the main study's, read from the recorded rows.
Intervals are the item-level paired bootstrap with the interval grouped by persona beside it.
Where the two disagree the wider one decides, and the tolerance is $\pm0.05$.

\begin{table}[H]
\caption{Own-name ablation. The superseded-names section of the ground-truth summary is re-rendered without the item's own attribute name and every name within its encoder neighbourhood ($102$ of $661$ summaries change); GPT-4.1 mini's obligation discrimination under that summary against the main study's row, per lifecycle state (positives of one state against the same $557$ negatives) and whole; the name lookup on the same two texts beside.}
\label{tab:resp-l1}\centering\scriptsize\setlength{\tabcolsep}{3pt}
\resizebox{\textwidth}{!}{\begin{tabular}{@{}lrrrrllrr@{}}\toprule
Population & $n$ / pos. & Ground-truth & Own name removed & $\Delta$ & item-level $95\%$ & by persona & Lookup, ground-truth & Lookup, removed \\ \midrule
whole population & 661 / 104 & $0.7339$ & $0.5016$ & $-0.2323$ & $[-0.2890,-0.1753]$ & $[-0.3054,-0.1540]$ & $0.7484$ & $0.4913$ \\
supersession-derived & 620 / 63 & $0.8572$ & $0.5310$ & $-0.3262$ & $[-0.3991,-0.2548]$ & $[-0.4327,-0.2167]$ & $0.9534$ & $0.5108$ \\
expiry-derived & 572 / 15 & $0.5110$ & $0.5136$ & $+0.0026$ & $[-0.0028,+0.0079]$ & $[-0.0041,+0.0103]$ & $0.4823$ & $0.4887$ \\
retraction-derived & 583 / 26 & $0.5638$ & $0.4234$ & $-0.1404$ & $[-0.2569,-0.0508]$ & $[-0.3305,-0.0028]$ & $0.4053$ & $0.4455$ \\
\bottomrule\end{tabular}}\end{table}

The draw-noise floor, the same $102$ prompts sent once more, moves the whole-population score by $+0.0179$ ($[-0.0117,+0.0523]$).
The ablation's move on the same items, $-0.3250$, therefore lies outside the noise on every state that has positives among them.
On retraction-derived items, which the summary only counts, the ablation moves GPT-4.1 mini's discrimination by $-0.1404$ with interval $[-0.2569,-0.0508]$, outside the tolerance.
The change cannot come from a name the positives carried, so it comes through the negatives whose names were removed, or through the model's response to a section that now reads ``(none)'' against non-zero counts.
Eleven of the $26$ retraction-derived obligations carry their own name in the section the ablation edits.

\begin{table}[H]
\caption{Retrieved-view and hybrid monitors. The five entries a similarity retriever returns from the item's store, active or not, each with the lifecycle flags the store's ground-truth join records; the retrieved-view monitor sees those five and no summary, the hybrid monitor sees both. Obligation rows on the $661$, per state and whole; coverage rows on the $307$ against the ground-truth summary; reach of the record that forbids use, a census over the $104$ obligations with a Wilson interval.}
\label{tab:resp-l2}\centering\scriptsize\setlength{\tabcolsep}{3pt}
\resizebox{\textwidth}{!}{\begin{tabular}{@{}llrrrrll@{}}\toprule
Monitor & Population & $n$ / pos. & Truthful & Monitor & $\Delta$ & item-level $95\%$ & by persona \\ \midrule
Retrieved view & obligation, whole & 661 / 104 & $0.7339$ & $0.6851$ & $-0.0488$ & $[-0.1274,+0.0285]$ & $[-0.1454,+0.0389]$ \\
Retrieved view & obligation, supersession & 620 / 63 & $0.8572$ & $0.6853$ & $-0.1719$ & $[-0.2641,-0.0734]$ & $[-0.2764,-0.0682]$ \\
Retrieved view & obligation, expiry & 572 / 15 & $0.5110$ & $0.7474$ & $+0.2364$ & $[+0.1053,+0.3829]$ & $[+0.0645,+0.4037]$ \\
Retrieved view & obligation, retraction & 583 / 26 & $0.5638$ & $0.6487$ & $+0.0850$ & $[-0.0682,+0.2235]$ & $[-0.1435,+0.2764]$ \\
Retrieved view & coverage, hard tier & 307 / 35 & $0.7003$ & $0.7870$ & $+0.0867$ & $[-0.0261,+0.1908]$ & $[+0.0041,+0.1683]$ \\
Summary + retrieved view & obligation, whole & 661 / 104 & $0.7339$ & $0.8303$ & $+0.0964$ & $[+0.0415,+0.1554]$ & $[+0.0227,+0.1635]$ \\
Summary + retrieved view & obligation, supersession & 620 / 63 & $0.8572$ & $0.9044$ & $+0.0472$ & $[-0.0337,+0.1206]$ & $[-0.0510,+0.1348]$ \\
Summary + retrieved view & obligation, expiry & 572 / 15 & $0.5110$ & $0.6858$ & $+0.1747$ & $[+0.0320,+0.3287]$ & $[+0.0195,+0.3335]$ \\
Summary + retrieved view & obligation, retraction & 583 / 26 & $0.5638$ & $0.7342$ & $+0.1704$ & $[+0.0430,+0.2926]$ & $[-0.0089,+0.3471]$ \\
Summary + retrieved view & coverage, hard tier & 307 / 35 & $0.7003$ & $0.8241$ & $+0.1238$ & $[+0.0206,+0.2265]$ & $[+0.0471,+0.1994]$ \\
Retrieved view, store's own tags & obligation, whole & 661 / 104 & $0.7339$ & $0.8144$ & $+0.0805$ & $[+0.0177,+0.1439]$ & $[-0.0081,+0.1660]$ \\
Retrieved view, store's own tags & obligation, supersession & 620 / 63 & $0.8572$ & $0.8189$ & $-0.0383$ & $[-0.1155,+0.0402]$ & $[-0.1192,+0.0442]$ \\
Retrieved view, store's own tags & obligation, expiry & 572 / 15 & $0.5110$ & $0.7866$ & $+0.2756$ & $[+0.1209,+0.4275]$ & $[-0.0346,+0.5094]$ \\
Retrieved view, store's own tags & obligation, retraction & 583 / 26 & $0.5638$ & $0.8194$ & $+0.2556$ & $[+0.1306,+0.3786]$ & $[+0.0608,+0.3916]$ \\
\addlinespace
\multicolumn{8}{@{}l}{\textit{Haiku 4.5, five calls averaged, against its own summary row; the own-name ablation on the $102$ changed items with the recorded score elsewhere}}\\
Summary + retrieved view, Haiku 4.5 & obligation, whole & 661 / 104 & $0.8210$ & $0.7863$ & $-0.0347$ & $[-0.0930,+0.0248]$ & $[-0.1109,+0.0433]$ \\
Summary + retrieved view, Haiku 4.5 & obligation, supersession & 620 / 63 & $0.8897$ & $0.7877$ & $-0.1020$ & $[-0.1739,-0.0265]$ & $[-0.2006,+0.0014]$ \\
Summary + retrieved view, Haiku 4.5 & obligation, expiry & 572 / 15 & $0.6874$ & $0.8270$ & $+0.1396$ & $[-0.0338,+0.3048]$ & $[-0.0914,+0.3385]$ \\
Summary + retrieved view, Haiku 4.5 & obligation, retraction & 583 / 26 & $0.7317$ & $0.7595$ & $+0.0278$ & $[-0.0928,+0.1408]$ & $[-0.1534,+0.1843]$ \\
Retrieved view, Haiku 4.5 & obligation, whole & 661 / 104 & $0.8210$ & $0.7134$ & $-0.1077$ & $[-0.1732,-0.0369]$ & $[-0.1939,-0.0203]$ \\
Retrieved view, Haiku 4.5 & obligation, supersession & 620 / 63 & $0.8897$ & $0.6911$ & $-0.1986$ & $[-0.2834,-0.1039]$ & $[-0.2994,-0.0967]$ \\
Retrieved view, Haiku 4.5 & obligation, expiry & 572 / 15 & $0.6874$ & $0.7743$ & $+0.0870$ & $[-0.0735,+0.2452]$ & $[-0.0493,+0.1954]$ \\
Retrieved view, Haiku 4.5 & obligation, retraction & 583 / 26 & $0.7317$ & $0.7322$ & $+0.0005$ & $[-0.1390,+0.1288]$ & $[-0.2181,+0.1773]$ \\
Own-name ablation, Haiku 4.5 & obligation, whole & 661 / 104 & $0.8210$ & $0.6882$ & $-0.1328$ & $[-0.1778,-0.0960]$ & $[-0.1940,-0.0790]$ \\
Own-name ablation, Haiku 4.5 & obligation, supersession & 620 / 63 & $0.8897$ & $0.6759$ & $-0.2139$ & $[-0.2752,-0.1584]$ & $[-0.3005,-0.1388]$ \\
Own-name ablation, Haiku 4.5 & obligation, expiry & 572 / 15 & $0.6874$ & $0.6914$ & $+0.0040$ & $[+0.0003,+0.0094]$ & $[-0.0010,+0.0122]$ \\
Own-name ablation, Haiku 4.5 & obligation, retraction & 583 / 26 & $0.7317$ & $0.7164$ & $-0.0153$ & $[-0.0516,+0.0164]$ & $[-0.0478,+0.0136]$ \\
\midrule
\multicolumn{8}{@{}l}{Reach of the forbidding record, five items: $54$ of $104$ ($0.5192$, Wilson $[0.4242,0.6128]$); supersession $22$ of $63$, retraction $20$ of $26$, expiry $12$ of $15$.} \\
\multicolumn{8}{@{}l}{Ten items: $74$ of $104$ ($0.7115$); supersession $37$ of $63$, retraction $25$ of $26$, expiry $12$ of $15$.} \\
\bottomrule\end{tabular}}\end{table}
\begin{table}[H]
\caption{Obligation AUROC of GPT-4.1 mini on the tagged view, on the full obligation set ($661$ questions, $104$ obligations). $\Delta$ is the monitor minus the comparison named in its row, with both intervals; the two indented rows score that state's obligations against the same $557$ negatives. Haiku 4.5 and the readers with no generative-model call are in Appendix~\ref{app:resp}.}
\label{tab:routes}
\centering
\scriptsize
\setlength{\tabcolsep}{3pt}
\begin{tabular}{@{}lrlll@{}}
\toprule
Monitor, tag source & AUROC & $\Delta$ (against) & item interval & persona interval \\
\midrule
Summary alone & $0.73$ & & & \\
Tagged view, ground-truth tags & $0.69$ & $-0.05$ (the summary alone) & $[-0.13,+0.03]$ & $[-0.15,+0.04]$ \\
\quad on expiry, 15 obligations & $0.75$ & $+0.24$ (the summary alone, $0.51$) & $[+0.11,+0.38]$ & $[+0.06,+0.40]$ \\
\quad on supersession, 63 obligations & $0.69$ & $-0.17$ (the summary alone, $0.86$) & $[-0.26,-0.07]$ & $[-0.28,-0.07]$ \\
Summary plus tagged view, ground-truth tags & $0.83$ & $+0.10$ (the summary alone) & $[+0.04,+0.16]$ & $[+0.02,+0.16]$ \\
Tagged view, the store's own tags & $0.81$ & $+0.13$ (the view under ground-truth tags) & $[+0.07,+0.18]$ & $[+0.06,+0.21]$ \\
\bottomrule
\end{tabular}
\end{table}

\paragraph{The tagged view under the store's own tags, and Haiku 4.5 on the retrieved-view monitors.}
A descriptive registration re-rendered the same five retrieved records with FR-Bank's own lifecycle tags in place of the ground-truth join: the store's inactive flag as superseded, its retracted flag as retracted, and an event time before the persona's cutoff as expired, the rule the store-written summary uses.
The store's own tags change the tag on $1{,}273$ of the $3{,}305$ lines shown, and against the ground-truth non-active flag the store's tag has precision $0.14$ and recall $0.87$ over lines and precision $0.67$ and recall $0.78$ over the forbidding records in view.
GPT-4.1 mini on that view scores $0.8144$ against $0.6851$ under ground-truth tags, a difference of $+0.1292$ with intervals $[+0.0666,+0.1826]$ and $[+0.0606,+0.2060]$, and $+0.0805$ against the ground-truth summary (Table~\ref{tab:resp-l2}); under the registration's reading rule the view does not collapse (the fall is not a reliable one beyond the tolerance), so the system-builder rule on tags stays unconditional.
The same registration ran Haiku 4.5 at five calls per question on the hybrid monitor, then the tagged view, then the own-name ablation, in that order; hybrid ran, retrieved ran, the own-name ablation ran (Appendix~\ref{app:economics}).
Against its own summary row Haiku 4.5 gains $-0.0347$ shown both and $-0.1077$ on the tagged view alone, so the two routes read for the second model as they do for the first.

Without the detection step of the two-step rubric, GPT-4.1 mini's blind obligation score takes $0.0$ on $604$ and $0.1$ on $57$ of the $661$ questions, two distinct values.
Its one-sample discrimination is $0.4888$ with interval $[0.4646,0.5165]$, so the constant was not the gate.

\begin{table}[H]
\caption{The tagged view, the rows Table~\ref{tab:routes} leaves out: Haiku 4.5 on the view and on summary plus view under ground-truth tags, and the four readers with no generative-model call under each tag source, each against the GPT-4.1 mini monitor that saw the same input. The readers are tag density (the fraction of lines tagged non-active), the tagged-name lookup (the question against the names of the tagged lines), the forbidding-record indicator (the item's forbidding record in view and tagged) and the hybrid of the name lookup with the tagged-name lookup. $\Delta$ is the first-named score minus the second, with both intervals.}
\label{tab:routes-full}
\centering
\tiny
\setlength{\tabcolsep}{2pt}
\begin{tabular}{@{}lrrrll@{}}
\toprule
Row & first & second & $\Delta$ & item interval & persona interval \\
\midrule
Haiku 4.5, view (ground-truth tags) minus its summary alone & $0.7134$ & $0.8210$ & $-0.1077$ & $[-0.1732,-0.0369]$ & $[-0.1939,-0.0203]$ \\
Haiku 4.5, summary plus view (ground-truth tags) minus its summary alone & $0.7863$ & $0.8210$ & $-0.0347$ & $[-0.0930,+0.0248]$ & $[-0.1109,+0.0433]$ \\
\addlinespace
\multicolumn{6}{@{}l}{\textit{Ground-truth tags: the GPT-4.1 mini monitor on the same input minus the reader with no generative-model call}}\\
View monitor minus tag density & $0.6851$ & $0.6157$ & $+0.0694$ & $[+0.0375,+0.1053]$ & $[+0.0257,+0.1103]$ \\
View monitor minus tagged-name lookup & $0.6851$ & $0.6313$ & $+0.0538$ & $[+0.0232,+0.0899]$ & $[+0.0110,+0.0951]$ \\
View monitor minus forbidding-record indicator & $0.6851$ & $0.6683$ & $+0.0168$ & $[-0.0192,+0.0572]$ & $[-0.0380,+0.0784]$ \\
Summary-plus-view monitor minus hybrid lookup & $0.8303$ & $0.7708$ & $+0.0595$ & $[-0.0005,+0.1228]$ & $[-0.0198,+0.1498]$ \\
\addlinespace
\multicolumn{6}{@{}l}{\textit{The store's own tags: the same, against the view monitor under those tags}}\\
View monitor minus tag density & $0.8144$ & $0.6209$ & $+0.1934$ & $[+0.1446,+0.2462]$ & $[+0.1280,+0.2536]$ \\
View monitor minus tagged-name lookup & $0.8144$ & $0.6523$ & $+0.1621$ & $[+0.1123,+0.2164]$ & $[+0.0967,+0.2320]$ \\
View monitor minus forbidding-record indicator & $0.8144$ & $0.7212$ & $+0.0932$ & $[+0.0440,+0.1453]$ & $[+0.0273,+0.1552]$ \\
View monitor minus hybrid lookup & $0.8144$ & $0.6857$ & $+0.1287$ & $[+0.0771,+0.1878]$ & $[+0.0597,+0.2003]$ \\
\bottomrule
\end{tabular}
\end{table}

\paragraph{Retrieval that returns tags, as configured.}
The retriever is the store's own cosine search over every entry the item's store variant holds, active or not, on the bank's cached embedding vectors.
It applies no lifecycle mask, no lexical channel, no category forcing, no multi-hop and no distillation, so every entry can be returned; the top ten are taken and the top five used.
Each returned line carries the lifecycle flags the store's ground-truth join records, the category, the scrubbed attribute, the age in days, and the text.
That is a modification of FR-Bank's stock configuration, which masks non-active records, and it is what makes the tags a route at all.

\begin{table}[H]
\caption{The downstream experiment. Seven registered policies assemble a view from the five retrieved entries and, where a rule fires, the store's non-active records for the attributes the summary names (the oracle: the item's own attribute; the gold labels decide abstention). Table~\ref{tab:ds-body-lower} carries each policy's fires, reach, view size, tagged reach and coverage; the first block here adds the abstention counts and the reach by the state the obligation arises from. Stage $2$ is one GPT-4.1 mini answer per distinct view, scored by the ground-truth token detector at its high-precision threshold ($0.80$) and at the ground-truth join's threshold ($0.50$), split by whether the forbidding record carried its tag. The literal-threshold lookup row is descriptive; its answer stage was registered after the first draft (Amendment 2) and prints in the second block, whose lower rows give the same counts under the answer stage run once more with an added instruction to treat tagged records as not to be used (Amendment 3; the Amendment 1 policy's views were not answered again). The last row of each block is the descriptive policy of the amendment registered after the study: the model rule with its obligation score elicited under the summary that names all three states, and the all-states fetch set, so that the counted-state gate can fire. Scored with the store's tags as joined, one retraction tag inverted (the negation paragraph).}
\label{tab:ds}\centering\scriptsize\setlength{\tabcolsep}{2.5pt}
\begin{tabular}{@{}lrrrrr@{}}\toprule
Policy & abstains & sup. (of $63$) & ret. (of $26$) & exp. (of $15$) & abstain, absent (of $383$) \\ \midrule
Top-5 only & $0$ & $22$ & $20$ & $12$ & $0$ \\
Oracle & $383$ & $63$ & $22$ & $14$ & $383$ \\
Model rule & $210$ & $53$ & $20$ & $12$ & $172$ \\
Lookup rule & $224$ & $53$ & $20$ & $12$ & $174$ \\
Model rule, all states named & $210$ & $53$ & $20$ & $12$ & $172$ \\
Lookup rule, all states named & $224$ & $53$ & $20$ & $12$ & $174$ \\
Lookup, literal threshold & $0$ & $62$ & $20$ & $12$ & $0$ \\
\addlinespace
Model rule, all states named and scored & $210$ & $54$ & $22$ & $13$ & $172$ \\
\bottomrule\end{tabular}\\[6pt]
\begin{tabular}{@{}lrrlr@{\hspace{10pt}}l@{}}\toprule
Policy & answered & stale at $0.80$ & stale at $0.50$, tagged / untagged & present answered & correct at $0.50$ \\ \midrule
Top-5 only & $54$ & $0$ & $2$ of $35$ / $7$ of $19$ & $174$ & $46$ ($0.2644$) \\
Oracle & $99$ & $0$ & $11$ of $82$ / $6$ of $17$ & $174$ & $46$ ($0.2644$) \\
Model rule & $74$ & $0$ & $5$ of $61$ / $4$ of $13$ & $155$ & $44$ ($0.2839$) \\
Lookup rule & $70$ & $0$ & $5$ of $61$ / $2$ of $9$ & $146$ & $40$ ($0.2740$) \\
Model rule, all states named & $74$ & $0$ & $5$ of $61$ / $4$ of $13$ & $155$ & $44$ ($0.2839$) \\
Lookup rule, all states named & $70$ & $0$ & $4$ of $61$ / $2$ of $9$ & $146$ & $39$ ($0.2671$) \\
Lookup, literal threshold & $94$ & $0$ & $7$ of $75$ / $7$ of $19$ & $174$ & $46$ ($0.2644$) \\
\addlinespace
Model rule, all states named and scored & $78$ & $0$ & $10$ of $67$ / $2$ of $11$ & $155$ & $44$ ($0.2839$) \\
\addlinespace
\multicolumn{6}{@{}l}{\textit{With the added instruction to treat tagged records as not to be used (Amendment 3)}}\\
Top-5 only & $54$ & $0$ & $2$ of $35$ / $8$ of $19$ & $174$ & $47$ ($0.2701$) \\
Oracle & $99$ & $0$ & $11$ of $82$ / $7$ of $17$ & $174$ & $47$ ($0.2701$) \\
Model rule & $74$ & $0$ & $6$ of $61$ / $4$ of $13$ & $155$ & $44$ ($0.2839$) \\
Lookup rule & $70$ & $0$ & $6$ of $61$ / $2$ of $9$ & $146$ & $40$ ($0.2740$) \\
Model rule, all states named & $74$ & $2$ & $6$ of $61$ / $4$ of $13$ & $155$ & $44$ ($0.2839$) \\
Lookup rule, all states named & $70$ & $1$ & $5$ of $61$ / $2$ of $9$ & $146$ & $40$ ($0.2740$) \\
Lookup, literal threshold & $94$ & $0$ & $8$ of $75$ / $8$ of $19$ & $174$ & $46$ ($0.2644$) \\
\bottomrule\end{tabular}\end{table}

The store's ground-truth join flags the forbidding record as non-active for $82$ of the $104$ obligations.
The other $22$ retrieve as active, so no fetch by lifecycle state reaches them and the oracle's ceiling is $0.7885$ by construction; its reach of $99$ counts the $17$ of those $22$ that the retriever returns anyway.
With the forbidding record in view and its tag shown, GPT-4.1 mini gives the stale value as the answer in $11$ of $82$ cases at the gold threshold, and in $6$ of $17$ where the record appeared as active.
The retention of the oracle's gain, $0.7561$, is a near miss under the stricter interval convention, with $[0.563,0.939]$ grouped by persona against $0.60$.
The registered text (supplement, Section~S.3) reports the same interval at four decimals.
The $22$ unflagged records split by the fault in FR-Bank's own write path.
For $17$ of them ($14$ retraction-derived, $3$ expiry-derived) the store's own metadata marks at least one forbidding record superseded and none retracted, so the retraction or expiry was recorded through the supersession mechanism.
For the other $5$ ($4$ retraction-derived, $1$ expiry-derived, four of the ten distinct questions) no forbidding record carries any lifecycle mark at all; this census was run after the results were known.
Two further faults of the write path came out of a forensic audit of the answer stage, run after the results were known and gated against every recorded per-row score (the downstream audit report in the artifact and its ledger, section 9).
Active duplicates: for $35$ of the $89$ supersession and retraction obligations ($21$ and $14$) the view holds an active entry whose text asserts the stale value, a second copy of the forbidding content that the ground-truth join tied to no lifecycle event, while the item's own forbidding record may sit tagged beside it.
Every stale answer of GPT-4.1 mini under the oracle ($16$ of $16$ on those $89$) and $14$ of Haiku 4.5's $19$ fall on such a view, and none of mini's falls where the stale content sits only on tagged lines.
Negation: the ground-truth join assigns the retracted state by token coverage of the original fact at $0.80$ and reads no negation, so for one question, in both store variants, the retraction statement itself (``decided not to launch \ldots\ Oud and Geometry'') carries the retracted tag while the plan records stay active, because ``launching'' is not ``launch''.
An answerer that obeys the tag then reads the withdrawal as withdrawn.
The tagged and untagged split of Table~\ref{tab:ds} counts the item's own forbidding records and does not see these duplicates.

The negation case is one of the $26$ retraction obligations, the full-store view of that question, carrying an inverted tag because the ground-truth join reads no negation; its slot-ablated view carries the same tag but holds no line asserting the stale value, so it is outside the $35$ either way. Correcting the tag, the retraction statement active and the plan records retracted, moves the $35$ of $89$ to $34$ and the retraction count from $14$ to $13$. The oracle's stale answers split by whether the view holds an active line asserting the stale value $16$ against $0$ for GPT-4.1 mini and $14$ against $5$ for Haiku 4.5 as joined, and $15$ against $1$ and $13$ against $6$ with the tag corrected, both answerers having answered that question from the plan lines the view showed as active. Every answer-stage table was scored with the tag as joined and says so in its caption; nothing was re-answered.

The fourth downstream prediction could not be met by its own gate, because the gate's score was elicited under the summary that only counts the retracted and expired states.
A descriptive repair registered afterwards elicited GPT-4.1 mini's obligation score once more under the summary that names all three states, on the same $661$ questions and under the main study's protocol, and drove the all-states fetch from it.
That score reaches an AUROC of $0.7354$ against the recorded $0.7339$, a paired difference of $+0.0015$ with intervals $[-0.0385,+0.0494]$ and $[-0.0473,+0.0503]$.
Naming the counted states in the summary therefore does not change GPT-4.1 mini's whole-population judgment; per state it reads $0.8790$ on supersession, $0.5941$ on retraction and $0.3770$ on expiry.
Thresholded at the same base rate the new gate fires on $110$ items ($63$ obligations, $22$ present, $25$ absent) against $75$ for the recorded gate, and the two agree on $65$.
The policy that reads it puts the forbidding record in view for $89$ of $104$ obligations ($0.8558$; $54$ of $63$ supersession, $22$ of $26$ retraction, $13$ of $15$ expiry; $72$ tagged).
It abstains as the model rule does, and it keeps coverage at $60$ of $174$.
On the $41$ counted-state obligations it reaches $0.8537$ against the recorded gate's $0.7805$, a difference of $+0.0732$ whose item-level interval $[+0.0000,+0.1707]$ and persona-clustered interval $[+0.0000,+0.1875]$ both reach zero.
The reason is that the new gate fires on only $5$ of the $17$ records naming makes fetchable, $3$ of them records the retriever misses.
In the answer stage its $78$ answered obligations give the stale value in $0$ cases at the high-precision threshold and in $10$ of $67$ tagged and $2$ of $11$ untagged cases at the gold threshold.

A second descriptive registration ran the answer stage for the literal-threshold lookup rule, which fetches whenever the name lookup's cosine is at or above the base rate and never abstains, under the protocol above.
Of its $268$ answered views, $238$ are byte-identical to a view a registered policy had already assembled and reuse the recorded answer, and $30$ were new calls.
Its $94$ reached obligations give the stale value in $0$ cases at the high-precision threshold and in $7$ of $75$ tagged and $7$ of $19$ untagged cases at the gold threshold, and its $174$ licensed items are answered correctly in $46$ cases, as under retrieval alone (Table~\ref{tab:ds}).
The same registration ran Haiku 4.5 as a second answerer on every distinct view of the seven policies, $452$ views, at temperature $0$ with the same two prompt texts (Table~\ref{tab:ds-h45}).
Haiku 4.5 writes longer answers, so the high-precision threshold no longer reads zero, and at the gold threshold it gives the stale value about twice as often as GPT-4.1 mini under every policy, with the untagged records again the worse half.
A third descriptive registration answered the same $452$ views once more with one instruction added to the prompt: to treat every entry tagged superseded, retracted or expired as not to be used, and to report the store as not holding the answer when only such entries bear on the question.
Under it GPT-4.1 mini gives the stale value at the gold threshold in $10$ of $54$ cases under retrieval alone and $18$ of $99$ under the oracle, against $9$ and $17$ without it, and Haiku 4.5 in $17$ and $24$ against $19$ and $27$.
Haiku 4.5's stale answers at the high-precision threshold fall from $7$ and $9$ to $3$ and $3$, and correct answers move by at most one under either answerer (the lower rows of Tables~\ref{tab:ds} and~\ref{tab:ds-h45}).
The instruction addresses tagged lines, and the audit finds the stale sources untagged: GPT-4.1 mini's answers are byte-identical under it on $310$ of the $452$ views.
A fourth amendment, registered on day $35$ of the registration record with two predictions, adds to the user message a header naming the persona and pinning the current date, because the view prints every entry as ``User'' and gives ages without a date.
The persona's name appears in $47$ of the $174$ licensed-item views while $172$ of the $174$ questions name the persona; $15$ of GPT-4.1 mini's $28$ in-view misses under retrieval alone are abstentions with the record on the page, and Haiku 4.5 gives that mismatch as its reason in $94$ of its $145$ abstentions on those items.
It predicted, before any call, that the correct count on the $65$ licensed items in view would rise reliably by at least $0.10$ for each answerer, and that the in-view abstentions would fall by more than half: from $15$ to $7$ or fewer for GPT-4.1 mini, from $38$ to $19$ or fewer for Haiku 4.5.
Both predictions cleared their bars for both answerers.
GPT-4.1 mini's correct count on the $65$ rose from $37$ to $48$, $+0.1692$ with intervals $[+0.0769,+0.2769]$ at item level and $[+0.0678,+0.2603]$ grouped by persona; Haiku 4.5's rose from $38$ to $59$, $+0.3231$ with $[+0.2000,+0.4462]$ and $[+0.1833,+0.4714]$.
In-view abstentions fell to $3$ for each answerer, and the abstentions the detector had counted correct fell from $12$ to $3$ for Haiku 4.5.
Over the $174$ licensed items the correct count rose from $46$ to $56$ and from $48$ to $73$; over the $221$ licensed-item views abstentions fell from $66$ to $42$ and from $145$ to $56$.
On the scorable obligations the header changed little for GPT-4.1 mini: $17$ of $85$ stale under the oracle against $16$ before, all $17$ on views whose active lines assert the stale value and none on the $50$ tagged-only or empty views.
Haiku 4.5, abstaining on $25$ of the $85$ against $44$ before, gave the stale value in $23$ of $85$ by the recorded detector, $15$ on the first sentence and $7$ from tagged-only views.
The expiry obligations, unscorable, show the pinned date at work: under the oracle GPT-4.1 mini abstains on $10$ of the $14$ and Haiku 4.5 on $13$, and Haiku 4.5's answers now say the arrangement's date has passed as of today's date.
Stale use prints by first sentence and by source beside the recorded column in the ledger (\texttt{AUX-DS-A4-*}).

\begin{table}[H]
\caption{The answer stage under Claude Haiku 4.5 as the answerer, every policy, on the same $452$ distinct views GPT-4.1 mini answered; the columns of the second block of Table~\ref{tab:ds}, counted over the answered subset, and the lower rows the same counts under the answer stage run once more with the added instruction to treat tagged records as not to be used (Amendment 3). Descriptive; both registered after the first draft. Scored with the store's tags as joined, one retraction tag inverted (the negation paragraph).}
\label{tab:ds-h45}
\centering
\scriptsize
\setlength{\tabcolsep}{2.5pt}
\begin{tabular}{@{}lrrlr@{\hspace{10pt}}l@{}}\toprule
Policy & answered & stale at $0.80$ & stale at $0.50$, tagged / untagged & present answered & correct at $0.50$ \\ \midrule
Top-5 only & $54$ & $7$ & $8$ of $35$ / $11$ of $19$ & $174$ & $48$ ($0.2759$) \\
Oracle & $99$ & $9$ & $17$ of $82$ / $10$ of $17$ & $174$ & $48$ ($0.2759$) \\
Model rule & $74$ & $9$ & $15$ of $61$ / $8$ of $13$ & $155$ & $45$ ($0.2903$) \\
Lookup rule & $70$ & $7$ & $14$ of $61$ / $5$ of $9$ & $146$ & $41$ ($0.2808$) \\
Model rule, all states named & $74$ & $5$ & $12$ of $61$ / $8$ of $13$ & $155$ & $45$ ($0.2903$) \\
Lookup rule, all states named & $70$ & $5$ & $13$ of $61$ / $5$ of $9$ & $146$ & $40$ ($0.2740$) \\
Lookup, literal threshold & $94$ & $10$ & $18$ of $75$ / $11$ of $19$ & $174$ & $47$ ($0.2701$) \\
\addlinespace
\multicolumn{6}{@{}l}{\textit{With the added instruction to treat tagged records as not to be used (Amendment 3)}}\\
Top-5 only & $54$ & $3$ & $5$ of $35$ / $12$ of $19$ & $174$ & $49$ ($0.2816$) \\
Oracle & $99$ & $3$ & $15$ of $82$ / $9$ of $17$ & $174$ & $49$ ($0.2816$) \\
Model rule & $74$ & $7$ & $16$ of $61$ / $8$ of $13$ & $155$ & $47$ ($0.3032$) \\
Lookup rule & $70$ & $5$ & $14$ of $61$ / $5$ of $9$ & $146$ & $40$ ($0.2740$) \\
Model rule, all states named & $74$ & $4$ & $10$ of $61$ / $8$ of $13$ & $155$ & $47$ ($0.3032$) \\
Lookup rule, all states named & $70$ & $3$ & $10$ of $61$ / $5$ of $9$ & $146$ & $40$ ($0.2740$) \\
Lookup, literal threshold & $94$ & $7$ & $16$ of $75$ / $12$ of $19$ & $174$ & $47$ ($0.2701$) \\
\bottomrule\end{tabular}\end{table}

\begin{table}[H]
\caption{The answer stage with the store's untagged duplicates tagged (DS Amendment 5, descriptive; no prediction was registered). For each of the $35$ obligations whose view carried an active line asserting the stale value, every such line is retagged with the obligation's state (superseded or retracted) and nothing else in any view changes; $74$ of the $452$ distinct views change ($132$ lines) and are re-answered by both answerers under the Amendment 4 header, the other $378$ keeping their Amendment 4 answers. A line counts as asserting the stale value by the item's own committed detector at the ground-truth join's threshold: at least two content tokens shared with the stale field, coverage of the stale-only tokens at least $0.50$, coverage of the successor-only tokens below $0.34$. Stale is the detector's flag on the answer at $0.50$ (the ground-truth join's threshold) and at $0.80$ (the registered threshold), over the reached obligations of each policy; the last two columns restrict to the scorable supersession- and retraction-derived obligations and give the flag on the whole answer beside the flag on the answer's first sentence alone (the expiry obligations' gold field is the expired fact, so they are unscorable there); correct is over the licensed items, whose views do not change. Every cell reads Amendment 4 (recorded) to Amendment 5; mini is GPT-4.1 mini and Haiku is Haiku 4.5. Scored with the store's tags as joined, one retraction tag inverted (the negation paragraph).}
\label{tab:ds-a5}\centering\tiny\setlength{\tabcolsep}{2pt}
\begin{tabular}{@{}llrrrrrrr@{}}\toprule
Answerer & Policy & Reached & Stale at $0.50$ & Stale at $0.80$ & Correct (licensed) & Rate at $0.50$ & Scorable, whole answer & Scorable, first sentence \\ \midrule
mini & Top-5 & $54$ & $11 \to 3$ & $3 \to 1$ & $56$ of $174$ & $0.2037 \to 0.0556$ & $10 \to 2$ of $42$ & $10 \to 2$ of $42$ \\
mini & Model & $74$ & $11 \to 2$ & $2 \to 1$ & $53$ of $155$ & $0.1486 \to 0.0270$ & $10 \to 1$ of $62$ & $10 \to 1$ of $62$ \\
mini & Lookup & $70$ & $9 \to 1$ & $2 \to 1$ & $48$ of $146$ & $0.1286 \to 0.0143$ & $8 \to 0$ of $58$ & $8 \to 0$ of $58$ \\
mini & Model, all states & $74$ & $12 \to 2$ & $2 \to 1$ & $53$ of $155$ & $0.1622 \to 0.0270$ & $11 \to 1$ of $62$ & $11 \to 1$ of $62$ \\
mini & Lookup, all states & $70$ & $9 \to 1$ & $1 \to 1$ & $47$ of $146$ & $0.1286 \to 0.0143$ & $8 \to 0$ of $58$ & $8 \to 0$ of $58$ \\
mini & Literal & $94$ & $17 \to 5$ & $3 \to 1$ & $55$ of $174$ & $0.1809 \to 0.0532$ & $16 \to 4$ of $82$ & $16 \to 4$ of $82$ \\
mini & Oracle & $99$ & $18 \to 3$ & $2 \to 1$ & $56$ of $174$ & $0.1818 \to 0.0303$ & $17 \to 2$ of $85$ & $17 \to 2$ of $85$ \\ \midrule
Haiku & Top-5 & $54$ & $18 \to 11$ & $5 \to 4$ & $73$ of $174$ & $0.3333 \to 0.2037$ & $10 \to 3$ of $42$ & $6 \to 0$ of $42$ \\
Haiku & Model & $74$ & $27 \to 17$ & $8 \to 6$ & $64$ of $155$ & $0.3649 \to 0.2297$ & $19 \to 9$ of $62$ & $12 \to 3$ of $62$ \\
Haiku & Lookup & $70$ & $24 \to 16$ & $7 \to 6$ & $58$ of $146$ & $0.3429 \to 0.2286$ & $16 \to 8$ of $58$ & $11 \to 3$ of $58$ \\
Haiku & Model, all states & $74$ & $23 \to 13$ & $7 \to 4$ & $64$ of $155$ & $0.3108 \to 0.1757$ & $15 \to 5$ of $62$ & $11 \to 1$ of $62$ \\
Haiku & Lookup, all states & $70$ & $20 \to 12$ & $6 \to 4$ & $58$ of $146$ & $0.2857 \to 0.1714$ & $12 \to 4$ of $58$ & $9 \to 1$ of $58$ \\
Haiku & Literal & $94$ & $31 \to 18$ & $8 \to 5$ & $72$ of $174$ & $0.3298 \to 0.1915$ & $23 \to 10$ of $82$ & $15 \to 5$ of $82$ \\
Haiku & Oracle & $99$ & $32 \to 22$ & $7 \to 5$ & $73$ of $174$ & $0.3232 \to 0.2222$ & $23 \to 13$ of $85$ & $15 \to 7$ of $85$ \\
\bottomrule\end{tabular}
\end{table}
On the $74$ views that change, stale answers at $0.50$ fall from $37$ to $8$ for GPT-4.1 mini (a rate difference of $-0.39$, intervals $[-0.51,-0.28]$ and $[-0.65,-0.18]$ with questions grouped by persona) and from $38$ to $12$ for Haiku~4.5 ($-0.35$, $[-0.46,-0.24]$ and $[-0.59,-0.14]$); every transition is from stale to not stale, none the other way, the flag at $0.80$ falls to zero for both, and abstentions rise from $2$ to $9$ and from $12$ to $27$. After the retagging no changed view carries an active line asserting the stale value, so every remaining stale answer on them has a tagged source ($8$ and $12$); GPT-4.1 mini's $8$ all flag on the first sentence, while $6$ of Haiku~4.5's $12$ flag only beyond it, a mention of the tagged content inside an abstention or an explanation.

\begin{table}[H]
\caption{The answer stage along the depth of the retrieval view (DS Amendment 6, descriptive; no prediction was registered). Views are the first $k$ lines of the recorded ten-deep ranking rendered by the recorded renderer (the five-line view is the recorded one), answered under the Amendment 4 header by both answerers; the gated view renders ten lines where GPT-4.1 mini's recorded coverage score is at or above the base-rate threshold ($366$ of the $557$ items: $155$ of the $174$ licensed, $211$ of the $383$ absent) and five elsewhere, a mean of $8.29$ lines. Per policy: the licensed record in view, correct answers at $0.50$ and at $0.80$, correct among the items whose record is in view, abstentions on the licensed items, and abstentions on the $383$ absent items, with Wilson intervals on the rates. The last rows are the budget-matched comparison: the gated view against the flat curve interpolated at $8.29$ lines between $k=8$ and $k=9$, the difference with the item-level and persona-grouped bootstraps ($10{,}000$ draws, the three policies resampled together). Scored with the store's tags as joined, one retraction tag inverted (the negation paragraph).}
\label{tab:ds-a6}\centering\tiny\setlength{\tabcolsep}{2pt}
\begin{tabular}{@{}lrlllrrl@{}}\toprule
Policy & Lines & Record in view, of $174$ (Wilson) & Correct at $0.50$ (Wilson) & at $0.80$ & Among in view & Abstain, licensed & Abstain, absent, of $383$ (Wilson) \\ \midrule
\multicolumn{8}{@{}l}{\textit{GPT-4.1 mini}} \\
Flat $5$ & $5.00$ & $65$ $[0.3051,0.4474]$ & $56$ $[0.2569,0.3945]$ & $20$ & $48$ of $65$ & $33$ & $241$ $[0.5798,0.6761]$ \\
Flat $6$ & $6.00$ & $72$ $[0.3432,0.4881]$ & $59$ $[0.2729,0.4122]$ & $21$ & $50$ of $72$ & $30$ & $236$ $[0.5666,0.6635]$ \\
Flat $7$ & $7.00$ & $77$ $[0.3708,0.5168]$ & $62$ $[0.2890,0.4299]$ & $23$ & $54$ of $77$ & $25$ & $228$ $[0.5454,0.6433]$ \\
Flat $8$ & $8.00$ & $82$ $[0.3985,0.5453]$ & $65$ $[0.3051,0.4474]$ & $20$ & $57$ of $82$ & $21$ & $218$ $[0.5192,0.6179]$ \\
Flat $9$ & $9.00$ & $88$ $[0.4321,0.5791]$ & $67$ $[0.3160,0.4591]$ & $22$ & $60$ of $88$ & $18$ & $219$ $[0.5218,0.6204]$ \\
Flat $10$ & $10.00$ & $89$ $[0.4378,0.5847]$ & $70$ $[0.3323,0.4765]$ & $23$ & $62$ of $89$ & $21$ & $214$ $[0.5087,0.6076]$ \\
Gated & $8.29$ & $87$ $[0.4265,0.5735]$ & $69$ $[0.3269,0.4707]$ & $23$ & $61$ of $87$ & $24$ & $219$ $[0.5218,0.6204]$ \\
\addlinespace
\multicolumn{8}{@{}l}{\textit{Haiku 4.5}} \\
Flat $5$ & $5.00$ & $65$ $[0.3051,0.4474]$ & $73$ $[0.3487,0.4938]$ & $27$ & $59$ of $65$ & $45$ & $280$ $[0.6845,0.7730]$ \\
Flat $6$ & $6.00$ & $72$ $[0.3432,0.4881]$ & $75$ $[0.3597,0.5053]$ & $26$ & $64$ of $72$ & $36$ & $268$ $[0.6520,0.7435]$ \\
Flat $7$ & $7.00$ & $77$ $[0.3708,0.5168]$ & $74$ $[0.3542,0.4996]$ & $28$ & $65$ of $77$ & $34$ & $263$ $[0.6386,0.7311]$ \\
Flat $8$ & $8.00$ & $82$ $[0.3985,0.5453]$ & $78$ $[0.3763,0.5225]$ & $31$ & $67$ of $82$ & $35$ & $261$ $[0.6332,0.7261]$ \\
Flat $9$ & $9.00$ & $88$ $[0.4321,0.5791]$ & $80$ $[0.3874,0.5339]$ & $34$ & $70$ of $88$ & $33$ & $257$ $[0.6225,0.7162]$ \\
Flat $10$ & $10.00$ & $89$ $[0.4378,0.5847]$ & $85$ $[0.4153,0.5622]$ & $33$ & $72$ of $89$ & $32$ & $255$ $[0.6171,0.7112]$ \\
Gated & $8.29$ & $87$ $[0.4265,0.5735]$ & $85$ $[0.4153,0.5622]$ & $33$ & $72$ of $87$ & $34$ & $261$ $[0.6332,0.7261]$ \\
\bottomrule\end{tabular}\\[4pt]
\begin{tabular}{@{}llrrrll@{}}\toprule
\multicolumn{7}{@{}l}{\textit{The budget-matched comparison: the gated view minus the flat curve at $8.29$ lines}} \\
Answerer & rate & gated & flat at $8.29$ & difference & item interval & persona interval \\ \midrule
GPT-4.1 mini & record in view & $0.5000$ & $0.4813$ & $+0.0187$ & $[-0.0091,+0.0473]$ & $[+0.0028,+0.0374]$ \\
GPT-4.1 mini & correct at $0.50$ & $0.3966$ & $0.3769$ & $+0.0197$ & $[-0.0106,+0.0517]$ & $[-0.0161,+0.0575]$ \\
GPT-4.1 mini & abstain, absent & $0.5718$ & $0.5699$ & $+0.0019$ & $[-0.0202,+0.0235]$ & $[-0.0191,+0.0239]$ \\
Haiku 4.5 & record in view & $0.5000$ & $0.4813$ & $+0.0187$ & $[-0.0091,+0.0473]$ & $[+0.0028,+0.0374]$ \\
Haiku 4.5 & correct at $0.50$ & $0.4885$ & $0.4516$ & $+0.0369$ & $[+0.0156,+0.0623]$ & $[+0.0136,+0.0639]$ \\
Haiku 4.5 & abstain, absent & $0.6815$ & $0.6784$ & $+0.0030$ & $[-0.0186,+0.0251]$ & $[-0.0169,+0.0235]$ \\
\bottomrule\end{tabular}
\end{table}

\begin{table}[H]
\caption{The composite per policy over the full obligation set ($661$ questions) with its three components (DS Amendment 7, descriptive; no prediction was registered; no generative-model call). A licensed item counts when its answer is correct at $0.50$. An obligation counts when it is reached, answered and reported: the answer names the lifecycle state from a closed word list (superseded, retracted, expired, no longer, not current, outdated, cancelled, replaced, withdrawn, discontinued, past, and previous or former in a sentence that also carries one of those) without giving the stale value by the committed detector at $0.50$, or, on a supersession item, gives the successor value; the count by the successor alone is printed beside it. An absent item counts when the policy's coverage rule abstained on it or, where the rule let it through, the answerer abstained on its recorded five-line view (the Amendment 6 answers under the header; the absent items a gated rule answers on a longer view, ten under the model rule, seventeen under the lookup rule and seventy-two under the literal rule, are read on that longer view, answered under the same header prompt after the registration, $99$ distinct views per answerer; read on the five-line view as a proxy instead, every composite is within one of the value printed). The last column is the rule-only reading, in which an absent item counts only when the coverage rule abstained. The oracle's rule is the label itself, so its absent component is $383$ of $383$; the row is printed and enters no comparison. Rows: retrieval alone and the literal rule (no coverage rule), the gated rules under the header prompt, the same rules with the store's untagged duplicates tagged (tagged), then the oracle; the denominators are $174$ licensed items, $104$ obligations, $383$ absent items and $661$ in all. Scored with the store's tags as joined, one retraction tag inverted (the negation paragraph).}
\label{tab:ds-a7}\centering\tiny\setlength{\tabcolsep}{2pt}
\begin{tabular}{@{}llrrlrr@{}}\toprule
Answerer & Policy (prompt) & Licensed correct & Reported (by the successor) & Absent abstained: rule $+$ answerer & Composite & Rule-only \\ \midrule
GPT-4.1 mini & Top-5 (header) & $56$ & $4$ ($1$) & $241 = 0 + 241$ & $301$ & $59$ \\
GPT-4.1 mini & Top-5 (tagged) & $56$ & $4$ ($1$) & $241 = 0 + 241$ & $301$ & $59$ \\
GPT-4.1 mini & Literal rule (header) & $55$ & $4$ ($1$) & $242 = 0 + 242$ & $301$ & $58$ \\
GPT-4.1 mini & Literal rule (tagged) & $55$ & $4$ ($1$) & $242 = 0 + 242$ & $301$ & $58$ \\
GPT-4.1 mini & Model rule (header) & $53$ & $2$ ($1$) & $265 = 172 + 93$ & $320$ & $226$ \\
GPT-4.1 mini & Lookup rule (header) & $48$ & $2$ ($1$) & $276 = 174 + 102$ & $326$ & $223$ \\
GPT-4.1 mini & Model rule, all states (header) & $53$ & $2$ ($1$) & $264 = 172 + 92$ & $319$ & $226$ \\
GPT-4.1 mini & Lookup rule, all states (header) & $47$ & $2$ ($1$) & $276 = 174 + 102$ & $325$ & $222$ \\
GPT-4.1 mini & Model rule (tagged) & $53$ & $2$ ($1$) & $265 = 172 + 93$ & $320$ & $226$ \\
GPT-4.1 mini & Lookup rule (tagged) & $48$ & $2$ ($1$) & $276 = 174 + 102$ & $326$ & $223$ \\
GPT-4.1 mini & Model rule, all states (tagged) & $53$ & $2$ ($1$) & $264 = 172 + 92$ & $319$ & $226$ \\
GPT-4.1 mini & Lookup rule, all states (tagged) & $47$ & $2$ ($1$) & $276 = 174 + 102$ & $325$ & $222$ \\
GPT-4.1 mini & Oracle, absent by label (header) & $56$ & $5$ ($1$) & $383 = 383 + 0$ & $444$ & $443$ \\
GPT-4.1 mini & Oracle, absent by label (tagged) & $56$ & $7$ ($1$) & $383 = 383 + 0$ & $446$ & $445$ \\ \midrule
Haiku 4.5 & Top-5 (header) & $73$ & $16$ ($1$) & $280 = 0 + 280$ & $369$ & $88$ \\
Haiku 4.5 & Top-5 (tagged) & $73$ & $20$ ($1$) & $280 = 0 + 280$ & $373$ & $92$ \\
Haiku 4.5 & Literal rule (header) & $72$ & $21$ ($1$) & $280 = 0 + 280$ & $373$ & $92$ \\
Haiku 4.5 & Literal rule (tagged) & $72$ & $29$ ($1$) & $280 = 0 + 280$ & $381$ & $100$ \\
Haiku 4.5 & Model rule (header) & $64$ & $16$ ($1$) & $292 = 172 + 120$ & $372$ & $251$ \\
Haiku 4.5 & Lookup rule (header) & $58$ & $17$ ($1$) & $303 = 174 + 129$ & $378$ & $248$ \\
Haiku 4.5 & Model rule, all states (header) & $64$ & $21$ ($2$) & $292 = 172 + 120$ & $377$ & $255$ \\
Haiku 4.5 & Lookup rule, all states (header) & $58$ & $19$ ($2$) & $303 = 174 + 129$ & $380$ & $249$ \\
Haiku 4.5 & Model rule (tagged) & $64$ & $22$ ($1$) & $292 = 172 + 120$ & $378$ & $257$ \\
Haiku 4.5 & Lookup rule (tagged) & $58$ & $22$ ($1$) & $303 = 174 + 129$ & $383$ & $253$ \\
Haiku 4.5 & Model rule, all states (tagged) & $64$ & $26$ ($2$) & $292 = 172 + 120$ & $382$ & $260$ \\
Haiku 4.5 & Lookup rule, all states (tagged) & $58$ & $24$ ($2$) & $303 = 174 + 129$ & $385$ & $254$ \\
Haiku 4.5 & Oracle, absent by label (header) & $73$ & $27$ ($2$) & $383 = 383 + 0$ & $483$ & $481$ \\
Haiku 4.5 & Oracle, absent by label (tagged) & $73$ & $33$ ($2$) & $383 = 383 + 0$ & $489$ & $487$ \\
\bottomrule\end{tabular}
\end{table}

\begin{table}[H]
\caption{Label propagation inside the view (E4; registered, DS Amendment 8). A rule that reads only the assembled view propagates a tagged line's lifecycle tag to the untagged lines of the same attribute that repeat it under the census function unchanged (shared tokens $\ge 2$, cross guard $0.34$, coverage $\ge 0.80$), with an age-field ordering (primary) or without it (secondary), under the ground-truth tags and under FR-Bank's own tags. Of the $272$ audited stale-line instances, $187$ sit under a different attribute key from their tagged twin, $45$ fall below $0.50$ coverage of any same-attribute tagged line, $40$ between $0.50$ and $0.80$, and none reaches the threshold. Detector counts; the re-answers under own tags are descriptive, no prediction was registered on them. Scored with the store's tags as joined, one retraction tag inverted (the negation paragraph).}
\label{tab:ds-a8}
\centering\tiny\setlength{\tabcolsep}{2pt}
\begin{tabular}{@{}lrrrr@{}}
\toprule
Measure & GT tags, primary & GT tags, secondary & own tags, primary & own tags, secondary \\
\midrule
Views changed, of $452$ & $1$ & $1$ & $90$ & $123$ \\
Lines propagated & $1$ & $1$ & $117$ & $152$ \\
Audited duplicate items fully tagged, of $35$ & $0$ & $0$ & $15$ & $15$ \\
False propagation on the licensed items' active lines & $3$ / $2333$ ($0.13\%$) & $3$ / $2333$ ($0.13\%$) & $69$ / $2333$ ($2.96\%$) & $83$ / $2333$ ($3.56\%$) \\
\midrule
\multicolumn{5}{@{}l}{\textit{Re-answers on the changed views under own tags (before $\to$ after; stale at $0.50$ over answered obligation views; correct licensed answers)}}\\
\multicolumn{5}{@{}l@{}}{GPT-4.1 mini: top-five: $1$/$7$ $\to$ $0$/$2$; correct $11 \to 7$; model: $7$/$12$ $\to$ $3$/$7$; correct $9 \to 7$; oracle: $6$/$17$ $\to$ $1$/$8$; correct $11 \to 7$} \\
\multicolumn{5}{@{}l@{}}{Haiku 4.5: top-five: $1$/$7$ $\to$ $0$/$2$; correct $11 \to 7$; model: $6$/$12$ $\to$ $3$/$5$; correct $9 \to 7$; oracle: $4$/$15$ $\to$ $2$/$5$; correct $11 \to 7$} \\
\bottomrule
\end{tabular}
\end{table}

\noindent A header naming the persona and the date raises in-view correct answers reliably for both answerers.
A composite over all three jobs rises under the model rule by a reliable $+0.03$ for GPT-4.1 mini and stays level for Haiku 4.5 (Table~\ref{tab:ds-a7}).

Under the header prompt the composite under the model rule minus under retrieval alone is $+19$ of $661$ for GPT-4.1 mini, a rate difference of $+0.03$ with intervals $[+0.01,+0.05]$ at item level and $[+0.01,+0.05]$ grouped by persona, and $+3$ for Haiku~4.5 ($+0.00$, $[-0.01,+0.02]$ and $[-0.01,+0.02]$); the lookup rule reads $+25$ ($+0.04$, reliable) and $+9$ ($+0.01$, both intervals including zero).
The rise, where there is one, comes from the absent items the coverage rule abstains on and the answerer would have answered: on the five-line view the answerer already abstains on $241$ of the $383$ absent items (GPT-4.1 mini) and $280$ (Haiku~4.5), the rule adds $24$ and $12$ to those, and the licensed component falls by $3$ and $9$ where the rule abstains on a licensed item.
In the rule-only reading the same difference is $+167$ and $+163$ ($+0.25$, reliable for both answerers), carried entirely by the rule's abstentions, because retrieval alone has no coverage rule and counts none of the answerer's own.
Reported answers are rare for GPT-4.1 mini ($2$ of $74$ reached under the model rule, one of them by the successor alone) and a fifth to a quarter of the reached obligations for Haiku~4.5 ($16$ of $74$ under the header, $22$ with the duplicates tagged), most of them inside an abstention.
Agreement with the second annotator's fourth question is not reported: the validation is registered, with its workbooks built and no label collected at submission.

\paragraph{Answer-level outcomes under the policies (exploratory).}
The table and the sentences below are exploratory: every stale and correct count is a token detector's, read by one of us where flagged, and the detector's human validation is registered, with its workbooks built and no label collected at submission (the validation sample is described at the end of this appendix).
\begin{table}[H]
\caption{The downstream policies for GPT-4.1 mini on the full obligation set: forbidding records put in view, detector-flagged stale answers at the $0.50$ threshold, and correct answers on the licensed items. The flagged rate holds under every gated policy until the store's untagged duplicates are retagged (Table~\ref{tab:ds-a5}, Appendix~\ref{app:resp}). Scored with the store's tags as joined, one retraction tag inverted (the negation paragraph).}
\label{tab:ds-body}
\centering
\scriptsize
\setlength{\tabcolsep}{3pt}
\renewcommand{\arraystretch}{0.9}%
\begin{tabular}{@{}lrrllr@{}}
\toprule
Policy & \hd{Forbidding records}{in view (of $104$)} & \hd{Stale answers}{(of $104$)} & \hdl{Stale, of the reached}{scorable obligations} & \hdl{Stale, of the answered}{subset; Wilson} & \hd{Correct}{(of $174$)} \\
\midrule
Top-5 retrieval & $54$ & $9$ & $8$ of $42$, $0.19$ & $8$ of $42$, $0.19$ $[0.10,0.33]$ & $46$ \\
Model rule & $85$ & $9$ & $8$ of $73$, $0.11$ & $8$ of $62$, $0.13$ $[0.07,0.23]$ & $44$ \\
Lookup rule & $85$ & $7$ & $6$ of $73$, $0.08$ & $6$ of $58$, $0.10$ $[0.05,0.21]$ & $40$ \\
Model rule, all states named and scored & $89$ & $12$ & $11$ of $76$, $0.14$ & $11$ of $65$, $0.17$ $[0.10,0.28]$ & $44$ \\
Literal lookup rule (descriptive) & $94$ & $14$ & $13$ of $82$, $0.16$ & $13$ of $82$, $0.16$ $[0.10,0.25]$ & $46$ \\
Oracle & $99$ & $17$ & $16$ of $85$, $0.19$ & $16$ of $85$, $0.19$ $[0.12,0.28]$ & $46$ \\
\bottomrule
\end{tabular}
\par\smallskip
\begin{minipage}{0.97\linewidth}\scriptsize
The stale and correct counts are a token detector's, read by one of us where flagged, with its limits in Appendix~\ref{app:resp}; its human validation is registered, with its workbooks built and no label collected at submission. The stale rate is printed over two denominators, every reached obligation the detector can score (supersession and retraction in view; expiry is unscorable) and the subset the policy answered, because the gated policies abstain on some reached obligations (the model rule answers $62$ of its $73$ reached scorable obligations, the lookup rule $58$; the top-five, literal and oracle policies abstain on none). The preregistered reading is stale use at the detector's $0.80$ threshold over the obligations whose forbidding record is in view and tagged, and it is zero under every policy. The counts and rates here are the descriptive reading at $0.50$; the Wilson intervals describe each policy alone and are not a paired test. The items on which each rule fires, the reach as a rate, the mean view size, tagged records and coverage are in Table~\ref{tab:ds-body-lower}.
\end{minipage}
\end{table}
Gating a lifecycle fetch on the obligation judgment reaches far more forbidding records than retrieval alone at the same budget.
On the full obligation set the reach rises from $0.52$ to $0.82$ of the obligations at a mean view of about seven records, where a seven-record retrieval view reaches $0.60$.
The lookup rule reaches the same $0.82$ and the oracle $0.95$ (Table~\ref{tab:ds-body}).
The lookup rule also costs correct answers on the licensed items, $40$ against $46$ under retrieval alone.
Stale answers do not fall under the gated fetch, because the store keeps an untagged active duplicate of the superseded content.
Tags in view are respected: the detector flags GPT-4.1 mini on $0$ of the $50$ views where every stale line is tagged or absent.
Each of its $16$ flagged answers under the oracle falls on a view with an untagged duplicate.
As a rate over the reached scorable obligations, the detector flags GPT-4.1 mini in $0.19$ of cases under retrieval alone and $0.19$ under the oracle, and the gated rules read $0.13$ and $0.10$ (Table~\ref{tab:ds-body}).
On the $42$ obligations retrieval alone reaches, the oracle's flagged count is $9$ against $8$, two answers turning stale and one turning clean, so no paired comparison separates the policies.
Retagging the store's untagged duplicates, and nothing else, removes most of the detector-flagged stale answers, which points to the write path.
The retagging is reference-assisted, the lines identified with the obligation's reference information and the detector, so it is a diagnostic intervention rather than a deployable repair.
Retagging the $35$ duplicates the audit found cuts flagged stale answers under the gated fetch from $0.15$ to $0.03$ for GPT-4.1 mini, over all reached obligations, a wider denominator than Table~\ref{tab:ds-body}'s scorable set.
For Haiku~4.5 the rate falls from $0.37$ to $0.23$; the licensed views were unchanged by construction, so this measures no cost on them.
Haiku~4.5's residual comes from tagged lines, mostly a mention of the tagged content beyond the answer's first sentence (Table~\ref{tab:ds-a5}).

\begin{table}[H]
\caption{The downstream policies, the columns Table~\ref{tab:ds-body} leaves out: the items on which each rule fires, the reach of the forbidding record as a rate (of $104$ obligations), the Wilson interval of the stale rate over the reached scorable obligations (GPT-4.1 mini, the detector at $0.50$), the mean view size on the obligations, tagged records, and coverage of the $174$ licensed items (record in view without abstention). Scored with the store's tags as joined, one retraction tag inverted (the negation paragraph).}
\label{tab:ds-body-lower}
\centering
\scriptsize
\setlength{\tabcolsep}{3pt}
\begin{tabular}{@{}lrlrrr@{}}
\toprule
Policy & fires & reach rate; Wilson interval of the stale rate & mean view & tagged & coverage (of $174$) \\
\midrule
Top-5 retrieval & $0$ & $0.5192$; $[0.100,0.333]$ & $5.00$ & $35$ & $65$ \\
Model rule & $75$ & $0.8173$; $[0.067,0.234]$ & $7.08$ & $66$ & $60$ \\
Lookup rule & $77$ & $0.8173$; $[0.048,0.208]$ & $6.88$ & $66$ & $56$ \\
Model rule, all states named and scored & $110$ & $0.8558$; $[0.097,0.278]$ & $11.42$ & $72$ & $60$ \\
Literal lookup rule (descriptive) & $184$ & $0.9038$; $[0.095,0.253]$ & $7.42$ & $75$ & $65$ \\
Oracle & $104$ & $0.9519$; $[0.119,0.284]$ & $7.07$ & $82$ & $65$ \\
\bottomrule
\end{tabular}
\end{table}

The coverage rule at the base rate costs $-0.03$ of coverage on the $174$ items, a small loss with both intervals excluding zero.
The two registered bars were a reach gain of $0.50$ for the gold-label fetch and $60\%$ retention of that gain for the gated fetch.
The two bars that did not hold, the reach gain of $0.50$ and the $60\%$ retention, were set under the masked-retrieval premise (the $5\%$ pass rate), which the reach of $0.52$ replaced.
The gated fetch retains $0.76$ of that gain, and under the weaker reading, the interval grouped by persona, the lower end of $0.56$ falls below the $60\%$ bar.

\paragraph{The registered paragraph.} The downstream experiment's registered paragraph reads, with its numbers, as follows.
Acting on the obligation judgment changes what a bounded view carries. The store's truthful metadata flags $82$ of the $104$ forbidding records; the other $22$ ($18$ retraction-derived, $4$ expiry-derived) are a store-side fact of the kind Table~\ref{tab:believed} records, retrieve as active, and are out of reach of any policy that acts on the metadata. On the $104$ obligation items, the five-entry similarity view contains the forbidding record for $0.5192$; a gold-label fetch of the store's lifecycle records for the obligation's own attribute reaches $0.9519$, $0.4327$ above retrieval alone and short of the $0.50$ bar; gated on the smallest model's committed obligation score at its base rate, the fetch reaches $0.8173$, retaining $0.7561$ of that gain on the supersession-derived items (interval $[0.610, 0.872]$; $[0.562, 0.939]$ with questions grouped by user profile), short of the $60\%$ bar under the weaker reading. The superseded-names lookup reaches $0.8173$ at the same fetch budget, within tolerance of the model. Naming the retracted and expired attributes raises the gated reach on those obligations only from $0.7805$ to $0.7805$: the gate elicited under the counting summary does not fire on them. The coverage gate at its base rate costs $0.0287$ of coverage on the items whose licensed record the store holds, within tolerance. With the forbidding record in view and tagged, the smallest model gives the stale value as the answer in $0$ of $82$ cases at the registered detector threshold ($11$ of $82$ at the join's gold threshold; $6$ of $17$ where the record printed as active) (Appendix~\ref{app:resp}).

\paragraph{Reach by ranking depth, and the rates over the whole populations.}
The recorded ranking is ten deep, so the reach of a $k$-item retrieval view is a census over the $104$ obligations for $k$ up to $10$ (Table~\ref{tab:reach-k}).
The gated policies' mean view on those items is $7.08$ records for the model rule and $6.88$ for the lookup rule, so the seven-item view is their matched comparison.
The all-states policies at $10.74$ and $11.42$ records are matched by the ten-item view, the deepest recorded.
Stale use and correctness in Table~\ref{tab:ds-body} are counted over all $104$ obligations and all $174$ licensed items, so that an abstention or an unreached record counts as neither a stale answer nor a correct one.
Table~\ref{tab:ds} counts them over the answered subset.
The registered detector threshold of $0.80$ was calibrated on entry texts.
On a two-sentence answer the stale value's unique tokens are paraphrased, with coverage $0.5$ to $0.67$ on the answers that give the stale value.
That row therefore reads zero under every policy, and the gold threshold of $0.50$ is the sensitive one.
The audit discloses ten limits of the instrument, in one place.
First, the $15$ expiry obligations are unscorable for stale use, because the reference the detector holds for them is the expired fact itself, so the recorded column counts a mention of an expired arrangement as its use; the audit's own columns run on the $89$ supersession and retraction obligations.
Second, five retraction obligations cannot be flagged even when the store's own line is echoed verbatim; third, $6$ successor entry texts would be flagged as stale if echoed; fourth, $3$ licensed items cannot be marked correct even when their gold line is echoed.
Fifth, the flag is one-directional token coverage and rewards length, with an odds ratio of $2.34$ for GPT-4.1 mini and $2.99$ for Haiku 4.5 per doubling of answer length; on the first sentence alone Haiku 4.5's oracle count is $8$ of $85$ against $19$, and cut to GPT-4.1 mini's length it is $7$.
Sixth, for $35$ of the $89$ scorable obligations the view carries an active line asserting the stale value; seventh, one retraction statement carries the retracted tag because the ground-truth join reads no negation; eighth, the persona's name is bound to ``User'' in none of the views.
Ninth, the licensed record's median best rank under the retriever is $8.5$; tenth, twelve of Haiku 4.5's $38$ in-view abstentions on the $65$ licensed items were counted correct because their explanation restates the gold line.
As rates on the reached scorable obligations, retrieval alone and the oracle read $0.19$ and $0.19$ for GPT-4.1 mini and $0.26$ and $0.22$ for Haiku 4.5; on the $42$ obligations both policies reach, GPT-4.1 mini's stale answers move from $8$ to $9$ and Haiku 4.5's stay at $11$, and the $43$ obligations only the oracle reaches add $7$ and $8$.
Read by one of us, GPT-4.1 mini gives the stale value in $11$ of its $16$ scorable oracle flags and Haiku 4.5 in $8$ of its $19$, the rest naming the value while stating it superseded or restating the question's words; those classes are one reader's and not an independent validation.
A validation sample of $100$ answers is stratified across retrieval alone, the gated policies and the oracle, and across tagged and untagged reach.
Two annotators label whether the answer gives the stale value, whether it surfaces it, whether a licensed answer is correct, and whether the answer correctly reports the obligation; their agreement with each other, and with the detector at both thresholds on the first three, is not reported: the validation is registered, with its workbooks built and no label collected at submission.
The second annotator's sheet also carries the $19$ answers the audit read as giving the stale value ($11$ GPT-4.1 mini, $8$ Haiku 4.5), so the decisive labels receive two human reads.

\begin{table}[H]
\caption{Reach of the forbidding record by the depth of the retrieval view, a census over the $104$ obligations from the recorded ten-deep ranking, whole and by the state the obligation arises from. Reach grows with depth and stays below the gated policies at their matched view size.}
\label{tab:reach-k}
\centering
\scriptsize
\setlength{\tabcolsep}{5pt}
\begin{tabular}{@{}rrrrrr@{}}
\toprule
$k$ & reach (of $104$) & rate & supersession (of $63$) & retraction (of $26$) & expiry (of $15$) \\
\midrule
$5$ & $54$ & $0.5192$ & $22$ & $20$ & $12$ \\
$6$ & $60$ & $0.5769$ & $26$ & $22$ & $12$ \\
$7$ & $62$ & $0.5962$ & $27$ & $23$ & $12$ \\
$8$ & $69$ & $0.6635$ & $32$ & $25$ & $12$ \\
$9$ & $70$ & $0.6731$ & $33$ & $25$ & $12$ \\
$10$ & $74$ & $0.7115$ & $37$ & $25$ & $12$ \\
\bottomrule
\end{tabular}
\end{table}

\paragraph{What the gate buys, reach only.}
Three policies with no monitor call and no answer call, assembled from the recorded ranking and the store's records, are read against the model rule at its mean view of $5.43$ lines over the $661$ questions (Table~\ref{tab:gatereach}); over the $104$ obligations alone the same view averages $7.08$ lines, the denominator of Table~\ref{tab:ds-body-lower}.
A fetch keyed on the name the lookup matches best, with no gate, reaches $0.8164$ at the matched budget against the model rule's $0.8173$, a difference of $-0.0009$ with both intervals including zero.
Unmatched, it spends $6.51$ lines per question against $5.43$, a ratio of $1.20$, and cut to the matched budget it loses licensed records reliably ($-0.0375$).
The recorded similarity view at seven lines reaches $0.5962$, and matched $0.5443$.
A view that keeps only the newest record of each attribute reaches $0.2337$ matched, $-0.5836$ below the rule, a reliable difference, and none of the $26$ retraction obligations, because a cancellation is never the newest record of what it cancels.
All three predictions came out as written.

\begin{table}[H]
\caption{What the gate buys, reach only (registered; ADDENDUM\_R122\_GATE\_REACH): three policies assembled from the recorded ranking and the store's records with no reader (monitor) call and no answer call, against the recorded model rule. The always-open fetch keys the lifecycle fetch on the name the lookup matches best and never abstains; the similarity view is the recorded top-$k$ retrieval; the recency-aware view keeps the newest record of each attribute and fills to $k$. Each is read at its named depth and budget-matched to the model rule's mean view over the $661$ questions ($5.43$ lines) by interpolating between the two bracketing depths (the always-open fetch between $k=3$ and $4$ with weight $0.91$ on the deeper, the similarity and recency-aware views between $k=5$ and $6$ with weight $0.43$). Upper panel: the forbidding record in view over the $104$ obligations with the Wilson interval and the per-state census, and the difference from the model rule with both intervals. Lower panel: the licensed record in view over the $174$ licensed items with its difference from the model rule, and the mean lines per question over the $661$ and over the $104$. The fetch keyed on the best-matched name at four similarity lines reaches $85$ of $104$ at $5.53$ lines; the matched always-open row loses licensed records reliably because the truncated fill drops them; the recency-aware view reaches no retraction obligation because the retracted record is never the newest of its attribute. All three predictions came out as written: the ungated fetch within $0.05$ of the model rule at the matched budget, its unmatched form spending more lines, and the recency-aware view below the rule by more than $0.05$.}
\label{tab:gatereach}
\centering\tiny\setlength{\tabcolsep}{1.5pt}
\begin{tabular}{@{}p{0.2\linewidth}llllll@{}}
\toprule
Policy & in view (of $104$) & Wilson & sup.\ / retr.\ / exp.\ & minus the model rule & item interval & persona interval \\
\midrule
Model rule (recorded) & $85$ = $0.8173$ & -- & 53 / 20 / 12 & -- & -- & -- \\
Always-open fetch, five lines plus the fetch & $86$ = $0.8269$ & $[0.743,0.888]$ & 54 / 20 / 12 & $+0.0096$ & $[-0.0577,+0.0673]$ & $[-0.0804,+0.0980]$ \\
Always-open fetch, matched & $0.8164$ & -- & -- & $-0.0009$ & $[-0.0700,+0.0664]$ & $[-0.0893,+0.0890]$ \\
Similarity view, seven lines (recorded rows) & $62$ = $0.5962$ & $[0.500,0.685]$ & 27 / 23 / 12 & $-0.2212$ & $[-0.3077,-0.1346]$ & $[-0.3530,-0.0850]$ \\
Similarity view, matched & $0.5443$ & -- & -- & $-0.2730$ & $[-0.3621,-0.1865]$ & $[-0.3976,-0.1378]$ \\
Recency-aware view, seven lines & $28$ = $0.2692$ & $[0.193,0.362]$ & 17 / 0 / 11 & $-0.5481$ & $[-0.6635,-0.4423]$ & $[-0.6970,-0.3924]$ \\
Recency-aware view, matched & $0.2337$ & -- & -- & $-0.5836$ & $[-0.6936,-0.4749]$ & $[-0.7300,-0.4370]$ \\
\bottomrule
\end{tabular}\\[4pt]
\begin{tabular}{@{}p{0.2\linewidth}llllrr@{}}
\toprule
Policy & licensed (of $174$) & minus the model rule & item interval & persona interval & lines, $661$ & lines, $104$ \\
\midrule
Model rule (recorded) & $65$ = $0.3736$ & -- & -- & -- & $5.43$ & $7.08$ \\
Top-five retrieval (recorded) & $65$ = $0.3736$ & -- & -- & -- & $5.00$ & $5.00$ \\
Lookup rule (recorded) & $65$ = $0.3736$ & -- & -- & -- & $5.46$ & $6.88$ \\
Oracle (recorded) & $65$ = $0.3736$ & -- & -- & -- & $5.33$ & $7.07$ \\
Always-open fetch, five lines plus the fetch & $65$ = $0.3736$ & -- & -- & -- & $6.51$ & $6.39$ \\
Always-open fetch, matched & $0.3361$ & $-0.0375$ & $[-0.0657,-0.0159]$ & $[-0.0667,-0.0142]$ & $5.43$ & -- \\
Similarity view, seven lines (recorded rows) & $77$ = $0.4425$ & -- & -- & -- & $7.00$ & $7.00$ \\
Similarity view, matched & $0.3910$ & $+0.0175$ & $[+0.0075,+0.0299]$ & $[+0.0055,+0.0312]$ & $5.43$ & -- \\
Recency-aware view, seven lines & $55$ = $0.3161$ & -- & -- & -- & $7.00$ & $7.00$ \\
Recency-aware view, matched & $0.2916$ & $-0.0820$ & $[-0.1394,-0.0270]$ & $[-0.1519,-0.0156]$ & $5.43$ & -- \\
\bottomrule
\end{tabular}
\end{table}

\paragraph{Registered texts.} The eleven prediction texts of the mechanism tests and the downstream experiment, written before any call and filled with their numbers, are in the supplement (Section~S.3), and the registered outcome sentence for the neutral-slice prediction is in its Section~S.9. They are the record behind the outcomes stated in Table~\ref{tab:registrations-b}, and every number they carry also appears in the tables of this appendix.

\subsection{Spend and caching}\label{app:economics}

Monitoring runs before every query, so its cost turns on prompt caching.
The projection and the billed micro-run are in the supplement (Section~S.8).
Every model run in this paper was capped before its first call and metered from its rows at list or batch prices; the amounts, in dollars, are: the mechanism tests $0.94$ and the downstream experiment $0.11$; Haiku 4.5 on the retrieved-view monitors $6.72$ of a $12$ cap; the downstream answer-stage registrations $0.46$, $0.41$, $0.56$, $0.54$ (of $8$), $0.10$ (of $2$), $3.63$ (of $6$) and $0$; the counterfactual set $1.08$ of $22$ (OpenAI) and $72.50$ of $130$ (Anthropic); its expiry extension $2.21$ of $5$; the list-length check and its amendment $7.89$ of $28$; the paraphrase drafting run $0.12$ over $413$ calls; the long-list check on the counterfactual set $8.11$ of $12$ (Anthropic) and $15.54$ of $20$ (OpenAI); Haiku 4.5 under the store's own metadata $9.36$ of $12$; the increment moved within the triple $3.23$ of $4$; the forty-persona Graphiti ingestion $11.88$ of a $13.22$ stop, twelve aborted partial runs included (Appendix~\ref{app:posthoc-graphiti}); two later runs, the composite's answers on the actual expanded views \$$0.14$ and Opus 5 on the moved-count control \$$4.52$, under a \$$40$ ceiling for that pass.
GPT-5.6 Luna and GPT-5.6 Sol on the full populations were metered at $0.20$ and $1.20$, and at $5$ and $30$, dollars per million input and output tokens, the price constants of the full-population run code, to $2.67$ and $12.39$, the Sol run stopped; OpenAI's model pages of 18 September 2026 list $0.20$ and $1.20$ for Luna and $4$ and $20$ for Sol, with the first send of a long prompt billed at $1.25$ times the input rate, so the recorded figures are metered estimates rather than invoices.
The Graphiti follow-up's cumulative token-accounted list cost is $22.025336$ under that same ceiling; conservative input allowances and retained failed-request reservations put the upper accounting total at $26.825194$.
This includes completed Graphiti monitoring and answer generation, the repair answers, and both automated judging passes; the cached-depth judging that remains was not run.

\subsection{Provenance of the headline numbers}\label{app:provenance}

Each headline number rests on a population, a label source, a record alignment, a tag source and an evaluator, and they are not the same across the paper.
Table~\ref{tab:provenance} lists them in one place, with the validation status of each evaluator, so that a number is never read with a stronger warrant than its row gives it.

\begin{table}[H]
\caption{Provenance of the headline numbers: the population, the label source, how records are aligned to labels, the tag source the reader saw, the evaluator, and its validation status. ``Manifest'' is the human-written persona manifest; ``join'' is the token-overlap alignment of manifest events to stored records at the $0.80$ bar; ``hand-verified'' means every \pn{} label was checked by hand from its constituent judgments.}
\label{tab:provenance}
\centering
\scriptsize
\setlength{\tabcolsep}{2pt}
\renewcommand{\arraystretch}{0.95}
\begin{tabular}{@{}p{0.18\linewidth}p{0.12\linewidth}p{0.135\linewidth}p{0.125\linewidth}p{0.125\linewidth}p{0.115\linewidth}p{0.115\linewidth}@{}}
\toprule
Number (where) & Population & Label source & Record alignment & Tag source & Evaluator & Validation \\
\midrule
Counterfactual NOK AUROC, lookup $0.77$ and models $0.76$ to $0.85$ (Section~\ref{sec:counterfactual}, Table~\ref{tab:cf-main}) & $444$ items, $148$ triples & manifest; \pn{} labels hand-verified; question fixed across versions & not needed: versions built from the manifest & ground-truth summary & AUROC, persona-grouped bootstrap & registered predictions held; PN-vs-PRESENT decomposition post hoc \\
Paraphrase check, the lookup's loss of $0.02$ (Section~\ref{sec:lookup}, Table~\ref{tab:para-pooled}) & $345$ items, $115$ of the $148$ triples & manifest; labels transferred from the original triples, whose store versions are unchanged; rewrites screened by a word rule and read by an author with disclosed AI assistance & not needed: versions built from the manifest & ground-truth summary & AUROC, paired bootstrap (item and persona) & descriptive registration, no prediction; readings as they came out; the matched metric and the supersession-only comparison post hoc \\
Long-list check, the lookup's loss of $0.18$ and the two leads of $0.13$ to $0.14$ (Section~\ref{sec:lookup}, Tables~\ref{tab:robust} and~\ref{tab:cfpad-sup}) & $444$ items, $148$ triples, with $60$-name lists; the $88$ supersession triples registered & manifest; labels unchanged, the store versions unchanged, the list lengthened with the persona's own names & not needed: versions built from the manifest & ground-truth summary, the superseded-names list padded & pooled AUROC primary and the matched score secondary, both fixed in advance; paired bootstrap (item and persona; triple and persona) & registration committed before the run; two of four predictions held; Opus 5 at one draw \\
Own-name ablation, $0.86$ to $0.53$ (Section~\ref{sec:names_counts}) & $661$, $63$ supersession obligations & manifest; hand-verified & join, for the names removed & ground-truth summary, edited & AUROC, paired bootstrap & registered prediction held \\
Moved-count control (Section~\ref{sec:names_counts}) & $60$ counted-state triples & manifest & not needed & ground-truth summary, one count moved & AUROC, paired bootstrap; band reading & descriptive; fired as neither \\
Tagged view $0.69$, $0.83$, $0.81$ (Section~\ref{sec:routes}, Table~\ref{tab:routes}) & $661$, $104$ obligations & manifest; hand-verified & join (ground-truth tags: $82$ of $104$ forbidding records marked); the store's own tags for the $0.81$ row & ground-truth tags; FR-Bank's own tags & AUROC, paired bootstrap & L2 predictions failed; the own-tags row descriptive \\
Census of the stores (Section~\ref{sec:backends}, Table~\ref{tab:census}) & every entry of each store; forty personas & manifest & join at $0.80$ (FR-Bank, Graphiti); the store's own fields (Mem0, A-MEM) & the store's own marks & counts & no sampling interval; Graphiti's population restricted to FULL-variant obligations (Appendix~\ref{app:posthoc-graphiti}) \\
Store-written summary, retention $72\%$ (Section~\ref{sec:backends}, Table~\ref{tab:levels}) & $661$ & manifest; hand-verified & join, for the summary; the store's fields for the store-written one & FR-Bank's own metadata & AUROC; ratio of paired differences & descriptive; the sixty-percent line conventional \\
Coverage FOK, easy tier $0.78$ (Appendix~\ref{app:baselines}) & $354$, $243$ positives & manifest; presence control & join & ground-truth summary & AUROC, paired bootstrap & the margin over the whole-map lookup not reliable (post hoc, Appendix~\ref{app:baselines}) \\
Downstream reach, flagged stale answers, retagging (Section~\ref{sec:downstream}, Table~\ref{tab:reach}; Appendix~\ref{app:resp}, Table~\ref{tab:ds-body}) & $104$ obligations, $174$ licensed; $89$ scorable & manifest; hand-verified & join; the audit's read of the views & ground-truth tags in the view; reference-assisted retags & token detector at $0.50$; one reader's classification of flags & the two-annotator validation registered, not labelled; the retagging diagnostic, not a deployable repair \\
\bottomrule
\end{tabular}
\end{table}

\section{Registration record}\label{app:registration}

This appendix is the registration record: each frozen artifact, its commit or file hash, its freeze day, scope and outcome.
Freeze days count from the first freeze, day $0$; the rows of each table follow the registration order, and the hashes identify the frozen files.
Registered texts are quoted verbatim and use the artifact's names; Appendix~\ref{app:glossary} maps them to this paper's terms.
The outcome vocabulary is defined at the head of Appendix~\ref{app:crossvendor}, because that is where held and failed first appear.
The September additions were registered before their runs but after earlier results on the same items were known; we claim no confirmatory design for them.

\begin{table}[H]
\caption{The registration record lists each frozen monitoring artifact, commit, freeze day (day 0 is the first freeze), scope, and outcome. It distinguishes held or failed predictions from N-A method freezes.}
\label{tab:registrations}
\centering
\scriptsize
\setlength{\tabcolsep}{1.5pt}
\begin{tabular}{@{}p{0.15\linewidth}p{0.13\linewidth}p{0.11\linewidth}p{0.42\linewidth}p{0.13\linewidth}@{}}
\toprule
Artifact & Commit hash & Freeze day & What it froze & Outcome \\
\midrule
E0 preregistration & \texttt{0f2fde2c3b} & day 0 & pilot universe, labels, summary, prompts, decision rules, bootstrap, leakage; both hypotheses held & held \\
E0 Amendment 1 & \texttt{d4ef3bee6a} & day 0 & contamination sensitivity, absence split, PN verification; a method freeze carrying no prediction & no prediction \\
E0 Amendment 2 & \texttt{c603f4a5ce} & day 0 & blind-deduplicated sensitivity; a method freeze carrying no prediction & no prediction \\
E1 preregistration & \texttt{875f726fab} & day 0 & taxonomy, tiers, summary source, prompts, decision rules, metrics; all four hypotheses held & held \\
E1 Amendment 1 & \texttt{307865f414} & day 0 & two-channel presence control and FR-only NOK scope; a method freeze carrying no prediction & no prediction \\
Frontier addendum & \texttt{3c3ef7dd6d} & day 0 & matched frontier condition and interpretation rule, and three mechanism hypotheses. The two carrying a bar held; the third was registered as descriptive and its reading was not supported, so the registration's own rule declared the mechanism unresolved & held \\
Redraw addendum & \texttt{c92deefd44} & day 0 & multi-draw estimator and supersession rule; a method freeze carrying no prediction & no prediction \\
Capability-ladder addendum & \texttt{6860ad0a8c} & day 1 & matched ladder and five predictions: four held and the threshold on the largest model's blind coverage failed & held / failed \\
Count-destruction addendum & \texttt{281c410293} & day 1 & count-only conditions and conditional interpretation; a method freeze carrying no prediction & no prediction \\
NOK-factorial mini-registration & \texttt{8597caa2dd} & day 1 & factors, endpoint, bootstrap, and three predictions: two held and the interaction failed & held / failed \\
E1-XV preregistration & \texttt{776bce07b3} & day 4 & the cross-vendor question set, its three models, four predictions and one descriptive item: three held and the universal blind-obligation baseline failed & held / failed \\
E1-QP preregistration & \texttt{10b3d4bff8} & day 4 & question-only probes, matched-set checks and failure texts: two held and the neutralization check failed & held / failed \\
R74 X1: FOK question-signal probe & \texttt{35f8416a55} & day 14 & probe target, populations and threshold branches. Branches were enumerated in advance and one fired; no pass bar was set & no prediction \\
R74 X2: summary-ablation probe & \texttt{35f8416a55} & day 14 & four ablation conditions, recovery rule and integration branches. Branches were enumerated in advance and one fired; no pass bar was set & no prediction \\
R74 X3: billed economics micro-run & \texttt{35f8416a55} & day 14 & telemetry-only runs and the as-billed replacement rule. The branch that fired requires both cost orderings to print, and both do; no pass bar was set & no prediction \\
R74 X4: expansion to the full populations & \texttt{35f8416a55} & day 14 & the full FR-Bank question sets, their models and five predictions, all of which held. Two runs stopped at the budget limit and are disclosed as stopped & held \\
R74 X5: authorship-matched set & \texttt{35f8416a55} & day 14 & the one-voice question set, its rendering protocol, a label-transfer bar and a pause rule. A defect condition was repaired before scoring; of two predictions one held and the other failed on GPT-4.1 mini while holding on the largest & held / failed \\
R76: authorship-matched follow-up & \texttt{692e379bdc} & day 14 & a third model on the one-voice question set with three descriptive outcomes, the label-structure census and the probe. All three were registered as descriptive; no pass bar was set & no prediction \\
R77: wrong-summary control & \texttt{41e31e70ab} & day 14 & the manipulated-summary conditions, populations, estimators and all branch text: the store-specificity prediction held and the correspondence one failed & held / failed \\
R81: superseded-names baseline (NB-6) & \texttt{5c3621b75e} & day 19 & the name lookup, its two variants, populations, tolerance and all outcome text, frozen before scoring: the coverage prediction failed and the obligation one held & held / failed \\
R82-CF: store-counterfactual set & \texttt{3147a05adf} & day 19 & thirty question and slot pairs, three store versions each, two summaries and four predictions: three held and the aggregate replication failed, with the second clause of its frozen failure text struck as defective (deviation ledger, Appendix~\ref{app:registration}) & held / failed \\
R82-CAL: post-hoc calibration & \texttt{ef351e9be5} & day 20 & three monotone maps under leave-one-persona-out cross-validation on two populations: one held, and the second named a quantity that does not exist at that base rate, which is itself a registered branch & held / undefined \\
\bottomrule
\end{tabular}
\end{table}

\begin{table}[H]
\caption{The registration record, continued: the side experiments, the powered counterfactual set, the mechanism tests and the downstream experiment; the Outcome column lists each prediction's outcome. Every row was committed before its first call except DS Amendment 4 and Graphiti Amendment 2, which are attested by the runner's lock (file hashes shown).}
\label{tab:registrations-b}
\centering
\scriptsize
\setlength{\tabcolsep}{1.5pt}
\begin{tabular}{@{}p{0.15\linewidth}p{0.13\linewidth}p{0.11\linewidth}p{0.42\linewidth}p{0.13\linewidth}@{}}
\toprule
Artifact & Commit hash & Freeze day & What it froze & Outcome \\
\midrule
Side experiments: name-bearing summary (S1) & \texttt{e8b2630ce3} & day 33 & a manifest-built, name-bearing summary for the two structureless stores, the entry map with its hand check, a coverage floor, and two predictions on A-MEM, both of which held; Mem0 fell below the floor and prints as descriptive & held / descriptive \\
Side experiments: store-written-state cap (S2) & \texttt{e8b2630ce3} & day 33 & two filter rules on the store's own metadata and a precision bar neither reached, so the repair was withdrawn before any call; one descriptive condition, the store-written summary with its fifteen-name cap lifted, carrying no prediction & no prediction \\
R94: CF-2 store-counterfactual set & \texttt{f016f033e9} & day 33 & $148$ question and slot pairs drawn under predicates fixed before any verdict, three store versions each, three summaries, three readers, five predictions and their branch texts; all five held, two of them printing either way: the lookups within tolerance and every reader inside the totals-randomization band; five recorded deviations & held \\
RESP: mechanism tests & \texttt{17e6328c81} & day 33 & the own-name ablation, the retrieved-view and hybrid monitors, and the rubric-gate test on the full obligation population, six predictions and their failure texts: L1-P1 held; L1-P2, L2-P1, L2-P2, L2-P3, L3-P1 failed, all printed & L1-P1 held; L1-P2, L2-P1, L2-P2, L2-P3, L3-P1 failed. \\
DS: downstream experiment & \texttt{a5a2036c3b} & day 34 & six policies over the obligation and coverage scores already on record, their assembled views, reach and abstention, and one answer call per distinct view; five predictions: DS-P3 and DS-P5 held; DS-P1, DS-P2 and DS-P4 failed; two bars set under the 5\% premise the study corrected at L2; one prediction unfulfillable by its own gate, recorded as a deviation & held / failed \\
DS Amendment 4: a view header binding the entries to the persona and pinning the date & \texttt{fb4b2ca8c8} (file hash) & day 35 & the same $452$ views answered once more by GPT-4.1 mini and Haiku 4.5 with a header naming the persona and the current date; the same detector with first-sentence and source columns beside; two predictions on the $65$ licensed items in view, with failure texts & two predictions; both held \\
Graphiti, forty personas: the event census, the forbidding-record census, a summary route and a tag route & \texttt{43132a3469} & day 36 & the unmodified release ingesting all forty personas; the two censuses against the manifests with four predictions and their failure texts, P1 on the census recall and precision, P2 to P4 on the summary and tag routes & P1 failed on recall, held on precision; P2 held, P3 failed, P4 held on the corrected full-store population; the three-persona readings read here: GR-1 failed, GR-2 not computable, GR-3 held, GR-4 and GR-5 descriptive (Appendix~\ref{app:posthoc-graphiti}) \\
Graphiti Amendment 2: untagged, tagged and active-only answer views & \texttt{b87401f3} (file hash) & day 38 & freeze by hash before calls; $178$ common full-store items and $100$ FR-Bank-only ablated items; cached generation complete & human scoring and prediction rulings not completed at submission \\
\bottomrule
\end{tabular}
\end{table}

\begin{table}[H]
\caption{The registration record, continued: twelve descriptive registrations, each hashed by the runner before its first call (the commits of day 39 do not date them) and carrying no prediction and no bar, so nothing in them could hold or fail. Every one has run; the paraphrase check ran last, on the pairs an author's read accepted, and its dated notes and amendment are the row below it.}
\label{tab:registrations-c}
\centering
\scriptsize
\setlength{\tabcolsep}{1.5pt}
\begin{tabular}{@{}p{0.17\linewidth}p{0.11\linewidth}p{0.11\linewidth}p{0.42\linewidth}p{0.13\linewidth}@{}}
\toprule
Artifact & sha256 (first bytes) & Freeze day & What it froze & Outcome \\
\midrule
Q4: the marker read as absence & \texttt{8b6ee9e3} & day 34 & one instruction line appended to the otherwise unchanged structureless summaries; the recorded coverage items of both stores; the paired differences against the recorded intact and blind rows (Table~\ref{tab:q4-marker}) & no prediction \\
DS Amendment 1: the gate elicited under the all-states summary & \texttt{72ab84da} & day 34 & GPT-4.1 mini's obligation score re-elicited under the summary naming all three states on the $661$; one further policy reading it; the counted-state reach beside the recorded gate (Table~\ref{tab:ds}) & no prediction \\
List-length padding & \texttt{93e33022} & day 34 & the superseded-names list of each of the $661$ summaries padded with correct attribute names to $15$, $30$ and $60$; both lookups and GPT-4.1 mini at each length, paired against the $1.75$-name row; three branches enumerated & no prediction \\
Paraphrased counterfactual set & \texttt{45cad63d} & day 34 & each of the $148$ questions rewritten to share no content token with the slot path or any printed name, screened and read by us; both lookups and three models on the same store versions, paired against the originals; four branches enumerated & no prediction; run on day $42$ on the $115$ accepted pairs (Table~\ref{tab:para-pooled}) \\
Paraphrased counterfactual set: the dated notes and the amendment & \texttt{2848f1ad} (the whole file) & day 42 & below the registered text's marker, dated and never editing the text above: the read disclosed as AI-assisted (an assistant flagged rows with reasons and a coding session printed persona evidence; every verdict an author's); the floor amended from $120$ to the $115$ accepted pairs before any scoring call, with the $124$ screen-passing pairs as a sensitivity analysis, not registered; GPT-5.6 Luna and GPT-5.6 Sol added as readers, not registered, declared after the lookup result and before any model result & no prediction \\
Expiry extension of the counterfactual set & \texttt{2f555c1a} & day 35 & the $14$ eligible expiry pairs the counterfactual draw left undrawn, fixed by the candidate census (\texttt{rows/cf3/CF3\_CENSUS.md} in the artifact); the same construction, renderer, leakage screen and readers; a pooled expiry row with the $444$'s $30$ pairs (Table~\ref{tab:cf3-expiry}) & no prediction \\
DS Amendment 2: the literal rule's answer stage, and a second answerer & \texttt{a3ff9d89} & day 35 & the answer stage of the literal-threshold lookup rule under the downstream protocol ($268$ views, $30$ new calls); Haiku 4.5 as answerer on every distinct view of the seven policies ($452$ views); the same detector and columns (Tables~\ref{tab:ds} and~\ref{tab:ds-h45}) & no prediction \\
DS Amendment 3: the answer stage with an instruction to treat tagged records as not to be used & \texttt{eca19607} & day 35 & the same $452$ views answered once more by GPT-4.1 mini and Haiku 4.5 with one instruction added to the prompt; the same detector; stale and correct beside the recorded rows (Tables~\ref{tab:ds} and~\ref{tab:ds-h45}) & no prediction \\
Haiku 4.5 under the store's own metadata & \texttt{77de1e37} & day 35 & Haiku 4.5 at five calls on the recorded store-written texts under the capped and the lifted rendering; retention against its recorded blind and ground-truth rows; the lookup on the same lists; three branches enumerated (Table~\ref{tab:bh-split}) & no prediction; at or above the sixty-percent line under both \\
The moved-count control & \texttt{de2dd0ba} & day 35 & for each counted-state pair, the obligation's own per-category increment moved from the present-negative version to a sibling under a recorded seed; GPT-4.1 mini and Haiku 4.5 against their recorded superseded-only rows; the state-cell and totals readers; the band reading (Table~\ref{tab:v2rcat}) & no prediction; the band reading fired as neither \\
DS Amendment 5: the store's untagged duplicates tagged & \texttt{a820195f} & day 35 & every active line asserting the stale value on the $35$ obligations the audit found retagged with the obligation's state; the $74$ changed views re-answered under the Amendment 4 header by both answerers; the same detector (Table~\ref{tab:ds-a5}) & no prediction \\
DS Amendment 6: coverage-gated depth of the retrieval view & \texttt{52effd8d} & day 36 & views of five to ten lines from the recorded ten-deep ranking and a gated view that expands to ten where the recorded coverage score is at or above its threshold; both answerers under the header; the budget-matched comparison at the gated view's mean depth (Table~\ref{tab:ds-a6}) & no prediction \\
DS Amendment 7: an obligation-reported detector and a composite & \texttt{6566bf09} & day 36 & a closed word list scored on every recorded answer under every policy; a composite per policy over the full obligation set; the detector's agreement with the second annotator's fourth question (Table~\ref{tab:ds-a7}) & no prediction; no model call \\
\bottomrule
\end{tabular}
\end{table}

\begin{table}[H]
\caption{The registration record, continued: the fourth backend's frozen readings, the five registrations of the counterfactual and downstream follow-ups, the intent crossover, the four registrations of the encoder, draw, rubric and reach checks, and the hybrid reader; the Outcome column lists each prediction's outcome, and a prediction registered after an interim read of other models' rows is labelled so. Every row was committed before its first call.}
\label{tab:registrations-d}
\centering
\scriptsize
\setlength{\tabcolsep}{1.5pt}
\begin{tabular}{@{}p{0.15\linewidth}p{0.11\linewidth}p{0.11\linewidth}p{0.42\linewidth}p{0.15\linewidth}@{}}
\toprule
Artifact & Commit hash & Freeze day & What it froze & Outcome \\
\midrule
Graphiti, three personas: the ingestion trial and its readings & \texttt{f2d7688a54} & day 20 & the unmodified release on three personas, its census, and five readings written before any run and read at forty personas: two with a bar, that the obligation gain under the store-written summary excludes zero and that the superseded-names lookup sits within tolerance of the model or above it; one uncomputable on that population; two descriptive (Appendix~\ref{app:posthoc-graphiti}) & GR-1 (the gain) failed, GR-2 (coverage) not computable, GR-3 (the lookup) held, GR-4 and GR-5 descriptive, all read at forty personas \\
Moved-count control, Amendment 1: Opus 5 at five draws & \texttt{177697a4ae} & day 40 & Opus 5 at five draws on the moved-count texts against its recorded five-draw row; one prediction, that the one-draw movement on retraction and expiry replicates reliably (Table~\ref{tab:v2rcat}) & held \\
Moved-count control, Amendment 2 & \texttt{b0bee9329d} & day 40 & GPT-5.6 Sol and GPT-5.6 Luna at five draws against their five-draw rows; one prediction, that each falls reliably on both counted states (Table~\ref{tab:v2rcat}) & failed for both on retraction \\
Names against counts on the counterfactual set & \texttt{177697a4ae} & day 40 & the default summary with its counts removed and with its names removed, GPT-4.1 mini and Haiku 4.5 against their recorded one-draw and five-draw rows; two predictions, that names only sits within tolerance of the recorded row and that counts only falls reliably below names only, and one descriptive reading (Table~\ref{tab:cf-names-counts}) & both held \\
The marker ablation on the structureless stores & \texttt{177697a4ae} & day 40 & the constrained rendering with its marker omitted and reworded, GPT-4.1 mini on both stores and GPT-5.6 Luna through the raw API; one prediction, that the coverage loss disappears under the neutral wording, and one descriptive reading (Table~\ref{tab:marker}) & failed on Mem0, held on A-MEM \\
DS Amendment 8: label propagation inside the view & \texttt{177697a4ae} & day 40 & a view-only rule propagating a tagged line's tag to same-attribute lines that repeat it, under ground-truth and own tags; five predictions on the views changed, false propagation and the flagged rate (Table~\ref{tab:ds-a8}) & one held (false propagation under ground-truth tags at most $5\%$); three failed; one not defined \\
GPT-5.6 Sol and GPT-5.6 Luna on the counterfactual set & \texttt{177697a4ae} & day 40 & both models at five draws through the raw API on the $444$ under the default summary; two predictions, that Sol exceeds the name lookup reliably and that Luna does not (Table~\ref{tab:cf-main}) & Sol's held; Luna's failed, Luna exceeds it reliably \\
The intent crossover & \texttt{648d67192c} & day 40 & the crossed questions, their author check and leak screen, two rubrics, four models, three lookups; three predictions with failure texts and two descriptive readings (Table~\ref{tab:cf-intent}): that every model sits within tolerance of $0.50$ on the same-store contrast under the main-study rubric (observed: four below the band, GPT-4.1 mini $0.40$, Haiku 4.5 $0.38$, GPT-5.6 Luna $0.35$, GPT-5.6 Sol $0.44$, the crossed question scoring higher on the same store, and Opus 5 above it, $0.90$); that Opus 5 and Haiku 4.5 separate the intents under the intent-explicit rubric (Opus 5 $0.99$; Haiku 4.5 $0.55$, both intervals including $0.50$); that the name lookup sits within tolerance of $0.50$ ($0.47$, $0.45$) & the first failed on all five models and the second on Haiku 4.5; the third held \\
The intent crossover, Amendment 1: GPT-5.6 Sol & \texttt{2c441a72ac} & day 40 & Sol as a fifth reader of the crossover, one draw; its prediction, that it separates the intents under the intent-explicit rubric with both intervals excluding $0.50$, was made after an interim read of three models' rows and is labelled so & held ($0.88$; a prediction made after the interim read) \\
Lookup encoders & \texttt{77bc0d9595} & day 40 & the name lookup under two further encoders and a random-token control (Table~\ref{tab:enc-lookup}); one prediction, both encoders within $0.05$ of MiniLM on the counterfactual set & held \\
GPT-4.1 mini at five draws & \texttt{77bc0d9595} & day 40 & five draws at the provider's default temperature under both summaries (Table~\ref{tab:mini-k5}); one prediction, within $0.05$ of the one-draw row with fewer tied triples & held \\
The counterfactual set under the intent-explicit rubric & \texttt{77bc0d9595} & day 40 & GPT-5.6 Luna and GPT-4.1 mini on the $444$ under the rubric that states the intent rule (Table~\ref{tab:cf-p1}); three predictions: each within $0.05$ of its recorded row, the same per-state order, Luna's margin over the lookup reliable & the first and third held; the second failed on GPT-4.1 mini, whose counted states swap order with neither move reliable \\
What the gate buys, reach only & \texttt{77bc0d9595} & day 40 & an always-open fetch, the similarity view and a recency-aware view, budget-matched to the model rule (Table~\ref{tab:gatereach}); three predictions on reach and lines & all three held \\
A hybrid reader with no model in it & \texttt{fe6ff84e12} & day 40 & a logistic regression of the label on the name lookup's cosine, the category-guessing reader's share and an assigned-category indicator, fitted under leave-persona-out on the $444$ (Table~\ref{tab:cf-hybrid}); two predictions, that it exceeds the name lookup reliably and that GPT-5.6 Sol stays reliably above it & the first failed ($+0.0201$, the item-level interval including zero); the second held \\
\bottomrule
\end{tabular}
\end{table}

\begin{table}[H]
\caption{The registration record, continued: the sibling-name control and the names-against-counts amendment on the three stronger models, both registered after the matched metric had been read, the long-list check on the counterfactual set, registered after the paraphrase check had been read, and the sibling-name control at $60$ names, registered after the long-list check had been read; the Outcome column lists each prediction's outcome. Every row was committed before its first call.}
\label{tab:registrations-e}
\centering
\scriptsize
\setlength{\tabcolsep}{1.5pt}
\begin{tabular}{@{}p{0.15\linewidth}p{0.11\linewidth}p{0.11\linewidth}p{0.42\linewidth}p{0.15\linewidth}@{}}
\toprule
Artifact & Commit hash & Freeze day & What it froze & Outcome \\
\midrule
The sibling-name control & \texttt{5cd8f809f2} & day 41 & a fourth store version per supersession pair with a same-category sibling in place of the target's name, the name lookup, five models, the random-token lookup and the count reader on the matched \pn{}-against-sibling comparison (Table~\ref{tab:cf-sibling}); four readings with a bar, two of them (P-SB1, P-SB3) disclosed at the freeze as recorded values from the scorer's dry run and not counted as predictions & P-SB2 (every model at or above $0.80$) and P-SB4 (no model reliably above the lookup) held; P-SB1 and P-SB3 as recorded \\
Names against counts, Amendment 1: the three stronger models & \texttt{5cd8f809f2} & day 41 & E2's names-only and counts-only renderings on Opus 5, GPT-5.6 Luna and GPT-5.6 Sol (Table~\ref{tab:cf-nc-strong}); three predictions: names only within tolerance of the intact row, counts only reliably below names only, counts only above $0.50$ on both counted states under the matched metric; Sol's lane dropped by the registration's own budget rule & the first failed on GPT-5.6 Luna and held on Opus 5; the second held on both; the third failed on both; not evaluable on Sol \\
The long-list check on the counterfactual set & \texttt{06c611f} & day 43 & the $444$ items' superseded-names lists padded to $60$ names, one pad per triple drawn from the persona's own printed attribute names, the three lookups and the five models at their Table~\ref{tab:cf-main} settings (Opus 5 at one draw against its recorded five-draw row), pooled AUROC primary and the matched score secondary on the supersession triples (Table~\ref{tab:cfpad-sup}); four predictions: each of GPT-5.6 Luna, GPT-5.6 Sol and Opus 5 reliably above the lookup (P-L1), neither GPT-4.1 mini nor Haiku 4.5 reliably above it (P-L2), the lookup reliably losing more than $0.05$ against its unpadded row (P-L3), each of the three stronger models losing less than the lookup (P-L4) & P-L2 and P-L3 held; P-L1 and P-L4 failed on Opus 5 alone, Luna and Sol as predicted \\
The sibling-name control at $60$ names & \texttt{f224177} & day 44 & the $88$ supersession pairs' padded \pn{} version against a fourth version with one name swapped (identical pad and counts), the no-model readers recorded before the freeze, GPT-5.6 Luna, GPT-5.6 Sol and GPT-4.1 mini at five draws (Table~\ref{tab:sb60}); four predictions: Luna reliably above the name lookup (P-S1), Sol reliably above it (P-S2), mini not reliably above it (P-S3), Luna and Sol at or above $0.80$ matched with the persona interval above $0.50$ (P-S4) & P-S2, P-S3 and P-S4 held; P-S1 failed \\
\bottomrule
\end{tabular}
\end{table}

\noindent\textbf{The count of predictions.} A prediction counts when it was registered with a bar, whether it held, failed or could not be evaluated; readings registered as descriptive are not counted, and the answer-stage predictions of Graphiti Amendment 2, not completed at submission, are not in the count. Table~\ref{tab:registrations} carries $47$ such predictions with $8$ that came out otherwise, Table~\ref{tab:registrations-b} $17$ with $10$, Table~\ref{tab:registrations-c} none, and %
Table~\ref{tab:registrations-d} $29$ with $11$ and Table~\ref{tab:registrations-e} $13$ with $5$: $106$ predictions, $34$ of which came out otherwise, the count Section~\ref{sec:setup} states. The sibling-name control's P-SB1 and P-SB3 were disclosed at its freeze as recorded values (the scorer's pre-freeze dry run had computed the no-model readers) and are not counted as predictions.\\

\noindent\textbf{How each freeze is attested.} Each runner wrote a lock record at its start holding the registration's sha256, and dated additions sit below the registered-text marker, outside the hash. For most registrations a commit holding the registered text also predates the run's first output. For CF-3, committed 33 minutes after its first call, and for DS Amendments 1 to 4, Q4, RESP Amendment 1 and Graphiti Amendment 2, first committed on day $39$, the evidence that the text predates the run is the runner's lock record and the file times.\\

\noindent\textbf{When Opus 5 entered.} Opus 5 entered the study at the cross-vendor preregistration (the cross-vendor addendum, Table~\ref{tab:registrations}; day $4$), carrying three of its predictions. That freeze came after the main-study freeze of day $0$ and before the counterfactual freeze of day $19$. In the counterfactual set it is a descriptive run carrying no prediction, because it was named at the freeze and funded before its run was submitted.\\

\noindent\textbf{Deviation index.} Every deviation named in this paper is indexed in the supplement (Section~S.2), with the ledger entry and the appendix that discusses it, because the index is a lookup aid for the artifact rather than a result.

\section{Analyses done after seeing the data}\label{app:posthoc}

This appendix collects analyses done after seeing the data, alongside the registered Graphiti follow-up in Section~\ref{app:posthoc-graphiti}.
The exploratory analyses carry no registered prediction and are not read as though they had been predicted.
The analyses marked ``specified before recomputation'' (the PN-versus-PRESENT decomposition, the intent-aware lookup, the whole-map coverage lookup and the reading of the forty-persona census) were written up as definitions with expected outcomes in one file, hashed before any of them ran (sha256 \texttt{ae2600fc}), and are post-hoc analyses, not predictions made before the original study.

\subsection{Calibration after the fact}\label{app:posthoc-cal}

\begin{figure}[h]
\centering
\begin{tikzpicture}
\node[anchor=south west,inner sep=0] (img) at (0,0) {\includegraphics[width=\linewidth]{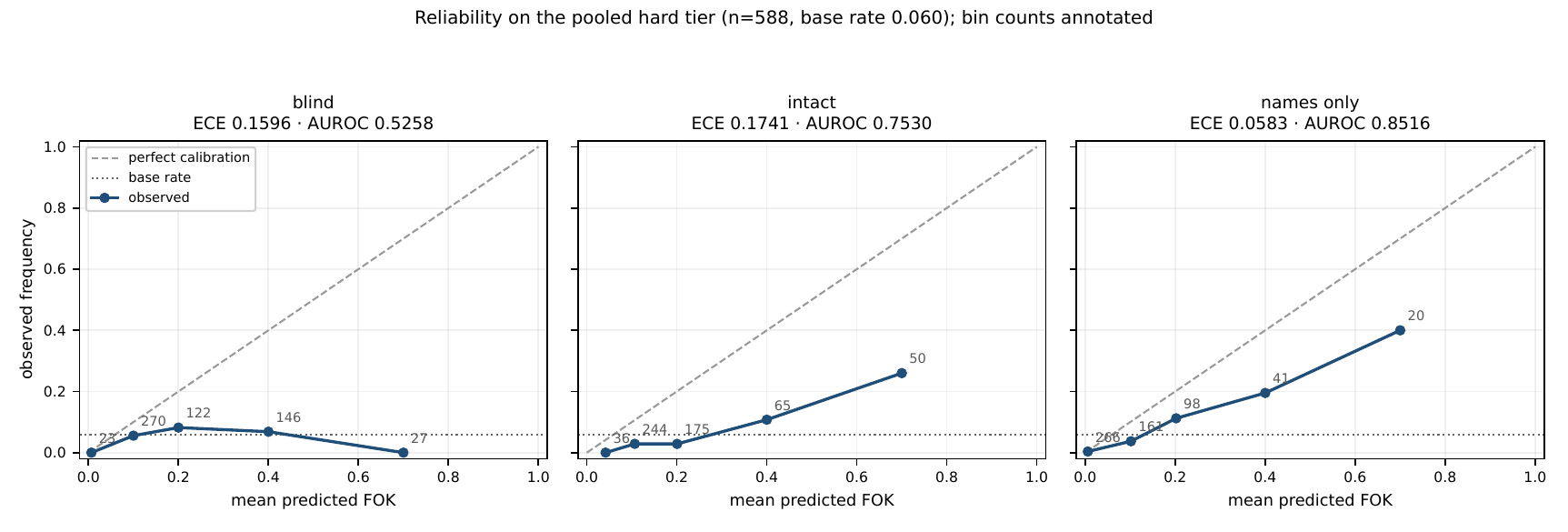}};
\begin{scope}[x={(img.south east)},y={(img.north west)}]
\node[anchor=south east,font=\scriptsize,text=black!60,inner sep=0.5pt] at (0.342,0.199) {base rate};
\node[anchor=south east,font=\scriptsize,text=black!60,inner sep=0.5pt] at (0.659,0.199) {base rate};
\node[anchor=south east,font=\scriptsize,text=black!60,inner sep=0.5pt] at (0.977,0.199) {base rate};
\end{scope}
\end{tikzpicture}
\caption{Reliability diagram for the pooled hard tier (bins with $n\ge10$). The indexed conditions rank well (Spearman $\rho=0.9000$ intact, $1.0000$ names-only) and scale badly (mean predicted $0.217$ against observed $0.056$, about $3.9\times$ over-confident), while the blind condition neither ranks ($\rho=0.1026$) nor calibrates: the elicited number behaves as a ranker rather than a probability.}
\label{fig:reliability}
\end{figure}
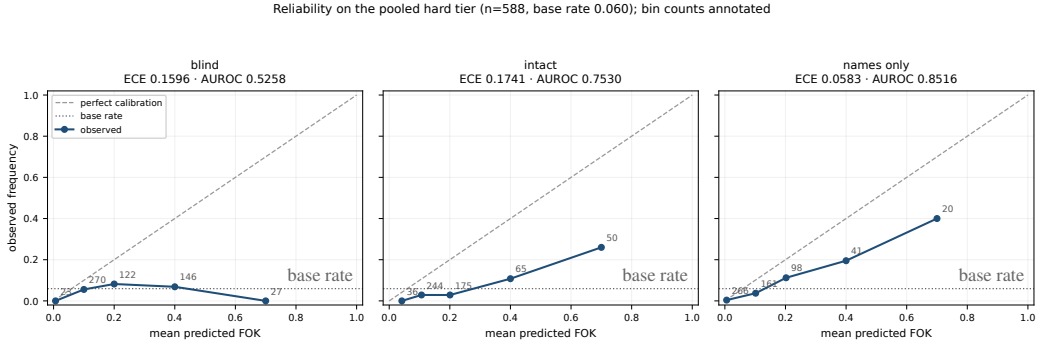

The elicited score is rescaled by three monotone maps \citep{platt2000,zadrozny2002,guo2017} under leave-one-persona-out cross-validation on the pooled hard-tier coverage set.
That set holds $588$ questions.
No question therefore contributes to the fit that predicts it.
Expected calibration error on the intact-summary score falls from $0.1741$ to $0.0146$ under an isotonic map and $0.0029$ under a logistic one.
That comparison needs its floor, and the floor is a constant predictor.
At a base rate of $5.95\%$ a constant predictor scores an expected calibration error of $0.0000$ and a Brier score \citep{brier1950} of $0.0560$.
The calibration numbers alone therefore understate how little is being asked.
Against that floor the rescaled score improves the Brier score to $0.0519$.
It also improves selective risk \citep{geifman2017} from $0.0615$ to $0.0373$, which is where its resolution shows.
Figure~\ref{fig:reliability} is the reliability diagram behind these numbers.

The rescaled probability never exceeds $0.375$ on the intact summary or $0.450$ on the names-only summary, so the value at which it would cross one half does not exist on either.
The comparison is therefore reported at each population's own base rate, where the crossings sit at $0.3$ and $0.2$ in score space.
At that threshold the rescaled intact-summary score flags $134$ of the $588$ hard questions and catches $24$ of the $35$ covered ones, a precision of $0.1791$ and a recall of $0.6857$ under the isotonic map.
Precision is low because the map's probabilities crowd the base rate, so the threshold that recovers most positives admits more than four false alarms for each true one.
Ordering is unchanged by construction, since every map is monotone.
Monotonicity is checked on each fitted map rather than on the pooled out-of-fold values.
Those pooled values are a mixture of forty maps, so their pooled ordering statistic moves.
On the question-only condition it reads $0.2572$ against a raw $0.4946$, a diagnostic of the pooling rather than a property of any map.

\subsection{A disclosure: the counterfactual set split by whether the question names its attribute}\label{app:posthoc-flags}

We read every question of the counterfactual set before scoring and flagged it when it names the plan or value its attribute is about (Appendix~\ref{app:counterfactual}).
The flag belongs to the question, so it is identical across the three store versions and cannot carry the label.
Table~\ref{tab:flag-split} splits the $444$ items by that flag, $89$ questions flagged and $59$ not, and reports every reader on each half.
The split answers whether the name lookup's score rides on questions that name what they ask about.
It does not: every reader scores higher on the questions that name only the topic, the name lookup at $0.8668$ against $0.7142$.
The split leaves the naming boundary in place on both halves.
It was made after the results were known, carries no prediction, and has no interval, since it is a deterministic census on the recorded rows.
It is a disclosure rather than a control, because the flag marks a question that names its attribute at the level of the topic while every value is scrubbed from every summary, so a flagged question need carry no more of the label than an unflagged one; the control is the paraphrase check reported in Appendix~\ref{app:counterfactual}, with this split recomputed on the rewrites (Table~\ref{tab:para-flags}).

\begin{table}[H]
\caption{Obligation AUROC on the counterfactual set by whether the question names the plan or value its attribute is about ($89$ flagged questions, $n=267$; $59$ unflagged, $n=177$), whole and by lifecycle state (supersession / retraction / expiry), for the two lookups on their own summaries and the three models under each summary. Every reader scores higher on the unflagged half. Exploratory: post hoc, no prediction, no interval.}
\label{tab:flag-split}
\centering
\scriptsize
\setlength{\tabcolsep}{3pt}
\begin{tabular}{@{}lllll@{}}
\toprule
Reader, summary & flagged, whole & flagged, by state & unflagged, whole & unflagged, by state \\
\midrule
Name lookup (sup.\ only) & $0.7142$ & $0.9047$ / $0.5000$ / $0.5000$ & $0.8668$ & $0.9789$ / $0.5000$ / $0.5000$ \\
All-states lookup (all three) & $0.8232$ & $0.8675$ / $0.7583$ / $0.8450$ & $0.9718$ & $0.9751$ / $0.9102$ / $1.0000$ \\
GPT-4.1 mini, sup.\ only & $0.6539$ & $0.7640$ / $0.5000$ / $0.5455$ & $0.8355$ & $0.9244$ / $0.5859$ / $0.5625$ \\
GPT-4.1 mini, all three & $0.7315$ & $0.7859$ / $0.8605$ / $0.5150$ & $0.9125$ & $0.9145$ / $0.9492$ / $0.8594$ \\
Haiku 4.5, sup.\ only & $0.6887$ & $0.8288$ / $0.5527$ / $0.5713$ & $0.8532$ & $0.9417$ / $0.5273$ / $0.6094$ \\
Haiku 4.5, all three & $0.8091$ & $0.8037$ / $0.8673$ / $0.7608$ & $0.9631$ & $0.9571$ / $0.9922$ / $0.9922$ \\
Opus 5, sup.\ only & $0.7997$ & $0.9217$ / $0.5863$ / $0.6245$ & $0.9060$ & $0.9593$ / $0.5781$ / $0.6055$ \\
Opus 5, all three & $0.8690$ & $0.9216$ / $0.7686$ / $0.8926$ & $0.9619$ & $0.9580$ / $0.9688$ / $1.0000$ \\
\bottomrule
\multicolumn{5}{@{}p{0.96\linewidth}@{}}{\footnotesize Positives by state and flag: supersession $45$ flagged / $43$ unflagged, retraction $22$ / $8$, expiry $22$ / $8$. The flags are our read of 2026-09-12, keyed by question; the whole-set values of Table~\ref{tab:counterfactual} are computed on the pooled items; a pooled AUROC also orders positives of one half against negatives of the other, so it is not a weighted average of the halves.}\\
\end{tabular}
\end{table}

\subsection{The lookup's loss under a real backend's faults, decomposed}\label{app:posthoc-faults}

A sweep run after every registered result was known asks how much of the name lookup's loss under FR-Bank's own list is due to each of three faults the census of Section~\ref{sec:backends} measured.
The faults are missing names (the list's recall of the ground-truth superseded names, $0.36$), wrong names (the share of negatives whose own name the list carries, $0.23$) and list length ($15$ names as rendered).
Each fault is injected into the ground-truth list at the store's measured rate, on the counterfactual set and on the $661$ questions, and the eight corners of the three-factor design give each factor's Shapley share of the total loss (Table~\ref{tab:faults}).
On the counterfactual set the embedding lookup falls from $0.77$ to $0.50$ under all three faults, and missing names carry about six tenths of that loss, wrong names about three tenths and length about one tenth.
On the $661$ questions it falls from $0.74$ to $0.61$; missing names carry about eight tenths and wrong names about four tenths, and length is a credit of about two tenths, because fillers drawn from the store's own names raise the absent probes less than the on-store items.
The three faults are sub-additive on both populations, since the single-factor losses sum past the total.
The sweep carries no prediction, and its report is in the artifact.

\begin{table}[H]
\caption{The lookup's loss under FR-Bank's measured faults, decomposed after the fact. Each row gives the ground-truth-list score, the score with all three faults injected at the store's measured rates (length $15$, contamination $0.232$, recall $0.3604$), and the Shapley share of the loss attributed to each fault; shares sum to one, and a negative share is a credit. Post hoc; no prediction.}
\label{tab:faults}
\centering
\scriptsize
\setlength{\tabcolsep}{4pt}
\begin{tabular}{@{}llrrrrr@{}}
\toprule
Population & lookup, slice & ground-truth & all faults & missing names & wrong names & length \\
\midrule
Counterfactual set ($444$) & embedding, whole set & $0.7663$ & $0.5005$ & $0.6010$ & $0.2943$ & $0.1047$ \\
Counterfactual set ($444$) & embedding, supersession & $0.9223$ & $0.5200$ & $0.5826$ & $0.1892$ & $0.2282$ \\
Counterfactual set ($444$) & lexical, whole set & $0.6698$ & $0.4949$ & $0.5682$ & $0.3753$ & $0.0565$ \\
Counterfactual set ($444$) & lexical, supersession & $0.7829$ & $0.5371$ & $0.6590$ & $0.2423$ & $0.0987$ \\
Full population ($661$) & embedding, whole set & $0.7442$ & $0.6096$ & $0.8127$ & $0.4167$ & $-0.2294$ \\
Full population ($661$) & embedding, supersession & $0.9599$ & $0.6593$ & $0.6423$ & $0.1675$ & $0.1902$ \\
Full population ($661$) & lexical, whole set & $0.5738$ & $0.5032$ & $0.6804$ & $0.3161$ & $0.0035$ \\
Full population ($661$) & lexical, supersession & $0.6241$ & $0.5345$ & $0.8936$ & $0.2833$ & $-0.1769$ \\
\bottomrule
\end{tabular}
\end{table}

\subsection{The fourth backend at forty personas}\label{app:posthoc-graphiti}

The three-persona trial of Appendix~\ref{app:census} was extended to every persona after the main results were known, under predictions written before the ingestion, and the reading of its population was fixed before any number here was printed.
The unmodified release (\texttt{graphiti-core} 0.28.1) ingested all forty personas at one episode per session: $1{,}365$ episodes rendered from FR-Bank's persona banks with every lifecycle field, status marker and supersession pointer removed, so the graph holds the same conversations FR-Bank holds and nothing of FR-Bank's own marks, and the ground truth is the persona manifest as everywhere in this paper.
The result is one graph per persona, $7{,}096$ entity nodes and $9{,}189$ edges; under the mapping of Appendix~\ref{app:census} (an invalidated edge with a successor is superseded, invalidated without one is expired, retraction inexpressible) $8{,}382$ edges are active, $130$ superseded and $677$ expired (the three sum to the $9{,}189$); inside those counts, $784$ carry no validity time and $72$ are dated after the persona's clock, an undated edge that no invalidation touches counting as active.
Every graph is the full store of its persona: the store variants of the main study, with categories or the successor removed, do not exist on it, so a Graphiti reading is valid only on the obligations of the full store, and the main text reports that population; the reading over all $104$ obligation rows, which repeats the same forbidding records on the same graph for the $57$ variant rows, prints below beside it because the prediction was written on it.

\begin{table}[H]
\caption{The forty-persona Graphiti census against the manifests. Recall is the share of ground-truth events, or forbidding records, on which some Graphiti edge asserting the old content carries a non-active state under the mapping above; stale-active means such an edge exists and every one is active; extraction miss means no edge asserts the content at the $0.80$ overlap bar. Precision is over all $807$ non-active edges of the forty graphs and is unchanged by the population. Retraction is inexpressible on this backend, so the eight obligations the ground truth marks retracted enter no recall.}
\label{tab:graphiti-g10}
\centering
\scriptsize
\setlength{\tabcolsep}{2.5pt}
\begin{tabular}{@{}>{\raggedright\arraybackslash}p{0.28\linewidth}rrrr>{\raggedright\arraybackslash}p{0.19\linewidth}>{\raggedright\arraybackslash}p{0.12\linewidth}@{}}
\toprule
Census & $n$ & recovered & stale-active & extraction miss & recall (bands $0.50$ / $0.75$) & precision \\
\midrule
Manifest supersession events & $109$ & $20$ & $12$ & $77$ & $0.1835$ ($57$ / $35$ of $109$) & $35$ of $807$, $0.0434$ \\
Forbidding records, obligations of the full store (the main-text reading) & $34$ of $47$ & $13$ & $10$ & $11$ & $0.3824$ & $54$ of $807$, $0.0669$ \\
\quad the same over distinct forbidding-record sets & $39$ of $56$ & $16$ & $11$ & $12$ & $0.4103$ & the same \\
Forbidding records, all $104$ obligation rows (the prediction's population) & $74$ of $104$ & $30$ & $21$ & $23$ & $0.4054$ ($48$ / $36$ of $74$); $0.3846$ over the $78$ by attack vector & the same \\
\bottomrule
\end{tabular}
\end{table}

By state on the full-store obligations, superseded preferences are recovered in $7$ of $16$, multi-version facts in $5$ of $11$ and expired arrangements in $1$ of $7$; over all $104$ rows the same three read $0.47$, $0.44$ and $0.09$, so expiry is the weak state on this backend under both readings.
The failure mode is extraction: of the $565$ manifest facts the graph asserts $134$ at the overlap bar, $77$ of the $109$ supersession events have no edge asserting the old value at all, and the user is split across two to four entity nodes in every persona, so a later fact about the same slot does not meet the earlier one.
The three trial personas, re-ingested into the forty-persona graphs, give the same recall on both graphs (elena $0.33$, tom $0.00$, yoli $0.25$) at $237$, $213$ and $224$ edges against the trial's $242$, $221$ and $222$.

The prediction fixed before the ingestion put the recall on the forbidding records at or below $0.25$ and the precision at or below $0.10$, the direction of the three-persona trial; the precision held and the recall did not, under the prediction's population and under both re-aggregations above.
Its outcome text, quoted verbatim with the artifact's names (Appendix~\ref{app:glossary}), reads: ``Over the forty personas Graphiti reaches 30 of the 74 forbidding records (the join's count; 30 of the attack-vector 78), above the registered quarter, at precision 0.0669.''
What the census establishes is therefore narrower than the trial suggested: some of the records the agent must not use are recovered by this backend's own invalidation, most are not, and almost every non-active mark it writes falls on a record the ground truth calls active.
The registered monitoring routes have since run on the $323$ full-store questions, with $47$ obligations (Table~\ref{tab:graphiti-completed}).
The originally named $661$-item form was not produced: the full graphs contradict $230$ ablated-variant labels, so those variants cannot be treated as their own reduced stores.
This population correction is recorded before those calls; results do not silently inherit the larger denominator.
The lifecycle mapping uses invalidation and matching successor edges for supersession, maps invalidated edges without a successor to expiry, and leaves retraction inexpressible.
Names reduce to relation types under the stock node labels.

\begin{table}[H]
\caption{Completed Graphiti monitoring on $323$ full-store questions, $47$ positive obligations. GPT-4.1 mini, one call per input; the name lookup uses the local embedding matcher. Paired intervals are item-level first, then persona-grouped. These are monitoring results, not final-answer improvements.}
\label{tab:graphiti-completed}
\centering\scriptsize
\begin{tabular}{@{}p{0.29\linewidth}rp{0.49\linewidth}@{}}
\toprule
Input or reader & NOK AUROC & Contrast and standing \\
\midrule
Store-written summary, capped & $0.5765$ & Minus ground truth $-0.1659$: $[-0.2593,-0.0764]$; $[-0.2533,-0.0876]$. P2 holds. \\
Store-written summary, lifted & $0.6511$ & Descriptive; the lifted name cap is $1{,}000$. \\
Ground truth on Graphiti's schema, capped / lifted & $0.7556$ / $0.6958$ & Descriptive schema controls. \\
Ground-truth summary / question only & $0.7424$ / $0.4964$ & Recorded comparators on the same population. \\
Store's superseded-name lookup & $0.6780$ & Minus capped monitor $+0.1014$ (from the unrounded scores): $[+0.0102,+0.2086]$; $[+0.0094,+0.1988]$. P3 fails; GR-3 holds. \\
Whole-map lookup, capped / lifted & $0.6994$ / $0.7959$ & Descriptive whole-map matcher. \\
Five retrieved edges, untagged & $0.6995$ & Descriptive. \\
Same five edges, native tags & $0.6854$ & Minus untagged $-0.0141$: $[-0.0786,+0.0554]$; $[-0.0613,+0.0325]$. P4 holds. \\
Tagged edges plus native summary & $0.7741$ & Minus tagged $+0.0887$: $[+0.0180,+0.1603]$; $[+0.0274,+0.1587]$. Descriptive. \\
\bottomrule
\end{tabular}
\end{table}

Read relative to what the graph retains rather than to the manifest, $23$ of the $34$ obligations it could express hold the forbidding content ($13$ with a non-active mark on it, $10$ with every such edge active) and $11$ were never extracted; Table~\ref{tab:graphiti-strata} re-scores every monitor row of Table~\ref{tab:graphiti-completed} on those strata.
On the $23$ retained, GPT-4.1 mini under the capped store-written summary reads $0.6920$ against $0.8433$ under the FR-Bank ground-truth summary on the same items, a loss that stays reliable, so part of the store-written loss is the summary's and part is extraction.

\begin{table}[H]
\caption{The rows of Table~\ref{tab:graphiti-completed} re-scored on strata of the $47$ full-store obligations, the $276$ negatives unchanged (post hoc; the edges read once from the forty graphs). Of the $34$ superseded-or-expired obligations the alignment marks, the graph retains the forbidding content for $23$ at the $0.80$ overlap bar ($13$ recovered, a non-active mark on a firing edge, and $10$ stale-active) and never extracted it for $11$; $28$ and $32$ are retained at the $0.75$ and $0.50$ bars; the $4$ retraction-derived obligations are inexpressible under the mapping and the $9$ the alignment does not mark have no stratum. On the $23$ retained, the capped store-written summary reads $-0.1512$ against the FR-Bank ground-truth summary on the same items with intervals $[-0.2866,-0.0160]$ and $[-0.2807,-0.0274]$, still a reliable loss, and the store's own name lookup $-0.1082$ ($[-0.2383,+0.0369]$; $[-0.1947,+0.0007]$); on the $13$ recovered the capped summary reads $-0.1047$ ($[-0.2913,+0.0921]$; $[-0.2905,+0.0803]$), not reliable at that size.}
\label{tab:graphiti-strata}
\centering\tiny\setlength{\tabcolsep}{1pt}
\begin{tabular}{@{}lrrrrrrrrrr@{}}
\toprule
Obligations & $n$ & \hd{store-written}{capped} & \hd{store-written}{lifted} & \hd{store's name}{lookup} & \hd{whole-map}{lookup} & \hd{five edges}{untagged} & \hd{native}{tags} & \hd{tagged +}{summary} & \hd{FR-Bank}{top-five} & \hd{FR-Bank}{ground truth} \\
\midrule
all $47$ (the printed rows) & $47$ & $0.5765$ & $0.6511$ & $0.6780$ & $0.6994$ & $0.6995$ & $0.6854$ & $0.7741$ & $0.6991$ & $0.7424$ \\
retained at $0.80$ (recovered or stale-active) & $23$ & $0.6920$ & $0.7043$ & $0.7350$ & $0.5983$ & $0.6360$ & $0.6159$ & $0.7307$ & $0.6160$ & $0.8433$ \\
recovered (a non-active mark on the edge) & $13$ & $0.7304$ & $0.7156$ & $0.7776$ & $0.5923$ & $0.6579$ & $0.7131$ & $0.8625$ & $0.6211$ & $0.8350$ \\
stale-active & $10$ & $0.6422$ & $0.6897$ & $0.6797$ & $0.6062$ & $0.6076$ & $0.4897$ & $0.5594$ & $0.6094$ & $0.8540$ \\
extraction miss & $11$ & $0.4559$ & $0.7574$ & $0.6574$ & $0.8337$ & $0.6354$ & $0.6313$ & $0.7413$ & $0.7724$ & $0.7624$ \\
retained at $0.75$ & $28$ & $0.6489$ & $0.7316$ & $0.7081$ & $0.6447$ & $0.6573$ & $0.6424$ & $0.7483$ & $0.6594$ & $0.8467$ \\
retained at $0.50$ & $32$ & $0.6318$ & $0.7399$ & $0.7191$ & $0.6741$ & $0.6470$ & $0.6325$ & $0.7560$ & $0.6642$ & $0.8261$ \\
retraction-derived (inexpressible) & $4$ & $0.6209$ & $0.4275$ & $0.6984$ & $0.8795$ & $0.8619$ & $0.9792$ & $0.9624$ & $0.6920$ & $0.5258$ \\
not marked by the alignment & $9$ & $0.4090$ & $0.4847$ & $0.5481$ & $0.7134$ & $0.8682$ & $0.7987$ & $0.8416$ & $0.8251$ & $0.5564$ \\
not retained (the three above) & $24$ & $0.4658$ & $0.6002$ & $0.6233$ & $0.7962$ & $0.7604$ & $0.7520$ & $0.8157$ & $0.7788$ & $0.6457$ \\
\bottomrule
\end{tabular}
\end{table}

Amendment 1's three monitoring predictions (P2 to P4 in Appendix~\ref{app:registration}) are that the store-written summary falls below the ground-truth summary by more than the tolerance, that the superseded-names lookup sits within tolerance of chance, and that the tagged view sits within tolerance of the untagged view.
Its five readings are GR-1, that the paired obligation gain on the store-written summary excludes zero; GR-2, the same for coverage; GR-3, that the superseded-names lookup sits within tolerance of the model or above it; GR-4, descriptive retention against FR-Bank's ground-truth gain; and GR-5, the census's failure mode.
The summary-route prediction holds; its registered text reads: ``Under Graphiti's store-written summary GPT-4.1 mini reads $0.5765$ against $0.7424$ under the ground-truth summary on the same questions, $-0.1659$ below with interval $[-0.2593,-0.0764]$; a matcher reading Graphiti's own list of superseded names reads $0.6780$.''
The lookup prediction fails: ``A matcher reading Graphiti's own list of superseded names reaches $0.6780$, $+0.1780$ from chance.''
The tag-route prediction holds: ``Shown the five retrieved edges with Graphiti's own states, GPT-4.1 mini reads $0.6854$ against $0.6995$ without them, within tolerance: the states this store writes do not reach the records that forbid use.''
GR-1's text reads: ``On a second store that records when a fact stopped being true, the obligation judgment does not rise beyond $+0.0801$ when the summary is supplied, so the gain measured on [the lifecycle store] does not reproduce where the lifecycle record is thinner.'' (The bracketed words stand in for the store's name as the registered text gives it.)
GR-3's reads: ``A matcher reading the store's own list of superseded names reaches $0.6780$ against the monitor's $0.5765$ on this store as well.''
The prediction that the native name lookup would remain near chance fails: its list carries useful matching signal despite weak summary monitoring.
For the capped summary minus question-only, the point difference is $+0.0801$; the item interval $[-0.0031,+0.1604]$ includes zero although the persona interval $[+0.0027,+0.1624]$ does not, so the weaker-interval rule makes GR-1 fail.
GR-2 is not computable: the full-store hard tier has $145$ questions and no coverage positives.
GR-4 is descriptive: native-summary gain retains $0.3257$ of the same-population ground-truth gain; GR-5 retains the extraction diagnosis.
The optional Haiku summary and hybrid readings remain unrun and their conditional rule is unevaluable.
The native tags' contrast does not establish equivalence, and detecting a correctly tagged obsolete edge is not discovering a backend defect.
The answer-stage comparison has generated its cached answers on $178$ common full-store items, plus $100$ FR-Bank-only ablated items, but its $622$-answer human sheet is blank.
Its registered human endpoints and prediction rulings were not completed at submission; the detector rows in the artifact are secondary and provisional.

\subsection{Exploratory source repair and practical-utility checks}\label{app:utility-completion}

These checks use repeatedly inspected development data and do not supply independent confirmation.
The original composite gains mainly reflect abstention on absent items; they do not establish correction of stale answers.
The existing repair comparison shares a coverage gate across five conditions: repaired top-five retrieval, an obligation-gated stale fetch, obligation-gated successor assembly, the same assembly without that gate, and matched-depth retrieval.
The complete population is $178$ items; all conditions retain $51$ coverage-gate exclusions.
The previously sampled $31$ questions span $24$ personas and include all five historical controls; the other $147$ questions remain semantically unjudged.
No answer or retrieval variant was regenerated.

We reproduced the original automated scores, then repaired every sampled reference packet by supplying all original user turns dated through the pinned query date, with session and turn identity retained.
The same reference rule applies to every condition, regardless of its answer or prior score.
Complete conversations remain available to human readers for context; assistant assertions are not personal-fact ground truth.
The original packets and judgments remain immutable.
The unchanged automated rubric was applied to $107$ distinct repaired inputs, representing $310$ answer cells.
For one question, a persona's hobbies, the original reference packet had omitted the session that supports the answer; with that source restored, the repaired judge marks every condition's answer correct for that question.
This is a new provisional judgment, not human adjudication or proof that every changed label was caused by the source addition alone.

\begin{table}[H]
\caption{Provisional automated outcomes on the fixed $31$-question repair sample. Strict correct / consequential stale endorsement; historical mention alone is not endorsement. All five conditions retain the common coverage gate. No registered human endpoint is replaced.}
\label{tab:repair-completion}
\centering\small
\begin{tabular}{@{}lcc@{}}
\toprule
Condition & GPT-4.1 mini & Haiku 4.5 \\
\midrule
Repaired top five & $9/5$ & $11/5$ \\
Obligation-gated stale fetch & $8/5$ & $10/3$ \\
Obligation-gated successor assembly & $9/5$ & $11/5$ \\
Successor assembly without obligation gate & $10/4$ & $12/4$ \\
Matched depth & $7/6$ & $11/5$ \\
\bottomrule
\end{tabular}
\end{table}

Assembly without the obligation gate versus repaired top five gains one correct answer and removes one stale endorsement for each answerer, each an absolute $1/31$.
The correctness difference is $+3.23$ percentage points with persona-grouped interval $[0.00,+10.01]$; it does not establish a reliable practical improvement.
Against matched depth, the differences are $+9.68$ points $[0.00,+18.92]$ for GPT-4.1 mini and $+3.23$ points $[0.00,+10.01]$ for Haiku.
Adding the obligation gate loses the one gain for both answerers.
All condition contrasts, adverse transitions and uncertain-reference sensitivity are retained in the artifact.
Fourteen distinct judgments combine a determinate truth label with an insufficient or conflicting reference rating, and three quotations fail exact-substring checks; these defects are flagged, not silently relabeled.
The original $32$ blind audit inputs and seven supplementary inputs have repaired human packets, but no human labels.
Record-identifier reach, evidence delivery, gate decisions and answer behavior are separate measurements; none substitutes for semantic evidence of improvement.

The cached depth application fixes $557$ questions ($174$ licensed, $383$ absent), two answerers, and $139$ expansions from five to ten records.
Coverage-score selection, the cheap name lookup and random selection share that expansion budget; always-five and always-ten are reference policies.
Boundary ties and random selection are expectations, not observed integer counts.
The selectors, source rule, endpoints and analysis were frozen before new semantic labels.
Judging covered $203$ of $1{,}289$ distinct semantic inputs across eight personas before the configured provider reported exhausted credits.
The full population remains fixed; the other outcomes are missing, and full-population semantic point differences are withheld rather than imputed.
The completed cached mechanical comparison and the partial semantic results are exploratory, not a validated deployment operating point.
At this budget, recorded answer-generation charges are $0.136574$ and $0.449085$ under coverage selection, compared with $0.174832$ and $0.555921$ under always-ten.
The corresponding maximum selector allowances are only $0.038258$ and $0.106836$ per batch, before unmeasured retrieval, latency and maintenance costs.
Applying the historically measured mean monitor cost to this batch would add $0.289083$; that is a scenario based on recorded monitoring telemetry, not a new charge or a free-cache assumption.
Quality retention is unestablished, and local name encoding and summary maintenance have not been fully priced.
Practical utility therefore remains unresolved; the data support monitoring discrimination and specific failure mechanisms, not a claim of lower total cost or fewer human-validated stale errors.
The registered human evaluation and detector validation remain explicit dependencies.


\begin{thebibliography}{66}


\bibitem[Asai et al.(2024)]{selfrag}
A.~Asai et al. Self-RAG: Learning to Retrieve, Generate, and Critique through Self-Reflection. \emph{ICLR}, 2024.

\bibitem[Brier(1950)]{brier1950}
G.~W. Brier. Verification of forecasts expressed in terms of probability. \emph{Monthly Weather Review}, 78(1):1--3, 1950.

\bibitem[Cameron et al.(2008)]{cameron2008}
A.~C. Cameron, J.~B. Gelbach, and D.~L. Miller. Bootstrap-based improvements for inference with clustered errors. \emph{Review of Economics and Statistics}, 90(3):414--427, 2008.

\bibitem[Chao et al.(2026)]{stale}
H.~Chao et al. STALE: Can LLM Agents Know When Their Memories Are No Longer Valid? \emph{arXiv preprint arXiv:2605.06527}, 2026.

\bibitem[Chhikara et al.(2025)]{mem0}
P.~Chhikara et al. Mem0: Building Production-Ready AI Agents with Scalable Long-Term Memory. \emph{arXiv preprint arXiv:2504.19413}, 2025.

\bibitem[Dhingra et al.(2022)]{dhingra2022}
B.~Dhingra et al. Time-Aware Language Models as Temporal Knowledge Bases. \emph{Transactions of the Association for Computational Linguistics}, 10:257--273, 2022.

\bibitem[Doyle(1979)]{doyle1979}
J.~Doyle. A truth maintenance system. \emph{Artificial Intelligence}, 12(3):231--272, 1979.

\bibitem[Efron \& Tibshirani(1993)]{efron1993}
B.~Efron and R.~J. Tibshirani. \emph{An Introduction to the Bootstrap}. Chapman \& Hall, 1993.

\bibitem[Fan et al.(2026)]{statemembench}
X.~Fan et al. Can Agent Memory Systems Track Evolving State? \emph{arXiv preprint arXiv:2608.19652}, 2026.

\bibitem[Geifman \& El-Yaniv(2017)]{geifman2017}
Y.~Geifman and R.~El-Yaniv. Selective Classification for Deep Neural Networks. \emph{NeurIPS}, 2017.

\bibitem[Guo et al.(2017)]{guo2017}
C.~Guo et al. On Calibration of Modern Neural Networks. \emph{ICML}, 2017.

\bibitem[Hanley \& McNeil(1982)]{hanley1982}
J.~A. Hanley and B.~J. McNeil. The meaning and use of the area under a receiver operating characteristic (ROC) curve. \emph{Radiology}, 143(1):29--36, 1982.

\bibitem[Hart(1965)]{hart1965}
J.~T. Hart. Memory and the feeling-of-knowing experience. \emph{Journal of Educational Psychology}, 56(4):208--216, 1965.

\bibitem[Hu et al.(2026)]{memoryagentbench}
Y.~Hu, Y.~Wang, and J.~McAuley. Evaluating Memory in LLM Agents via Incremental Multi-Turn Interactions. \emph{ICLR}, 2026.

\bibitem[Jiang et al.(2023)]{flare}
Z.~Jiang et al. Active Retrieval Augmented Generation. \emph{EMNLP}, 2023.

\bibitem[Kadavath et al.(2022)]{kadavath2022}
S.~Kadavath et al. Language Models (Mostly) Know What They Know. \emph{arXiv preprint arXiv:2207.05221}, 2022.

\bibitem[Kamath et al.(2020)]{kamath2020}
A.~Kamath, R.~Jia, and P.~Liang. Selective Question Answering under Domain Shift. \emph{ACL}, 2020.

\bibitem[Kirichenko et al.(2025)]{abstentionbench}
P.~Kirichenko, M.~Ibrahim, K.~Chaudhuri, and S.~J. Bell. AbstentionBench: Reasoning LLMs Fail on Unanswerable Questions. \emph{arXiv preprint arXiv:2506.09038}, 2025.

\bibitem[Koriat(1993)]{koriat1993}
A.~Koriat. How do we know that we know? The accessibility model of the feeling of knowing. \emph{Psychological Review}, 100(4):609--639, 1993.

\bibitem[Koriat \& Goldsmith(1996)]{koriat1996}
A.~Koriat and M.~Goldsmith. Monitoring and control processes in the strategic regulation of memory accuracy. \emph{Psychological Review}, 103(3):490--517, 1996.

\bibitem[Lewis et al.(2020)]{lewis2020}
P.~Lewis et al. Retrieval-Augmented Generation for Knowledge-Intensive NLP Tasks. \emph{NeurIPS}, 2020.

\bibitem[Lin et al.(2022)]{lin2022}
S.~Lin, J.~Hilton, and O.~Evans. Teaching Models to Express Their Uncertainty in Words. \emph{Transactions on Machine Learning Research}, 2022.

\bibitem[Liu \& Gabriel(2026)]{pmbench}
G.~Liu and S.~Gabriel. PM-Bench: Evaluating Prospective Memory in LLM Agents. \emph{arXiv preprint arXiv:2607.12385}, 2026.

\bibitem[Longpre et al.(2021)]{longpre2021}
S.~Longpre et al. Entity-Based Knowledge Conflicts in Question Answering. \emph{EMNLP}, 2021.

\bibitem[Maharana et al.(2024)]{locomo}
A.~Maharana et al. Evaluating Very Long-Term Conversational Memory of LLM Agents. \emph{ACL}, 2024.

\bibitem[Maini et al.(2024)]{maini2024}
P.~Maini et al. TOFU: A Task of Fictitious Unlearning for LLMs. \emph{COLM}, 2024.

\bibitem[Mallen et al.(2023)]{mallen2023}
A.~Mallen et al. When Not to Trust Language Models: Investigating Effectiveness of Parametric and Non-Parametric Memories. \emph{ACL}, 2023.

\bibitem[Mireshghallah et al.(2025)]{cimemories}
N.~Mireshghallah et al. CIMemories: A Compositional Benchmark for Contextual Integrity of Persistent Memory in LLMs. \emph{arXiv preprint arXiv:2511.14937}, 2025.

\bibitem[Mullick \& T{\"u}z{\"u}n(2026)]{fr}
A.~Mullick and E.~T{\"u}z{\"u}n. Fortunate Recall: Ontology-Driven Memory Lifecycle Management for Persistent Coherence in LLMs. \emph{arXiv preprint arXiv:2609.10413}, 2026.

\bibitem[Nelson \& Narens(1990)]{nelson1990}
T.~O. Nelson and L.~Narens. Metamemory: a theoretical framework and new findings. \emph{Psychology of Learning and Motivation}, 26:125--173, 1990.

\bibitem[Nissenbaum(2004)]{nissenbaum2004}
H.~Nissenbaum. Privacy as contextual integrity. \emph{Washington Law Review}, 79(1):119--158, 2004.

\bibitem[Packer et al.(2023)]{packer2023}
C.~Packer et al. MemGPT: Towards LLMs as Operating Systems. \emph{arXiv preprint arXiv:2310.08560}, 2023.

\bibitem[Patel(2026)]{supersede}
V.~Patel. Supersede: Diagnosing and Training the Memory-Update Gap in LLM Agents. \emph{arXiv preprint arXiv:2606.27472}, 2026.

\bibitem[Platt(2000)]{platt2000}
J.~C. Platt. Probabilities for SV Machines. In \emph{Advances in Large-Margin Classifiers}, pages 61--74. MIT Press, 2000.

\bibitem[Rajpurkar et al.(2018)]{rajpurkar2018}
P.~Rajpurkar, R.~Jia, and P.~Liang. Know What You Don't Know: Unanswerable Questions for SQuAD. \emph{ACL}, 2018.

\bibitem[Rasmussen et al.(2025)]{zep}
P.~Rasmussen et al. Zep: A Temporal Knowledge Graph Architecture for Agent Memory. \emph{arXiv preprint arXiv:2501.13956}, 2025.

\bibitem[Ren et al.(2026)]{gatemem}
Z.~Ren et al. GateMem: Benchmarking Memory Governance in Multi-Principal Shared-Memory Agents. \emph{arXiv preprint arXiv:2606.18829}, 2026.

\bibitem[Shapiro et al.(2011)]{shapiro2011}
M.~Shapiro, N.~Pregui\c{c}a, C.~Baquero, and M.~Zawirski. Conflict-free replicated data types. In \emph{Stabilization, Safety, and Security of Distributed Systems (SSS)}, LNCS, pages 386--400. Springer, 2011.

\bibitem[Snodgrass \& Ahn(1986)]{snodgrass1986}
R.~Snodgrass and I.~Ahn. Temporal databases. \emph{Computer}, 19(9):35--42, 1986.

\bibitem[Su et al.(2024)]{su2024}
W.~Su et al. DRAGIN: Dynamic Retrieval Augmented Generation based on the Real-time Information Needs of Large Language Models. \emph{ACL}, 2024.

\bibitem[Tian et al.(2023)]{tian2023}
K.~Tian et al. Just Ask for Calibration: Strategies for Eliciting Calibrated Confidence Scores from Language Models Fine-Tuned with Human Feedback. \emph{EMNLP}, 2023.

\bibitem[Uddin et al.(2026)]{memora}
M.~N. Uddin et al. From Recall to Forgetting: Benchmarking Long-Term Memory for Personalized Agents. \emph{arXiv preprint arXiv:2604.20006}, 2026.

\bibitem[Wu et al.(2025)]{longmemeval}
D.~Wu et al. LongMemEval: Benchmarking Chat Assistants on Long-Term Interactive Memory. \emph{ICLR}, 2025.

\bibitem[Wu et al.(2026)]{automem}
S.~Wu et al. AutoMem: Automated Learning of Memory as a Cognitive Skill. \emph{arXiv preprint arXiv:2607.01224}, 2026.

\bibitem[Xiong et al.(2024)]{xiong2024}
M.~Xiong et al. Can LLMs Express Their Uncertainty? An Empirical Evaluation of Confidence Elicitation in LLMs. \emph{ICLR}, 2024.

\bibitem[Xu et al.(2024)]{xu2024conflicts}
R.~Xu et al. Knowledge Conflicts for LLMs: A Survey. \emph{EMNLP}, 2024.

\bibitem[Xu et al.(2025)]{amem}
W.~Xu et al. A-MEM: Agentic Memory for LLM Agents. \emph{NeurIPS}, 2025.

\bibitem[Xu et al.(2026)]{hushbench}
L.~Xu et al. HUSH-Bench: Measuring Memory-Use Boundaries for Sensitive History in Conversational Agents. \emph{arXiv preprint arXiv:2606.06055}, 2026.

\bibitem[Yang et al.(2024)]{yang2024}
Y.~Yang et al. Alignment for Honesty. \emph{NeurIPS}, 2024.

\bibitem[Zadrozny \& Elkan(2002)]{zadrozny2002}
B.~Zadrozny and C.~Elkan. Transforming classifier scores into accurate multiclass probability estimates. \emph{KDD}, 2002.

\bibitem[Zhang et al.(2025)]{zhang2025survey}
Z.~Zhang et al. A Survey on the Memory Mechanism of Large Language Model-based Agents. \emph{ACM Transactions on Information Systems}, 43(6):1--47, 2025.

\bibitem[Zhong et al.(2024)]{zhong2024}
W.~Zhong et al. MemoryBank: Enhancing Large Language Models with Long-Term Memory. \emph{AAAI}, 2024.

\end{thebibliography}
\end{document}